\documentclass[12pt,  openany, cmyk, dvipsnames, svgnames]{book} 
\usepackage[utf8]{inputenc}
\usepackage[a4paper, top=25mm,bottom=35mm,right=2.5cm, left=2.5cm]{geometry} 

\usepackage{emnlp2021}
\usepackage{graphicx}
\graphicspath{{images/}}
\usepackage{float}
\usepackage{caption}
\usepackage{subcaption}
\usepackage{amsmath}
\usepackage{linguex}
\usepackage{kotex}
\usepackage{indentfirst}

\usepackage{cancel}
\usepackage{stfloats}
\usepackage{amsbsy}
\usepackage{amssymb}
\usepackage{graphicx}
\usepackage[export]{adjustbox}
\usepackage{footnote}
\usepackage{multirow}
\makesavenoteenv{tabular}
\makesavenoteenv{table}
\usepackage{url}
\usepackage{tikz}
\usetikzlibrary{fit,positioning}
\usepackage{booktabs}
\usepackage{tabularx}
\usepackage{makecell}
\usepackage[ruled,vlined]{algorithm2e}
\usepackage[english]{babel} 
\usepackage{arabtex}
\usepackage{utf8}
\setcode{utf8}

\DeclareMathOperator*{\KL}{KL}
\DeclareMathOperator*{\ELBo}{ELBo}
\DeclareMathOperator*{\IWAE}{IWO}

\DeclareMathOperator*{\MHA}{MHA}
\DeclareMathOperator*{\TransEnc}{Enc}
\DeclareMathOperator*{\TransDec}{Dec}
\DeclareMathOperator*{\FF}{F}
\DeclareMathOperator*{\SA}{SA}
\DeclareMathOperator*{\CA}{CA}
\DeclareMathOperator*{\PE}{PE}
\DeclareMathOperator*{\ARTransDec}{\overline{Dec}}
\DeclareMathOperator*{\QKVDec}{QKVDec}
\DeclareMathOperator*{\ARQKVDec}{\overline{QKVDec}}
\DeclareMathOperator*{\Categorical}{Categorical}
\DeclareMathOperator*{\Concat}{Cat}
\DeclareMathOperator*{\SoftPlus}{SoftPlus}

\DeclareMathOperator*{\bos}{bos}
\DeclareMathOperator*{\eos}{eos}

\DeclareMathOperator*{\Gammaopenc}{\Gamma^{enc}}

\DeclareMathOperator*{\Gammaopdec}{\Gamma^{dec}}
\DeclareMathOperator*{\argmax}{argmax}
\DeclareMathOperator*{\argr}{arg_{\textit r}}
\DeclareMathOperator*{\rop}{\rho_{\textit r}}

\DeclareMathOperator*{\Attention}{Attention}
\DeclareMathOperator*{\softmax}{softmax}
\usepackage{makecell}
\usepackage{tikz-dependency}
\usetikzlibrary{decorations}
\usetikzlibrary{decorations.pathreplacing}
\usepackage[ruled,vlined]{algorithm2e}
\usetikzlibrary{shapes.geometric}
\pgfdeclarelayer{bg}    
\pgfsetlayers{bg, depgroups, main}  

\usepackage{roboto}  
\usepackage[T1]{fontenc}

\newcommand{\textBF}[1]{%
    \pdfliteral direct {2 Tr 0.35 w} 
     #1%
    \pdfliteral direct {0 Tr 0 w}%
}
\definecolor{DDbule}{RGB}{45, 72, 207}

\usepackage[hidelinks]{hyperref}
\hypersetup{
    colorlinks=true,
    linkcolor=DDbule,
    citecolor=DDbule,
    urlcolor=DDbule,
    linktoc=all
}

\usepackage{fancyhdr}
\newcommand{\changefont}{%
    \fontsize{10}{12}\selectfont
}
\usepackage[Lenny]{fncychap} 
\ChTitleVar{\LARGE\bfseries} 

\usepackage{setspace}

\usepackage{chemformula}

\usepackage[nottoc,notlof,notlot]{tocbibind} 

\usepackage{ctable}

\usepackage{longtable}
\usepackage{tabularx, blindtext}

\usepackage{array}
\newcommand{\PreserveBackslash}[1]{\let\temp=\\#1\let\\=\temp}
\newcolumntype{C}[1]{>{\PreserveBackslash\centering}p{#1}}
\newcolumntype{R}[1]{>{\PreserveBackslash\raggedleft}p{#1}}
\newcolumntype{L}[1]{>{\PreserveBackslash\raggedright}p{#1}}
\usepackage{multirow} 
\usepackage{color}
\definecolor{lightgray}{rgb}{0.95, 0.95, 0.95}
\definecolor{darkgray}{rgb}{0.4, 0.4, 0.4}
\definecolor{editorGray}{rgb}{0.95, 0.95, 0.95}
\definecolor{editorOcher}{rgb}{1, 0.5, 0} 
\definecolor{editorGreen}{rgb}{0, 0.5, 0} 
\definecolor{orange}{rgb}{1,0.45,0.13}		
\definecolor{olive}{rgb}{0.17,0.59,0.20}
\definecolor{brown}{rgb}{0.69,0.31,0.31}
\definecolor{purple}{rgb}{0.38,0.18,0.81}
\definecolor{lightblue}{rgb}{0.1,0.57,0.7}
\definecolor{lightred}{rgb}{1,0.4,0.5}
\usepackage{upquote}
\usepackage{listings}
\lstdefinelanguage{CSS}{
  keywords={color,background-image:,margin,padding,font,weight,display,position,top,left,right,bottom,list,style,border,size,white,space,min,width, transition:, transform:, transition-property, transition-duration, transition-timing-function},	
  sensitive=true,
  morecomment=[l]{//},
  morecomment=[s]{/*}{*/},
  morestring=[b]',
  morestring=[b]",
  alsoletter={:},
  alsodigit={-}
}

\lstdefinelanguage{JavaScript}{
  morekeywords={typeof, new, true, false, catch, function, return, null, catch, switch, var, if, in, while, do, else, case, break},
  morecomment=[s]{/*}{*/},
  morecomment=[l]//,
  morestring=[b]",
  morestring=[b]'
}

\lstdefinelanguage{HTML5}{
  language=html,
  sensitive=true,	
  alsoletter={<>=-},	
  morecomment=[s]{<!-}{-->},
  tag=[s],
  otherkeywords={
  >,
	<!DOCTYPE,
  </html, <html, <head, <title, </title, <style, </style, <link, </head, <meta, />,
	</body, <body,
	</div, <div, </div>, 
	</p, <p, </p>,
	</script, <script,
  <canvas, /canvas>, <svg, <rect, <animateTransform, </rect>, </svg>, <video, <source, <iframe, </iframe>, </video>, <image, </image>, <header, </header, <article, </article
  },
  ndkeywords={
  =,
  charset=, src=, id=, width=, height=, style=, type=, rel=, href=,
  fill=, attributeName=, begin=, dur=, from=, to=, poster=, controls=, x=, y=, repeatCount=, xlink:href=,
  margin:, padding:, background-image:, border:, top:, left:, position:, width:, height:, margin-top:, margin-bottom:, font-size:, line-height:,
  transform:, -moz-transform:, -webkit-transform:,
  animation:, -webkit-animation:,
  transition:,  transition-duration:, transition-property:, transition-timing-function:,
  }
}

\lstdefinestyle{htmlcssjs} {
  basicstyle={\footnotesize\ttfamily},   
  frame=b,
  xleftmargin={0.75cm},
  numbers=left,
  stepnumber=1,
  firstnumber=1,
  numberfirstline=true,	
  identifierstyle=\color{black},
  keywordstyle=\color{blue}\bfseries,
  ndkeywordstyle=\color{editorGreen}\bfseries,
  stringstyle=\color{editorOcher}\ttfamily,
  commentstyle=\color{brown}\ttfamily,
  language=HTML5,
  alsolanguage=JavaScript,
  alsodigit={.:;},	
  tabsize=2,
  showtabs=false,
  showspaces=false,
  showstringspaces=false,
  extendedchars=true,
  breaklines=true,
  literate=%
  {Ö}{{\"O}}1
  {Ä}{{\"A}}1
  {Ü}{{\"U}}1
  {ß}{{\ss}}1
  {ü}{{\"u}}1
  {ä}{{\"a}}1
  {ö}{{\"o}}1
}

\lstdefinestyle{py} {%
language=python,
literate=%
*{0}{{{\color{lightred}0}}}1
{1}{{{\color{lightred}1}}}1
{2}{{{\color{lightred}2}}}1
{3}{{{\color{lightred}3}}}1
{4}{{{\color{lightred}4}}}1
{5}{{{\color{lightred}5}}}1
{6}{{{\color{lightred}6}}}1
{7}{{{\color{lightred}7}}}1
{8}{{{\color{lightred}8}}}1
{9}{{{\color{lightred}9}}}1,
basicstyle=\footnotesize\ttfamily,
numbers=left,              
numbersep=5pt,             
tabsize=4,                 
extendedchars=true,        
breaklines=true,           
keywordstyle=\color{blue}\bfseries,
frame=b,
commentstyle=\color{brown}\itshape,
stringstyle=\color{editorOcher}\ttfamily, 
showspaces=false, 
showtabs=false,   
xleftmargin=17pt,
framexleftmargin=17pt,
framexrightmargin=5pt,
framexbottommargin=4pt,
showstringspaces=false,
}

\begin{document}
\mainmatter
\pagestyle{empty}
\begin{titlepage}
    \begin{center}
    
        \begin{figure}[!htb]
           \begin{minipage}{0.32\textwidth}
             \centering
             \includegraphics[width=\linewidth]{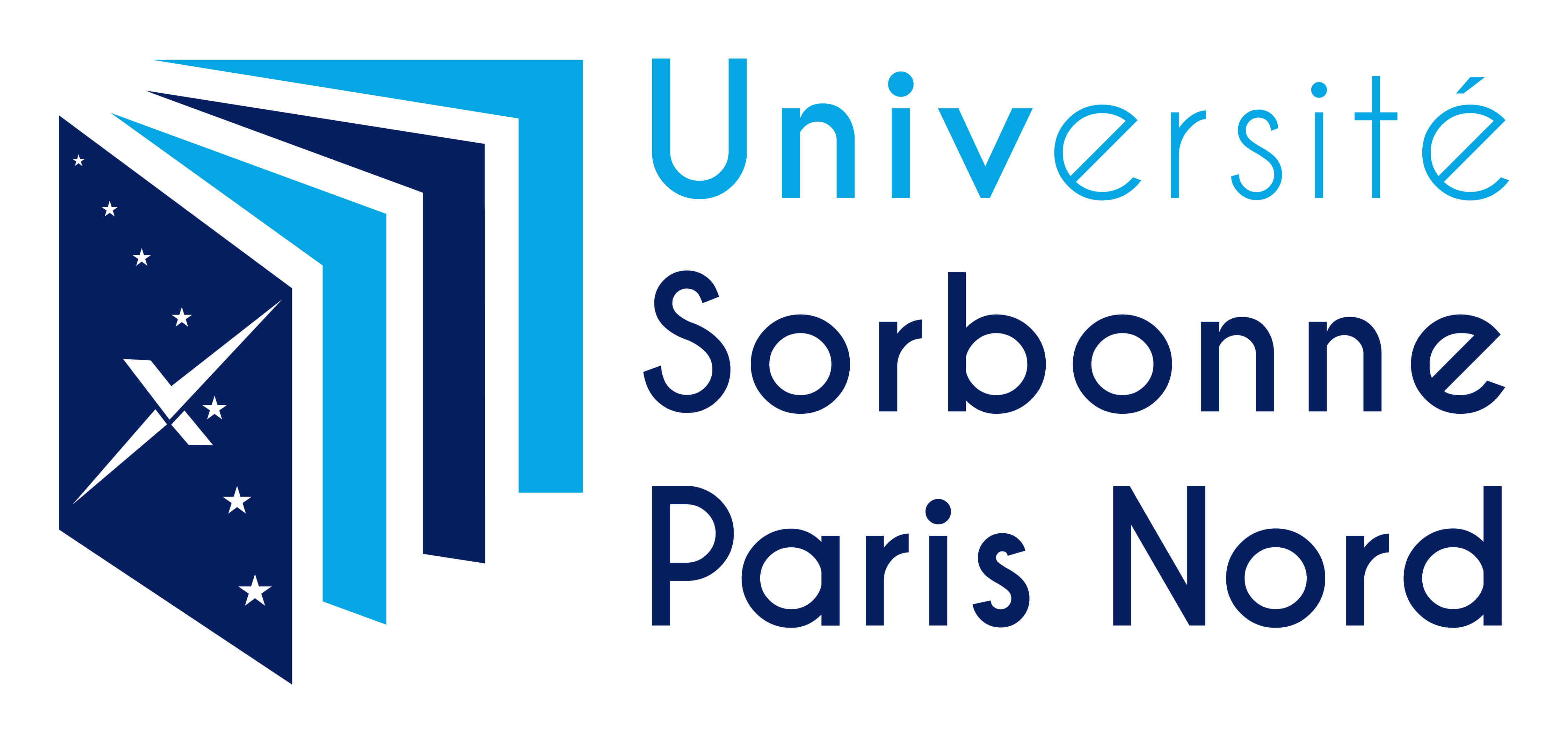}
           \end{minipage}\hfill
           \begin{minipage}{0.16\textwidth}
            
           \end{minipage}\hfill
           \begin{minipage}{0.16\textwidth}
             
           \end{minipage}\hfill
           \begin{minipage}{0.32\textwidth}
             \centering
             \includegraphics[width=\linewidth]{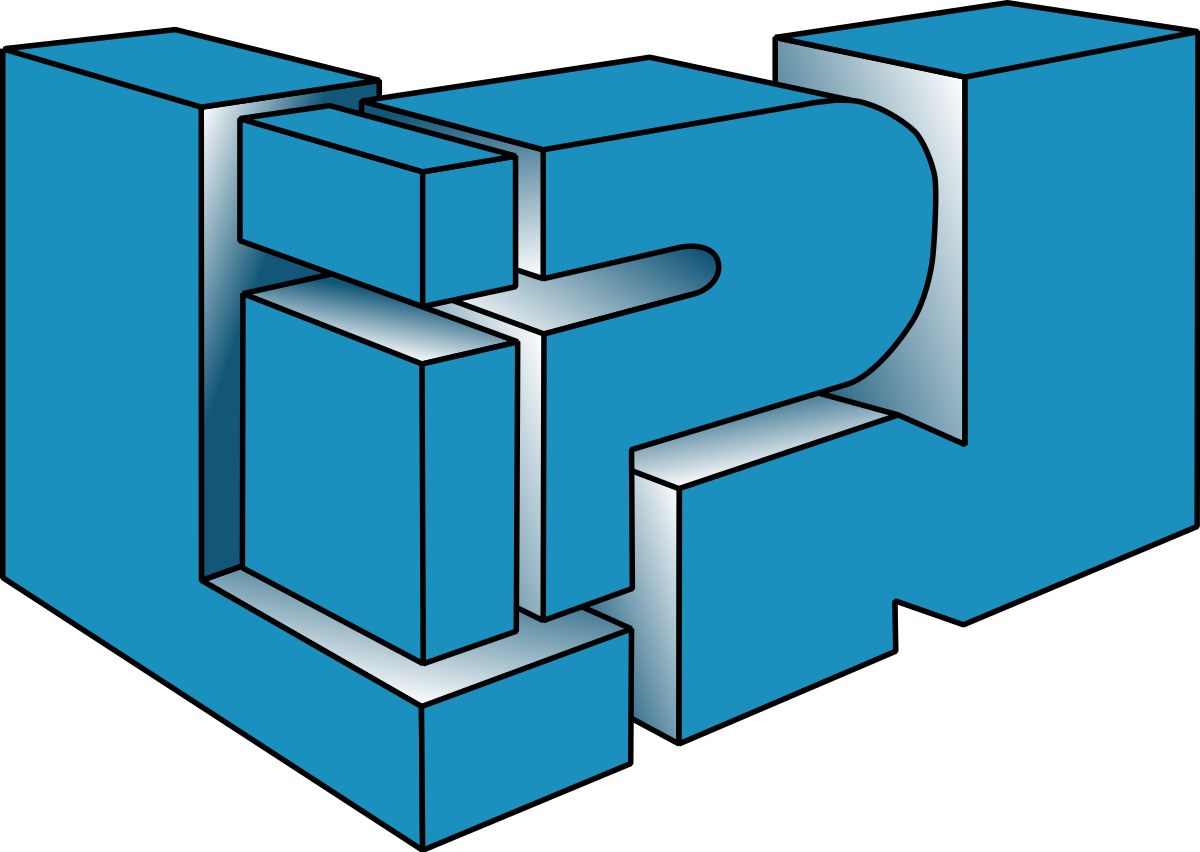}
           \end{minipage}
        \end{figure}
        
        \vspace*{0.3cm}
        \normalsize
        \href{https://www.univ-paris13.fr/}{Université Sorbonne Paris-Nord}\\
        \href{https://lipn.univ-paris13.fr/}{Laboratoire d'Informatique de Paris-Nord}\\
        \vspace*{2.5cm} 
        \LARGE  
        \textBF{Interpretable Sentence Representation with Variational Autoencoders and Attention}\\
        {\normalsize \textBF{Représentations de Phrases Interprétables avec Autoencodeurs \\Variationnels et Attention}}

        \vspace{1cm}
        \normalsize
        Presentée par \href{https://ghazi-f.netlify.app/}{Ghazi Felhi}\\
        
        \vspace{1.5cm}
        \large
        Soutenue, le 26 Janvier 2023, devant le jury composé de :\\
        \begin{table}[h!]
            \centering
            \begin{tabular}{l l l}
               Mme. Adeline Nazarenko  & Sorbonne Paris Cité & Directrice de thèse\\
                M. Joseph Le Roux & Université Sorbonne Paris-Nord & Encadrant \\
                M. Djamé Seddah & Université Paris Sorbonne & Co-encadrant \\
                M. François Yvon & Université Paris-Saclay & Rapporteur \\
                M. Benjamin Piwowarski & Sorbonne Université & Rapporteur \\
                M. Laurent Besacier & Université Grenoble Alpes & Examinateur
            \end{tabular}
        \end{table}

        
    \end{center}
\end{titlepage}










\chapter*{Abstract}
Recent progress in NLP can largely be attributed to progress in Deep Representation Learning techniques, which are based on opaque statistical learning methods. The success of these methods is due to the availability of large volumes of data which they leverage in pre-training schemes to improve generalization when fine-tuned on specific NLP tasks. Producing such high performance representations with an understandable meaning most often calls for the availability of annotated data. However, annotated data comes at great costs, and incurs recurring annotation effort which is impractical when deploying models at scale, i.e. for different languages, different tasks, and on different domains.

The purpose of this thesis is therefore to develop methods that enhance the interpretability of recent representation learning techniques, while accounting for the unavailability of annotated data. We choose to leverage Variational Autoencoders (VAEs), given their established efficiency in learning to relate observations to latent generative factors. VAEs have also been proven effective for semi-supervised learning, and interpretable representation learning, which makes them suitable for the research we tackle in this thesis.  

As a first contribution, we identify two unnecessary components in the functioning scheme of Semi-Supervised VAEs, namely the Kullback-Leibler Divergence and the unobserved latent variable, and perform ablation experiments on them. Our experiments show that these components are indeed of no use to the semi-supervised VAE framework, and that removing them speeds-up computations. We also show that without these components, the model is easier to define and to train.

Our second contribution is based on VAEs, together with Transformers. Recent Deep Learning-based NLP is largely based on variants of the Transformer architecture, which is built using as a main component attention, a learning module which allows exchanging information between a set of elements in a parallel fashion. In previous works, attention has been shown to excel at spontaneously aligning structures from different languages, and to process language in a manner that exhibits patterns resembling understandable NLP concepts such as dependency trees. Based on such observations, we use Transformers attention to build two models with inductive bias to separate information in latent representations into understandable concepts without annotated data. This information separation in neural representations is a process called disentanglement. 

The first model we present is an Attention-Driven VAE (ADVAE). It is the first VAE to use Transformers Cross-Attention to encode and decode vectorial latent variables. We experimentally demonstrate the ability of this model to separately represent, and separately control information about the realizations of core syntactic roles in sentences. 

The second model we present, called QKVAE, is based on ADVAE and uses separate latent variables to form keys and values for the Transformer decoder it uses. The name QKVAE is a contraction of the Query, Key, Value (QKV) abstraction used in Transformers attention and the VAE acronym. We empirically demonstrate the ability of this model to separate syntactic information from semantic information in its neural representations. In experiments involving transfer of syntactic or semantic properties between sentences, QKVAE exhibits competitive performance when compared to previous supervised models, and equivalent performance to a previous supervised model that uses 50K annotated samples. Moreover, this final model displays improved syntactic role disentanglement capabilities compared to ADVAE.

In a context where text data is abundant but annotations are scarce, our work demonstrates that it is feasible to enhance the interpretability of state-of-the-art deep learning architectures for language modeling using only unannotated data.

\paragraph{Keywords:} Variational Autoencoders, Transformers, Interpretability, Disentanglement, Semi-Supervised Learning, Unsupervised Learning, Language Modeling, Syntax.

\chapter*{Français}
\paragraph{Titre :}
Représentations de Phrases Interprétables avec Autoencodeurs Variationnels et Attention
\paragraph{Résumé :}
Les progrès récents en Traitement Automatique de Langues (TAL) peuvent en grande partie être attribués aux progrès des techniques d'apprentissage de représentations profondes, qui reposent sur des méthodes d’apprentissage statistique opaques. Le succès de ces méthodes est dû à la disponibilité de grandes quantités de données utilisées en pré-entraînement afin d’améliorer leur généralisation lorsqu'elles sont adaptées (fine-tunées) à des tâches de TAL spécifiques. Produire de telles représentations qui soient à la fois performantes et compréhensibles nécessite souvent la disponibilité de données annotées. Or, ces données annotées sont très coûteuses, et doivent être collectées séparément pour les différents cas d’usage d’un modèle (i.e. langues, domaines, et tâches), ce qui n’est souvent pas envisageable pour des déploiements à grande échelle.

Le but de cette thèse est  de développer des méthodes qui améliorent l'interprétabilité des techniques récentes d'apprentissage de représentation, tout en prenant en compte l'indisponibilité de données annotées. Nous avons choisi de nous appuyer sur les Auto-Encodeurs Variationnels (Variational Autoencoders ; VAE), compte tenu de leur efficacité établie dans l'apprentissage de la relation entre des observations et des facteurs génératifs latents. Il a également été montré que les VAE sont efficaces pour l'apprentissage semi-supervisé et l'apprentissage de représentations interprétables, ce qui les rend pertinents pour cette thèse.

Dans la première partie des contributions présentées dans cette thèse, nous identifions deux composants superflus dans le schéma de fonctionnement des VAE semi-supervisés, à savoir la divergence de Kullback-Leibler et la variable latente non observée, et nous effectuons des expériences d'ablation sur ces composants. Nos expériences montrent que ces composants n’ont effectivement pas d'utilité dans le cadre VAE semi-supervisé et que leur suppression accélère les modules utilisés. Nous montrons également que sans ces composants, le modèle devient plus facile à définir et à entraîner.

Notre seconde contribution repose sur les VAE, associés aux Transformers. Les techniques récentes de TAL qui reposent sur l'apprentissage profond s’appuient largement sur des variantes de l'architecture Transformer, laquelle est construite en utilisant comme principale brique l'attention, un module d'apprentissage qui permet d'échanger des informations entre un ensemble d'éléments de manière parallèle. Des travaux précédents ont montré que l'attention excelle dans l'alignement spontané des structures de différentes langues et qu’elle présente des motifs ressemblant à des concepts linguistiques interprétables tels que les arbres de dépendance. Sur la base de ces observations, nous utilisons l'attention des Transformers pour construire deux modèles dont le biais inductif permet de séparer l’information dans les représentations latentes en concepts compréhensibles sans données annotées. Cette séparation d'information dans les représentations neuronales est un processus appelé désenchevêtrement (disentanglement).

Le premier modèle que nous présentons est un VAE à attention (Attention-Driven VAE ; ADVAE). C'est le premier VAE à utiliser l'attention croisée des Transformers pour encoder et décoder des variables latentes vectorielles. Nous démontrons expérimentalement la capacité de ce modèle à représenter séparément les informations sur les réalisations de rôles syntaxiques essentiels (core syntactic roles) dans les phrases et à les contrôler.

Le second modèle que nous présentons, appelé QKVAE, dérive de ADVAE et utilise des variables latentes séparées pour formuler des clés et des valeurs pour le décodeur Transformer qu'il utilise. Le nom QKVAE est une contraction du triplet Query (Requête), Key (Clé), Value (Valeur) utilisé dans l'attention des Transformers et de l'acronyme VAE. Nous démontrons empiriquement la capacité de ce modèle à séparer les informations syntaxiques des informations sémantiques dans ses représentations neuronales. Dans des expériences de transfert de propriétés syntaxiques ou sémantiques entre les phrases, QKVAE présente une performance compétitive par rapport aux précédents modèles supervisés et une performance équivalente à un précédent modèle supervisé utilisant 50 000 échantillons annotés. De plus, ce modèle final présente des capacités de désenchevêtrement de rôles syntaxiques améliorées par rapport à ADVAE.

Dans le contexte actuel où les données textuelles sont abondantes et où les données annotées sont difficiles à obtenir, notre travail démontre qu'il est possible d'améliorer l'interprétabilité des architectures de deep learning de pointe pour les modèles de langue à partir de données non annotées.

\paragraph{Mots-clés :} Autoencodeurs Variationnels, Transformers, Interpretabilité, Désenchevêtrement, Apprentissage Semi-Supervisé, Apprentissage Non-Supervisé, Modèle de Langue, Syntaxe.

\chapter*{Acknowledgements}
I would like to express my sincere gratitude to my thesis jury members, Laurent Besacier, Benjamin Piwowarski and François Yvon for their meticulous review of my thesis and insightful comments.

I feel deeply grateful for having had my supervisor Joseph Le Roux as the person with whom I worked most closely. His deep insights into my work, his extensive knowledge, and his unwavering positivity were crucial to both my work \textit{and} well-being.
I also wish to thank my supervisor Djamé Seddah for helping me improve on various regards, for providing invaluable links and references to parts of the linguistics literature I could not navigate, and for bearing with my often unbearable stubbornness. I also have my Ph.D. director Adeline Nazarenko to thank for her scientific, emotional, and even administrative guidance throughout the 3 past years.

I am thankful to my colleagues at LIPN to whom I owe a lot of good memories, and with whom I wish I spent more time than COVID allowed us to. I thank Victor for not getting tired of me asking questions about physics, Alex for teaching me lots of cool ice skating tricks, and both of them plus Francesco for indulging in our 3 hour One Piece-centered debates every Friday Morning. I thank Dasha for teaching me a dance I still can't name and letting me practice my mediocre Russian on her, and Théo for taking interest in my Trello system (I hope you still use it) and for showing me that french people can in fact pronounce \<ق> . I would also like to thank Mathieu for doing the most random things every time I looked his way, and Will for teaching me that Australian spiders were actually cool and that the real danger was snakes.

To my friends who collectively managed to make 2 years of pandemic as pleasant as could be, thank you. Namely, Léo, who first made me want to pursue a Ph.D., Firas who embarked on a similar journey and helped me through the difficult times by simply saying he was going through the same things, Dali, my roommate, who bore with my unwashed dishes and my chaotic sleeping schedule, Sami who's had the life sucked out of him by his startup but still found time to be a good friend, and numerous other people who oftentimes broke the curfew to come and help me get more noise complaints from my neighbors.

Finally, I extend my heartfelt appreciation to my family to whom I owe my birth, this work, and everything in between. I would like to express my gratitude to my two sisters, Hiba and Rym, whom I was proud to introduce to my friends after the presentation, and my parents Souad and Miled who came all the way from Tunisia to see me present my work in Paris despite the horrid travel logistics.

\tableofcontents
\listoftables
\listoffigures

\pagestyle{fancy}
\fancypagestyle{plain}{%
  \fancyhead{}%
  \renewcommand*{\headrule}{}%
  \fancyfoot{}%
  \fancyfoot[LE,RO]{\thepage}%
}

\part{Preamble}
\chapter{Introduction}\label{chap:chap1}

Natural Language Processing (NLP) has come a long way in the recent years due to a paradigm shift towards Deep Learning-based systems. A major catalyst for the advances it has seen, is an intense focus on Representation Learning for language, \textit{i.e.} designing embedding techniques for language (\textit{e.g.} word embeddings or sentence embeddings). These embeddings are the cornerstone to all state-of-the-art NLP systems, since mapping textual observations to numerical vectors is the starting point of all these systems. Consequently, over the past years, better representation learning for language has consistently translated to better performance on downstream tasks such as  parsing~\cite{chen-manning-2014-fast}, translation~\cite{devlin-etal-2014-fast}, or natural language inference~\cite{bowman-etal-2015-large}. Over the last decade, representation learning in NLP evolved from static vectors for context-agnostic word-level representation~\cite{Mikolov2013}, to contextual, and thus dynamic, subword-level representations~\cite{mccann2017learned}. 

An essential ingredient to the rapid growth of the literature on contextual language representation is the Transformer architecture~\cite{Vaswani2017}. This architecture has been built to provide sequence elements (tokens) with context information while solely relying on a mechanism called attention. Contrary to the serial processing occurring in Recurrent Neural Networks (RNNs), attention allows sharing information between sequence elements in a completely parallel fashion, which allows faster learning and inference. The first Transformer-based language representation model, BERT~\cite{devlin-etal-2019-bert}, led to a significant improvement on language understanding benchmarks and paved the way for a surge of similar models~\citep{Yang2019XLNet:Understanding, Lan2020ALBERT, Liu2019RoBERTa:Approach}, later dubbed BERT-like models.

The race to performance caused representation learning models to grow larger, deeper, and thus less and less interpretable. In response to this, great efforts have been deployed to read into the inner workings of these models~\citep{jawahar-etal-2019-bert, Hu2020AModels, Kodner2020OverestimationModels, Marvin2020TargetedModels, Kulmizev2020DoFormalisms, Rogers2020AWorks}, and procedures aimed at grounding language representation into understandable concepts have grown to become a major research direction in the NLP community. Here again, attention, the cornerstone of BERT-like models, was essential to a great number of insights into state-of-the-art NLP models. As a matter of fact, a byproduct of its design is that it calculates values that determine how much context information is pulled from a token to another, which provides a reading into the interaction between these tokens. For instance, \citet{clark-etal-2019-bert} have shown that attention in BERT spontaneously specializes in dependency parsing at different parts of the network, an observation that was the basis for the current state-of-the-art unsupervised dependency parsing system~\cite{shen-etal-2022-unsupervised}. 

As an alternative to \textit{post-hoc} analyses on black-box models, interpretability in NLP models can be tackled by building models which are understandable by design. This built-in interpretability can be pursued at different levels in an NLP system. For instance, there have been efforts to plug knowledge graphs into language models~\cite{logan-etal-2019-baracks, Peters2020KnowledgeRepresentations} so as to freely \textit{control} the knowledge used by these models for generation. Other types of interpretability-oriented interventions on NLP models include \textit{interpretable} inference schemes for multi-hop question answering where, for example,~\citet{Weber2019NLProlog:Language} integrate a Prolog prover into their system and \citet{Saha2019ComplexPrograms} turn questions into series of programmatical queries. In essence, built-in interpretability efforts are channelled into solving two research questions:
\begin{itemize}
    \item How can we obtain \textit{interpretable} inner-representations in neural models ? \textit{(i.e. interpretable encoding)}
    \item How can we control \textit{interpretable} aspects of the output of neural models ? \textit{(i.e. interpretable decoding)}
\end{itemize}

To tackle the above problems and design a system that is interpretable with regard to a certain aspect, the classical procedure is to collect samples annotated with this aspect. Subsequently, one correlates these annotations to some inner-representations in the system, or uses them as a conditioning factor for the outputs of the system. For example, provided text samples annotated with their style, one could train neural representations to separately encode information that correlates with style, modify it, then decode it to obtain a sentence with similar meaning but different style~\cite{wang-etal-2019-harnessing}. However, this procedure assumes access to annotated samples, which are often costly, especially when the annotation procedure requires expertise. For instance, \cite{Seddah2020BuildingHell} report an annotation cost of 87k€ for 1500 sentences to be fully annotated in morpho-syntax and Universal Dependency\citep{nivre-etal-2016-universal} syntax. The cost of labeled data is not only high, but also recurrent since adapting to data that continuously evolves such as language  or to new domains\footnote{For reviews on concept drift and domain adaptation, readers may refer, respectively, to \citet{Lu2019LearningReview} and \citet{Farahani2021AAdaptation}} entails a fine-tuning of the learned models, and therefore a new round of annotation. Given the considerable volumes of data needed for training neural models and the plurality of languages and domains, there is clearly little hope for classical supervised learning to provide coverage with interpretable language technologies to a sufficient proportion of their possible use cases.

To summarize the above, there is a dire need for methods that would improve interpretability for neural NLP models while requiring little-to-no annotated text samples. Accordingly, we turn to Latent Variable Models (LVMs), a framework where it is possible to relate observations to a carefully crafted latent structure of probabilistic variables. This framework provides a principled means to inject beliefs in models through the choice of distributions (\textit{e.g.} Gaussian, Categorical, etc) and latent structures (\textit{e.g.} trees, lattices, independent variables) which can be leveraged to counteract the lack of annotations.

The flagship of LVMs in the Deep Learning era is the Variational Autoencoder (VAE; \citealp{Kingma2014Auto-encodingBayes}). In that sense, the purpose of this thesis is to explore and improve on the current usage of VAEs in NLP, so as to ease their application, and to create neural representations of language with a clear grounding to understandable linguistic factors, while requiring little-to-no supervised learning.


\section{VAEs for Explainable NLP}

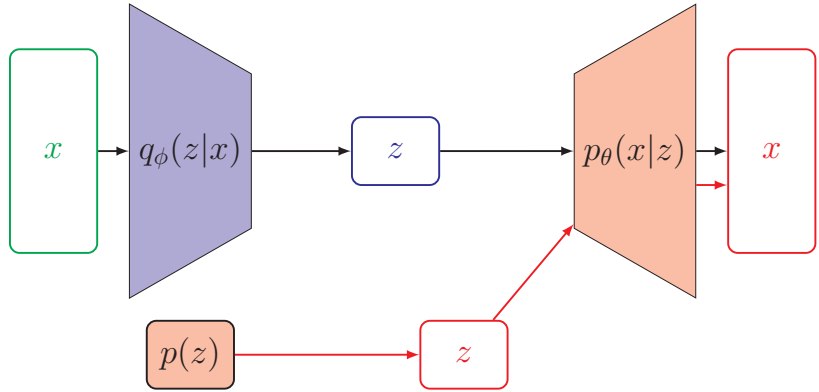
\begin{figure}
    \raggedright
    \hskip 2 cm
    \begin{minipage}[b]{0.33\textwidth}
    \begin{adjustbox}{minipage=\textwidth,scale=0.9}
        \begin{tikzpicture}[node distance = 2cm, auto]
        \tikzstyle{main}=[rounded corners, minimum size = 10mm, thick, draw =black!80, node distance = 2cm, font=\Large, text centered]
        \tikzstyle{connect}=[-latex, thick]

        
        \node[main,text width=1cm,minimum height = 3cm, color=green!100] (xdat) [] {$x$};
        
        \node [trapezium, trapezium left angle=60, trapezium right angle=60,text centered,text width = 2cm,minimum height=1cm, minimum width=2cm, draw=black, fill=blue!30, rotate around={-90:(0,0)}](encoder) [above of=xdat]{\rotatebox{90}{\Large $q_\phi(z|x)$}};
        
        \node[main,text width=1cm,minimum height = 1cm, color=black!100, below of=encoder, fill=red!30, yshift=-1cm] (pz) [] {$p(z)$};

        \node[main,text width=1cm,minimum height = 1cm, color=blue!100, right of= encoder, yshift=0cm, xshift=1cm] (zinf) [] {$z$};
        
        \node[main,text width=1cm,minimum height = 1cm, color=red!100, right of= encoder, yshift=-3.0cm, xshift=2cm] (zgen) [] {$z$};
        
        \node [trapezium, trapezium left angle=60, trapezium right angle=60,text centered,text width = 2cm,minimum height=1cm, minimum width=2cm, draw=black, fill=red!30, rotate around={90:(0,0)}, yshift=-6.5cm, xshift=2cm](decoder) [left of= encoder ]{\rotatebox{-90}{\Large $p_\theta(x|z)$}};
        \node[main,text width=1cm,minimum height = 3cm, color=red!100] (xgen) [right of= decoder] {$x$};
        
      \path   (xdat) edge [connect] (encoder)
              (encoder) edge [connect] (zinf)
              (zinf) edge [connect] (decoder)
              (decoder) edge [connect] (xgen)
              (decoder) edge [connect, color=red!100,transform canvas={yshift=-0.5cm}] (xgen)
              (pz) edge [connect, color=red!100] (zgen)
              (zgen) edge [connect, color=red!100] (decoder);
        \end{tikzpicture}
    \end{adjustbox}
    \vskip -5px
    \end{minipage}
    \caption{A diagram sketching the functioning scheme of a Variational Autoencoder. $x$ is an observation, $q_\phi(z|x)$ is an encoder, $p_\theta(x|z)$ is a decoder, and $p(z)$ is a prior.}
    \label{SSVAEFIGINTRO}
\end{figure}

We give, in this section, a quick overview as to what constitutes a VAE. We also give a broad description of its usages in NLP in general, and how it proved useful for explainable NLP in particular. The brief explanations given here are elaborated upon in a dedicated background chapter (Chapter~\ref{VAEBGCHAP}).

\subsection{What is a VAE ?}

The components forming a VAE are depicted in Figure~\ref{SSVAEFIGINTRO}. Given a set of observations $x$, VAEs are a class of Deep Learning models that train a generative model $p_\theta(x) = \int_z p_\theta(x|z)p(z)dz$, where $p(z)$ is a prior distribution on latent variables $z$ that serve as a seed for generation, and $p_\theta(x|z)$, called the decoder, generates an observation $x$ from each latent variable value $z$.

Since directly maximizing the likelihood $p_\theta(x)$ to train a generative model is rarely tractable\footnote{Maximizing this likelihood is possible for cases where the support of $z$ is small enough for summation. However latent variables with such a small support are, in general, not expressive enough to describe the distribution of target observations $x$.}, an approximate inference distribution $q_\phi(z|x)$, called the encoder, is used to formulate a lower-bound to the exact log-likelihood of the model, called the Evidence Lower-Bound (ELBo). This lower-bound is formulated as follows:
\begin{multline}
    \log p_\theta(w) \geq \mathbb{E}_{(z) \sim q_\phi(z|w)}\left[ \log p_\theta(w|z) \right] -
    \KL[q_\phi(z|w)||p(z)] = \ELBo(w;z) \label{ELBOEQ}
\end{multline} 
\noindent where $\KL$ is the Kullback-Leibler divergence. Although this objective still requires calculating an expectation over $z$, in practice, it can be approximated with sampling-based estimates and used to efficiently learn a generative model.

The VAE framework combines the efficiency of Variational Inference~\cite{jordan1999introduction} with the representational power of Deep Learning. Its generative capabilities, as well as its encoder-decoder architecture made it into a Swiss Army knife for NLP, as is explained in the next section.

\subsection{An Overview of VAEs in NLP}
The first work in NLP exploring the use of VAEs was \citet{bowman-etal-2015-large}'s work on language generation from smooth continuous representations, where these continuous representations are VAE latent variables. The smoothness mentioned by Bowman was a byproduct of the fact that VAE encoders are probability distributions, as opposed to the deterministic scalars yielded by classical Sequence Autoencoders~\cite{Sutskever2014}. As a matter of fact, the regularization provided by the sampling procedures occurring during training produces \textit{smooth} vicinities in the VAE latent space where intervening on representations doesn't damage the well-formedness of output sentences. Moreover, many recent works have shown that VAEs have a natural tendency toward producing disentangled representations with their encoders~\cite{Higgins2019-VAE:Framework, Rolinek2019VariationalAccident}, \textit{i.e.} representations where understandable concepts are localized in identified neurons. This ability to disentangle, together with the smoothness property, allowed various works on interpretability to produce understandable representations with VAE encoders, and to intervene on these representations and decode them to obtain well-formed samples where they successfully modify the factors they explicited in their representations
~\cite{Chen2019ARepresentations, Cheng2020ImprovingGuidance, huang-chang-2021-generating}. Although these works produced encouraging results, they largely assume the availability of annotated data and therefore mostly rely on supervised learning. In contrast to these works, the present thesis focuses on data-efficient methods to minimize the need for annotated data.

After the introduction of VAEs, \citet{Kingma2014a} also showed that they could be used for semi-supervised learning, where a classifier is trained normally on annotated samples, and as a VAE encoder on non-annotated samples. This use of VAEs also propagated to NLP improving data efficiency across several tasks~\cite{Wolf-Sonkin2018AInflection,Habib2019Semi-SupervisedSynthesis, Corro2019DifferentiableAutoencoder, Chen2018VariationalLearning}. Semi-Supervised Learning with a VAE produces an Autoencoder where the latent representation is understandable while requiring relatively few annotated samples. It is therefore an important avenue for research on both interpretable representations learning (with the encoder), and controllable generation (with the decoder).

\section{An Outline of this Thesis}

Posterior to the present introductory part of this thesis, we dedicate a part to background notions which are necessary to understand our contributions, we present our semi-supervised learning-specific contribution in a third part, then our disentanglement-specific contribution in a fourth part. A fifth and final part concludes our thesis with a summary of our findings and a few perspective research directions.




\paragraph{Part II: Background}

This part details all the notions necessary to understanding the contributions presented in this thesis. First, in Chapter~\ref{LMBGCHAP}, we go over neural language modeling, as it is an essential brick to generative NLP systems, and the basis to VAE language models which are what we aim to build in an interpretable way throughout the thesis. The chapter first provides explanations on how to build a standard Language Model (LM), how to build a conditional LM (to lay the groundwork for the introduction of latent variables), the techniques to sample from LMs, and the most common evaluation protocols for these models. Masked Language Models are also be explained as they are an important component of recent representation learning techniques.

Chapter~\ref{VAEBGCHAP} is about VAEs, since they are the machine learning framework used throughout all contributions in this thesis. It covers its theoretical foundations (\textit{i.e.} architecture and loss function), the implementation details necessary for it to work, and how importance weighting is used to approximate its perplexity, the main evaluation metric for language modeling. It also discusses posterior collapse, an issue that is specific to language modeling with VAEs,  and the different solutions that were proposed to deal with it in the literature. Finally, we discuss in this chapter the way VAEs were adapted for semi-supervised learning.

Chapter~\ref{DISENTCHAP} is about a core notion for this thesis : Disentanglement, \textit{i.e.} the process of separating understandable concepts in neural representations. The chapter justifies the need for disentanglement and explains how it works. It also provides technical justifications for the use of VAEs for disentanglement, and gives precise reasons as to why and when unsupervised disentanglement works with VAEs. This chapter also describes the evaluation protocols used to measure disentanglement, and previous works in NLP pertaining to this area of research.

Chapter~\ref{TRANSCHAP} revolves around Transformers, the architecture used to implement the inductive bias used for our disentanglement-specific contributions. We first discuss in this chapter attention, the core component of Transformers, then the way the Transformer architecture is built with different uses for attention and other carefully crafted components. The remainder of this chapter consists in a summary of Transformer-related works throughout NLP, namely (causal and masked) language modeling with Transformers, and Transformer-specific model analyses.

The final background chapter, Chapter~\ref{SYNBGCHAP}, goes over a few notions regarding syntactic analysis that we leverage to explain our disentanglement-specific contributions. Specifically, this chapter summarizes Context-Free Grammars and how they produce constituency trees and presents dependency parsing and the notion of syntactic role. The explanations on dependency analysis also focus on the distinction between oblique and core syntactic roles and how the latter relate to the predicate-argument structure. 

\paragraph{Part III: Semi-Supervised Learning with VAEs}

As a first contribution, we chose to study the use of VAEs in semi-supervised learning (Chapter~\ref{chap:SSVAEchap}). In this context, textual observations are described with 2 latent variables: \textit{i)} a partially-observed latent variable with a known meaning for which we have a small amount of annotations, \textit{ii)} an unobserved latent variable that describes factors other than that of the partially-observed variable, and for which we have no annotated data. Semi-Supervised VAEs (SSVAE) can be used to leverage unlabeled data to improve the inference of a factor for which we only have a small amount of annotations. In this chapter, we provide a detailed description of the source of this improvement, and use it to ablate two sources of over-complexity in the SSVAE framework. These ablations yield a model that is smaller, faster, and easier to define, while preserving the performance of the original framework. 

\paragraph{Part IV: Unsupervised Disentanglement of Sentence Representations}
The bulk of our contributions pertains to disentanglement of sentence representations where we build, with Transformers attention, latent variable models with inductive bias that are capable of giving rise to understandable concepts in the latent representations without annotations \textit{i.e.} in an unsupervised fashion. These contributions are presented in this part over two chapters. 

In chapter~\ref{chap:SynRoleDisentChap}, we tackle unsupervised learning of sentence representations that display separation (disentanglement) with regard to the realizations of their \emph{core} syntactic roles. Drawing from the observation that attention-based Neural Machine Translation systems are capable of aligning spans between languages in a coherent fashion, we introduce an Attention-Driven Variational Autoencoder (ADVAE). ADVAE is the first VAE that uses Cross-Attention to encode and decode latent variables. ADVAE also enables using attention to quantify the interaction of latent variables with inputs. The assessment of this model required designing an evaluation procedure that enables quantifying the disentanglement of realizations of syntactic roles, both in the encoder and in the decoder of our model. Our experimental results show that it is indeed possible with ADVAE to obtain sentence representations where the realizations of syntactic roles exhibit pronounced separation without using supervised learning. We also show that our model is capable of separately changing the realizations of core syntactic roles in sentences it generates,  and that it does it better than classical LSTM-based or Transformer-based text VAEs. However, we show that the success of this separation is limited to the case where a dataset consists of regularly structured sentences. 

In the same line of work, we tackle in Chapter~\ref{chap:QKVCHAP} disentanglement of structure from content in sentence embeddings. By leveraging the internal variables of the cross-attention mechanism in the decoder of our previously built ADVAE, we define a variable that is enforced to only control the \emph{keys} in the decoder's Cross-Attention, and control the \textit{values} using the remaining latent variable. We experimentally demonstrate the ability of the key-specific latent variable to channel syntactic information, while leaving semantic information to the value-specific latent variable. In experiments where the resulting model, QKVAE, is set to transfer syntax and semantics, we show that it performs competitively compared to supervised counterparts, and that a previous supervised model needs more than 50K annotated samples to outperform it. Experiments measuring its ability to disentangle realizations of syntactic roles also show that it improves upon ADVAE, and that it mitigates its inability to disentangle information from different syntactic roles on datasets where sentences do not display regular structure.

\paragraph{Part IV: Conclusion and Perspectives}
This part concludes this thesis by summarizing our findings, and describing a few research directions which may be pursued on the basis of the contributions presented in this thesis. Specifically, it elaborates the way a structured latent variable model, as opposed to the independent latent variable models described in this thesis, can alleviate some of the present shortcomings of our last model, QKVAE. As a last note, we emphasize the ease of applicability of our models to non-text data, since they are unsupervised and modality-agnostic, and thus argue their potential for producing explainable representations for other modalities (\textit{e.g.} images) or in the multi-modal setup.

\section{Publications Related to this Thesis}
\begin{itemize}

    \item Ghazi Felhi, Joseph Le Roux, and Djamé Seddah. 2021. Challenging the Semi-Supervised VAE Framework for Text Classification. In Proceedings of the Second Workshop on Insights from Negative Results in NLP, pages 136–143, Online and Punta Cana, Dominican Republic. Association for Computational Linguistics.~\cite{felhi2021challenging} \textbf{(Chapter~\ref{chap:SSVAEchap})}

    \item Ghazi Felhi, Joseph Le Roux, and Djamé Seddah. 2021. Towards Unsupervised Content Disentanglement in Sentence Representations via Syntactic Roles. Presented at CtrlGen: Controllable Generative Modeling in Language and Vision. Online, co-located with NeurIPS 2021.~\cite{Felhi2021TowardsRoles} \textbf{(Chapter~\ref{chap:SynRoleDisentChap})}

    \item Ghazi Felhi, Joseph Roux, and Djamé Seddah. 2022. Exploiting Inductive Bias in Transformers for Unsupervised Disentanglement of Syntax and Semantics with VAEs. In Proceedings of the 2022 Conference of the North American Chapter of the Association for Computational Linguistics: Human Language Technologies, pages 5763–5776, Seattle, United States. Association for Computational Linguistics.~\cite{felhi-etal-2022-exploiting} \textbf{(Chapter~\ref{chap:QKVCHAP})}
\end{itemize}

\part{Background}

\chapter{Neural Language Modeling}
\label{LMBGCHAP}

Language Modeling is a classical NLP task where one builds generative models for a language. These models can be learned using text corpora and functions with learnable parameters that allow assigning probabilities to text samples quantifying how likely these samples are to belong to these corpora. Tractably assigning probabilities to text samples implies that these models enable \textit{sampling} text samples, \textit{i.e.} \textit{generating} samples which are likely to come from the training data distribution.

This resulting ability to generate text samples is essential to most \textit{generative} (as opposed to \textit{discriminative}) tasks in NLP: namely the design of Dialogue Systems~\cite{zhang-etal-2018-personalizing}, Machine Translation~\cite{Bahdanau2015NeuralTranslate}, Textual Style Transfer~\cite{li-etal-2018-delete}, Abstractive Summarization~\cite{chopra-etal-2016-abstractive}, \textit{inter alia}.  

Besides the utility of Language Models (LMs) as a brick for the above NLP systems, \citet{Radford2018LanguageLearners} have shown through their the large-scale training of an LM, GPT-2\footnote{The GPT-2 experiment is explained in more detail in \S~\ref{TRANSFORMERCLMBG}}, that modeling the distribution of text corpora requires learning a wide set of skills such as Reading Comprehension, Question Answering, and even Machine Translation. Subsequent large-scale LMs such as GPT-3~\citep{Brown2020LanguageLearners} have also shown that LMs can exhibit proficiency at advanced reasoning skills such as few-digit arithmetics and fake news generation. This stresses the centrality of LMs to NLP in particular, and to AI in general.

In this thesis, we aim to build \textit{interpretable} Neural LMs. To lay the groundwork for our contributions, we explain the intuitions behind LMs, and formally define the learning modules and objectives used to obtain them (\S~\ref{LMDEFSEC}). The particular class of LMs that we base our work on are Variational Auto-Encoder-based LMs which condition LMs on latent variables. In that regard, we introduce Conditional LMs (\S~\ref{CLMDEFSEC}) that are later used to plug latent variables into LMs in Chapter~\ref{VAEBGCHAP}. Next, we introduce sampling techniques we use to obtain output text samples from LMs(\S~\ref{SAMPLMDEFSEC}), and metrics that are usually applied to quantify the quality of an LM(\S~\ref{METRLMDEFSEC}). Finally, we explain MLMs(\S~\ref{MLMDEFSEC}), which are the basis to recent Transformer-based representation learning techniques.  

\section{Learning Language Models}
\label{LMDEFSEC}
 Consider a corpus $U$\footnote{The notation $U$ refers to the fact that samples are \textit{unlabeled}, \textit{i.e.} not paired with annotations on some explaining factor for each observation.} of text samples $w$. An LM, learned on the basis of this corpus, consists in a probability distribution $p_\theta$, where $\theta$ is a set of parameters, that should assign high probabilities to samples that belong to $U$, and low probabilities to samples that are unlikely to come from $U$. To represent a text sample $w$, we break it down to a series of tokens $w_{i} \in V$ where $V$ is a predefined vocabulary (\textit{e.g.} words, characters, sub-word units, etc). Using these tokens the probability of a sample is then measured as $p_\theta(w) = \prod_i p_\theta(w_i|w_{<i})$, \textit{i.e.} one token $w_i$ at a time conditioned  on the tokens $w_{<i}$ that precede $w_i$ in $w$. This model is referred to as an \textit{autoregressive} or \textit{causal} LM\footnote{Non-autoregressive language modeling is outside of the scope of this thesis. Despite being less common, it is an active area of research~\cite{Li-2022-DiffusionLM},}, and the default learning strategy to estimate $\theta$ is simply to maximize the probability (or log-probability) of samples in $U$ according to $p_\theta$ as follows\footnote{The objective described in Equation~\ref{MLLMObj} may be accompanied by a regularization term on the parameters $\theta$ which is added to the objective as an auxiliary term $R(\theta)$.}:
\begin{align}
    \argmax_\theta \sum_{w \in U} \log p_\theta(w)=  \argmax_\theta \sum_{w \in U} \sum_{i}^{|w|} \log p_\theta(w_i|w_{<i}) \label{MLLMObj}
\end{align}
The above learning strategy is called Maximum Likelihood Estimation (MLE)\footnote{Alternative learning strategies have been proposed in the literature (\textit{e.g.} Sequential Generative Adversarial Networks; \citealp{Yu2017SeqGAN:Gradient}), but MLE remains dominant for language modeling.}. Given this strategy, all that remains is to define $p_\theta(w_i|w_{<i})$.

LMs can be approached in a naïve way by restricting the dependency of the current word's probability estimation to the $n-1$ previous words, \textit{i.e} $p_\theta(w_i|w_{<i})=$\\ $p_\theta(w_i|w_{i-(n-1)},\dots,w_{i-1})$. This approach, called n-gram language modeling, is limited in that it can only account for a \textit{fixed} number of previous words but it is fairly simple to implement. 

Among other design choices, n-gram LMs can be tackled with a linear model as follows: Let $E$ and $C$ respectively a word embedding matrix and a context embedding matrix that store for each word $w_i$ in the vocabulary $V$ representations $E_{w_i} \in \mathbb{R}^{D_E}$, and $C_{w_i} \in \mathbb{R}^{D_C}$. Using a linear transformation defined by the matrix $M \in \mathcal{M}(n D_C, D_E)$, and the concatenation operator $\Concat$ one can define an n-gram LM as follows:
\begin{align}
    p_\theta(w_i|w_{<i}) = p_\theta(w_i|w_{i-(n-1)},\dots, w_{i-1}) &= {\softmax}_{w_{i}}(s)\\
    \text{s.t. :\hspace{0.5cm}}
    {\softmax}_{w_{i}}(s) &=     \frac{exp(s_{w_{i}})}{\sum_{s_{w_{j}} \in s} exp(s_{w_{j}})}\\
    & s= \{s_{w_{j}}; \text{ s.t. } s_{w_{j}}=E_{w_j}M\hat{C}, w_j \in V)\\
    & \hat{C} = \Concat(C_{w_{i-(n-1)}},\dots, C_{w_{i-1}})
\end{align}
The above procedure can be summarized in the following steps:
\begin{itemize}
    \item Calculate a representation $\hat{C}$ for the context or the word $w_i$.
    \item Score all word in the vocabulary with regard to the current context using a score function $s$.
    \item Calculate a normalized version of the score $s$ that will be used to \textit{parameterize} the categorical distribution $p_\theta(w_i|w_{<i})$.
\end{itemize}
This procedure is the standard procedure used to define an LM even beyond n-gram language modeling. LMs have been improved across various stages of this procedure compared to the above naïve model. For instance, the work of \citet{Yang2017BreakingModel} has shown that the $\softmax$ operator can only model as many contexts as the number of dimensions used for its input matrix, and therefore represents a bottleneck for the scoring of contexts with regard to words. However, research on LMs is most active on the context representation $\hat{C}$. As a matter of fact, a version of the above n-gram LM has been explored by \citet{bengio2000neural} with neural networks to calculate word representations and achieved, at the time, state of the art language modeling performance with $n=5$.

In the early 2010's, LMs started incorporating Recurrent Neural Networks (RNNs) to represent context over indefinitely long series of previous words~\cite{Mikolov2010}. These neural modules enable estimating a fixed size representation $h_N$ for context windows of arbitrary length $N$ using the following recursive rule:
\begin{align}
    a_N &= b + W h_{N-1} + U C_{w_N}\\
    h_N &= \tanh(a_N) 
\end{align}
\noindent where $b$, $W$ and $U$ are learnable parameters, and $\tanh$ is the hyperbolic tangeant function. In a similar spirit, previous works have shown that state-of-the-art language modeling performance can be achieved using variants of RNNs, for instance LSTM~\cite{Merity2018RegularizingModels} or Mogrifier LSTM~\cite{Melis2019MogrifierLSTM}), and neural models thus became the \textit{de facto} approach to parameterize LMs.


\section{Conditional Language Models}
\label{CLMDEFSEC}
In contrast to vanilla LMs, controllable or conditional LMs require an input constraint on $p_\theta$ that steers the probability estimation towards a desired property. A simple example can be that of an LM trained on movie reviews, where we try to \textit{control} the sentiment (positive or negative) of the generated reviews. Another example, that is rarely referred to as a conditional LM, is that of Machine Translation where we aim to generate text samples in a certain language conditioned on their counterpart in another language.

To model controllable language generation, one simply \textit{conditions} the output probability distribution on an additional factor $y$ such that word probabilities at each generation step become parameterized by $p_\theta(w_i|w_{<i}, y)$. Together with the previous words $w_{<i}$, $y$ is used to estimate a context representation $\hat{C}$ that will be trained to assign high probabilities to words that realize text samples with the desired underlying attribute. Given a labeled dataset $L$ consisiting in text samples $w$ coupled with annotations $y$ on the attribute that one aims to account for, learning a conditional LM is straightforward using a slightly modified version of the Maximum Likelihood objective introduced in~\ref{MLLMObj}:

\begin{align}
    \argmax_\theta  \sum_{w \in U} \log p_\theta(w|y)= \argmax_\theta  \sum_{(w, y) \in L} \sum_{i}^{|w|} \log p_\theta(w_i|w_{<i}, y) \label{CLMObj}
\end{align}

However, when the underlying factor we aim to model is only partially-observed in the data at hand, or when it is never observed\footnote{Later in this thesis, contrary to observed or partially-observed latent variables that we denote $y$, we denote unobserved latent variables $z$.}, the above objective must be swapped for another that accounts for the data points that only consist of an unlabeled observation\footnote{For example, one could marginalize the objective described in~\ref{CLMObj} over $y$, if this marginalization is tractable (which is rarely the case).} $w$. An objective that enables working with this constraint is described later in this thesis in the context of semi-supervised learning with VAEs (\S~\ref{SSVAEBG}).

\section{Sampling Techniques for Autoregressive Language Models}
\label{SAMPLMDEFSEC}
The canonical sampling procedure to generate text from LMs, random sampling, is to pick words according to their probabilities $p_\theta(w_i|w_{<i})$ until a maximum length is reached, or an end-of-sequence flag is triggered. LMs are typically trained on text samples bracketed by a beginning of sentence token $<\bos>$ that plays the role of $w_{<i}$ for the first generation step, and an end of sentence token $<\eos>$ that signals the end of the text sample, and thus the token sampling loop.

Random sampling makes sense when the model needs to generate diverse samples. An instance where this can be useful is data augmentation where the model needs to augment a dataset with diverse samples in order to improve the generalization of methods learned on this dataset~\cite{Han2019NeuralDetection}. In contrast to such case, many applications such as Machine Translation, require picking \textit{the most likely} sequence of tokens rather than \textit{a likely} sequence of tokens. In this case, generation is most often referred to as \textit{decoding}\footnote{This most frequently used term, \textit{decoding}, is short for \textit{Maximum A Posteriori (MAP)}  decoding, where one tries to produce the sequence with highest overall probability. Other decoding strategies, such as \textit{minimum Bayes risk decoding}~\citep{kumar-byrne-2004-minimum} may be encountered in the literature~\citep{eikema-aziz-2020-map} but the MAP decoding is by far the dominant paradigm.} since it is required to roll out an expected response within a narrow range of possibilities based on precise contextual clues.

\begin{figure}[!t]
\centering
    \begin{minipage}[b]{1.0\textwidth}
            \centering
            \begin{minipage}[b]{\textwidth}
            \begin{adjustbox}{minipage=\textwidth,scale=0.3}
                \includegraphics{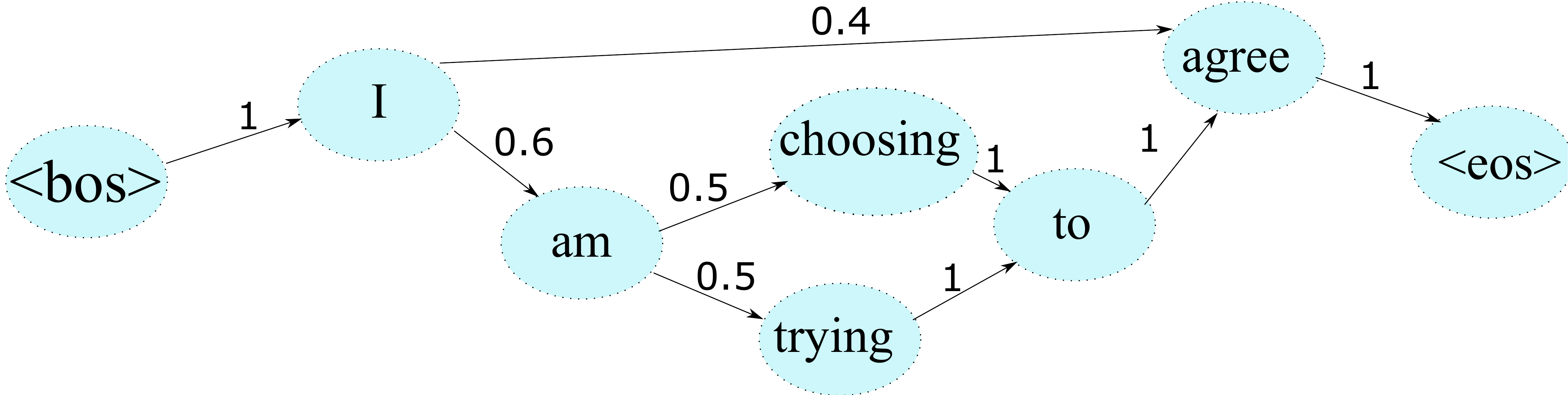}
            \end{adjustbox}
            \end{minipage}
            \caption{\centering Example transition probabilities for a decoding process.$<\bos>$ and $<\eos>$ are respectively beginning-of-sentence and end-of sentence tokens. Transition probabilities are transcribed over arrows linking different tokens. The absence of an arrow means a null transition probability.}
            \label{fig:GSvsBM}
    \end{minipage}
\end{figure}

Finding the most likely sequence of tokens requires estimating the probabilities of all possible sequences, which is intractable. Therefore, decoding is usually tackled using heuristics, where the 2 most commonly used heuristics are Greedy Sampling and Beam Search. Greedy sampling is fairly straightforward: one simply picks, at each generation step, the word with the highest probability. However, picking the most likely word at each generation step does not guarantee sampling the highest-probability sequence. An example of how this can occur is illustrated in Figure~\ref{fig:GSvsBM}. When only the token \textit{I} has been picked, $p_\theta(am|I)>p_\theta(agree|I)$. However, the probability of the sentence that follows from choosing \textit{agree} is:
\begin{flalign}
    &p_\theta(I, agree, &&\hskip -1.9cm<eos>|<bos>) = && &&\nonumber\\
    &p_\theta(I|<bos>) &&p_\theta(agree| <bos>, I) &&p_\theta(<eos>|<bos>, I,&&\hskip -0.9cm agree)\nonumber\\
    = &1* &&0.4* &&1 &&=0.4
\end{flalign}

On the other hand, the probability of a sentence that follows the choice \textit{am}\footnote{Branching on \textit{choosing} or \textit{trying} leads to the same sentence probability here.} is:
\begin{align}
    p_\theta(I, am, choosing&, to, agree, <eos>|<bos>)=\nonumber\\
    &p_\theta(I|<bos>)p_\theta(am|<bos>, I)\nonumber\\
    &p_\theta(choosing|<bos>, I, am)p_\theta(to|<bos>, I, am, choosing)\nonumber\\
    &p_\theta(agree|<bos>, I, am, choosing, to)\nonumber\\
    &p_\theta(<eos>|<bos>, I, am, choosing, to, agree)\nonumber\\
    &=1*0.6*0.5*1*1*1=0.3
\end{align}

Therefore, although picking the best option at each generation step favors the token \textit{am} at the second step, the overall highest probability sentence is rather obtained by picking \textit{agree}.

To better explore word transition graphs such as the one represented in Figure~\ref{fig:GSvsBM}, we use Beam Search. This method is initialized by picking the $k$ highest-probability tokens to initialize $k$ sequences, where $k$ is called the beam width. In what follows, Beam Search measures for the $k$ sequences the probability for each token in the vocabulary $V$ to be the next token. The combination of each sequence with each word from the vocabulary yields an array of sequences of size $k*|V|$ from which $k$ highest-probability \textit{sequences} will be selected. The process is repeated until all $k$ sequences have reached the $<\eos>$ flag or a predetermined maximum length. Here we emphasize that, contrary to Greedy Decoding, Beam Search keeps or drops sequences on the basis of the entire sequence probability (\textit{i.e.} $p_\theta(w_i, w_{i-1}, \dots, w_1)$) instead of the next token probability (\textit{i.e.} $p_\theta(w_i |w_{i-1}, \dots, w_1)$). While guaranteeing equal or higher-probability decoded sentences, Beam Search requires using $k$ times the memory required for Greedy Sampling where typical values for $k$ range from 5 to 20.

Next to Beam Search and Greedy decoding, other decoding schemes have been developed in order to better imitate human language. Specifically, \citet{Holtzman2020The} show that aiming for high probability sentences through Beam Search and Greedy Decoding, even from high quality LMs, leads to sentences that have low diversity compared to human-generated text. On the other hand, unhinged random sampling results in erratic and often unnatural sequences. A middle-ground can be reached using methods like Top-k Sampling, or \citet{Holtzman2020The}'s Nucleus Sampling which truncate the range of possible tokens for random sampling to a restricted high-probability list. Although the output of these methods may exhibit more human-like patterns, we restrict our study to outputs from Greedy Sampling or Beam Search, which are designed with the intent\footnote{We stress here that neither Greedy Sampling, nor Beam Search are guaranteed to produce the most likely sequence of tokens.} to produce maximally likely sequences, and are therefore better suited for inquiries about what the model has learned. 

\section{Evaluation the Generative Capabilities of Language Models}
\label{METRLMDEFSEC}
Similar to sampling techniques, evaluation strategies for generative models depend on whether the model is required to describe a range of outputs with its samples, or to produce a specific output given some contextual clues like previous words used as a prompt, or extrinsic factors encoded in the context representation.

In the case where no specific output is required from the LM, we aim to assess the well-formedness\footnote{Here, \textit{well-formedeness} is relative to the data that we aim to learn. In that sense, a non-normalized text sample (\textit{e.g.} \textit{c'mon} instead of \textit{come on}), is considered well-formed if it adheres to the distribution of the LM's training data, which could allow for non-normalized text.} of its outputs. Ideally, an oracle would look at samples from the LM, and score their syntactic and semantic soundness. But since such an oracle is rarely available, the norm is to evaluate the model with a score that depends on the likelihood that this model assigns to a held-out collection of unseen ground truth text data. The idea here is that, since we know that our held-out data was not seen by the model during training, and that it is well-formed, the higher the probability assigned by our model to this data, the higher the chances that it will, in general, generate well-formed samples. This principle is formalized through a metric called perplexity. Given a test set $U_{test}$ consisting of samples $w=\{w_1, \dots, w_{|w|}\}$, perplexity is calculated as follows:
\begin{align}
    PP(U_{test}) &= p_\theta(U_{test})^{-\frac{1}{N}}\\
    \text{s.t:\hspace{0.5 cm}} p_\theta(U_{test}) &= \prod_{w \in U_{test}} \prod_{i}^{|w|} p_\theta(w_i|w_{<i})\\
                  N &= \sum_{w \in U_{test}} |w|
\end{align}
In the above formula, $N$ is the total number of words in our corpus, and it is used to normalize the inverse-probability score we measure. As shown in this formula, perplexity is a decreasing function of the token-wise probability $p_\theta(w_i|w_{<i})$. Therefore, lower perplexity means better language modeling performance. It is also important to note that perplexity \textit{does not} account for the sampling strategy used at test time.

Now, in case we need to measure how close a model is to generating a specific output, a range of metrics have been developed in order to compare series of tokens in a linguistically meaningful way. The metrics we present here are metrics that were mainly aimed at measuring performance in the task of Machine Translation, and therefore designed to correlate with human judgment on sentence or phrase semantic similarity. To get the gist of how these metrics work, we describe, in what follows, the most commonly used metric: BiLingual Evaluation Understudy (BLEU; \citealp{Papineni2001BLEU:Translation}).

Given a generated (or candidate) sentence $w$ and a set of reference translations \\$\{w_{ref}^1, \dots, w_{ref}^R\}$ (\textit{e.g.} alternative target translations for a single source sentence), this metric computes a \textit{modified $n$-gram precision} of the $n$-grams present in the candidate, where $n$-grams are $n$-tuples of tokens. A standard $n$-gram precision measure consists in counting the number of $n$-grams in the candidate sentence that are present in a reference sentence, and then dividing the result by the total number of n-grams in this candidate. To see the problem with such a measure, observe the following example:
\begin{itemize}
    \item \textbf{Candidate:} food food food food food.
    \item \textbf{Reference 1:} the dog ate its food.
    \item \textbf{Reference 2:} the canine consumed its nourishment.
\end{itemize}
Using standard precision, the above sentence that consists only in repeats of an $n$-gram that is present in a reference sentence (the unigram \textit{food}) gets a perfect score. To deal with the above issue, BLEU modifies precision by clipping the total occurrence count for an $n$-gram in references by the maximum number of times it occurs in a reference. This measure is formalized as follows:
\begin{align}
    p_n =& \frac{\sum\limits_{\mathcal{C}\in Candidates}\sum\limits_{\text{\textit{n-gram}}\in C}Count_{clip}(\text{\textit{n-gram}})}{
    \sum\limits_{\mathcal{C'}\in Candidates}\sum\limits_{\text{\textit{n-gram}}'\in C'} Count(\text{\textit{n-gram}}')}\\
    \text{s.t.:}& \hskip 0.5cm  Count_{clip}(\text{\textit{n-gram}}) = \min(Count, Max\_Ref\_Count)
\end{align}

In practice, BLEU is measured through a geometric mean of the above measures for different values of $n$, typically for $n\leq 4$.

In the same line of work, ROUGE~\citep{lin2004rouge} was engineered to maximally correlate with human judgment of summarization quality, and improvements over BLEU were brought about with METEOR~\cite{Banerjee2005METEOR:Judgments} and METEOR 1.5~\citep{denkowski-lavie-2014-meteor} which match words exactly, through stemming and through synonymy, and includes an explicit account for grammaticality through a chunking penalty. 

More recently, efforts have been deployed to use neural representations as a continuous estimate of meaning, and to correlate similarity metrics between these representations (\textit{e.g.} cosine similarities) with human judgments. A number of works have shown that state of the art correlations with human judgements could be obtained either with unsupervised representation learning techniques~\cite{Zhang2020BERTScore:, kamal-eddine-etal-2022-frugalscore} or with supervised versions of these techniques where the representation is trained to be invariant to paraphrasing~\cite{huang-etal-2021-disentangling}. 

\section{Masked Language Modeling}
\label{MLMDEFSEC}
As discussed earlier in this chapter, the term \textit{Language Model} generally refers to models that assign probabilities to text samples. An exception to this rule is a recent class of models trained to perform a task dubbed \textit{masked language modeling}. This task consists in recovering a sequence of tokens from a corrupted version of this sequence. More precisely, to train a Masked Language Model (MLM), one takes a sequence of tokens, replaces a proportion of the tokens with a special $[MASK]$ token, replaces another proportion of tokens with a random token, and trains a neural module $p_\theta$ to recover the original sequence using the following objective:

\begin{align}
    \argmax_\theta \sum_{w \in L} \sum_{i}^{|w|} \log p_\theta(w_i|w_{corr}) \label{MLMObj}\\
    \text{s.t. :\hskip 0.5cm} w_{corr} = corrupt(w) 
\end{align}
\noindent where the $corrupt$ operator applies the random replacements described above. Contrary to classical LMs, the above $p_\theta$ cannot be used to measure the probability of a sentence. In that sens, although it can be used to fill-in missing words in a sequence, it can not be used to generate sequences of tokens. In fact, it is important to notice that sampling a token from  $p_\theta(w_4|w_{corr,1}=t_1, w_{corr,2}=t_2, w_{corr,3}=t_3, w_{corr,4}=[MASK])$ does not yield the \textit{next} token for the sequence $\{t_1, t_2, t_3\}$ but rather the \textit{last} token since $p_\theta$ takes \textit{bidirectional} information as a context. Therefore, using only previous tokens as a context for an MLM implicitly gives it the information that there are no other words in the sentence.  

The concept of Masked Language Models (MLM) has been formulated and applied first in the work of \citet{devlin-etal-2019-bert} to train their Bidirectional Encoder Representations from Transformers (BERT). The objective was to obtain neural representations that feed off the entirety of the context in sentences through a self-supervised learning scheme, \textit{i.e.} a learning scheme where the target information for inference is the input data itself. The intent behind this objective was to \textit{transfer} the information channeled in these representations to other tasks. Transferring representations was popularized in NLP by \citet{Mikolov2013}, and contextual representations were explored before MLMs (see for example \citealp{Peters2018}), but the work of \citet{devlin-etal-2019-bert} has shown that the ability of MLMs to factor context from the entire sentence enables contextual representations to display significantly higher transfer performance when applied to new tasks. The success of \citet{devlin-etal-2019-bert}'s model, BERT, was due to masked language modeling but also to the Transformer architecture. This architecture, which was highly popularized by BERT, is better explicited in Chapter~\ref{TRANSCHAP}, together with its implementation details, and some of the most prolific BERT variants that were also based on it. 

\section{Conclusion}
Since the purpose of our work is to build interpretable neural LMs, we dedicated this chapter to introducing neural language modeling. First, we explained how Auto-regressive LMs can be formulated, and how they can be parameterized by learnable modules $p_\theta$ (\S~\ref{LMDEFSEC}). We then explained how, in addition to previous tokens, the probability assigned by $p_\theta$ to a token can be conditioned on other factors to explicitly model dependency on extrinsic attributes(\S~\ref{CLMDEFSEC}).

Using LMs for generation requires sampling techniques, among which random sampling, Greedy Sampling and Beam Search are described in this Chapter(\S~\ref{SAMPLMDEFSEC}). Subsequently, we gave intuitive then formal descriptions of a few evaluation metrics, namely perplexity and BLEU, which are respectively used to measure the quality of unconstrained randomly generated samples, and the similarity of conditionally-generated samples with regard to a specific desired output(\S~\ref{METRLMDEFSEC}). 

As a final point, we presented masked language modeling(\S~\ref{MLMDEFSEC}), a training objective that deviates from classical language modeling in order to produce representations from a bidirectional textual context, and that provides the basis for modern high-performance NLP representation learning. 

\chapter{Variational Autoencoders}
\label{VAEBGCHAP}
Variational Autoencoders~\cite{Kingma2014Auto-encodingBayes} are models which leverage Deep Learning components by using the Autoencoder architecture~\cite{rumelhart1985learning} to learn a generative model using Variational Inference~\cite{jordan1999introduction}. Contrary to what its name implies, its primary purpose is, therefore, not to learn encodings but rather to learn to generate observations that are likely to come from a given dataset $U=\{x_1, \dots, x_{|U|}\}$. For a generative model $p(x) = \int_z p(x|z)p(z)dz$, where $x$ is an observation and $z$ is a latent variable with a known prior distribution p(z), Variational Inference can be used to approximate the posterior distribution $p(z|x)$, which is often intractable, by optimizing a different distribution $q(z|x)$. Using Neural Networks, or more precisely Autoencoders, this Variational Inference-based approach can be implemented using a neural encoder $q_\phi(z|x)$ and a neural decoder $p_\theta(x|z)$ to form what later came to be called Varational Autoencoders (VAEs). The presence of both an encoder and a decoder in VAEs led to a plethora of works that employ them outside of the strict generative modeling setup, namely for semi-supervised learning~\cite{Wolf-Sonkin2018AInflection,Habib2019Semi-SupervisedSynthesis, Corro2019DifferentiableAutoencoder, Chen2018VariationalLearning}, or interpretable representation learning and controllable generation~\cite{Chen2019ARepresentations, Cheng2020ImprovingGuidance, Xu2020OnSupervision, John2020DisentangledTransfer, Bao2020, huang-chang-2021-generating, huang-etal-2021-disentangling}. 

This chapter is dedicated to presenting VAEs, the machine learning framework used throughout all of our contributions. We first describe the architecture it uses as a backbone, the Autoencoder architecture (\S~\ref{AEARCHIBG}). Then we explain how the encoder and decoder, together with a prior distribution on the latent variables, are used to formulate a lower-bound on the exact log-likelihood of a generative model (\S~\ref{VAETHBGSEC}). This generative model requires a few implementation tricks to be learned efficiently which we explain in a dedicated section (\S~\ref{VAEIMPLEMBGSEC}). Language modeling with VAEs is discussed in Section~\ref{VAELMSec}, followed by the method used to evaluated perplexity for such models (\S~\ref{VAEIWBGSEC}). VAEs, when applied to language modeling, exhibit a particular type of failures called \textit{posterior collapse} which we describe, together with common techniques to deal with it, in Section~\ref{PColSec}. Finally, we lay out the principles governing semi-supervised learning with VAEs (\S~\ref{SSVAEBG}). The reasons why VAEs are capable of producing disentangled representations as well as the way they are used for that purpose are later detailed in Chapter~\ref{DISENTCHAP}. 

\section{The Autoencoder Architecture}
\label{AEARCHIBG}
\begin{figure}
    \raggedright
    \hskip 2 cm
    \begin{minipage}[b]{0.33\textwidth}
    \begin{adjustbox}{minipage=\textwidth,scale=0.9}
        \begin{tikzpicture}[node distance = 2cm, auto]
        \tikzstyle{main}=[rounded corners, minimum size = 10mm, thick, draw =black!80, node distance = 2cm, font=\Large, text centered]
        \tikzstyle{connect}=[-latex, thick]

        
        \node[main,text width=1cm,minimum height = 3cm, color=green!100] (xdat) [] {$x$};
        
        \node [trapezium, trapezium left angle=60, trapezium right angle=60,text centered,text width = 2cm,minimum height=1cm, minimum width=2cm, draw=black, fill=red!30,
        rotate around={-90:(0,0)}](encoder) [above of=xdat]{\rotatebox{90}{\Large $\hat{q}_\phi(x)$}};

        \node[main,text width=1cm,minimum height = 1cm, color=red!100,
        right of= encoder, yshift=0cm, xshift=1cm] (zinf) [] {$z$};
        
        \node [trapezium, trapezium left angle=60, trapezium right angle=60,text centered,text width = 2cm,minimum height=1cm, minimum width=2cm, draw=black, fill=red!30, rotate around={90:(0,0)}, yshift=-6.5cm, xshift=2cm](decoder) [left of= encoder ]{\rotatebox{-90}{\Large $p_\theta(x|z)$}};
        \node[main,text width=1cm,minimum height = 3cm, color=red!100] (xgen) [right of= decoder] {$x$};
        
      \path   (xdat) edge [connect] (encoder)
              (encoder) edge [connect] (zinf)
              (zinf) edge [connect] (decoder)
              (decoder) edge [connect] (xgen);
        \end{tikzpicture}
    \end{adjustbox}
    \vskip -5px
    \end{minipage}
    \caption{A diagram sketching the Autoencoder architecture. $x$ is an observation, $\hat{q}_\phi(x)$ is an encoder and $p_\theta(x|z)$ is a decoder. Green denotes the ground truth variable, while red denotes inferred variables.}
    \label{fig:AE1}
\end{figure}
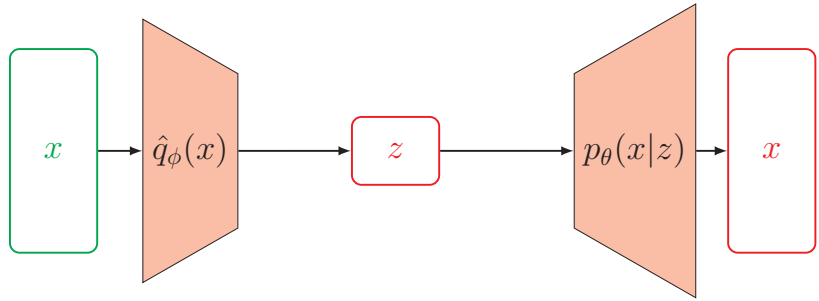

Autoencoders~\cite{rumelhart1985learning} are a class of Deep Learning models which aim to learn lossy compressions or \textit{encodings} of high dimensional data through low dimensional representations. These models usually consist of an encoder function $\hat{q}_\phi$ which produces representations $z$ from a samples $x$, and a decoder distribution $p_\theta$ which aims to reconstruct the input sample $x$ from its representation $z$. This architecture is sketched in Figure~\ref{fig:AE1}.

Given observations $x$ coming from a ground truth distribution $p_{data}$, the parameters $\phi$ and $\theta$ of Autoencoders can be learned by maximizing the following reconstruction objective:

\begin{align}
    \argmax_{\theta, \phi}  \mathbb{E}_{x \sim p_{data}(x)} \left[ \log p_\theta(x|z=\hat{q}_\phi(x)) \right] \label{AEObj}
\end{align}
Notice that for \textit{textual} observations, the above objective is exactly that of a conditional LM (\textit{cf.} Eq.~\ref{CLMObj} in Section~\ref{CLMDEFSEC}), where the conditional LM $p_\theta$ is jointly learned with $q_\phi$, the module that provides the conditioning variables $z$. For the sake of the forthcoming explanations, we also point out the fact that the deterministic encoding function $\hat{q}_\phi(x)$ can be written as a conditional Dirac probability density function $q_\phi(z|x)$ such that:

\[q_\phi(z|x)= 
  \left \{
    \begin{array}{l l}
        +\infty &\text{ if }z=\hat{q}_\phi(x)\\
            0 &\text{ if } z\neq \hat{q}_\phi(x)
    \end{array}
    \right.
\]
Which enables rewriting the reconstruction loss as:
\begin{align}
    \argmax_{\theta, \phi}  \mathbb{E}_{x \sim p_{data}(x)} \mathbb{E}_{z \sim q_\phi(z|x)}[ \log p_\theta(x|z)] \label{AEObj2}
\end{align}

\section{Theoretical Foundation of Variational Autoencoders}
\label{VAETHBGSEC}
A generative model that generates observations $x$ from a latent variable $z$ with a known \textit{prior} distribution, is modeled using this prior $p(z)$ and a \textit{decoder} $p_\theta(x|z)$. Directly learning it through Likelihood Maximization is inefficient since the summation over all values of $z$ to calculate $p_\theta(x)=\int_z p_\theta(x|z) p(z) dz$ is intractable in general\footnote{The integral can be made tractable if the support of $p(z)$ is small enough for summation, but such a choice of prior makes for an extremely weak generative model.}. For an observation x, Variational Inference uses a distribution $q(z)$\footnote{We stress that each observation $x$ is assigned with a dedicated $q(z)$. To lighten notations and to adhere to conventions adopted by other resource on this matter, $q$ is not indexed with $x$.} to formulate a lower-bound on the exact log-likelihood of the model $\log p_\theta(x)$. This lower-bound finds its origin in the following derivations:
\begin{align}
    \text{Given that } p_\theta(x) = \frac{p_\theta(x, z)}{p_\theta(z|x)}&:\nonumber\\
    \log p_\theta(x) &= \mathbb{E}_{z\sim q(z)}\log \frac{p_\theta(x, z)}{p_\theta(z|x)} 
\end{align}
By splitting $p_\theta(x, z)$ to $p_\theta(x|z)p(z)$ and simply multiplying and dividing by $q(z)$, we obtain:
\begin{align}
\mathbb{E}_{z\sim q(z)}\log \frac{p_\theta(x, z)}{p_\theta(z|x)}&= \mathbb{E}_{z\sim q(z)}\left[\log \left( p_\theta(x|z)\frac{p(z)}{q(z)}
    \frac{q(z)}{p_\theta(z|x)}\right)\right] 
\end{align}
Then, using the Kullback-Leibler Divergence operator defined by $\KL[q(z)||p(z)]=$\\ $\mathbb{E}_{z\sim q(z)}\left[\log \frac{q(z)}{p(z)}\right]$, we obtain:
\begin{align}
\mathbb{E}_{z\sim q(z)}\left[\log \left( p_\theta(x|z)\frac{p(z)}{q(z)}
    \frac{q(z)}{p_\theta(z|x)}\right) \right]
    &= \mathbb{E}_{z\sim q(z)}\left[\log p_\theta(x|z)\right] - \KL[q(z)||p(z)]\nonumber\\
    &+\KL[q(z)||p_\theta(z|x)]
\end{align}
The above derivations lead to the equality:
\begin{align}
    \log p_\theta(x) - KL[q(z)||p_\theta(z|x)] &= \mathbb{E}_{z\sim q(z)}\left[\log p_\theta(x|z)\right] - KL[q(z)||p(z)]\label{ELBoEqua}
\end{align}
Since KL divergences are positive, the above equality allows writing the inequality that enables learning with VAEs\footnote{$z$, the latent variable, is marginalized throughout $\ELBo$ with expectations. Therefore, the value of $\ELBo$ does not depend on $z$. Nevertheless, we abuse notations and specify it as an argument in $\ELBo(x; z)$ to clarify the role of each variable in different instances of $\ELBo$ throughout the thesis. }:
\begin{align}
    \log p_\theta(x) \geq \mathbb{E}_{z\sim q(z)}\left[\log p_\theta(x|z)\right] - \KL[q(z)||p(z)]= \ELBo(x;z) \label{ELBOBGEQ}
\end{align}
The right-hand side of the above inequality, called the Evidence Lower-Bound (ELBo), is a tractable lower-bound to the exact log-likelihood of our model. According to Equation~\ref{ELBoEqua}, maximizing this lower-bound either maximizes the log-likelihood of our model, or brings $q(z)$ closer to the true posterior $p_\theta(z|x)$. Since $q(z)$ is intended to mimic the true posterior, it is referred to as the \textit{approximate posterior}.

Classical Variational Inference assumes the estimation of an approximate posterior $q(z)$ for each observation $x$. However, one can use a single model $q_\phi(z)=\int_x q_\phi(z|x)p_{data}(x) dx$ to learn approximate posteriors for all observations $x$. This is called Amortized Variational Inference (AVI; \citealp{pmlr-v80-kim18e}).

Variational Autoencoders' learning is simply an application of the AVI learning scheme where $q_\phi(z|x)$, the encoder, and $p_\theta(x|z)$, the decoder, are neural networks which are learned through some form of Stochastic Gradient Descent (SGD).

After training, VAEs are used to generate samples with a two-step sampling process: First, a latent variable must be sampled from the prior distribution $p(z)$, then this latent variable is decoded using the decoder $p_\theta(x|z)$ to obtain a generated sample $x$. 

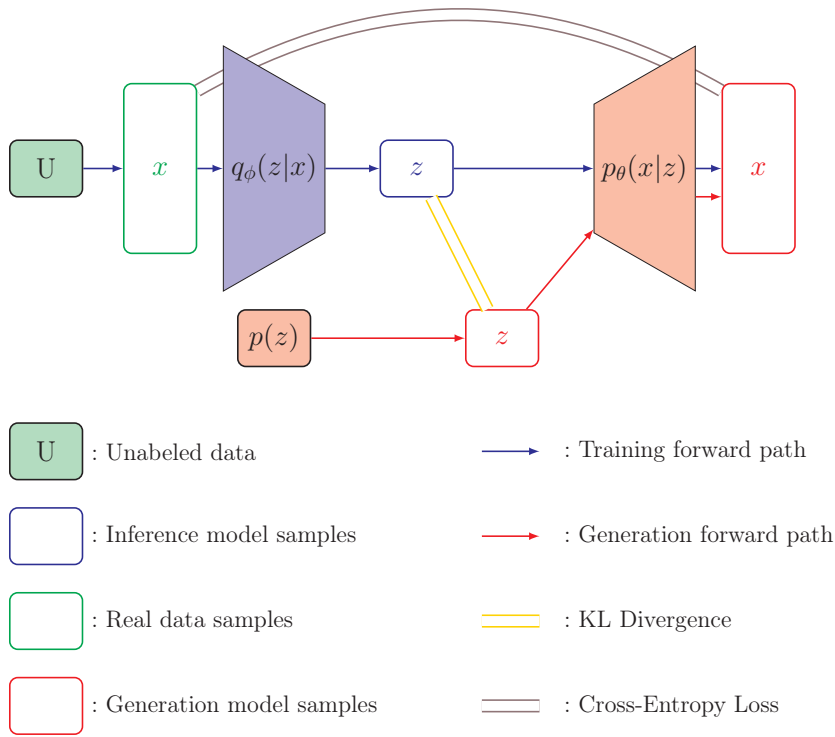
\begin{figure}[!t]
    \hskip 2cm
    \begin{minipage}[b]{1.0\textwidth}
    \begin{adjustbox}{minipage=\textwidth,scale=0.75}
        \begin{tikzpicture}[node distance = 2cm, auto]
        \tikzstyle{main}=[rounded corners, minimum size = 10mm, thick, draw =black!80, node distance = 2cm, font=\Large, text centered]
        \tikzstyle{connect}=[-latex, thick]

        
        \node[main,text width=1cm,minimum height = 3cm, color=green!100] (xdat) [] {$x$};
        \node[main,text width=1cm,minimum height = 1cm,  color=black!100, fill=green!30, left of=xdat] (U) [] {U};
        
        \node [trapezium, trapezium left angle=60, trapezium right angle=60,text centered,text width = 2cm,minimum height=1cm, minimum width=2cm, draw=black, fill=blue!30, rotate around={-90:(0,0)}](encoder) [above of=xdat]{\rotatebox{90}{\Large $q_\phi(z|x)$}};
        
        \node[main,text width=1cm,minimum height = 1cm, color=black!100, below of=encoder, fill=red!30, yshift=-1cm] (pz) [] {$p(z)$};

        \node[main,text width=1cm,minimum height = 1cm, color=blue!100, right of= encoder, yshift=-0cm, xshift=0.5cm] (zinf) [] {$z$};
        
        \node[main,text width=1cm,minimum height = 1cm, color=red!100, right of= encoder, yshift=-3.0cm, xshift=2cm] (zgen) [] {$z$};
        
        \node [trapezium, trapezium left angle=60, trapezium right angle=60,text centered,text width = 2cm,minimum height=1cm, minimum width=2cm, draw=black, fill=red!30, rotate around={90:(0,0)}, yshift=-6.5cm, xshift=2cm](decoder) [left of= encoder ]{\rotatebox{-90}{\Large $p_\theta(x|z)$}};
        \node[main,text width=1cm,minimum height = 3cm, color=red!100] (xgen) [right of= decoder] {$x$};
        
      \path 
              (U) edge [connect, color=blue!100] (xdat)
              (xdat) edge [connect, color=blue!100] (encoder)
              (encoder) edge [connect, color=blue!100] (zinf)
              (zinf) edge [connect, color=blue!100] (decoder)
              (decoder) edge [connect, color=blue!100] (xgen)
              (decoder) edge [connect, color=red!100,transform canvas={yshift=-0.5cm}] (xgen)
              (pz) edge [connect, color=red!100] (zgen)
              (zgen) edge [connect, color=red!100] (decoder)
              
              (zinf) edge [connect, -, double distance = 5pt, color=Gold!100] (zgen)
              ;
              
              \begin{pgfonlayer}{bg}    
                \path (xdat) edge [connect, -, double distance = 5pt, color=Gold!50!Blue, bend left=30,transform canvas={yshift=1cm}] (xgen);
              \end{pgfonlayer}

        \node[main,text width=1cm,minimum height = 1cm,  color=black!100, fill=green!30, left of=U, yshift=-5cm, xshift=2cm] (ULeg) [] {U};
         \node[right] at (ULeg.east) {: Unabeled data};
         \node[main,text width=1cm,minimum height = 1cm, color=blue!100, below of= ULeg, yshift=0.5cm] (infvar) [] {};
         \node[right] at (infvar.east) {: Inference model samples};
         \node[main,text width=1cm,minimum height = 1cm, color=green!100, below of= infvar, yshift=0.5cm] (datvar) [] {};
         \node[right] at (datvar.east) {: Real data samples};
         \node[main,text width=1cm,minimum height = 1cm, color=red!100, below of= datvar, yshift=0.5cm] (genvar) [] {};
         \node[right] at (genvar.east) {: Generation model samples};
         
         \node[right of=ULeg, xshift=5.5cm] (Train1) [] {};
         \node[right of=ULeg, xshift=6.8cm] (Train2) [] {};
         \path (Train1) edge [connect, color=blue!100] (Train2);
         \node[right] at (Train2.east) {: Training forward path};
         \node[below of=Train1, yshift=0.5cm] (Gen1) [] {};
         \node[below of=Train2, yshift=0.5cm] (Gen2) [] {};
         \path (Gen1) edge [connect, color=red!100] (Gen2);
         \node[right] at (Gen2.east) {: Generation forward path};
         \node[below of=Gen1, yshift=0.5cm, xshift=0.0cm] (KL1) [] {};
         \node[below of=Gen2, yshift=0.5cm, xshift=0.0cm] (KL2) [] {};
         \path (KL1) edge [connect, -, double distance = 5pt, color=Gold!100] (KL2);
         \node[right] at (KL2.east) {: $\KL$ Divergence};
         \node[below of=KL1, yshift=0.5cm] (CE1) [] {};
         \node[below of=KL2, yshift=0.5cm] (CE2) [] {};
         \path (CE1) edge [connect, -, double distance = 5pt, color=Gold!50!Blue] (CE2);
         \node[right] at (CE2.east) {: Cross-Entropy Loss};
        \end{tikzpicture}
    \end{adjustbox}
    \vskip -5px
    \end{minipage}
            \caption{\centering Variational Autoencoder Functioning scheme. Inference paths and components are colored in red, while components only used during training are colored in blue. Loss functions linking the different variables are sketched using links with two lines.}
            \label{fig:VAE1}
\end{figure}

Figure~\ref{fig:VAE1} summarizes the way a VAE is built, trained, and used for generation after training. It should be emphasized that, in the context of generative modeling, the input samples from the ground truth distribution together with the inference network (the blue components and the green components in Figure~\ref{fig:VAE1}) are only used during training. At test time, only the prior distribution and the decoder (the red components in Figure~\ref{fig:VAE1}) are used. 

Figure~\ref{fig:VAE1} also illustrates the way ELBo is calculated. The first term in ELBo is a reconstruction term where the probability of the input sample $x\sim p_{data}(x)$ is evaluated according to $p_\theta(x|z)$, the output distribution on observations conditioned on a sample from the encoder's distribution $q_\phi(z|x)$. The second term  brings this encoder distribution, or approximate posterior, $q_\phi(z|x)$, closer to the prior distribution. In simple terms a VAE is trained to \textit{reconstruct} input samples from latent codes that are \textit{made likely to come from} the prior distribution.


\section{Implementation Details}
\label{VAEIMPLEMBGSEC}
The objective ELBo requires calculating expectations over the distribution $q_\phi(z|x)$. This expectation is classically approximated using a Monte-Carlo estimate, \textit{i.e.} an average of the term inside the expectation applied to a few samples from $q_\phi(z|x)$. Fortunately, empirical investigations~\cite{Kingma2014Auto-encodingBayes, Bowman2016GeneratingSpace} have shown that VAEs can be trained using single-sample estimations of the aforementioned expectations, and that learning with multiple samples displays no significant gain compared to learning with a single sample. This means that the reconstruction term in $\ELBo$ can be approximated during training as follows:

\begin{align}
         \mathbb{E}_{z\sim q_\phi(z|x)}\left[\log p_\theta(x|z)\right]&\approx \sum_{i=1}^N \frac{1}{N}\log p_\theta(x|z_i)\approx \log p_\theta(x|z_0)\\ 
      \text{s.t. }&\hspace{0.2 cm} z_0, z_1, \dots, z_N \sim q_\phi(z|x) 
\end{align}
Similarly, the Kullback-Leibler term can be efficiently estimated as follows:

\begin{align}
        \KL[q_\phi(z|x)||p(z)]&=\nonumber\\
      \mathbb{E}_{z\sim q_\phi(z|x)}\left[\log\frac{p(z)}{q_\phi(z|x)}\right]&\approx \sum_{i=1}^N \frac{1}{N} \log\frac{p(z_0)}{q_\phi(z_0|x)} \approx \log\frac{p(z_0)}{q_\phi(z_0|x)}\\
      \text{s.t. }&\hspace{0.2 cm} z_0, z_1, \dots, z_N \sim q_\phi(z|x) 
\end{align}

Despite the fact that it is usable, the Monte-Carlo sampling approximation can be avoided for the Kullback-Leibler divergence term in ELBo in cases where a closed form is available for this divergence, \textit{e.g.} if the prior $p(z)$ and the approximate posterior $q_\phi(z|x)$ are both Gaussian. 

Although sampling enables efficiently estimating expectations, it requires the sampling operations to be differentiable in order to propagate the gradients to the encoder and to learn the parameters $\phi$. This is made possible through this simple observation: given $\mu\in \mathbf{R}^{d}$ and $\sigma\in \mathbf{R}^{d*d}$ if $z$, a $d$-dimensional random variable, follows a standard normal distribution $\mathcal{N}(0, I_d)$, then $\sigma z+\mu$ follows $\mathcal{N}(\mu, \sigma)$. Accordingly, the expectation of a function $f(z)$ over a Gaussian distribution $q_\phi(z|x)$ with mean vector $\mu(x)$ and standard deviation $\sigma(x)$ can be reformulated as follows:
\begin{align}
    \mathbb{E}_{z\sim q_\phi(z|x)}[f(z)] = \mathbb{E}_{z\sim \mathcal{N}(0, I_d)}[f(\sigma(x)*z+\mu(x))] 
\end{align}
Using the above formula decouples the parameterization from the sampling procedure and introduces it through differentiable operations (addition and multiplication). This trick, called the \textit{reparameterization trick}~\citep{Kingma2014Auto-encodingBayes}, is applicable to any distribution from the \textit{location-scale} family (Gaussian, Laplacian, $\dots$). Posterior to the initial work of \citet{Kingma2014Auto-encodingBayes} on Variational Autoencoders where the reparametrization trick was introduced, differentiable sampling schemes have also been developed for Categorical distributions such as the Gumbel-Softmax Trick~\citep{Jang2017CategoricalGumbel-softmax}, and direct optimization through $\argmax$~\cite{lorberbom2019direct}.

Besides the above implementation tricks, a few techniques have also been developed to reduce the variance of the stochastic gradient estimation during the optimization process. Namely, throughout the experiments carried in this thesis, we use a technique called Sticking The Landing (STL-ELBo; \citealp{Roeder2017StickingInference}) which helps reduce the variance of the gradients estimated during the learning process. The techniques stems from a derivation on the gradient of ELBo which identifies a term which is null in expectation. To identify this term, let us first derive $\ELBo$ as follows:
\begin{align}
     \ELBo &= \mathbb{E}_{z\sim q_\phi(z|x)}\left[\log p_\theta(x|z)\right] - KL[q_\phi(z|x)||p(z)]\\
     &= \mathbb{E}_{z\sim q_\phi(z|x)}\left[\log p_\theta(x|z) + \log p_\theta(z) - \log q_\phi(z|x)\right]\\
     &= \mathbb{E}_{z\sim q_\phi(z|x)}\left[\log p_\theta(z|x) + \log p_\theta(x) - \log q_\phi(z|x)\right]
\end{align}
\noindent where the last line exploits the equality $p_\theta(x|z)p(z)=p_\theta(x, z)=p_\theta(z|x)p_\theta(x)$. In the following, we calculate the total derivative of the expression inside the expectation operator w.r.t $\phi$, which yields partial derivatives w.r.t $\phi$ and $z$ since $z$ is a function of $\phi$:
\begin{align}
    &\frac{d}{d\phi}\left[\log p_\theta(z|x) + \log p_\theta(x) - \log q_\phi(z|x)\right]\\
    =&\frac{\partial}{\partial z}\left[\log p_\theta(z|x)  - \log q_\phi(z|x)\right] \frac{\partial z}{\partial \phi} - \frac{\partial }{\partial \phi}\log q_\phi(z|x)
\end{align}
In the last line, the expectation of the last term can be proven null as follows:
\begin{align}
    &\mathbb{E}_{z\sim q_\phi(z|x)}\frac{\partial }{\partial \phi}\log q_\phi(z|x) = \int_z q_\phi(z|x) \frac{\partial }{\partial \phi}\log q_\phi(z|x) dz \\
    = &\int_z q_\phi(z|x) \frac{1}{q_\phi(z|x)} \frac{\partial }{\partial \phi} q_\phi(z|x) dz = \int_z \frac{\partial }{\partial \phi} q_\phi(z|x) dz \\
    =  &\frac{\partial }{\partial \phi} \int_z q_\phi(z|x) dz =
     \frac{\partial }{\partial \phi} 1 = 0
\end{align}
Since this term has a null expectation, it can be dropped from the gradient estimator without biasing it. The STL-ELBo technique consists in dropping this term, which leads to better log-likelihood results than the standard ELBo estimator, as shown by \citet{Roeder2017StickingInference}.


\section{Language Modeling with a VAE}
\label{VAELMSec}
VAEs were first explored as LMs by \citet{Bowman2016GeneratingSpace}. Prior to this work, Neural LMs were mainly tackled through purely auto-regressive generation schemes such as the one presented in Section~\ref{LMDEFSEC}. Purely auto-regressive generation schemes, mainly implemented in recent works with LSTMs~\cite{Melis2019MogrifierLSTM} or Transformers~\cite{Radford2018LanguageLearners}, achieve low perplexity values and model long-term dependencies through step-wise updates to a context vector, which does  not clearly separate sequence-level factors of variation for word-level factors of variation. When need arises, the sequence-level factors are usually modeled through the introduction of conditioning variables to the generative model, where the model is trained with supervised learning as was explained in Section~\ref{CLMDEFSEC}. However, ELBo provides a means to train a generative LM in the form $p_\theta(x)=\int p_\theta(x|z)p(z)dz$ while waving the need for any specifics on $z$ beyond the prior $p(z)$. Therefore, in order to train generative models with explicit sequence-level representations $z$, \citet{Bowman2016GeneratingSpace} explored the VAE option.

\citet{bowman-etal-2015-large} trained VAEs for language modeling, using single-layer LSTMs as encoders and RNNs conditioned on the latent codes as decoders. They obtained competitive results with RNN LMs on the Penn Treenbank~\cite{marcinkiewicz1994building} dataset. 
Their work also led to a few interesting observations on such VAE LMs. They first observed that, given that a VAE encoder is trained to concentrate all representations in the same vicinity through the $\KL$-Term, and to generate well-formed sentences from this vicinity through the reconstruction term, VAEs had what they called a \textit{smooth latent space}. This smoothness manifests in the fact that intervening on a sample $z$ (\textit{e.g.} resampling components or interpolating with another vector) from an encoded sentence and decoding the resulting latent vector resulted in well-formed sentences. In contrast to this, operating on representation from classical Sequence-to-Sequence Autoencoders~\citep{Sutskever2014} in their latent spaces often leads to malformed decoded sentences. To illustrate this smoothness, \citet{Bowman2016GeneratingSpace} produce sentences by decoding latent codes corresponding to progressive linear interpolations of the representations of two sentences\footnote{This process is often refered to as a \textit{homotopy}.}. The original sentences and the sentences generated from the progressive interpolation are displayed in Table~\ref{tab:homotopy}.

\begin{table}
\normalsize
\centering
\begin{tabular}{l}
\hline
\textbf{i went to the store to buy some groceries .}\\
i store to buy some groceries .\\
i were to buy any groceries .\\
horses are to buy any groceries .\\
horses are to buy any animal .\\
horses the favorite any animal .\\
horses the favorite favorite animal .\\
\textbf{horses are my favorite animal .}\\
\hline
\end{tabular}
\caption{Example sentence interpolations using VAE latent variables from \citet{Bowman2016GeneratingSpace}. Original sentences constituting the edges of the linear interpolation are given in bold.}
\label{tab:homotopy}
\end{table}

The sentences displayed in Table~\ref{tab:homotopy} clearly show progressive change from the first sentence to the second sentence.
This smoothness property is especially valuable when it comes to combining characteristics from different sentences in the context of controllable generation.

A second valuable observation highlighted by this work was a phenomenon dubbed \textit{posterior collapse}, where all the posterior distributions $q_\phi(z|x)$ converge (or \textit{collapse}) to the prior distribution $p(z)$, and therefore become uninformative on the input $x$. This phenomenon stems from the fact that auto-regressive decoders used in VAE LMs are powerful enough to learn good LMs without the need for latent variables. Solutions to avoid this problem are detailed in Section~\ref{PColSec}

\section{A Tighter Lower-Bound to the Log-Likelihood with Importance Weighting}
\label{VAEIWBGSEC}
VAEs allow optimizing a generative model via ELBo, a lower-bound to the exact log-likelihood. ELBo is a convenient strategy when it comes to training, but testing generative models often requires estimating the exact likelihood of data according to a model. In fact, as explained in Section~\ref{METRLMDEFSEC}, assessing the generative capabilities of an LM requires estimating its perplexity, which is based on the exact likelihood measure. In order to provide such a measure for VAEs, so as to be able to compare their generative capabilities to standard Maximum Likelihood-based LMs, one needs a tight estimate of the exact likelihood.

Posterior to \citet{Kingma2014Auto-encodingBayes}'s work on VAEs, the following Importance Weighting-based lower-bound to the exact log-likelihood of a VAE has been developed by \citet{Burda2016ImportanceAutoencoders}:
\begin{multline}
    \log p_\theta(x) \geq 
     \mathbb{E}_{z_1, ...,z_k \sim q_\phi(z|x)} \Bigl[\log\frac{1}{k}\sum_{i=1}^k\frac{p_\theta(x, z_i)}{q_\phi(z_i|x)}\Bigr]
     =\IWAE(x; z; k)\label{IWAEObj}
\end{multline}
VAEs using this Importance Weighted Objective (IWO) during training are referred to as Importance Weighted Autoencoders (IWAEs). Notice that this objective is equal to ELBo when it uses a single sample (\textit{i.e.} $k=1$). More importantly, \citet{Burda2016ImportanceAutoencoders} provided a proof establishing that this objective satisfies the following inequality:
\begin{align}
    \log p_\theta(x) \geq \IWAE(x; z; k+1)\geq \IWAE(x; z; k)\label{IWAEIneq}
\end{align}
and that, under mild assumptions, $\IWAE(x; z; k)$ converges to $\log p_\theta(x)$ when k goes to infinity. This means that using more importance samples (\textit{i.e.} a higher $k$) allows monotonously approaching an exact estimate of the log-likelihood, and therefore approaching the exact value of perplexity\footnote{Since perplexity is a decreasing function of likelihood, our estimate here would be an upper-bound to the perplexity.} for a language model. 

Other tight lower-bounds to the exact log-likelihood were developed after IWAE, such as the Thermodynamic Variational Objective (TVO; \citealp{Masrani2019TheObjective}), but Importance Weighting remains the standard method to assess VAE-based LM perplexities.

\section{Dealing with Posterior Collapse}
\label{PColSec}
As mentioned in Section~\ref{VAELMSec}, VAEs suffer from posterior collapse, a phenomenon where the VAE LM converges to a local minimum where the LM is only learned through the autoregressive decoder, and the distributions $q_\phi(z|x)$ parameterized by the encoder all converge exactly to $p(z)$ minimizing the Kullback-Leibler divergence term in ELBo. A first solution to this problem was proposed in the work of \citet{Bowman2016GeneratingSpace}, and consists in \textit{jump-starting} the encoder by multiplying the $\KL$ term by $\beta=0$ for a few optimization steps, so that the whole network is only optimized for reconstruction. The $\KL$ term is then slowly introduced over a number of \textit{annealing} steps by linearly increasing $\beta$ to 1. Another trick they implemented which mitigated posterior collapse was \textit{word-dropout}: A technique where a proportion (set to 40\% in their work) of the previous words in the auto-regressive generation scheme of decoders is replaced with a null vector during training to enforce reliance on the latent codes.  

In the later work of \citet{Kingma2016ImprovedFlow}, this solution was improved upon through a strategy called Free-bits strategy or $\KL$-thresholding strategy. Given that, for a $d$-dimensional latent variable $z=\{z_1, \dots, z_d\}$, the prior $p(z)$ and the approximate posterior $q_\phi(z|x)$ are most often respectively set to a standard normal distribution $\mathcal{N}(0, I_d)$ and to a diagonal Gaussian distribution $\mathcal{N}(\mu_\phi(x), \sigma_\phi(x))$, The KL-term can be written as the sum of dimension-wise $\KL$ divergences between prior and approximate posterior distributions, \textit{i.e.} $\KL[q_\phi(z|x), p(z)]=\sum_i^d \KL[q_\phi(z_i|x), p(z_i)] $. The $\KL$-thresholding strategy consists in minimizing these dimension-wise $\KL$ terms as part of the ELBo optimization only down to a global threshold $\gamma$.

Other works have proposed diverse solutions to the posterior collapse problem throughout the years such as a cyclic annealing scheme for the $\KL$ term~\citep{Fu2019CyclicalVanishing}, or aggressive optimization phases that focus on the encoder~\citep{He2019LaggingAutoencoders}, but \citet{li-etal-2019-stable} have shown that simply combining $\KL$ annealing with $\KL$-thresholding leads to the most effective fix to the posterior collapse problem while still requiring minimal modification to the original VAE learning scheme. In fact, the modification simply consists in an ELBo that takes the following form:

\begin{align}
     \ELBo(x;z) = \mathbb{E}_{z\sim q(z)}\left[\log p_\theta(x|z)\right] - \beta \sum_i^d \max(\gamma, \KL[q_\phi(z_i|x)||p(z_i)])
\end{align}
\noindent where $\gamma$ is fixed, and $\beta$ is used for the progressive introduction of the $\KL$ term to the optimization procedure as described above. It is worth noting that a common practice in VAE-based language modeling to counteract posterior collapse is simply to lower the final value of $\beta$ so as to allow $q_\phi(z|x)$ to drift further from $p(z)$ in order to better optimize the model for reconstruction and therefore to absorb more information about $x$. This is especially common in works that employ VAEs as \textit{regularized} Autoencoders for controllable generation (\textit{e.g.} \citealp{chen-etal-2019-controllable}).

More recent works like those of \citet{Wu2020OnBeyond} and \citet{menon2022forget} propose solutions that further mitigate posterior collapse, but they both employ auxiliary networks with sizes similar to that of the original generative model. Since designing competitive LMs in the Deep Learning era is a memory sensitive endeavor, we stick to the above combination proposed by \citet{li-etal-2019-stable} for the works presented in this thesis as it constitutes a suitable middle ground between effectiveness and memory efficiency.   

\section{Semi-Supervised Learning with VAEs}
\label{SSVAEBG}
Standard supervised learning considers the setup where an inference module $q_\phi(y|x)$ is tasked with learning to predict labels $y$ from observations $x$ using a dataset of labeled samples $L=\{(x_1, y_1), ..., (x_{|L|}, y_{|L|})\}$. The semi-supervised learning setup however, also makes use of an auxiliary set of unlabeled data points $U=\{x'_1, ..., x'_{|U|}\}$ which are typically available in much larger quantities than their labeled counterparts. The use of VAEs in semi-supervised learning has first been explored by \citet{Kingma2014a}. 

The idea here, is to use the VAE encoder as a classifier by introducing the target label as a latent variable to the VAE-based generative model. Specifically, 
besides the usual unobserved latent variable $z$, the semi-supervised VAE framework also uses a partially-observed latent variable $y$. These variables are typically modeled as independent latent variables such that $q_\phi(y, z|x)=q_\phi(y|x)q_\phi(z|x)$ and $p_\theta(x)=\int p_\theta(x|y, z)p(y) p(z)$.
The encoder $q_\phi(y|x)$ serves both as the inference module for the supervised task, and as an approximate posterior (and encoder) for the $y$ variable in the VAE framework. Accordingly, $y$ is learned on data from $U$ in a similar manner to $z$, \textit{i.e.} as an unobserved latent variable as part of the VAE generative model. But when presented with data from $L$, the learning procedure uses $y$ both as a target for supervised learning with the encoder $q_\phi(y|x)$ and as an observed conditioning variable for the decoder $p_\theta(x|y, z)$ on $L$.  
Formally, the training objective $ \mathcal{J}^\alpha$~\citep{Kingma2014a} for a Semi-Supervised VAE (SSVAE)  is expressed as follows:
 \begin{align}
  \mathcal{J}^\alpha &= \sum_{(x, y) \in L}\Bigr( \ELBo((x, y);z)+ \alpha \hspace{1mm} \log q_\phi(y|x)\Bigl)\nonumber\\
  &+ \sum_{x \in U} \ELBo(x;(y, z)) \label{JALPHA}
 \end{align}
 \noindent where the first argument of $\ELBo$ is the set of observed variables, and the second argument is the set of unobserved variables. As can be seen in the formula above, for samples coming from $L$, $q_\phi(y|x)$ is trained to predict $y$ using a Cross-Entropy\footnote{Recall, here, that the Cross-Entropy objective is an application of Maximum-Likelihood Estimation. In fact, maximizing the likelihood of ground truth labels according to a classifier is equivalent to minimizing the Cross-Entropy between the ground truth label distribution, and the label distribution defined by the classifier.} objective that is weighted with regard to the remaining terms of the overall objective using a scalar $\alpha$.
\section{Conclusion}
This chapter is an introduction to the main technical framework for this thesis: Variational Autoencoders. We explained the motivation behind generative modeling with VAEs in general, and Language Modeling with VAEs in particular. Starting from the standard Autoencoder architecture (\S~\ref{AEARCHIBG}), we explained the additions and modifications required to obtain the basic components of a VAE, and to derive its objective ELBo (\S~\ref{VAETHBGSEC}). To give better practical insight into VAEs, we explained in Section~\ref{VAEIMPLEMBGSEC} the basic tricks required to implement it and to train it. Given the NLP-specific context of this thesis, we also summarized the main findings of the first work on VAE-based LMs (\S~\ref{VAELMSec}). We also layed-out the IWO, a tighter lower-bound to the exact log-likelihood which constitutes the basis for estimating perplexity in order to measure language modeling performance (\S~\ref{VAEIWBGSEC}). The particular case of language modeling  with VAEs  suffers from an issue called posterior collapse, which we explained and for which we described a few solutions while pointing to the solution we retain for the works described in this thesis (\S~\ref{PColSec}).

Finally, we broke down semi-supervised learning with VAEs in the final section of this chapter (\S~\ref{SSVAEBG}) so as to provide the prerequisites for our semi-supervised learning-related contribution. The second part of our contributions, which is dedicated to disentanglement, relies on the disentanglement capabilities of VAEs. As this is better explained following a few introductory notions, we elaborate on the relation between VAEs and disentanglement in the chapter dedicated to disentanglement (Chapter~\ref{DISENTCHAP}).

\chapter{Disentangled Representation Learning}
\label{DISENTCHAP}

A neural network is a function $f_\theta$ with parameters $\theta$, where the parameters are estimated so as to \textit{learn} a relation between an input to the function and a desired output. This function is most often built as a composition of functions such that $f_\theta=f^1_{\theta_1}\circ f^2_{\theta_2}, \dots, \circ f^n_{\theta_n}$ where each $f^i_{\theta_i}$ is called a \textit{layer} and $n$ is the number of layers in the neural network $f_\theta$. The individual $f^i_{\theta_i}$ have multidimensional outputs, where each output is called a neuron, and any combination of neurons (\textit{e.g.} a layer, part of a layer, a concatenation of layers) is called a \textit{neural representation}. Neural representations have shown great promise as automatic feature extractors in and outside of NLP~\cite{Mikolov2013, simonyan2014very, he2016deep, devlin-etal-2019-bert}. However, aside from output neurons, it is difficult to separate and identify understandable concepts in neural representations. Investigating methods which yield neural representations with identified interpretable concepts is the very purpose of disentanglement. More precisely, disentangled representation learning aims at designing methods which produce neural representations where each dimension, or set of dimensions, relates significantly to an understandable concept.

Interest in disentangled neural representation learning started emerging in the early 2010's where, for instance,  \citet{Bengio2013Representation} argue in their review of representation learning that it is crucial for the future of AI to disentangle understandable factors of variation within neural representations. It persists as a crucial research direction in recent works such as \citet{Rudin2022InterpretableChallenges} which lists supervised and unsupervised disentanglement within the ten great challenges for interpretable machine learning. Besides interpretability, disentanglement is also sought as a means to improve sample complexity under the intuition that reducing inputs to a minimalistic set of human-validated interpretable concepts should improve generalization. This intuition was empirically confirmed by a number of studies such as the large-scale experiments carried by \citet{locatello2020weakly} which show that disentangled representations improve  sample efficiency and generalization under covariate shift(\textit{i.e.} a change in the distribution of observations).

From a practical point of view, disentanglement is induced either with supervised learning where neural representations are constrained to be informative on a given factor, or through unsupervised learning which is the paradigm chosen for this thesis. An underlying assumption to unsupervised disentanglement, is that the data can be factored into a set of \textit{independent} generative factors. This was the main guiding principle for the design of unsupervised disentanglement methods, and also the main reason behind the successes of VAEs at disentanglement.

As a starting point, we describe the mechanisms inside VAEs that encourage obtaining independent generative factors (\S~\ref{INDPLVDISENTBG}). In the following section, we clarify the relation between the factorization induced by VAEs and disentanglement using a few recent theoretical results(\S~\ref{ALIGNDISENTBGSEC}). Subsequently, we describe the few common ways inductive bias is introduced to neural networks in order for them to exhibit disentanglement, specifically in a supervised or in an unsupervised fashion(\S~\ref{DISENTINDUCSEC}). Moving on to evaluation, in Section~\ref{DisentMeasSec}, we summarize the intuition behind measures of disentanglement, and detail a few measures that serve here as an example, and later in the thesis (Chapter~\ref{chap:SynRoleDisentChap}) as the basis for disentanglement measures of our own design. Finally, we give in Section~\ref{DISENTAPPPBGSEC} an overview of the \textit{disentanglement in NLP} landscape with an emphasis on a few shortcomings of the current literature in regard to which we aim to contribute in this thesis. 

\section{Finding Independent Generative Factors with VAEs}
\label{INDPLVDISENTBG}
VAEs are a natural candidate method for disentanglement as their primary purpose is to relate observations to latent variables. The conditional or marginal distributions of these latent variables are set by the practitioner, which most often chooses them to be independent. In fact, the most common VAE is built with a standard Gaussian prior $p(z)$ and a diagonal Gaussian approximate posterior $q_\phi(z)$. To the best of our knowledge, choosing Gaussians for both distributions mostly stems from the fact that practitioners usually default to Gaussians when no information is available about the underlying distribution of a variable. This default choice also enables using a closed form of the $KL$ divergence instead of its Monte-Carlo estimate. It can be argued, for the standard Normal distribution often set to be the prior, that it works as a regularization on latent variables~\cite{Chen2018VariationalLearning,Wolf-Sonkin2018AInflection,Yacoby2020FailureTasks}, similar to a Ridge regularization~\citep{hoerl1970ridge}. As for the posterior distribution, the choice of a diagonal Gaussian was due to computational gains, since it only requires estimating the diagonal elements of the covariance matrix instead of the entire matrix. Given these design choices, latent variable modeling in VAEs aims to retrieve a set of independent generative factors from the observations, which makes it an intuitive choice for disentanglement.


The first approach to disentanglement using VAEs is called $\beta$-VAE~\cite{Higgins2019-VAE:Framework}, as it simply consists in adding a scalar $\beta$ to scale the $\KL$ term in $\ELBo$ (Eq.~\ref{ELBOBGEQ}):

\begin{align}
    \mathbb{E}_{z\sim q_\phi(z|x)}\left[\log p_\theta(x|z)\right] - \beta\KL[q_\phi(z|x)||p(z)] \label{BETAVAEEQ}
\end{align}
The argument behind this modification is fairly straightforward: The $\KL$ term pushes the distribution of the encoded representations $q_\phi(z|x)$ towards $p(z)$, which is most often a standard Normal distribution that encourages the different dimensions in $q_\phi(z|x)$ to be independent. Therefore, $\beta$-VAE allows choosing a compromise between high disentanglement with a high $\beta$, or high reconstruction fidelity with a low $\beta$.

The subsequent work of \citet{Chen2018c} further pinpoints the source of disentanglement in ELBo through a decomposition of its $\KL$ term. Assuming that observations $x$ in the training set are indexed with an integer $n\in \{1, \dots, N\}$, we can define a uniform distribution $p(n)$ over that range, and subsequently consider the distributions $q_\phi(z|n)=q_\phi(z|x_n)$, $q_\phi(z, n)=q_\phi(z|n)p(n)=q_\phi(z|n)\frac{1}{N}$ and $q_\phi(z)=\sum_1^Nq_\phi(z|n)p(n)$. These distributions can be used to write the following decomposition\footnote{The full derivation leading to this decomposition can be found in Appendix~\ref{TCDERIVAPP}.}:
\begin{align}
\mathbb{E}_{n\sim p(n)}KL[q_\phi(z|n)||p(z)] =&  KL[q_\phi(z, n)||q_\phi(z)p(n)] &\hspace{0.5cm} \text{\textcircled{1}} \nonumber\\
+& KL[q_\phi(z)||\prod_j q_\phi(z_j)] &\hspace{0.5cm} \text{\textcircled{2}}\nonumber\\
+& \sum_j KL[q_\phi(z_j)||p(z_j)] &\hspace{0.5cm} \text{\textcircled{3}} \label{TCVAEKLEQ}
\end{align}
The decomposition of the $\KL$ term of ELBo in equation~\ref{TCVAEKLEQ} displays three components: \textcircled{1} The index-code Mutual Information, which is the mutual information between inputs $x$ (indexed by $n$ in the equation) and latent variables $z$; \textcircled{2} The Total-Correlation (TC), which is a measure of dependence between components of the latent variable vector; \textcircled{3} The dimension-wise $\KL$ divergence. \citet{Chen2018c} argue that component \textcircled{2}, TC, is responsible for disentanglement, since it measures the dependence between components of latent vectors, and propose to control a weight on this component instead of the entire $\KL$ term, leading to the TC-VAE objective. 

Driven by the same intuition, other objectives were designed in order to force independence between the components of latent vectors, namely Annealed VAE~\cite{Burgess2018UnderstandingIn-VAE}, FactorVAE~\cite{Kim2018DisentanglingFactorising}, and DIP-VAE~\cite{Kumar2018VariationalObservations}. However, the large-scale experiments carried by \citet{locatello2020commentary} have shown that no objective from the above encourages disentanglement significantly more than the others, and that, in general, random seeds influence disentanglement results more than the choice of method. Therefore, for its simplicity, $\beta$-VAE is the most commonly chosen objective in recent disentanglement works. It is also the objective chosen to carry out the works described in this thesis.

\section{About the Alignment Between Independent Generative Factors and Understandable Concepts}
\label{ALIGNDISENTBGSEC}
\begin{figure*}[!h]
\centering
    \begin{minipage}[b]{0.3\textwidth}
            \centering
           \begin{minipage}[b]{\textwidth}
            \begin{adjustbox}{minipage=\textwidth,scale=0.45}
            \hspace{ 1cm} \includegraphics[trim={1.3cm 0.7cm 2.2cm 1.3cm},clip] {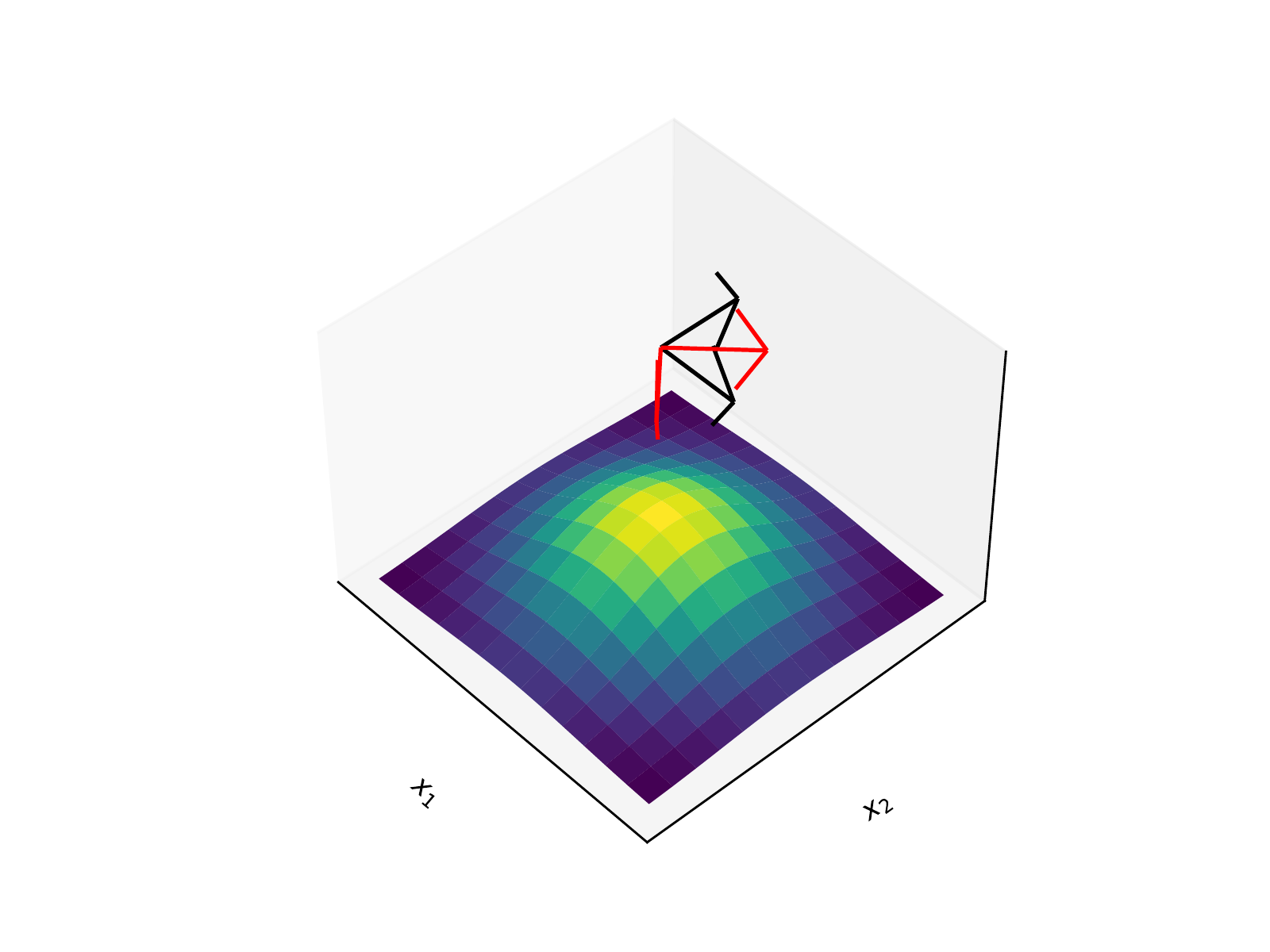}
            \end{adjustbox}
            \end{minipage}
    \end{minipage}
    \begin{minipage}[b]{0.3\textwidth}
            \centering
            \begin{minipage}[b]{\textwidth}
            \begin{adjustbox}{minipage=\textwidth,scale=0.45}
             \hspace{ 1cm} \includegraphics[trim={1.4cm 0.7cm 2.2cm 1.3cm},clip] {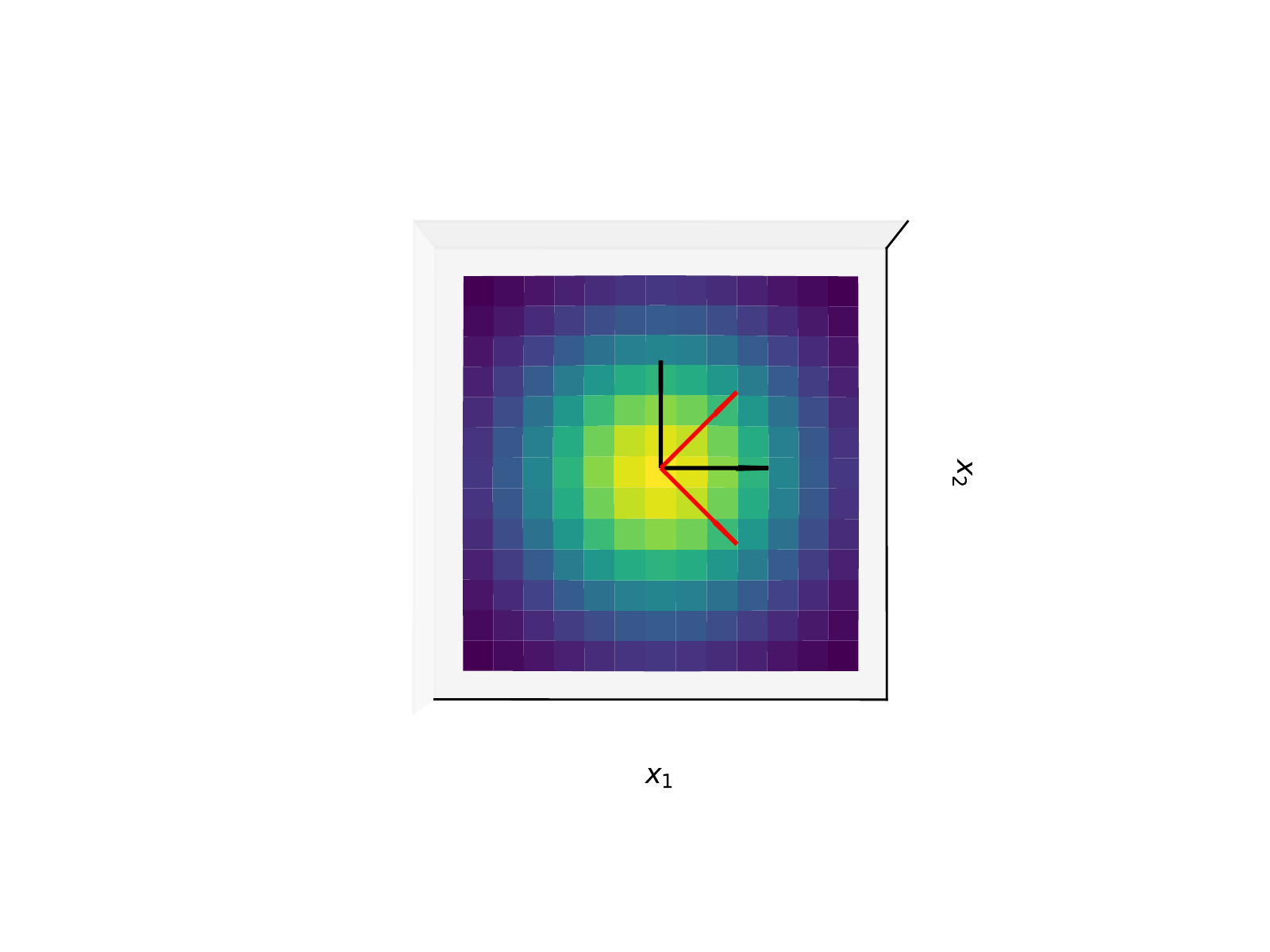}
            \end{adjustbox}
            \end{minipage}
    \end{minipage}
    \begin{minipage}[b]{0.3\textwidth}
            \centering
            \begin{minipage}[b]{\textwidth}
            \begin{adjustbox}{minipage=\textwidth,scale=0.45}
             \hspace{ 1cm} \includegraphics[trim={1.4cm 0.7cm 2.2cm 1.3cm},clip] {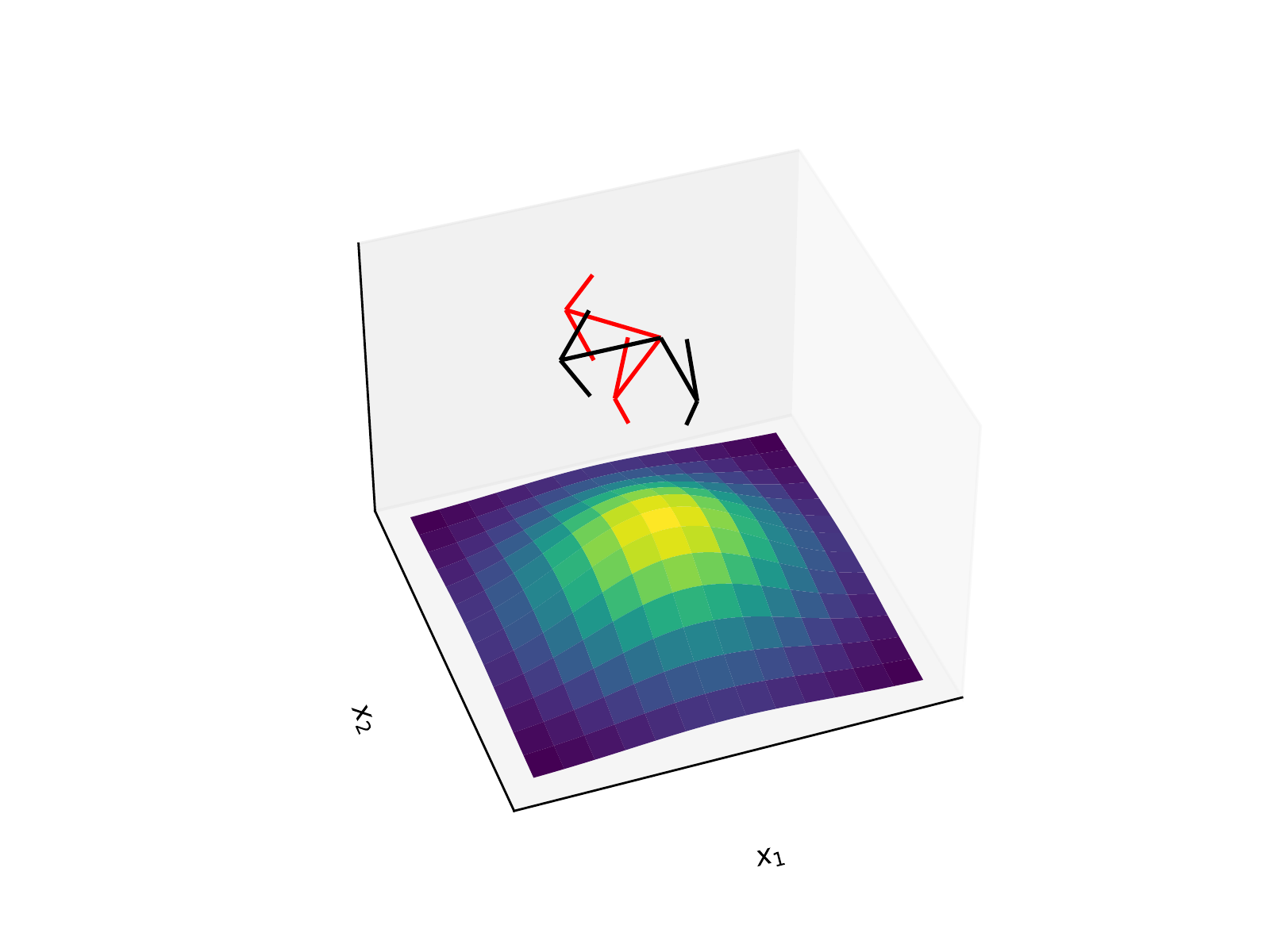}
            \end{adjustbox}
            \end{minipage}
    \end{minipage}
            \caption{Different perspectives on a 2D Gaussian distribution. Arrows in black represent directions $x_1$ and $x_2$, which are two independent axes of variation for this Gaussian distribution. Arrows in \textcolor{red}{red} represent directions $(x_1+x_2)$ and $(x_1-x_2)$, which also represent independent axes of variation.}
            \label{fig:GaussVectors}
\end{figure*}

Crucially, the above methods \textit{only} enable relating observations $x$ to independent generative factors $z$. However, does this mean that such methods will converge to \textit{THE} target independent generative factors ? As will be argued below using Figure~\ref{fig:GaussVectors}, the answer is "no".

Figure~\ref{fig:GaussVectors} is a plot of the density of a 2D standard Normal distribution. Coordinates $x_1$ and $x_2$ of 2D samples from this distribution have independent distributions. Therefore, $x_1$ and $x_2$ represent independent generative factors for samples from this distribution. However, notice that $(x_1+x_2)$ and $(x_1-x_2)$ also have independent realizations, and thus equally represent independent generative factors for samples from this distribution. The above simple case demonstrates that simply seeking independent generative factors \textit{cannot alone} yield generative factors that align with a predefined set of such factors. This rationale is the basis of the \textit{unsupervised disentanglement impossibility} result presented in the work of \citet{locatello2020commentary}.

Through a formal proof that unsupervised factorization into independent generative factors cannot align different latent variables with identifiable disentangled generative factors,  \citet{locatello2020commentary} argue that disentanglement requires an additional form of inductive bias in order to align latent variables with identifiable generative factors.

Posterior to that work, \citet{Rolinek2019VariationalAccident} presented yet another important result: standard VAE implementation guidelines induce a behavior that mimics Principled Component Analysis (PCA): As mentioned earlier, the most commonly used distributions for a VAE are a standard Normal prior, and a diagonal Normal approximate posterior. \citet{Rolinek2019VariationalAccident} show that a PCA-like behavior is displayed by VAEs and that it is caused by diagonal posteriors. Most importantly, PCA factors observations into independent axes of variation, and greedily select the highest direction of variation at each factorization step. Contrary to a plain search of independent factors, this process converges to a fixed set of generative factors (up to a permutation operation).

 Rolinek's result requires the covariance matrix of the different generative factors to have distinct singular values. This means that the different generative factors must contribute differently to the overall variance of the samples. In fact, for generative factors with equal contribution to the overall variance, \textit{e.g.} $x_1$ and $x_2$ for the isotropic Gaussian in Figure~\ref{fig:GaussVectors}, Locatello's impossibility result stands unquestioned. In practice, Rolinek shows that disentanglement works out better when the gap in variance is higher between generative factors, which partly explains the difference in disentanglement performance across different datasets. In light of these observations, and as concluded by Locatello, VAEs encourage disentanglement but require additional inductive bias for disentanglement either from the model or from the data. The nature of such inductive bias is explicited in the next section. 

\section{Disentanglement as a Result of Inductive Bias}
\label{DISENTINDUCSEC}
In this section, we describe different methods to implement bias inducing separation in the latent representations of neural networks. The most explicit method is of course a form of supervision. To implement separation, a latent variable $z$ is usually split into two latent variables $z_1$ and $z_2$, where we try to associate one of the variables (\textit{e.g.} $z_1$ in what follows) to the target factor of variation $y$. Given generative model $p_\theta(x)=\int_{z_1, z_2} p_\theta(x|z_1, z_2)p(z_1, z_2)dz_1dz_2$ with approximate posterior $q_\phi(z_1, z_2|x)$ and labeled data $L$ aligning observation $x$ with generative factors $y$, the latent variables $z_1$ can be made informative about generative factor $y$ simply by means of the objective:
\begin{align}
    \sum_{(x, y) \in L} \ELBo(x; (z_1, z_2)) + \alpha \mathbb{E}_{z_1\sim q_\phi(z_1|x)}[q_\phi(y|z_1)] \label{SUPDISENTEQ}
\end{align}
\noindent where $q_\phi(y|z_1)$\footnote{If the random variable $z_1$ has the same support as $y$, $z_1$ can be directly trained to take the values of $y$ by setting $q_\phi(y|z_1)$ to an identity function. In this case, one can even use the Semi-Supervised VAE objective for labeled data from Equation~\ref{JALPHA} in Section~\ref{SSVAEBG}. } is an additional inference module relating $z_1$ to $y$, and $\alpha$ is a weighting coefficient. The objective described in Equation~\ref{SUPDISENTEQ} encourages $z_1$ to be fully informative on $y$. Given that VAEs also encourage latent variables to be independent, encouraging all the information about $y$ to be in $z_1$ also discourages $z_2$ from including any information on $y$.

\begin{figure}[!h]
\centering
    \begin{minipage}[b]{1.0\textwidth}
            \centering
           \begin{minipage}[b]{\textwidth}
            \begin{adjustbox}{minipage=\textwidth,scale=0.8}
            \vskip 2cm
             \includegraphics
            {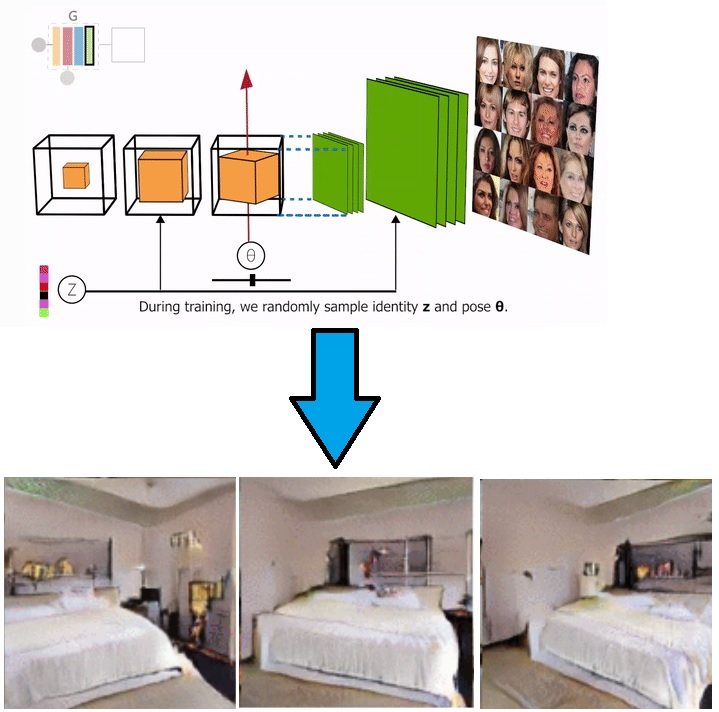}
            \end{adjustbox}
            \end{minipage}
            
    \end{minipage}
    \caption{\label{fig:HOLOGAN}A graphic by \citet{HoloGAN2019} summarizing the generation steps in HoloGAN together with a few generated sample at different angles.}
\end{figure}

Alternatively, bias towards disentanglement can be induced through the way a model is built. To better understand this, we take the example of HoloGAN~\cite{HoloGAN2019}, a model for natural image generation. During generation, this model generates a 3D structure from a latent vector $z$, chooses an angle $\theta$, then projects the 3D structure on a plane according to $\theta$ in order to obtain a 2D image. This model is trained as a Generative Adversarial Network (GAN; \citealp{Goodfellow2014}) to generate natural images. Without any information on the latent 3D structure or its orientation in samples from the training set, this model naturally learns to decouple elements represented in natural images from their pose. To better summarize the idea, Figure~\ref{fig:HOLOGAN} shows a graphic by \citet{HoloGAN2019} as well as a few samples from their generative model.

HoloGAN's example shows that disentanglement can be induced in neural models given a \textit{shared inductive bias} between the model and the data it is trained on. Namely here, the model infers 3D structures and projects them on 2D planes in accordance with the data which consists in 3D structures projected on 2D planes.

\section{Measuring Disentanglement}
\label{DisentMeasSec}
Perfectly disentangled neural representations are supposed to display 3 characteristics~\cite{Ridgeway2018LearningLoss, Eastwood2018ARepresentations}: \begin{enumerate}
    \item \textit{Modularity/Disentanglement:} Each portion\footnote{\textit{Portions} here emphasizes the fact that one can aim for disentanglement along distinct single dimensions in the neural representations, or a multi-dimensional partition of neural representations.} of the latent code must capture at most one factor of variation.
    \item \textit{Compactness/Completeness}: Each factor of variation must be captured by at most one portion of the latent code. 
    \item \textit{Informativeness/Explicitness}: The representations must be maximally informative on the factors of variation.
\end{enumerate}

\paragraph{A generic measure for disentanglement:}

Multiple \textit{generic} metrics have been developed in the literature with the above characteristics in mind~\cite{Higgins2019-VAE:Framework, Kim2018DisentanglingFactorising, Ridgeway2018LearningLoss}. As an example, we detail the design of Mutual Information Gap (MIG; \citealp{chenisolating}) which is fairly generic, accounts for axis alignment between latent variable dimensions and generative factors, and requires no \textit{post-hoc} classifier training.
To define this metric, \citet{chenisolating} first define the joint distribution $q_\phi(z_j, v_k)=\sum_{n=1}^N p(v_k)p(n|v_k)q_\phi(z_j |n)$, where $v_k\in \{v_1, \dots, v_K\}$ is a generative factor, $n$ is the index of a ground truth sample among $N$ available samples, and $z_j\in \{z_1, \dots, z_J \}$ is a component of the latent vector.  
Given that joint distribution, the following empirical estimate of mutual information between latent dimension $z_j$ and generative factor $v_k$ is defined :
\begin{align}
I_n(z_j; v_k) = \mathbb{E}_{z_j, v_k  \sim q(z_j, v_k)}\left[log \sum_{n\in \mathcal{X}_{v_k}}q_\phi(z_j|n)p(n|v_k)\right]+H(z_j)
\end{align}
\noindent where $\mathcal{X}_{v_k}$ is the support of $p(n|v_k)$; \textit{i.e.} the available samples verifying the sampled $v_k$. Using this estimator, MIG sums over all generative factors the gap between the highest calculated $I_n$ with regard to a component of the latent vector and the second highest of such measures, normalized by the entropy of the present generative factor:
\begin{align}
    \frac{1}{K}\sum_{k=1}^K \frac{1}{H(v_k)}\left( I_n(z_{j^{(k)}}; v_k) - \max_{j\neq j^{(k)}} I_n(z_j; v_k) \right) \label{MIGEQ}
\end{align}
\noindent where $j^{(k)}= \argmax_{j} I_n(z_j; v_k)$. For each generative factor $v_k$, the corresponding component in the above sum is high if it has high mutual information with \textit{a single latent dimension}, since the value of this component is penalized by the second highest mutual information. Ideally, the component is equal to 1 if the most informative latent variable is perfectly informative on $v_k$ (meaning $ I_n(z_{j^{(k)}}; v_k)=H(v_k)$) and if all other latent variables contain no information on $v_k$ (meaning $\max_{j\neq j^{(k)}} I_n(z_j; v_k)=0$). 

Notice that, among the 3 criteria explicited above, MIG only captures Compactness and Informativeness. In fact, a representation where only one latent dimension $z_j$ captures all the information from all generative factors $v_k$ would obtain a perfect score of MIG=1. Therefore, to complement MIG on Modularity, \citet{Li2020ProgressiveRepresentations} define MIG-sup, which simply differs with MIG through a partial\footnote{The permutation is only partial in that the normalization in MIG-sup is still done with regard to the Entropy of $v_k$.} permutation of the roles of latent variables and generative factors:
\begin{align}
    \frac{1}{J}\sum_{j=1}^J \frac{1}{H(v_k)}\left( I_n(z_j; v_{k^{(j)}}) - \max_{k\neq k^{(j)}} I_n(z_j; v_k) \right)
\end{align}
\noindent where $k^{(j)}= \argmax_{k} I_n(z_j; v_k)$.

\paragraph{\textit{Ad hoc} disentanglement measure:}

The above measure assumes the tractability of its mutual information estimate, which assumes knowledge of the distribution of generative factors in the data at hand $p(v_k)$. This is a mild assumption when the target factor of variation is simple, such as a categorical factor. However, generative factors underlying textual content are often ill-defined and difficult to encapsulate in a simple probability distribution. In fact, disentangled textual representations must account for meaning, and therefore require estimating semantic similarity for evaluation which is still an open problem. 

Textual disentanglement works often employ procedures that correlate representations with human semantic similarity judgement to report on semantic disentanglement.  Two examples of such procedures can be found in the work of \citet{Huang2021DisentanglingModelsb}. Namely, they measure correlation of embedding similarity with human similarity judgment, and train linear classifiers over their embeddings for paraphrase detection. 

Alternatively, works employing Autoencoding models for disentanglement can leverage outputs from the decoder to quantify disentanglement. More precisely, for an embedding $z=\{z_1, \dots, z_J\}$ corresponding to an observation $x$ for which we want to study the relation between a generative factor of interest $v_k(x)$ and portion of interest $z_p$ such that $p=\{j, \dots, j'\} \subset J$, a common procedure employing the decoder consists in resampling $z_p$, either from a prior $p(z_p)$ or from $q_\phi(z_p|x')$, the encoding of a distinct sample $x'$. The rest of the representation $z_{\bar{p}}$ is then concatenated to the newly sampled  ${z'}_p$ and decoded to produce a sample $x_1$. The analogous sampling and decoding procedure is also conducted in order to  generate $x_2$ with the original $z_p$ and a newly sampled ${z'}_{\bar{p}}$. Disentanglement is then studied by comparing $v_k(x)$ to $v_k(x_1)$ and $v_k(x_2)$. Given a similarity measure $sim_{v_k}$ over realizations of the generative factor $v_k$, one concludes that $v_k$ is disentangled in $z$ and represented by $z_p$ if, on average, $sim_{v_k}(v_k(x), v_k(x_2)) >> sim_{v_k}(v_k(x), v_k(x_1))$.

An instance of this \textit{perturb-and-measure} procedure can be found in \citet{Xu2019OnSupervision}. Conveniently this work provides 2 examples of similarity measures for assessing disentanglement: An exact match between the predictions of a pre-trained classifier for the sentiment factor, and a continuous score between 0 and 1 for content preservation (BLEU).

\section{Applications in NLP}
\label{DISENTAPPPBGSEC}
The main line of work in the area of disentanglement in NLP revolves around using multitask learning to separate concepts in neural representations. Works on unsupervised learning of disentangled representations remain rare compared to their supervised counterparts. Disentanglement without supervision on text data has first been explored in the work of \citet{Xu2020OnSupervision} where they proposed a method to improve meaning preservation and applied it to separate sentiment or topic information from content in sentence representation. Other examples of works addressing unsupervised disentanglement on text data include that of  \citet{Mercatali2021DisentanglingAutoencoders} who studied the use of discrete latent variables for such purpose, the work of \citet{Behjati2021InducingAttention} who worked on inducing morphemes-level representations using character-level Seq2Seq models, and \citet{Tjandra2021UnsupervisedRepresentation}'s work in which they disentangle \textit{speech} content features from its style features by leveraging the intuition that content is local and style is global. 

Works on \textit{supervised} disentanglement in NLP are however abundant. Looking at their respective contribution types, these works can be divided in two categories: works with task-specific contributions and works which present contributions that aim to improve disentanglement in the general case.  

Withing task-specific disentanglement, particular effort has been invested in separating semantics from form in general (meaning for example style) or syntax in particular. This is due to the fact that this separation proved useful for many applications, namely paraphrase detection~\cite{Chen2019ARepresentations, Bao2020, Huang2021DisentanglingModelsb} often accompanied by paraphrase generation~\cite{Romanov2019AdversarialRepresentation, chen-etal-2019-controllable, Zhang2020Syntax-infusedGeneration, Huang2021GeneratingPairs, Hosking2021FactorisingParaphrasing, Li2021ARepresentation, Hosking2022HierarchicalGeneration} as well as for Machine Reading Comprehension~\cite{Wu2022LearningComprehension}. Additionally, 
task-specific disentanglement has been used to enforce fairness by dissociating sensitive attributes from task relevant information in the prediction process~\cite{colombo-etal-2022-learning}. 

Concerning contribution that aim to improve disentanglement, methods generally fall within two categories: Methods that use Information Theoretic constraints to enforce dependence or independence between latent variables and generative factors~\cite{Cheng2020ImprovingGuidance, Mercatali2021DisentanglingAutoencoders, Colombo2021ARepresentations}, and Methods that use adversarial learning schemes to enforce separation between information in latent variables~\cite{Romanov2019AdversarialRepresentation, john-etal-2019-disentangled}.

Concerning the types of inductive biases discussed in section~\ref{DISENTINDUCSEC}, the vast majority of NLP works addressing disentanglement seem to employ supervised learning as inductive bias. Only a few attempts have been made at using unsupervised inductive bias, \textit{i.e.} using known characteristics of language to direct information towards specific parts of latent representations. For instance \citet{Chen2019ARepresentations}, \citet{Li2021ARepresentation} and \citet{Huang2021GeneratingPairs} discard word order information from semantic embedding modules in order to encourage syntactic embeddings to contain this information\footnote{This invariance of semantic representations to word order is, of course, not a viable design choice since it leads to sentences such as \textit{"John loves Mary."} and \textit{"Mary loves John."} to have the same semantic representations.}. 

\section{Conclusion}

This chapter served laying the groundwork for a major component of the contributions described in this thesis: Disentanglement. In Section~\ref{INDPLVDISENTBG}, after defining and motivating disentanglement, we explained how VAEs factor observations into independent generative factors, as dictated by the intuitive understanding of what disentanglement requires. To provide deeper understanding of how VAEs truly operate on data, and how that may relate to a set of target generative factors, we explained in Section~\ref{ALIGNDISENTBGSEC} Locatello's impossibility result regarding unsupervised disentanglement, and the relation put forward by Rolinek's work between VAEs and PCA.

Since VAEs \textit{only} factor observations into independent latent factors, we explained in Section~\ref{DISENTINDUCSEC} alternative inductive biases used in the literature to induce disentanglement. Moving on to more practical notions, we summarized in Section~\ref{DisentMeasSec} the different ways disentanglement can be measured and described MIG and MIG-sup to give examples, but also to provide better understanding of disentanglement metrics designed in this thesis (Chapter~\ref{chap:SynRoleDisentChap}) which were inspired by MIG.

Finally, we provided in Section~\ref{DISENTAPPPBGSEC} an overview of the current NLP disentanglement landscape with an emphasis on two important observations:
\begin{itemize}
    \item Unsupervised disentanglement in NLP is a research direction that needs more exploration.
    \item Most of the effort expanded towards disentanglement in NLP relies on labeled data, information theory-based or adversarial training-based objectives. In fact, little-to-no contributions in this area address the design of linguistically inspired built-in inductive bias for models to spontaneously induce (or help induce) disentanglement.
\end{itemize}

\chapter{Transformers and their NLP applications}
\label{TRANSCHAP}

Transformers have been introduced by \citet{Vaswani2017} as an architecture to process \textit{sets} of observations in general, or \textit{sequences} of observations in the particual case of linguistic observations. Contrary to RNNs, the \textit{de facto} language modeling architecture before Transformers, this alternative architecture computes a contextualized representation of each element of the input sequence using a \textit{parallelizable} computation method, as opposed to the recursive computations implemented within RNNs. The core component that enables this parallelizable contextualisation in Transformers is called \textit{attention} and has been used in Deep Learning way before the introduction of Transformers~\cite{Schmidhuber1991, Bahdanau2015NeuralTranslate, Luong2015EffectiveTranslation}.

Besides parallelizability, the quick adoption of attention was driven by the clear meaning behind its computation: Assigning a \textit{weight} to each element in a sequence in order to calibrate the extent to which it participates in the prediction at hand. Simply put, this mechanism calculates the degree to which \textit{attention} must be paid to each element of the sequence, hence its name. 
On top of performance gains often brought by attention, its clear meaning enables inspecting attention weights to understand the rationales behind predictions at test time. An instance where this feature can prove crucial is, for example, that of Deep Learning systems subject to strict fairness regulation such as the attention-based hireability prediction system of \citet{Felhi2019}.

Transformers were widely adopted after the introduction of the first Transformer-based MLM: Bidirectional Encoder Representations from Transformers (BERT; \citealp{devlin-etal-2019-bert}). The ensuing prolific research on Transformer-based language modeling can be described along three Major research directions: \begin{itemize}
    \item \textit{Pre-training and transfer}: This area pertains to self-supervised Transformer training schemes which yield models that perform well on tasks for which they are only fine-tuned with minimal data and computational resources.
    \item \textit{Language Modeling}: This area looks into the use of Transformers for causal (classical) language modeling, and takes special interest in scaling them to unusually large corpora with an unusually large number of parameters.
    \item \textit{Model Interpretability and Analysis}: This area looks into the inner-working of Transformer-based NLP systems with a special focus on BERT and its variants.
\end{itemize}

The disentanglement-related contributions we present in this thesis are all built upon Transformers attentions as an inductive bias. In that sense, we dedicate this chapter to providing in-depth explanations on attention and Transformers. We first describe the initial motivation behind attention, as well as the first formulations proposed to calculate it (\S~\ref{ATTBGSEC}). Moving on to the subject matter, we formally describe the Transformer architecture introduced by \citet{Vaswani2017} (\S~\ref{TRANSFORMERBGSEC}). The subsequent section (\S~\ref{TRANSFORMERLMBG}) consists in an overview of Transformer-based language modeling where we present MLMs such as BERT and its variants(\S~\ref{TRANSFORMERMLMBG}), prominent \textit{causal } LMs and VAE-based Transformer LMs(\S~\ref{TRANSFORMERCLMBG}), and a summary of findings from BERT-specific studies (a.k.a. \textit{Bertology}) in particular and Transformer-based LM analysis in general (\S~\ref{TRANSFORMERLMBGANAL})
.
\section{The Core Component of Tranformers: Attention Mechanisms}
\label{ATTBGSEC}
As explained above, the purpose of an attention mechanism is to produce scores that are used to weight the contribution of each sequence element in a sequence-level representation. Formally, a sequence of token-level representations $h=\{h_1, \dots, h_N\}$ can be aggregated to a sequence-level representation $\bar{h}$ using attention as follows:
\begin{align}
    \bar{h} =& \sum_{i=1}^N h_i f_i(h)\\
    \text{s.t.: }\hspace{0.5cm} f_i(h) =& \softmax{}_i(\{s(h_1, c), \dots, s(h_N, c)\})\\
    =&\frac{exp^{s(h_i, c)}}{ \sum_j exp^{s(h_j, c)} }
\end{align}
\noindent where $s$ is a scoring function that calculate the \textit{unnormalized} weight of each sequence element, and $c$ is a context element with regard to which attention is calculated. Across the literature, attention mechanisms mainly differ in the way $s$ and $c$ are formulated.

\paragraph{Attention before Transformers:}

In the pre-Transformer era, for applications such as sequence classification, $c$ is was usually posited to be a vector (or multiple vectors) of constant learnable parameters~\cite{Yang2016}. The intuition behind this design choice was to calculate \textit{relevance scores} with regard to the task at hand, which was constant and therefore represented by a fixed context vector $c$.

The more sophisticated uses of attention, which led to the Transformer architecture, arose from sequence-to-sequence prediction tasks such as Machine Translation~\cite{Bahdanau2015NeuralTranslate} or Speech Recognition~\cite{Chorowski2014}. In this setup, the informativeness of each element of the source sequence depends on the step reached in the target sequence generation. Accordingly, a context vector $c_j$ is calculated at each generation step $j$ as a function of the current state of the generator (or decoder). This use of attention in the sequence-to-sequence setup was introduced by \citet{Bahdanau2015NeuralTranslate}, and the benefit of it was twofold:
\begin{itemize}
    \item Seeking the most relevant source token to generate the current target token is similar to predicting \textit{alignments} between source and target tokens. Simultaneously learning to translate and align was actually the main motivation behind \citet{Bahdanau2015NeuralTranslate}'s pioneering use of attention for sequence-to-sequence transduction.
    \item Contrary to previous sequence-to-sequence models in the literature~\cite{Sutskever2014}, attention mechanisms allow calculating a representation of the source sequence that accommodates information pertaining each decoding step. In fact, the performance of previous models which used a single sequence representation for all generation steps was know to deteriorate for long sequences~\cite{cho-etal-2014-properties}.
\end{itemize}

Concerning the scoring function $s$, \citet{Bahdanau2015NeuralTranslate} used a feed forward neural network with a $\tanh$ activation function on the concatenation of context $c_j$, representing the current target generation step, and $h_i$  which is a representation of the $i^{th}$ source token:
\begin{align}
    s(h_i, c_j) = \tanh(W_s.\Concat(h_i, c_j)+b_s)
\end{align}
\noindent where $W_s$ and $b_s$ are respectively a matrix and a bias term with learnable parameters and $\Concat$ is the concatenation operator. This formulation of attention later came to be known as \textit{additive} attention. It was referred to as such in contrast to \textit{multiplicative} formulation of the score function developed by \citet{Luong2015EffectiveTranslation}. In this formulation attention is rather computed as a dot product between the current target context, and a linear transformation of the source token representation:

\begin{align}
    s(h_i, c_j) = c_j^T\cdot W_s\cdot h_i
\end{align}
\noindent where $W_s$ is a matrix with learnable parameters.

\begin{figure}[!h]
\centering
    \begin{minipage}[b]{1.0\textwidth}
            \centering
            \begin{minipage}[b]{\textwidth}
            \begin{adjustbox}{minipage=\textwidth,scale=1.0}
                \hskip 0cm \includegraphics{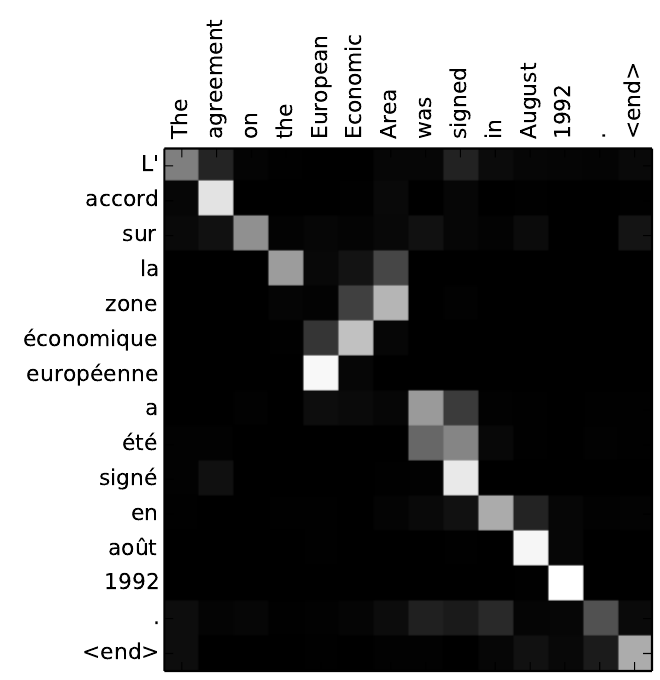}
                \hskip 0cm \includegraphics{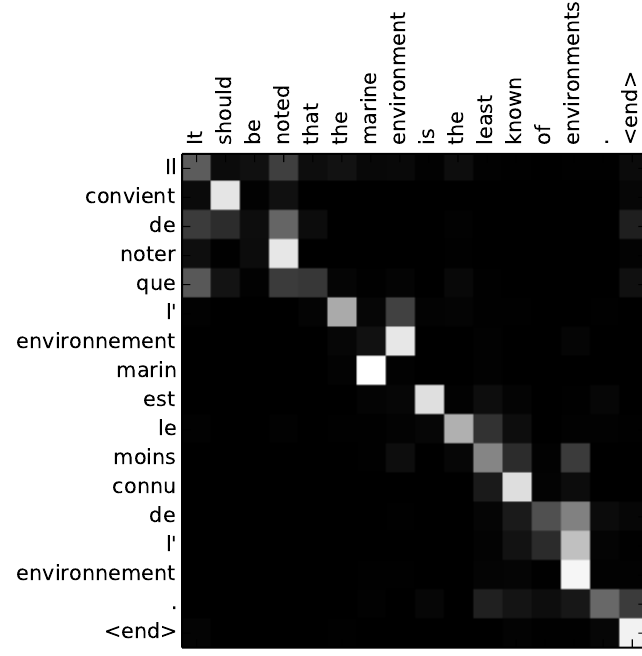}
            \end{adjustbox}
            \end{minipage}
            \caption{Two example alignments between source and target tokens induced by attention values in the work of \citet{Bahdanau2015NeuralTranslate}. The brightness of each cell is proportional to the value of $s(h_i, c_j)$, where the columns $i$ correspond to source tokens and the rows $j$ correspond to target tokens.}
            \label{fig:bahdanau_att}
    \end{minipage}
\end{figure}

As mentioned above, \citet{Bahdanau2015NeuralTranslate}'s use of attention was, in part, an attempt at learning alignments between source and target words in Machine Translation without using annotations on alignment. Their work, as well as subsequent works using attention, was successful at inducing such alignments and led to results such as the ones depicted in Figure~\ref{fig:bahdanau_att}.

Quantifying the influence of source tokens on target tokens using attention, as is done in Figure~\ref{fig:bahdanau_att}, is important technique which serves as a basis for quantitative evidence presented later in Chapter~\ref{chap:SynRoleDisentChap}.

\paragraph{Attention within Transformers:}

Transformers attention is multiplicative attention augmented with a few changes. A first change, we deem important, is a change in terminology. Although it was hinted prior to Transformer, that attention is a "\textit{soft}-search" module~\cite{Bahdanau2015NeuralTranslate}, this interpretation of its functioning scheme is explicited in Transformers terminology. In fact, source hidden states $h_i$ are said to formulate \textit{keys} ($K$) and \textit{values} ($V$), and target hidden states, previously referred to as contexts, are said to formulate \textit{queries} ($Q$) that call for the source hidden states according to their keys, and aggregate their values using the obtained weights. Formally, for a given query $Q_j$ and a set of source keys  $K=\{K_1, \dots, K_N\}$ and values $V=\{V_1, \dots, V_N\}$, the result of attention for the target with index $j$ is formulated as follows:
\begin{align}
    \Attention(Q_j, K, V) =& \sum_{i=1}^N f_i(Q_j, K) V_i \\
    \text{s.t. :}\hskip 0.5cm f_i(Q_j, K) =& \softmax{}_i(\frac{Q_j\cdot K^T}{\sqrt{d}}) 
\end{align}
\noindent where $d$ is the size of the representations $Q_j$ and $K_i$. Normalizing by the size of the representation is one of the changes brought by Transformers attention, and benefits the model with more numerical stability according to \citet{Vaswani2017}. When considering the full range of targets $Q=\{Q_1, \dots, Q_{|L|}\}$, Transformers attention can be abbreviated in its efficient matrix multiplication formulation as follows: 
\begin{align}
    \Attention(Q, K, V) = softmax(\frac{Q.K^T}{\sqrt{d}}).V
\end{align}
In the above, we insist on the importance of the change in terminology as it constitutes the basis of the intuition behind the last work described in this thesis (Chapter~\ref{chap:QKVCHAP}).

The last difference between Transformers attention and standard multiplicative attention is called \textit{Multi-Head Attention} (MHA). Specifically, in contrast to previous Neural Networks employing attention, Transformers utilize a multitude of attention mechanisms in parallel and concatenate their results:

 \begin{align}
     \MHA(\tilde{Q}, \tilde{K}, \tilde{V}) =& \Concat(head_1, ...head_H)W^O \hskip 2mm \\
    \hskip 2mm \text{s.t :}\hskip 1mm head_i =& \Attention(\tilde{Q}W_i^Q, \tilde{K}W_i^K, \tilde{V}W_i^V) \label{MHAInternals}
 \end{align}
 \noindent where $W_i^Q$, $W_i^K$, $W_i^V$, and $W^O$ are matrices with trainable parameters. 

\section{Transformer Architecture}
\label{TRANSFORMERBGSEC}
The Transformer architecture~\citep{Vaswani2017} has been proposed for sequence-to-sequence processing, and therefore consists of a sequence encoder and a sequence decoder. 

Throughout encoding and decoding, each $\MHA$ layer and each feed-forward layer are followed by a residual connection~\cite{He2016ResNet}, which is crucial for scaling to deeper networks, and  a Layer Normalization~\cite{Ba2016LayerN} which helps improve training stability and speed.  

\paragraph{Transformer encoders:}
A Transformer encoder uses only one sequence of input elements $S$, which is typically a source language sequence of tokens in Machine Translation. Attention is used in this encoder for the different elements of the input sequence to exchange information; \textit{i.e.} for contextualization. Specifically, Multi-Head Attention internally calculates its queries, keys, and values (\textit{cf.} Eq. \ref{MHAInternals}) from this single input sequence, and thus produces contextualized sequence elements.
This single input Multi-Head Attention is called Self-Attention ($\SA$):
 \begin{align}
  \SA(S) = \MHA(S, S, S)
 \end{align}
 
 Given this operator, a Transformer encoder with $D^{enc}$ layers can be formulated as follows\footnote{Here, we omit layer normalization and residual connections to lighten these equations, although they are essential to learn Transformers reliably~\cite{He2016ResNet, Ba2016LayerN}.}:
 
  \[
    \TransEnc(S) =\tilde{S}_{D^{enc}}, \ \text{s.t. } \tilde{S}_{d} =
    \left \{
      \begin{array}{l}
        S \text{ if } d=0\\
        \FF_d(\SA_d(\tilde{S}_{d-1})) \text{ if } d>0
      \end{array}
      \right.
  \]
 \noindent where each $\FF_d$ is a feed-forward neural network. Feed-forward neural networks in Transformers are formulated as two linear transformations with  a Rectified Linear Unit (ReLU) activation in between\footnote{The ReLU activation is the function $f: x \longrightarrow \max(0, x)$. Subsequent Transformer-based models such as GPT~\cite{Radford2018ImprovingLU} and BERT~\cite{devlin-etal-2019-bert} replaced this function by a Gaussian Error Linear Unit (GELU) activation function, which is a continuous version of ReLU introduced by  \citet{hendrycks2016gelu} who show that it empirically outperforms ReLU across various NLP tasks, among others.}:

 \begin{align}
      \FF(x) = \max(0, x.W_1 + b_1).W_2 + b_2 \label{TRANSFFEQ}
 \end{align}

\noindent where $W_1$ and $W_2$ are learnable parameter matrices, and $b_1$ and $b_2$ are learnable bias terms.

\paragraph{Transformer decoders:}
A transformer decoder takes as an input a source sequence of elements $S$, and a target sequence of elements $T$. In the decoder, attention is used in two manners. The first is an $\SA$ module that exchanges information between elements in the target sequence. The second is a \textit{Cross-Attention} ($\CA$) module which \textit{pulls} information from source \textit{values} through their \textit{keys} with target \textit{queries}:

 \begin{align}
  \CA(T, S) = \MHA(T, S, S)
 \end{align}

A Transformer decoder with  $D^{dec}$ layers can be formulated as follows\footnote{Layer Normalization and Residual connections are also omitted here.}:

  \[
    \TransDec(T, S) =\tilde{T}_{D^{dec}}, \text{s.t. }:\ 
    \]
  \[ \tilde{T}_{d}  = 
    \left \{
      \begin{array}{l}
       T \text{ if } d=0\\
       \FF(\CA(\SA(\tilde{T}_{d-1}),S)) \text{ if } d>0
      \end{array}
      \right.
  \]

\paragraph{NLP-specific processing in Transformers:}\hspace{-0.4cm}\footnote{We refer to the implementation details described here as \textit{NLP-specific} since they were originally engineered for textual inputs. Nevertheless, they remain applicable to any discrete times series, and the positional encoding, described later in this section, remain applicable to any type of ordered input sets (\textit{e.g.} image patches; \citealp{NEURIPS2019_c74d97b0}).}
To process textual sequences with Transformers, textual tokens must be embedded and their positional information must be encoded withing their embedding vectors. The token embedding step simply employs an embedding lookup layer, which is a standard component in Deep Learning-based NLP. However, contrary to previous sequence processing architectures such as LSTMs~\cite{Hochreiter1997LongMemory}, Transformers are order-invariant, and therefore require engineering a mechanism to encode word order information. \citet{Vaswani2017} chose a mechanism,  which they called \textit{positional encoding}\footnote{This is neither the first, nor the best attempt at encoding positions. For instance, \citet{Gehring2017} train a separate embedding for each position while experimenting with attention-augmented Convolutional Neural Networks for sequence-to-sequence modeling, and \citet{shaw-etal-2018-self} show that embedding relative, as opposed to absolute, positions improves results for Machine Translation. }, that consists in injecting order information by adding the amplitudes of \textit{sine} and \textit{cosine} functions applied to different position-dependant frequencies:
\begin{align}
    \PE{}_{pos, 2i} = \sin(\frac{pos}{10000^{\frac{2i}{d_{model}}}})\\
    \PE{}_{pos, 2i+1} = \cos(\frac{pos}{10000^{\frac{2i}{d_{model}}}})
\end{align}
\noindent where $pos$ is the position of the token, and $i$ is the dimension at which addition is being performed in the token embedding vector. This positional encoding operation is applied to both source and target input tokens. The intuition behind \citet{Vaswani2017} choosing sinusoidal functions is that the model should better attend to relative positions since, for any fixed offset $k$, $PE_{pos+k}$ is a linear function of $PE_{pos}$.

To model textual sequences in an auto-regressive manner, Transformers are also required to restrict the information exchange between target tokens so that target sequence elements only pull information from previous tokens. In practice, this is enforced through a binary mask $M$ on attention, leading to what is called \textit{masked attention}:
\begin{align}
    \Attention(Q, K, V) =& (\softmax(QK^T)\odot M)V\label{MaskedAttEq}\\
    \text{s.t. : \hskip 0.5 cm} M_{ij} =& \left \{
      \begin{array}{l}
       1 \text{ if } i\leq j\\
       0 \text{ if } i > j\\
      \end{array}
      \right.
\end{align}

This masked attention is typically used in Transformer decoders for SA between the target elements. 

Given the above adaptations, NLP-specific Transformers only further require a linear layer with softmax activations in order to produce word probabilities for sequence modeling. 

\paragraph{The Transformer architecture for textual sequence-to-sequence modeling:}

\begin{figure}[!h]
\centering
    \begin{minipage}[b]{1.0\textwidth}
            \centering
            \begin{minipage}[b]{\textwidth}
            \begin{adjustbox}{minipage=\textwidth,scale=0.48}
                \hskip 0cm \includegraphics{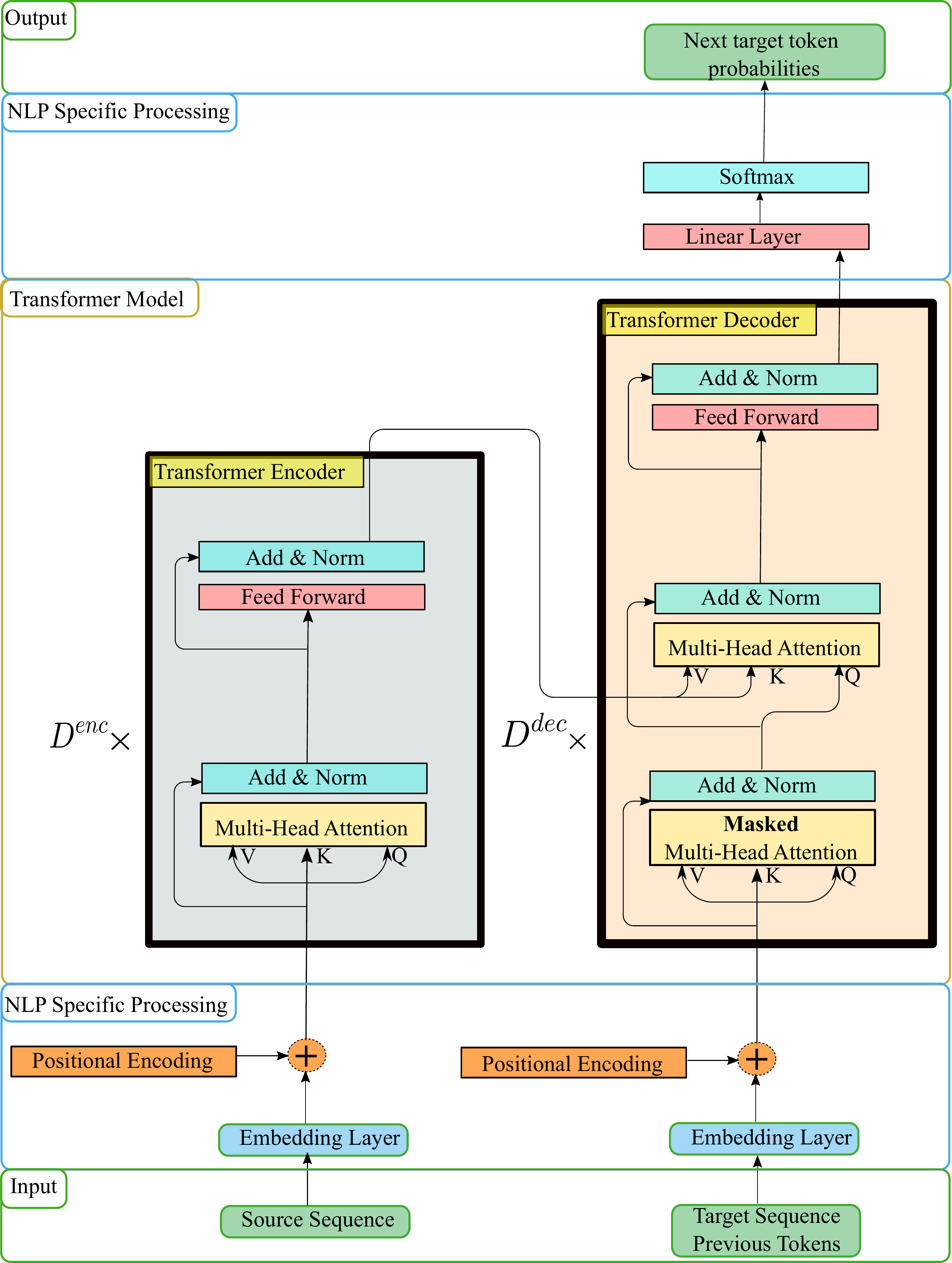}
            \end{adjustbox}
            \end{minipage}
            \caption{The Transformer architecture for textual sequence to sequence modeling. The encoder and decoder are recursively applied $D_{enc}$ and $D_{dec}$ times respectively. As emphasized in the figure, the middle part \textit{"Transformer Model"} constitutes the core Transformer architecture, which is input agnostic, while the parts above and below this core part represent NLP-specific adaptations.}
            \label{fig:TransFigBG}
    \end{minipage}
\end{figure}

Figure~\ref{fig:TransFigBG} displays how the Transformer encoder and decoder are assembled with NLP-specific processing steps for the textual sequence-to-sequence modeling in the work of \citet{Vaswani2017}. 

\section{Language Modeling with Transformers}
\label{TRANSFORMERLMBG}
In this section, we go over recent Transformer-based trends in NLP research, namely through the 3 axes along which they can be categorized: 
masked language modeling and pre-training, causal language modeling, and LM Analysis. 

\subsection{Masked Language Modeling and Pre-training with Transformers}
\label{TRANSFORMERMLMBG}
The flagship of MLMs is the very first model to use masked language modeling: BERT~\cite{devlin-etal-2019-bert}. This model encodes tokens using only the Transformer encoder from the architecture depicted in Figure~\ref{fig:TransFigBG}. BERT is trained to encode consecutive sentences $A=\{a_1, \dots, a_{|A|}\}$ and $B=\{b_{1}, \dots, b_{|B|}\}$ tokenized with a WordPiece Tokenizer~\cite{Schuster2012}. These sentences are provided to the model under the form \\$\{[CLS], a_{1}, \dots, a_{|A|}, [SEP], b_{1}, \dots, b_{|B|}\}$\footnote{Notice that BERT does not use and end-of-sentence token, since it does not perform language generation.} where $[CLS]$ and $[SEP]$ are special tokens. Apart from summing token embeddings with their positional encodings, BERT also sums them with a an additional segment embedding, where the segment is either $A$ or $B$.

Pre-training BERT consists in optimizing it for two tasks on 16 GB of text data from BookCorpus~\cite{Zhu2015Book} and english Wikipedia. The first task is masked language modeling as described in Section~\ref{MLMDEFSEC} (Eq.~\ref{MLMObj}). The second is a Next Sentence Prediction (NSP) task, where $C$, the output representation of the $[CLS]$ token, is fed to a binary classifier which predicts whether sentence B is the sentence that follows $A$ in the corpus, or a randomly sampled sentence from this corpus. 

After pre-training, BERT is \textit{finetuned} on tasks from the General Language Understanding Evaluation suite (GLUE; \citealp{wang-etal-2018-glue}). BERT outperformed the previous state-of-the-art on the GLUE benchmark by a large margin, which opened an avenue for research on Transformer-based Pre-trained LMs.

 Numerous works after BERT succeeded in establishing new state-of-the-art performance on language understanding using variants of its pre-training scheme. \citet{Yang2019XLNet:Understanding} use a Transformer architecture optimized for longer sequences~\cite{Dai2020Transformer-XL:Context}, and reformulate the masked language modeling objective as an auto-regressive language modeling objective that goes through tokens in a randomized order.  \citet{Liu2019RoBERTa:Approach} find that training BERT longer, with larger batches, more data (160 Gb) and without the NSP objective substantially improves performance on downstream tasks.

Besides the above works, a number of studies explored modifying BERT while aiming for improvements other than better downstream task transfer. For instance, ALBERT~\cite{Lan2020ALBERT} outperforms BERT on language understanding benchmarks with less parameters by using a single layer that applies to the inputs multiple times in order to simulate a multi-layer Transformer. Also on the parameter efficient side, \citet{Sanh2019DistilBERTAD} distill the original BERT to 60\% of its original size while retaining 97\% of its original performance. To improve training speed, \citet{clark2020electric} train a Transformer encoder (ELECTRA) to distinguish real tokens from fake tokens, rather than to generate tokens in masked placeholders, and achieve performance similar to that of \citet{Liu2019RoBERTa:Approach}'s approach while using only $\sim 22$\% of their compute\footnote{Compute, here, was measured in Floating Point Operations (FLOPs).}.

Pre-trained Transformers have also been proven effective outside of BERT's \textit{token embedding} paradigm, namely in the sequence-to-sequence setup~\cite{Raffel2020t5, Lewis2020BART:Comprehension}, and the multimodal (language and vision) setup~\cite{Lu2019Vilbert,Li2021ALBEF, huang2021seeing}.

\subsection{Causal Language Modeling with Transformers}
\label{TRANSFORMERCLMBG}
In addition to lower perplexity scores, the ability of Transformers to scale to high volumes of data benefited causal language modeling in that it enabled verifying a crucially important hypothesis for the future of AI as a whole that we discuss in what follows: "\textit{Language models are unsupervised multitask learners"}~\cite{Radford2018LanguageLearners}\footnote{The quoted sentence is actually the title of Radford's paper}.

Through the relatively large-scale\footnote{GPT-2 used 40Gb of text for its training which, at the time, was considered "large-scale".} LM GPT-2, \citet{Radford2018LanguageLearners} showed that language modeling is a task that involves learning, and therefore requires solving, multiple non-trivial reasoning subtasks for which there are no labels in the data apart from the text itself, hence the \textit{unsupervised multitask learning} mentioned by Radford. For instance, a large-scale LM is likely to encounter many times the phrase "TL;DR"\footnote{This phrase is internet lingo for "Too Long; Didn't Read". Although the phrase was originally meant to be commented on \textit{long} posts, it came to be used as a marker for small abstracts accompanying long internet posts that may be skipped by readers due to their length.}, which is a cue signaling the presence of a small summary. Consequently, one can probe the ability of an LM to summarize by prompting it with a text appended with the phrase "TL;DR", and scoring the completion generated by the LM with regard to a gold standard summary. Similar prompting strategies can be developed for a number of tasks like translation (\textit{e.g.} \textit{<French sentence>}'s translation in English is \textit{<LM completion>}), reading comprehension (often investigated with conversational prompts), or even basic arithmetics (\textit{e.g.} \textit{<x> times <y> is <LM completion>}). \citet{Radford2018LanguageLearners} succeed in showing that LMs learn these tasks underlying language modeling by demonstrating the ability of GPT-2 to score at least significantly above random chance at the aforementioned tasks.

The recent years have seen the emergence of Transformer-based LMs larger than GPT-2 by multiple orders of magnitude. The first instance of such models, which was GPT-3~\cite{Brown2020LanguageLearners}, consisted of 175 billion parameters (approximately 116 times the size of GPT-2) and was trained on 540 Gb of data (approximately 11 times the data used for GPT-2). GPT3 demonstrated strong few-shot learning performance through pure prompting on a wide range of tasks, and even generated short news articles ($\sim$200 words) which were nearly unidentifiable\footnote{When asked to identify the model-generated article between a GPT-3 article and a human-generated article, human operators had a 52\% accuracy with a confidence interval of 49\%-54\%. Random chance accuracy (50\%) being in this interval, the experiment could not conclude that humans were capable of discerning GPT-3 articles from human articles.} by humans.  Extremely large LMs subsequently started emerging as a research direction primarily fueled by industrial fundings~\cite{rae2021scaling, Chowdhery2022Palm, Smith2022Megatron, du2022glam, thoppilan2022lamda} with a specific focus on how they can solve tasks through prompting with few-to-no training steps~\cite{su-etal-2022-transferability}.

Due to the difficulties incurred by posterior collapse, Transformers have rarely been trained as VAEs for language modeling, and latent variable LMs have therefore been relatively left out of the large-scale language modeling race. To the best of our knowledge, Optimus~\cite{Li2020Optimus:Space} and DELLA~\cite{hu-etal-2022-fuse} constitue the only attempts at building large-scale VAE LMs. Optimus uses BERT as an encoder and GPT-2 as a decoder to demonstrate that Transformer-based VAE LMs can outperform their standard Transformer-based LM counterparts when similar effort is expanded to train them. But seeing that Optimus still suffers from posterior collapse to a certain extent, \citet{hu-etal-2022-fuse} proposes through DELLA a hierarchical latent variable modeling scheme which they prove to be effective in dealing with posterior collapse to a large extent. Given that DELLA is very recent, its potential effect on the language modeling landscape is yet to be seen.

\subsection{Bertology and Transformer-Based Model Analysis}
\label{TRANSFORMERLMBGANAL}
The previous sections enumerated the successes Transformers have had across various NLP tasks, which necessarily raises the question: How do they do it ? A flurry of studies, following the pioneering work of \citet{devlin-etal-2019-bert}, have dived into the inner working of Transformers in order to make sense out of the information flow in this architecture, and how it relates to what we know about linguistics. A digest of the BERT-specific portion of these works (a.k.a \textit{Bertology}) can be found in \citet{rogers-etal-2020-primer}.

The first series of works on the subject tackled identifying the content of each layer through \textit{probing}, a technique that consists in training a classifier on top of a representation and measuring the performance of the classifier with the intuition that the higher it is, the more informative the representation is about the target concept. When applying this methodology to BERT, \citet{Tenney2020BERTPipeline} observed that it tends to process information in a manner that bears resemblance to the classical NLP pipeline, meaning that the first layers tend to encode low-level surface form information such as PoS Tags, the middle layers appear to be informative on more advanced syntactic relations such as dependencies, while semantic information is either in the final layers (\textit{e.g.} for co-reference resolution) or spread out throughout the entire network as is the case for relation classification. These findings were corroborated by later studies on BERT~\cite{Jawahar2019, hewitt-manning-2019-structural}, but do not seem to generalize to all BERT-like models. As a matter of fact, \citet{fayyaz-etal-2021-models} have shown that linguistic information emerges earlier compared to BERT in the layers of XL-Net~\cite{Yang2019XLNet:Understanding}, and later in the layers of ELECTRA~\cite{clark2020electric}. 

In order to better understand the way Transformers encode linguistic information, a number of works have focused on the extent to which these models grasp syntactic notions. To that end, \citet{Marvin2020TargetedModels} developed a test suite with minimally modified pairs designed to assess specific syntactic capabilities in LMs such as subject/verb agreements and  correct selection of negative polarity items. By compiling this evaluation suite together with similar syntactic and psycholinguitic evaluation toolsets~\cite{wilcox-etal-2018-rnn, Futrell2018, wilcox-etal-2019-structural}, \citet{Hu2020AModels} compared a range of popular architectures for language modeling and observed that the syntactic capabilites of the different models they tested \textit{was not} correlated with their respective perplexities, meaning that advances in language modeling measured by perplexity do not translate to models that better understand the syntax underlying their training corpora. Additionally, their experiments revealed that syntactic capabilities were more influenced by architectures than they were by the size of the training corpora, with the Transformer architecture tested with GPT-2 coming on top of previous state-of-the art RNN-based architectures such as RNNG~\cite{Dyer2016RecurrentGrammars} and ON-LSTM~\cite{Shen2019OrderedNetworks}. Subsequent studies on the presence of syntactic information in Transformers have backed and better justified the ability of Transformers to process syntactic information by showing that Transformers encode information in tree-like structures~\cite{Jawahar2019, hewitt-manning-2019-structural}. 

Concerning the role played by different Transformer components in formulating output representations, \citet{geva-etal-2021-transformer} proposed and leveraged an interesting perspective on feed-forward layers in Transformers to provide a unified view on information processing in Transformers: the two matrices forming these layers  ($W_1$ and $W_2$ in Eq.~\ref{TRANSFFEQ}) function as keys and values where rows from the first matrix are used to weight (or call) lines from the second matrix. Using this insight they show that the values called by keys in these matrices correspond to human-understandable textual patterns. Then, In \citet{Geva2022FFVoc}, the same authors show that these feed-forward layers operate on outputs by gradually promoting concepts, which are also largely understandable, in their outputs while forming the word representations. Also pertaining to the association between Transformer components and understandable concepts, \citet{clark-etal-2019-bert} have shown that many attention heads in BERT emulate identifiable syntactic functions such as dependency relations and co-references with high accuracy. They also show that BERT attention heads learn to default to special tokens ([CLS] and [SEP]) when they can't find the target argument for their specific linguistic functions. 

It is worth noting that, alongside a better understanding of Transformers, this line of works also led to numerous advances in methodology concerning, for instance,
the use of attention as an explanation~\cite{Jain2019AttentionExplanation,Wiegreffe2020AttentionExplanation}, the validity
of probing~\cite{Pimentel2020Information-TheoreticStructure}, or contrastive evaluation with minimal pairs~\cite{Kodner2020OverestimationModels, Vamvas2021OnEvaluation}.

\section{Conclusion}
This background chapter is aimed at explaining crucial architectural components for our contributions on disentanglement: Transformers and attention.
We first motivated attention mechanisms and introduced the first attempts at formulating them (additive and multiplicative), as well as the first alignment results they exhibited in Machine Translationng (\S~\ref{ATTBGSEC}). We then showed in the same section how Transformers normalized multiplicative attention, and duplicated it into Multi-Head Attention to form their basic building block. Given this attention mechanism, we explained in section~\ref{TRANSFORMERBGSEC} how Transformers exchange information between sequence elements (\textit{i.e.} Self-Attention) or pull information from source sequences to target sequences (\textit{i.e} Cross-Attention). In the same section, we also explained how these attention blocks, together with other Transformer components come together to form the complete multi-layer Transformer architecture, and how NLP-specific measures, such as positional encoding and masked attention, must be applied in order to deal with textual inputs and outputs.

Moving to their applications, we painted in Section~\ref{TRANSFORMERLMBG} a picture of the current landscape of Transformer-based works in NLP. Specifically, we descibed the first MLM, BERT, and gave a summary of the subsequent models that modify and improve upon it (\S~\ref{TRANSFORMERMLMBG}). In the following subsection (\S~\ref{TRANSFORMERCLMBG}), we outlined works on large-scale language modeling with Transformers and how they exhibited the ability of LMs to solve a range of reasoning tasks underlying simple text completion. We also provide a summary of the few works attempting large-scale VAE-based language modeling with Transformers.
Finally, we give an overview of the different conclusions drawn by works attempting to understand the inner-working of Transformer-based LMs (\S~\ref{TRANSFORMERLMBGANAL}).

\chapter{Syntactic Structure of Sentences}
\label{SYNBGCHAP}

The approach to interpretable machine learning in the context of NLP is peculiar in that language, the input data, is a media that has been subjected to meticulous scrutiny for millennia\footnote{Dependency grammars for instance, approached in Section~\ref{DEPBGSEC}, have first been formulated by the Indian scholar Panini around the $5^{th}$ century B.C.E.~\cite{joshi1991a}.
 }. The myriad works investigating and theorizing about language provide solid foundations for interpretable NLP to lean on. As briefly discussed in the previous chapter (\S~\ref{TRANSFORMERLMBGANAL}), works analyzing the behavior of Transformers are largely built on relations to linguistic notions~\cite{hewitt-manning-2019-structural, Tenney2020BERTPipeline, Hu2020AModels}. Similarly, we build in this thesis interpretable representation learning techniques for NLP which are largely motivated by their meaningfulness with regard to linguistics in general, and \textit{syntax} in particular.
 
 Syntax is the area of linguistics that looks into the way words combine in order to produce phrases or sentence that abide by the grammar of a language. Although theoretical frameworks formalizing this set of rules may vary, it is generally agreed upon that words in a sentence can be embedded in a tree-like structure describing its syntax, where these words combine into phrases.  Figure~\ref{fig:BGSYNALL}\footnote{These dependency relations can be obtained using Spacy's online dependency parser which can be used at this URL: \href{https://explosion.ai/demos/displacy}{https://explosion.ai/demos/displacy}} is a minimal example illustrating this combination process through the dependency tree of a sentence. 

\begin{figure*}[!h]
\centering
    \begin{minipage}[b]{0.48\linewidth}
    \begin{adjustbox}{minipage=\linewidth,scale=1.2}
        \begin{dependency}[theme = default]
           \begin{deptext}[column sep=1em, row sep=.1ex]
              A \& talented \& musician \& holds \& his \& nice \& guitar \\
           \end{deptext}
           \deproot[
            ]{4}{ROOT}
           \depedge[
            ]{4}{3}{nsubj}
           \depedge[
            ]{4}{7}{dobj}
           \depedge{3}{2}{amod}
           \depedge{3}{1}{det}
           \depedge{7}{6}{amod}
           \depedge{7}{5}{poss}
        \end{dependency}
            \end{adjustbox}
    \end{minipage}
    \begin{minipage}[b]{0.48\linewidth}
    \begin{adjustbox}{minipage=\linewidth,scale=0.9}
        \begin{tikzpicture}
            
        \end{tikzpicture}
    \end{adjustbox}
    \end{minipage}

    \caption{A sentence and its dependency structure. The structure is built as a tree over words in the sentence, where arrows go from \textit{heads} (or \textit{parents}) to \textit{dependents} (or \textit{children}).}
    \label{fig:BGSYNIntro}
    
\end{figure*}
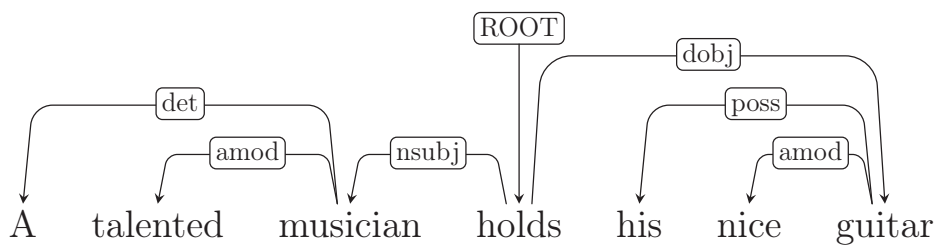

Beyond enumerating valid constructions in a certain language, syntax reflects the way minimal semantic units are \textit{composed} to convey complex meaning. For instance, dependency relations illustrate the compositional nature of language and the role of syntax in this composition by relating words through \textit{head-to-dependent} relations. These relations model the way words in sentences \textit{feed} off the meaning of other words in a hierarchical modification process that yields the overall meaning of the composed phrase or sentence \citep[Chapter~21]{Tesniere1959}. With that in mind, recall that the core component of Transformers, discussed in the previous chapter (\S~\ref{ATTBGSEC}), is attention: a mechanism that organizes the exchange of information between token representations, \textit{i.e.} the way representations are \textit{composed}. Understanding the relation between the way attention behaves and the linguistic theory describing the rules for phrase and sentence composition, \textit{i.e.} syntax, is therefore crucial in making sense of NLP systems in the Transformers era.

In light of the above, we dedicate this chapter to presenting the syntactic notions which inspired our contributions, and provided grounding for the evaluation protocols used to validate them. First, we introduce \textit{constituency analysis} (\S~\ref{CONSTBGSEC}) and its backbone, Context-Free Grammars(CFGs). Subsequently, we present an alternative to the constituency view on syntax: \textit{dependency analysis} (\S~\ref{DEPBGSEC}). This second section also emphasizes the strong relation of dependencies to semantics and predicate-argument structures. 

\section{Constituency Analysis}
\label{CONSTBGSEC}  
 The constituents, referred to in \textit{constituency analysis}, are groups of words that can behave as a single unit(\textit{cf.} \citealp[Chapter~4]{Chomsky1957} and \citealp[Chapter~12]{Jurafsky2022spl3}). In other words, constituency analysis is a description of:\begin{itemize}
     \item the way words form constituents, \textit{i.e.} multi-word units.
     \item the \textit{type} of each of these constituents.
 \end{itemize}
 Classifying constituents into different types is useful for describing rules governing word grouping, \textit{e.g.} A \textit{determiner} ($DT$) and a \textit{noun} ($NN$) form a \textit{Nominal Phrase} ($NP$) in English, which in turn may be followed by a \textit{Verb Phrase} ($VP$) to form a sentence ($S$). The set of such \textit{production rules} for a certain language can be formalized through what is called a Context-Free Grammar\footnote{Note that CFGs are simplified models for the syntax of natural languages. In fact, as shown by \citet{shieber1985evidence}, some syntactic phenomena exhibited by natural languages fall outside of the range of CFGs.} (CFG). For instance, the aforementioned production rules for English would appear in a CFG as follows:
 \begin{align}
    NP &\longrightarrow DT~NN\\
    S &\longrightarrow NP~VP
 \end{align}
 The first line from the above rules means "a nominal phrase may be formed by a determiner and a noun (in that order)". Symbols used in production rules belong to two categories: \textit{terminal} and \textit{non-terminal} symbols. terminal symbols are the words that actually appear in sentences (\textit{e.g.} car, he, draw, ...) and non-terminal symbols are the abstractions (or types) that may produce words or other non-terminals (\textit{e.g.} noun, preposition, verb phrase, ...). The subset of non-terminals that produce terminals, such as \textit{determiners} ($DT$) and nouns ($NN$), are called Parts-of-Speech (PoS). With the rules defined over terminal and non-terminal nodes, the set of possible strings of terminals that can be generated by a CFG from the starting symbol $S$ forms a formal language. Conversely, sentences that can not be arrived at from the rules within a CFG are said to be \textit{ungrammatical} with regard to that CFG.
 
\begin{figure*}[!b]
\centering
            \begin{minipage}[b]{\textwidth}
            \begin{adjustbox}{minipage=\textwidth,scale=0.26}
                \includegraphics{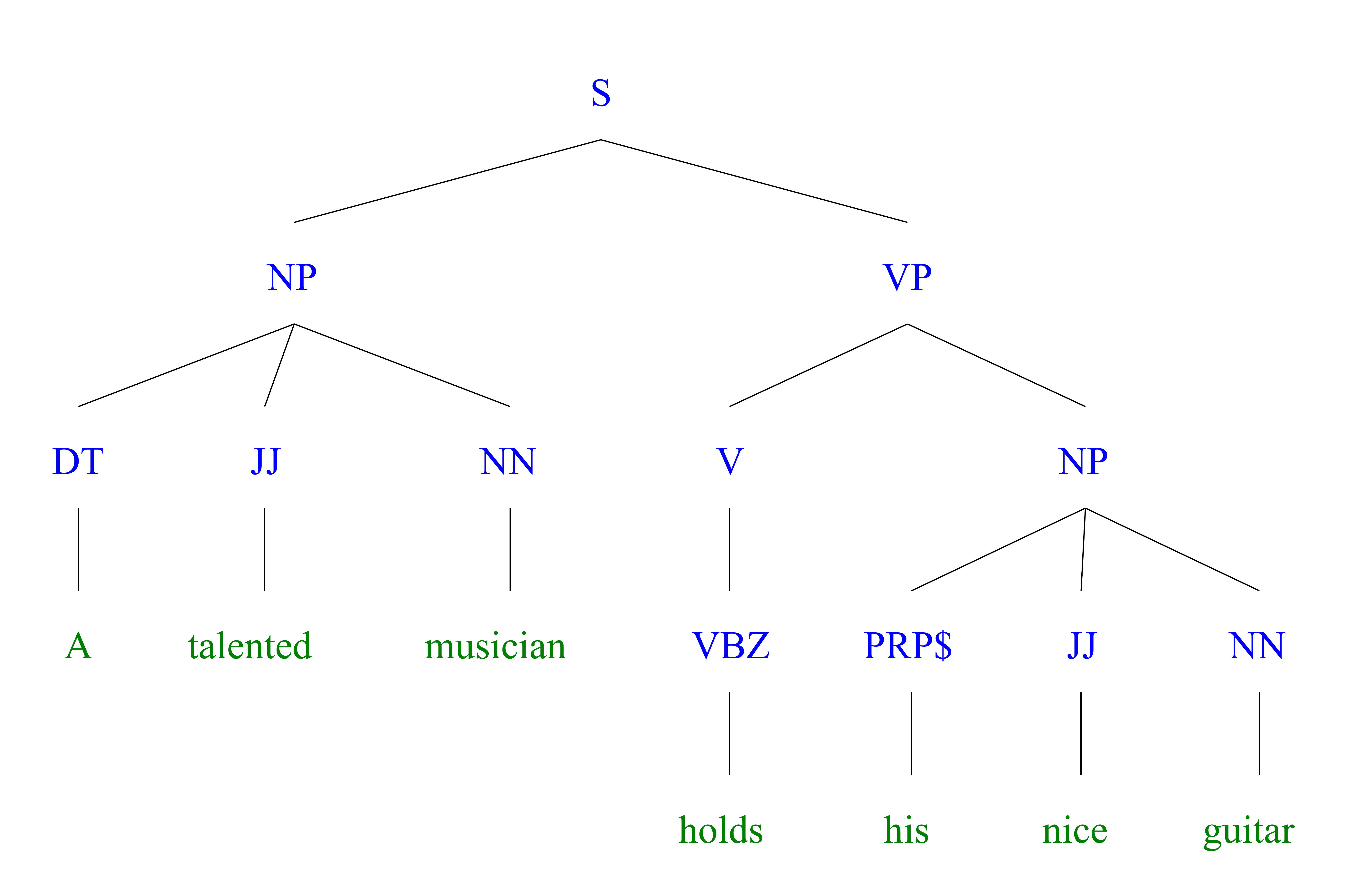}
            \end{adjustbox}
            \end{minipage}
            \caption{Example constituency tree for the sentence "A talented musician holds his nice guitar". Terminal nodes are labeled in \textcolor{editorGreen}{green} while non-terminal nodes are labeled in \textcolor{blue}{blue}. $JJ$, $VBZ$ and $PRP$  are respectively "adjective, numeral or ordinal", "verb, present tense, 3rd person singular" and "pronoun, personal".}
            \label{fig:CONSTIT}
\end{figure*}
 
 Given these rules, constituency trees are constructions over sentences that depart from  the starting symbol node $S$ and use production rules from the CFG of the language at hand to arrive at the leaf nodes containing the terminal symbols in the sentence. An example of such construction can be seen in Figure~\ref{fig:CONSTIT}\footnote{This constituency tree was obtained through AllenAI's online NLP demo platform available at this URL: \href{https://demo.allennlp.org/constituency-parsing}{https://demo.allennlp.org/constituency-parsing}}.
 
\section{Dependency Analysis}
\label{DEPBGSEC}
As discussed in the introduction, dependency analysis illustrates sentence structure through syntactic relations between words~\citep{mel1988dependency}. These relations form a tree where each node is a word, in contrast to the constituency tree where words are only leaf-nodes.

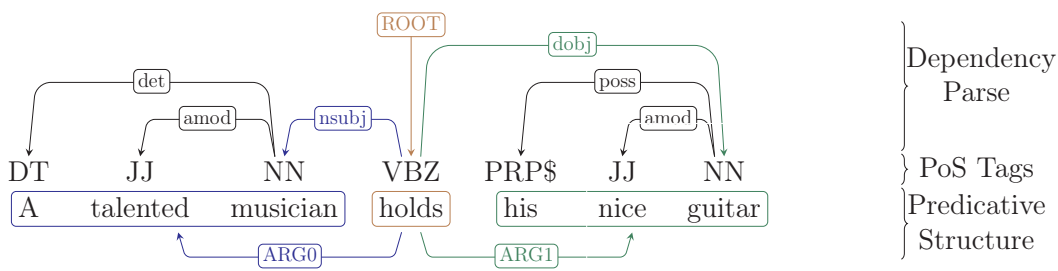
\begin{figure*}[!h]
\centering
    \begin{minipage}[b]{0.48\linewidth}
    \begin{adjustbox}{minipage=\linewidth,scale=0.9}
        \begin{dependency}[theme = default]
           \begin{deptext}[column sep=1em, row sep=.1ex]
               DT  \&JJ  \&NN  \&VBZ  \&PRP\$  \&JJ  \&NN \\
              A \& talented \& musician \& holds \& his \& nice \& guitar \\
           \end{deptext}
           \deproot[edge style={orange!70!black},
            label style={orange!70!black, fill=white, text=orange!70!black}]{4}{ROOT}
           \depedge[edge style={blue},
            label style={blue, fill=white, text=blue}]{4}{3}{nsubj}
           \depedge[edge style={green!60!black},
            label style={green!60!black, fill=white, text=green!60!black}]{4}{7}{dobj}
           \depedge{3}{2}{amod}
           \depedge{3}{1}{det}
           \depedge{7}{6}{amod}
           \depedge{7}{5}{poss}
           \wordgroup[style={orange!70!black, fill=white}]{2}{4}{4}{pred}
           \wordgroup[style={blue, fill=white}]{2}{1}{3}{a0}
           \wordgroup[style={green!60!black, fill=white}]{2}{5}{7}{a1}
           \groupedge[edge style={blue},
            label style={blue, fill=white, text=blue}, edge below]{pred}{a0}{ARG0}{2ex} 
           \groupedge[edge style={green!60!black},
            label style={green!60!black, fill=white, text=green!60!black}, edge below]{pred}{a1}{ARG1}{2ex} 
        \end{dependency}
            \end{adjustbox}
    \end{minipage}
    \begin{minipage}[b]{0.48\linewidth}
    \begin{adjustbox}{minipage=\linewidth,scale=0.9}
        \begin{tikzpicture}
            \draw [color=white](0,0) rectangle (3,2);
            \draw[decorate,decoration={brace,raise=0.1cm}]
            (4.5,3.5) -- (4.5,1.6) node[align=center,above=0.2cm,pos=0.8, xshift=1.2cm] {~Dependency \\Parse};
            
            \draw[decorate,decoration={brace,raise=0.1cm}]
            (4.5,1.55) -- (4.5,1.1) node[align=center,above=-0.2cm,pos=0.8, xshift=1.2cm] {PoS Tags};
            \draw[decorate,decoration={brace,raise=0.1cm}]
            (4.5,1.05) -- (4.5,0.0) node[align=center,above=-0.2cm,pos=0.8, xshift=1.2cm] {Predicative\\ Structure};
        \end{tikzpicture}
    \end{adjustbox}
    \end{minipage}

    \caption{A sentence and its syntactic roles. The correspondence between syntactic roles and elements of the predicative structure is highlighted with colors. $ROOT$ is the root node of the dependency tree, $nsubj$ is a nominal subject, $dobj$ is a direct object,$det$ is a determiner, $poss$ is a possessive determiner, and $amod$ is an adjectival modifier. $ARG0$ and $ARG1$ are semantic \textit{proto-roles}~\cite{bonial2012english} designating respectively the entity that performs the action (a.k.a the \textit{agent}) and the entity affected by the action (a.k.a the \textit{patient}).}
    \label{fig:BGSYNALL}
\end{figure*}

The words forming subtrees in a dependency tree (\textit{i.e.} a node and all its descendants) correspond to words that form sentence constituents from the constituency perspective~\citep{de-marneffe-etal-2006-generating}. However, the dependency perspective offers different information on this constituent. First, in a dependency tree we label arcs (relations between words) rather than nodes (constituent types). Second, a dependency subtree features a \textit{head} word that is targeted by the syntactic functions of its direct children in the subtree; \textit{e.g.} in "talented musician", "talented" is an \textit{adjectival modifier (amod)} targeting the word "musician". Third, dependency trees are less rigid when it comes to word order\footnote{This property is interesting for languages with a \textit{free word order} since dependency trees do not need to register different rules for different arrangements of the same syntactic roles.}. Figure~\ref{fig:BGSYNALL}\footnote{The semantic roles describing the predicative structure here can be obtained through AllenAI's online semantic role labeling demo available at this URL: \href{https://demo.allennlp.org/semantic-role-labeling}{https://demo.allennlp.org/semantic-role-labeling}} shows the same dependency tree displayed in the introduction augmented with PoS tags, and semantic annotation concerning its predicative structure which is explicited below.

The dependency structure of a sentence is especially interesting in that it strongly relates to the semantic structure of the sentence. In fact, formal theories of semantics, such as Generative Lexicon Theory~\cite{pustejovsky1998generative}, often posit that sentences translate to \textit{lambda expressions}\footnote{Refer to \citet{rojas2015tutorial} for an introdution to lambda calculus.} which describe semantic composition as function applications over some abstract arguments. The lambda expression corresponding to the semantic processing underlying the sentence in Figure~\ref{fig:BGSYNALL} can be written as follows\footnote{The possession information expressed by the pronoun "his" is ignored here for simplicity.}:
\begin{align}
    \lambda x.\lambda y. [musician(x)\wedge talented(x)\wedge nice(y)\wedge guitar(y)\wedge holds(x,y)]
\end{align}
This expression reads \textit{"Let $x$ and $y$ such that $x$ verifies $musician$, $x$ verifies $talented$, $y$ verifies nice, $y$ verifies $guitar$, and $(x, y)$ verifies $holds$"}. 
Notice that, in Figure~\ref{fig:BGSYNALL}, the part of the sentence characterizing the argument $x$ corresponds to the $nsubj$ dependency subtree, while the $dobj$ subtree contains all the information about the argument $y$. The $ROOT$ node which is directly above $dobj$ and $nsubj$ is anchored to the verb \textit{holds}, which informs about the main predicate expressed by the sentence. Identifying elements of the predicative structure, most often called \textit{semantic role labeling}, is an NLP task introduced by \citet{gildea2002automatic} where portions of the sentence are labeled according to \textit{semantic roles} which identify arguments of the semantic structure of the sentence. Although semantic role labeling started out with a large vocabulary of fine-grained semantic roles, subsequent annotation guidelines such as those of PropBank~\cite{bonial2012english} collapse semantic roles into a minimalistic set of \textit{proto-roles} such as the ones used in Figure~\ref{fig:BGSYNALL}. 

The correspondence of high-level syntactic roles such as $ROOT$, $nsubj$, and $dobj$ with the main predicate and its arguments illustrated in Figure~\ref{fig:BGSYNALL} is a well-marked pattern. As a matter of fact, statistics on the coNLL 2008 shared task on joint parsing of syntactic and semantic dependencies~\citep{surdeanu-etal-2008-conll} calculated by \citet{Lang2010UnsupervisedRoles} show, for instance, that 84\% of constituents having the \textit{agent} (ARG-0) semantic role have the \textit{subject} syntactic role, and that 58\% of constituents having the \textit{patient} semantic role have the \textit{object} syntactic role. This correspondance is strong enough for it to be emphasized by dependency annotation guidelines such as Universal Dependencies~\cite{nivre-etal-2020-universal} through a distinction between \textit{core} syntactic roles, \textit{i.e.} syntactic roles which directly relate to an element of the predicative structure such as the ones we color in Figure~\ref{fig:BGSYNALL}, and \textit{non-core} syntactic roles\footnote{As an example, Universal Dependencies provide a comprehensive inventory of syntactic roles they use with definitions and examples at the following URL: \href{https://universaldependencies.org/u/dep/index.html}{https://universaldependencies.org/u/dep/index.html}}. 

The distinction between \textit{core} and \textit{non-core} syntactic roles, as well as the fact that core syntactic roles are placeholders segmenting the main information carried by the sentence are of particular importance to this thesis as they constitute the basis for the intuition behind the work presented in Chapter~\ref{chap:SynRoleDisentChap}.
\section{Conclusion}
\begin{figure*}[!h]
\centering
            \begin{minipage}[b]{\textwidth}
            \begin{adjustbox}{minipage=\textwidth,scale=0.26}
                \includegraphics{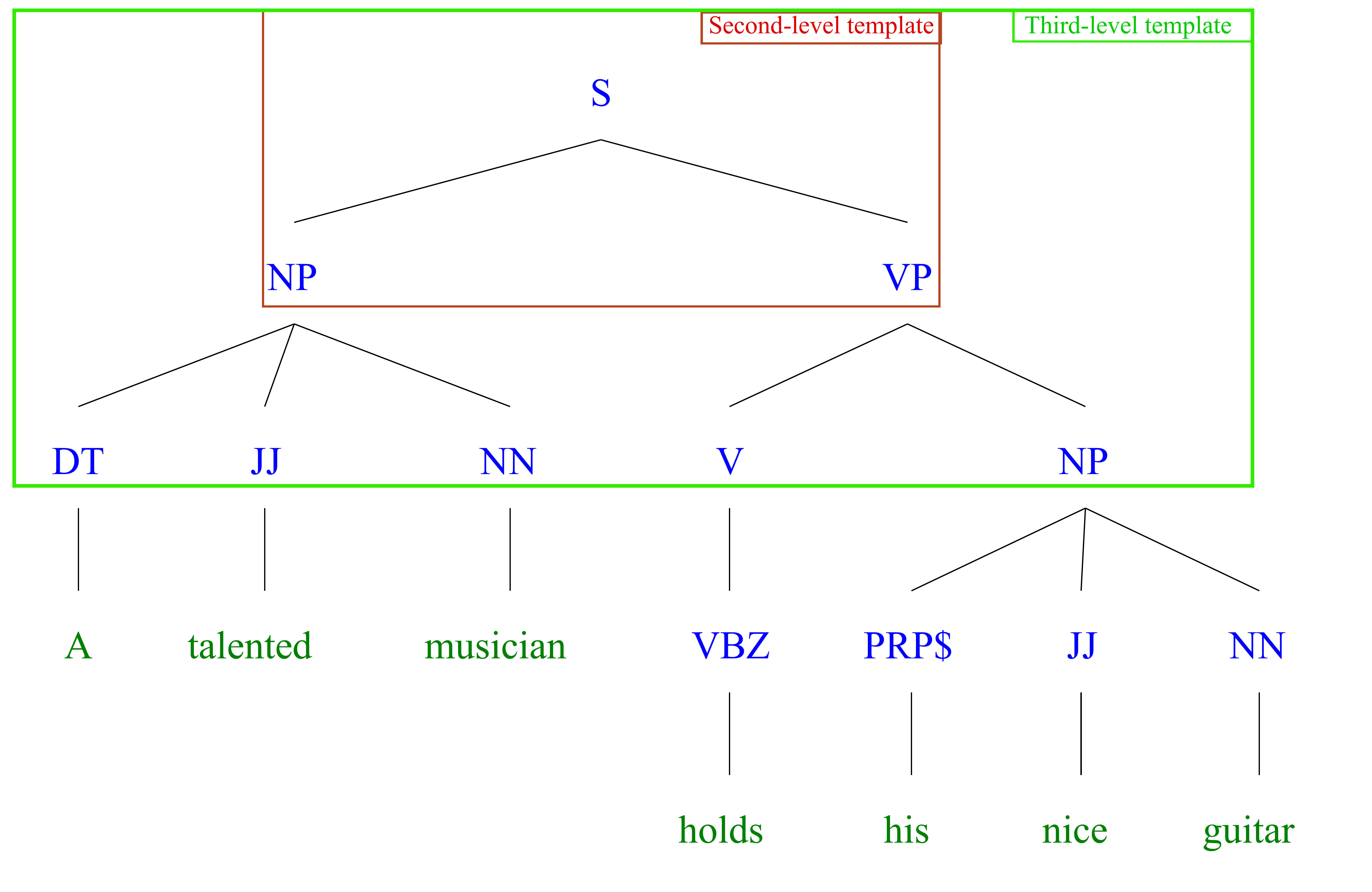}
            \end{adjustbox}
            \end{minipage}
            \caption{Example syntactic templates obtained by cutting a constituency tree at a certain level. The red frame delimits a syntactic tree cut at the $2^{nd}$ level ($2^{nd}$ level template), while the green frame delimits a syntactic tree cut at the $3^{rd}$ level ($3^{rd}$ level template).}
            \label{fig:CONSTITCUT}
\end{figure*}

In this chapter, we introduced two syntactic analysis schemes, namely constituencies and dependencies. We explained that constituencies find their roots in CFGs, and describe sentences as the result of the successive application of the rules registered in the generative grammar describing the language at hand (\S~\ref{CONSTBGSEC}). We use them to quantify the effectiveness of the method described in Chapter~\ref{chap:QKVCHAP} of this thesis. More precisely, we study the degree to which syntactic information is present in latent variables where syntactic information ranges from shallow constituency structures, \textit{i.e.} constituency trees cut at the $2^{nd}$ or $3^{rd}$ level, to deep constituency structures, \textit{i.e.} entire trees. Example cuts of a constituency tree, also called syntactic templates, are displayed in Figure~\ref{fig:CONSTITCUT}. 

Concerning dependencies, we explained that they offer a perspective on sentences which explicits the role played by each word in the sentence structure and therefore better relates it to the manner in which it participates in building the meaning behind the sentence (\S~\ref{DEPBGSEC}). A particular notion that was emphasized in the section pertaining to dependencies is the relation between core syntactic roles and elements of the predicative structure, which is important to explain the intuition behind disentanglement with regard to syntactic roles in Chapter~\ref{chap:SynRoleDisentChap}.


\part{Semi-Supervised Learning with VAEs}
{\fancyhead[RO]{CHAPTER \theHchapter.  CHALLENGING THE SSVAE FRAMEWORK}
\chapter{ Challenging the Semi-Supervised VAE Framework for Text Classification}
\label{chap:SSVAEchap}

Obtaining labeled data to train NLP systems is a process that has often proven to be costly and time-consuming, and this is still largely the case~\cite{bohmova2003prague, martinez-alonso-etal-2016-noisy,choudhary2018cost,Seddah2020BuildingHell}.
Consequently, semi-supervised approaches are appealing to improve performance while alleviating dependence on annotations.
To that end, Variational Autoencoders (VAEs; \citealp{Kingma2014Auto-encodingBayes}) have been adapted to semi-supervised learning~\cite{Kingma2014a}, and subsequently applied to several NLP tasks~\cite{Chen2018VariationalLearning,Corro2019DifferentiableAutoencoder,Gururangan2020VariationalClassification}. As mentioned in~\ref{DISENTINDUCSEC}, semi-supervised learning with VAEs is an instance of the case where supervised learning signal is used as an inductive bias to obtain a latent variable that has a clear meaning. This learning framework therefore yields an Autoencoder where the encoded representations are partly understandable, and where the decoded observations are decoded from this partly understandable representation, hence its importance for interpretability in general and for the present thesis in particular.

A notable difference between the generative model case from where VAEs originate, and the semi-supervised case is that only the decoder (generator) of the VAE is kept after training in the first case, while in the second, it is the encoder (classifier) that we keep. This difference, as well as the auto-regressive nature of text generators has not sufficiently been taken into account in the adaptation of VAEs to semi-supervised text classification.
In this chapter, we show that some components can be ablated from the long used semi-supervised VAEs (SSVAEs) when only aiming for text classification. These ablations simplify SSVAEs and offer several practical advantages while preserving their performance and theoretical soundness.

The usage of unlabeled data through SSVAEs is often described as a \emph{regularization}
on representations~\cite{Chen2018VariationalLearning,Wolf-Sonkin2018AInflection,Yacoby2020FailureTasks}.
More specifically, we explain in Section~\ref{ABLATING} that SSVAEs add to the supervised learning signal, a conditional generation learning signal that is used to train on unlabeled samples.
From this observation, we study two changes to the standard SSVAE framework.
The first simplification we study (\S~\ref{ABLATINGKL}) is the removal of a term from the objective of SSVAEs: the Kullback-Leibler term.
This encourages the flow of information into latent variables, frees the users from choosing priors for their latent variables, and is harmless to the  theoretical soundness of the semi-supervised framework.
The second simplification we study (\S~\ref{ABLATINGUNOBS}) is made to account for the auto-regressive nature of text generators. In the general case, input samples in SSVAEs are described with two latent variables: a partially-observed latent variable, which is also used to infer the label for the supervised learning task, and an unobserved latent variable, which describes the rest of the variability in the data.
However, auto-regressive text generators are powerful enough to converge without the need for latent variables. Therefore, removing the unobserved latent variable is the second change we study in SSVAEs. The above modifications can be found in some rare works throughout the literature, \textit{e.g.} \citet{Corro2019DifferentiableAutoencoder}.  We, however, aim to provide justification for these changes beyond the empirical gains that they exhibit for some tasks.

Our experiments on four text classification datasets show no harm to the empirical classification performance of SSVAE in applying the simplifications above. Additionally, we show that removing the unobserved latent variable leads to a significant speed-up.

The contributions presented in this chapter are the following: we justify two simplifications to the standard SSVAE framework, we explain the practical advantage of applying them (\S~\ref{ABLATING}), and we provide empirical results showing that they speed up the training process while causing no harm to the classification performance(\S~\ref{SSVAEEXPSEC}).

\section{Simplifying SSVAEs for Text Classification}
\label{ABLATING}
As explained in \S~\ref{SSVAEBG}, semi-supervised learning with VAEs was introduced by \citet{Kingma2014a}. Keeping the same terminology, we consider here a set of labeled examples $L=\{(x_1, y_1), ..., (x_{|L|}, y_{|L|})\}$, and a set of unlabeled examples $U=\{x'_1, ..., x'_{|U|}\}$. Also recall that \textit{i)} besides the usual unobserved latent variable $z$, the semi-supervised VAE framework uses a partially-observed latent variable $y$; \textit{ii)}
the encoder $q_\phi(y|x)$ serves both as the inference module for the supervised task, and as an approximate posterior (and encoder) for the $y$ variable in the VAE framework. 

~\cite{Kingma2014a} formulate the semi-supervised objective with $L$, $U$, $x$, $y$ and $z$ as follows:

 \begin{align}
  \mathcal{J}^\alpha &= \sum_{(x, y) \in L}\Bigr( \ELBo((x, y);z)+ \alpha \hspace{1mm} \log q_\phi(y|x)\Bigl)
  &+ \sum_{x \in U} \ELBo(x;(y, z)) \label{JALPHAChapSSVAE}
 \end{align}
The first term in Equation~\ref{JALPHAChapSSVAE} is the part used to train the network on the labeled data $L$, where $\ELBo$ considers $x$ and $y$ to be observed and $z$ to be non-observed, and $q_\phi(y|x)$ is also trained with a Cross-Entropy objective weighted by a hyper-parameter $\alpha$. The second term in $\mathcal{J}^\alpha$ is the term to be used with unlabeled samples from $U$, which is simply an $\ELBo$ that only considers $x$ to be observed. 

The simplifications we propose to the above SSVAE framework stem from an analysis of the alternative form under which ELBO can be written (Eq.~\ref{ELBoEqua} in Chapter~\ref{VAEBGCHAP}). 
Although it is valid for any arguments of $\ELBo(.;.)$, we display it here for an observed variable $x$, and the couple of latent variables $(y,z)$:\
\begin{align}
\ELBo(&x; (y, z))= 
\log p_\theta(x) - \KL[q_\phi(y, z|x)||p_\theta(y, z|x)] \label{AltELBO}
\end{align}
For the case of SSVAEs, this form provides a clear reading of the additional effect of ELBo on the learning process: \textit{i)} maximizing the log-likelihood of the generative model $p_\theta(x)$, \textit{ii)} bringing the parameters of the inference model $q_\phi(y, z|x)$ closer to the posterior of the generative model $p_\theta(y, z|x)$.
Since $p_\theta(y, z|x)$ is the distribution of the latent variables expected by the generative model $p_\theta$ for it to be able to generate $x$,
we can conclude that ELBo trains \emph{both} latent variables for conditional generation on the unsupervised dataset $U$. If we write $\ELBo((x, y), z)$ under the same form:
\begin{align}
\ELBo(&(x, y); z)= 
\log p_\theta(x, y) - \KL[q_\phi(z|x, y)||p_\theta(z|x, y)] \label{AltELBO2}
\end{align}
it can also be seen that only $z$ is trained on conditional generation for the labeled examples in $L$.

\subsection{Dropping the Unobserved Latent Variable}
\label{ABLATINGUNOBS}
Building on observations from equations~\ref{AltELBO} and~\ref{AltELBO2}, we question the usefulness of training both latent variables for conditional generation when semi-supervised learning only aims for an improvement on the inference of the partially-observed latent variable $y$.

VAEs' generative capabilities have first been exhibited on image datasets~\cite{Kingma2014Auto-encodingBayes}. For such datasets, the reconstruction loss in $\ELBo$ is an L2 loss (resp. L1 loss). This stems from the fact that $p_\theta(x|z)$ is modeled by a Gaussian (resp. a Laplace) distribution. Given a single sample $z$, the blur modeled by such distributions cannot (in the general case) model an entire image dataset. The reader may refer to \citet{Zhao2017TowardsModels} for a full discussion of this issue. As a consequence, it is necessary for such datasets to incorporate a latent variable $z$ besides the partially-observed variable $y$ to model the characteristics that $y$ does not describe.

As explained in Chapter~\ref{VAEBGCHAP}, most decoders used for language are auto-regressive, \textit{i.e.} of the form $p_\theta(x|y, z) = \prod_i p_\theta(x_i|y, z, x_{<i})$. Such decoders are able to generate realistic samples when trained on a target text corpus, in fact so much that it causes them to ignore latent variables (\textit{cf.} posterior collapse discussed in section~\ref{PColSec}). 
Given the above, $p_\theta(x)=\int_y p_\theta(x_0|y)\prod_i p_\theta(x_i|y, x_{<i}) dy$, an autoregressive LM that incorporates only $y$ as a latent variable is an expressive enough generative model to provide quality learning signal for $y$'s training on conditional generation (Eq.~\ref{AltELBO}). 
We therefore propose to keep only $y$ and to drop $z$ from the model avoiding its presence in
the Kullback-Leibler divergence in Equation~\ref{AltELBO} and saving some
parameters.

With this simplification, the training objective shown in Equation~\ref{JALPHAChapSSVAE} becomes:

 \begin{align}
 & \sum_{(x, y) \in L}\Bigr( \log p_\theta(x|y)+ \alpha \hspace{1mm} \log q_\phi(y|x)\Bigl)
  &+ \sum_{x \in U} \ELBo(x;y) \label{JALPHANoZ}
 \end{align}
 
There is a caveat regarding this modification: Since using only $y$ often makes integration over the latent variables possible\footnote{For instance, integration becomes possible when the support of the discrete latent variable is small, as is the case for binary classification.}, one may be tempted to optimize the exact log-likelihood instead of ELBo. This should not be done as it would remove the second term from Eq.~\ref{AltELBO}, decoupling the learning processes of the generative model from that of the inference model, and therefore discarding the benefit provided by semi-supervised learning with VAEs. Nevertheless, keeping only $y$ still often enables calculating exactly the reconstruction term which presents no harm to the learning process.

\subsection{Dropping the Kullback-Leibler Term}
\label{ABLATINGKL}
As discussed in Section~\ref{PColSec}, the KL divergence in $\ELBo$ sometimes discourages the model from using latent variables and makes them useless in practice~\cite{Bowman2016GeneratingSpace, DBLP:journals/corr/ZhaoSE17a, Chen2018c}.

An interesting result from \citet{DBLP:journals/corr/ZhaoSE17a} is that ELBo without KL divergence (\emph{KL-free}) is still a theoretically sound objective for generative modeling with VAEs.
The difference between the generative model resulting from a regular ELBo and a KL-free ELBo is the prior of the model.
A KL-free ELBo results in a generative model that uses as a prior $q_\phi(z)=\int_z q_\phi(z|x)p_{data}(x)dx$.
This prior is intractable which makes the resulting model impractical for generation, but causes no problem for semi-supervised VAEs.
We therefore propose, as a second change to the standard SSVAE framework, the removal of the KL-divergence in $\ELBo$.

Note that using the prior $q_\phi(z)$ mentioned above means that the network formulates its own prior instead of requiring the user to choose it.
 Priors for partially-observed latent variables are delicate to choose as it is highly preferred for them to \textit{i)} yield a closed form or the KL-divergence in $\ELBo$ to stabilize training, \textit{ii)} realistically model the \emph{default} behavior of the latent variables. The latter requirement can be particularly tedious for non-trivial latent variable models (\textit{e.g.} trees; \citealp{Corro2019DifferentiableAutoencoder}). 

If we apply this removal of the KL-divergence to both $\ELBo$'s in the standard SSVAE objective $\mathcal{J}^\alpha$ in Equation~\ref{JALPHAChapSSVAE}, it becomes:
 \begin{align}
 & \sum_{(x, y) \in L}\Bigr( \mathbb{E}_{z\sim q_\phi(z|x)}\left[\log p_\theta(x|y, z)\right]+ \alpha \hspace{1mm} \log q_\phi(y|x)\Bigl)\nonumber\\
  &+ \sum_{x \in U} \mathbb{E}_{(y, z)\sim q_\phi(y, z|x)}\left[\log p_\theta(x|y, z)\right] \label{JALPHANoKL}
 \end{align}

\subsection{Resulting Objective}
Applying both of the previous simplifications to the semi-supervised objective in Eq.~\ref{JALPHA} leads to the following objective:

 \begin{align}
 & \sum_{(x, y) \in L}\Bigr( \log p_\theta(x|y)+ \alpha \hspace{1mm} \log q_\phi(y|x)\Bigl)\nonumber\\
  &+ \sum_{x \in U} \mathbb{E}_{y\sim q_\phi(y|x)}\left[\log p_\theta(x|y)\right] \label{JALPHAMOD}
 \end{align}
 As can be seen, the first ELBo in Eq.~\ref{JALPHA} turns into a supervised conditional generation objective, while the second ELBo turns into a reconstruction term that relies only on $y$. Nevertheless, we stress that the second term is still an ELBo since removing the $KL$ term still trains a generative model~\citep{DBLP:journals/corr/ZhaoSE17a}, and the whole objective is therefore still a VAE-based semi-supervised learning objective. 
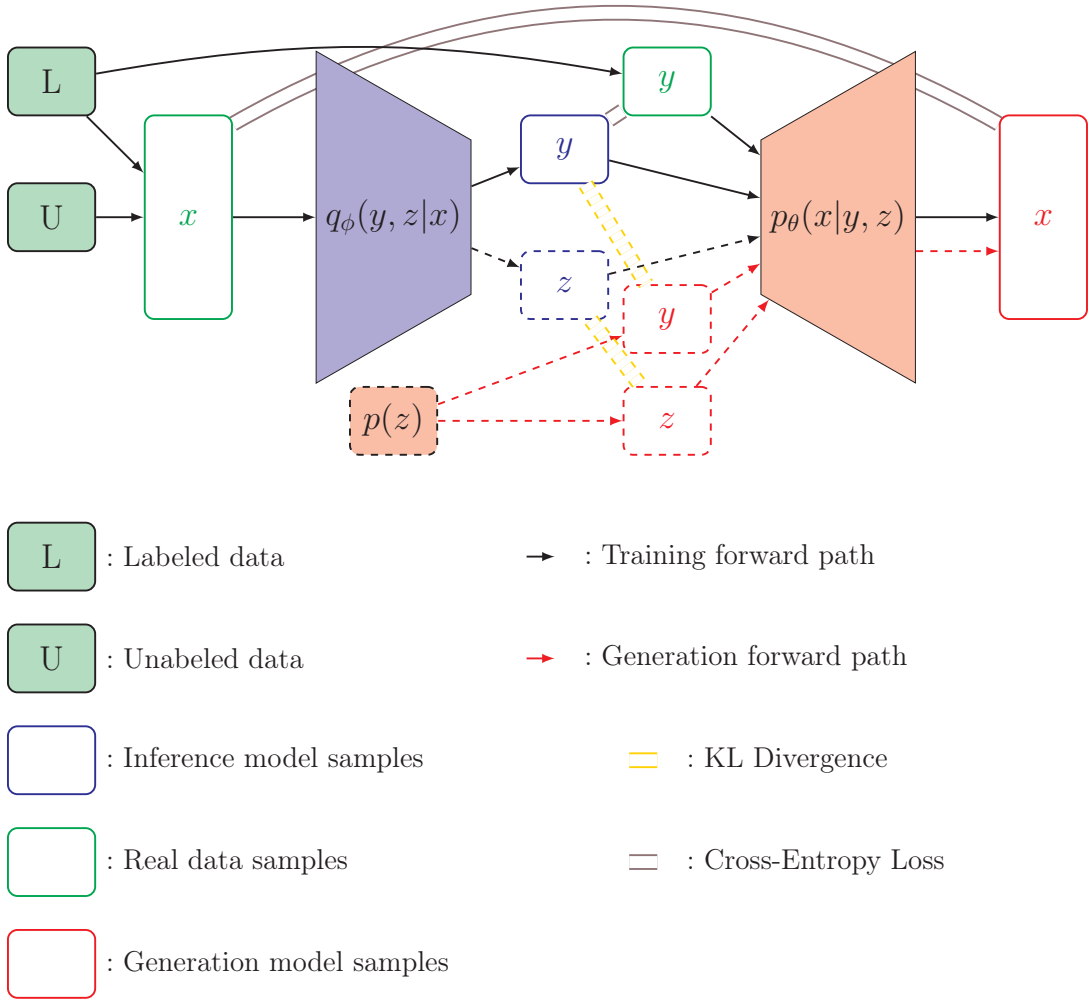
\begin{figure}
    \raggedright
    \begin{minipage}[b]{0.33\textwidth}
    \begin{adjustbox}{minipage=\textwidth,scale=0.9}
        \begin{tikzpicture}[node distance = 2cm, auto]
        \tikzstyle{main}=[rounded corners, minimum size = 10mm, thick, draw =black!80, node distance = 2cm, font=\Large, text centered]
        \tikzstyle{connect}=[-latex, thick]

        
        \node[main,text width=1cm,minimum height = 3cm, color=green!100] (xdat) [] {$x$};
        \node[main,text width=1cm,minimum height = 1cm, color=black!100, fill=green!30, left of=xdat, yshift=2cm] (L) [] {L};
        \node[main,text width=1cm,minimum height = 1cm,  color=black!100, fill=green!30, below of= L] (U) [] {U};
        
        \node [trapezium, trapezium left angle=60, trapezium right angle=60,text centered,text width = 2cm,minimum height=1cm, minimum width=2cm, draw=black, fill=blue!30, xshift=1.0cm, rotate around={-90:(0,0)}](encoder) [above of=xdat]{\rotatebox{90}{\Large $q_\phi(y, z|x)$}};
        
        \node[main,text width=1cm,minimum height = 1cm, color=black!100, below of=encoder, fill=red!30, yshift=-1cm, dashed] (pz) [] {$p(z)$};

        \node[main,text width=1cm,minimum height = 1cm, color=blue!100, right of= encoder, yshift=-1cm, xshift=0.5cm, dashed] (zinf) [] {$z$};
        \node[main,text width=1cm,minimum height = 1cm, color=blue!100, right of= encoder, yshift=1cm, xshift=0.5cm] (yinf) [] {$y$};
        
        \node[main,text width=1cm,minimum height = 1cm, color=red!100, right of= encoder, yshift=-3.0cm, xshift=2cm, dashed] (zgen) [] {$z$};
        \node[main,text width=1cm,minimum height = 1cm, color=red!100, right of= encoder, yshift=-1.5cm, xshift=2cm, dashed] (ygen) [] {$y$};
        \node[main,text width=1cm,minimum height = 1cm, color=green!100, right of= encoder, yshift=2.0cm, xshift=2cm] (ydat) [] {$y$};
        
        \node [trapezium, trapezium left angle=60, trapezium right angle=60,text centered,text width = 2cm,minimum height=1cm, minimum width=2cm, draw=black, fill=red!30, rotate around={90:(0,0)}, yshift=-6.5cm, xshift=2cm](decoder) [left of= encoder ]{\rotatebox{-90}{\Large $p_\theta(x|y, z)$}};
        \node[main,text width=1cm,minimum height = 3cm, color=red!100, xshift=1.0cm] (xgen) [right of= decoder] {$x$};
        
      \path (L) edge [connect] (xdat)
              (L) edge [bend left=10][connect] (ydat)
              (U) edge [connect] (xdat)
              (xdat) edge [connect] (encoder)
              (encoder) edge [connect] (yinf)
              (encoder) edge [connect, dashed] (zinf)
              (yinf) edge [connect] (decoder)
              (ydat) edge [connect] (decoder)
              (zinf) edge [connect, dashed] (decoder)
              (decoder) edge [connect] (xgen)
              (decoder) edge [connect, dashed, color=red!100,transform canvas={yshift=-0.5cm}] (xgen)
              (pz) edge [connect, dashed, color=red!100] (ygen)
              (pz) edge [connect, dashed, color=red!100] (zgen)
              (ygen) edge [connect, dashed, color=red!100] (decoder)
              (zgen) edge [connect, dashed, color=red!100] (decoder)
              
              (yinf) edge [connect, -, double distance = 5pt, color=Gold!50!Blue] (ydat)
              (yinf) edge [connect, -, double distance = 5pt, color=Gold!100, dashed] (ygen)
              (zinf) edge [connect, -, double distance = 5pt, color=Gold!100, dashed] (zgen)
              ;
              
              \begin{pgfonlayer}{bg}    
                \path (xdat) edge [connect, -, double distance = 5pt, color=Gold!50!Blue, bend left=30,transform canvas={yshift=1cm}] (xgen);
              \end{pgfonlayer}

         \node[main,text width=1cm,minimum height = 1cm, color=black!100, fill=green!30, left of=L, xshift=2cm, yshift=-7cm] (LLeg) [] {L};
         \node[right] at (LLeg.east) {: Labeled data};
        \node[main,text width=1cm,minimum height = 1cm,  color=black!100, fill=green!30, below of= LLeg, yshift=0.5cm] (ULeg) [] {U};
         \node[right] at (ULeg.east) {: Unabeled data};
         \node[main,text width=1cm,minimum height = 1cm, color=blue!100, below of= ULeg, yshift=0.5cm] (infvar) [] {};
         \node[right] at (infvar.east) {: Inference model samples};
         \node[main,text width=1cm,minimum height = 1cm, color=green!100, below of= infvar, yshift=0.5cm] (datvar) [] {};
         \node[right] at (datvar.east) {: Real data samples};
         \node[main,text width=1cm,minimum height = 1cm, color=red!100, below of= datvar, yshift=0.5cm] (genvar) [] {};
         \node[right] at (genvar.east) {: Generation model samples};
         
         \node[left of=L, xshift=8.8cm, yshift=-7cm] (Train1) [] {};
         \node[left of=L, xshift=9.5cm, yshift=-7cm] (Train2) [] {};
         \path (Train1) edge [connect] (Train2);
         \node[right] at (Train2.east) {: Training forward path};
         \node[below of=Train1, yshift=0.5cm] (Gen1) [] {};
         \node[below of=Train2, yshift=0.5cm] (Gen2) [] {};
         \path (Gen1) edge [connect, color=red!100] (Gen2);
         \node[right] at (Gen2.east) {: Generation forward path};
         \node[below of=Gen1, yshift=0.5cm, xshift=1.5cm] (KL1) [] {};
         \node[below of=Gen2, yshift=0.5cm, xshift=1.5cm] (KL2) [] {};
         \path (KL1) edge [connect, -, double distance = 5pt, color=Gold!100] (KL2);
         \node[right] at (KL2.east) {: $\KL$ Divergence};
         \node[below of=KL1, yshift=0.5cm] (CE1) [] {};
         \node[below of=KL2, yshift=0.5cm] (CE2) [] {};
         \path (CE1) edge [connect, -, double distance = 5pt, color=Gold!50!Blue] (CE2);
         \node[right] at (CE2.east) {: Cross-Entropy Loss};
         \node[below of=CE1, yshift=-0.2cm, xshift=7cm] (ABL1) [] {};
         \node[below of=CE2, yshift=-0.2cm, xshift=7cm] (ABL2) [] {};
         \path (ABL1) edge [-, color=white!50!white, dashed] (ABL2);
        \end{tikzpicture}
    \end{adjustbox}
    \vskip -5px
    \end{minipage}
    \caption{A diagram sketching the functioning scheme of a vanilla SSVAE. Dashed components are components we argue are unnecessary in our study.}
    \label{SSVAEGLOBALDIAG}
\end{figure}

  It should also be noted that, without $z$, the latent variables cannot provide the decoder with  the full information about a sentence and, therefore, cannot reach a state where each sample is \emph{reconstructed}. To avoid confusion, instead of \emph{reconstructing} from $y$, the role of the reconstruction term is better read in our case as \emph{raising the probability} of the sample at hand under the associated label $y$.

To better envision the impact of our modifications on the SSVAE framework, Figure~\ref{SSVAEGLOBALDIAG} presents a holistic view of the ablations we presented in this section.


\section{Experiments}
\label{SSVAEEXPSEC}
In this section, we display comparisons between instances of standard SSVAEs and the same SSVAEs after applying the changes we propose.
\subsection{Setup}
\paragraph{Datasets} We consider 4 datasets for our study: the IMDB~\cite{Maas2011LearningAnalysis} and Yelp review~\cite{li-etal-2018-delete} binary sentiment analysis datasets, and the AG News and DBPedia~\cite{zhang2015character} topic classification datasets. The datasets have been chosen to represent a range over different tasks (Sentiment Analysis and Topic Classification), different numbers of classes, and different sentence lengths.  A summary of dataset statistics is in Table~\ref{tab:Data}. 

\begin{table}[!htbp]
    \centering
    \resizebox{1.0\textwidth}{!}{
    \begin{tabular}{c|c c c c c c c }
    \hline
    dataset & Labels & Av. Sample length &  N° Classes\\
    \hline \hline
    AG News & Topic  & 37.85$\pm$10.09 & 4\\
    DBPedia& Topic & 46.13$\pm$22.46 & 14\\
    IMDB & Sentiment & 233.79$\pm$173.72& 2 \\
    Yelp & Sentiment & 8.88$\pm$3.64 & 2\\
    \hline
    \end{tabular}}
    \caption{Dataset properties.}
    \label{tab:Data}
  \end{table}
As was done in \citet{Chen2020MixText:Classification}, we measure performance on the different datasets with equal numbers of samples. Accordingly, for each dataset, we randomly subsample 10K samples from the original training set as \emph{unlabeled} data. We also use 4K labeled samples as training set and 1K as development set. We use the original test sets from each dataset. All the samples are tokenized using a simple whitespace tokenizer. 

\paragraph{Network architecture}
 The size of $z$ is set to 32. For experiments without $z$, we simply drop all the components associated to it from the network.
 
 The encoder consists of a pre-trained $300$-dimensional fastText~\cite{bojanowski-etal-2017-enriching} embedding layer, and 2 Bidirectional LSTM networks with $100$ hidden states each, one for each of the latent variables $y$ and $z$. The logits of $y$ are then obtained by passing the last state of its Bidirectional LSTM through a linear layer. Similarly the last state of the Bidirectional LSTM for $z$ is passed through a linear layer to obtain its mean parameter, and a linear layer with a softplus activation to obtain its standard deviation parameter.
 
As for the decoding step, to allow backpropagation, $z$ is sampled using the reparameterization trick~\cite{Kingma2014Auto-encodingBayes}, and $y$ is sampled using the Gumbel-Softmax trick~\cite{Jang2017CategoricalGumbel-softmax}. \citet{Xu2017VariationalClassification} have shown that latent variables are best exploited in SSVAEs when concatenated with the previous word at each generation step to obtain the next word. We design our decoder accordingly and use a 1-layered LSTM with size $200$. The only hyper-parameter we tune on the development set is $\alpha$, the coefficient weighting the supervised learning objective in Eq.~\ref{JALPHAChapSSVAE}, which is selected in the set $\{10^{0}, 10^{-1}, 10^{-2}, 10^{-3}\}$.

\paragraph{Training and validation data splits}
We sample 5 \emph{labeled} data splits of size 1K. Each of these 5 splits will play, in turn, the role of validation set for one experiment, while the other 4 splits are used for training. Looping over these splits yields 5 runs for each experiment. The results we display are the average (and standard deviation) of the results for each of these runs. The validation score serves selecting hyper-parameters (in our case only $\alpha$ from Eq.~\ref{JALPHAChapSSVAE}). The final test scores are measured on the original test set of each dataset. 
\paragraph{Probabilistic graphical model}
For models that use both $z$ and $y$, we consider the latent variables to be conditionally independent in the inference model (\textit{i.e.} $q_\phi(y, z|x)=q_\phi(y|x)q_\phi(z|x)$) ) and independent in the generation model (\textit{i.e} $p_\theta(y, z)= p(y)p(z)$). 

\paragraph{Training procedure}
 The network is optimized using ADAM~\cite{Kingma2015}, with a learning rate of 4e-3 and a dropout rate of 0.5. If the accuracy on the validation set doesn't increase for 4 epochs, the learning rate is divided by 4. If it doesn't increase for 8 epochs, the training is stopped. For objectives that include a KL-divergence, we scale it with a coefficient that is null for 3K steps then linearly increased to 1 for the following 3K steps to avoid posterior collapse~\cite{Li2020AText}. 
 
\begin{table*}[!ht]
    \centering
    \resizebox{1.0\textwidth}{!}{%
    \begin{tabular}{c c|c c c c c }
    \hline
    Dataset&  Objective & 1\% &  3\% & 10\% & 30\% & 100\% \\
    \hline \hline
    \multirow{5}{*}{IMDB}
&Supervised& \textbf{54.62}\textcolor{gray}{(3.30)}& 56.47\textcolor{gray}{(1.02)}& 62.01\textcolor{gray}{(2.75)}& 69.65\textcolor{gray}{(2.02)}& 81.02\textcolor{gray}{(0.64)}\\ 
&SSVAE& 53.92\textcolor{gray}{(2.34)}& 56.03\textcolor{gray}{(4.20)}& 62.15\textcolor{gray}{(5.03)}& 75.39\textcolor{gray}{(0.49)}& 83.34\textcolor{gray}{(0.91)}\\ 
&SSVAE-\{KL\}& 52.70\textcolor{gray}{(1.72)}& 54.95\textcolor{gray}{(0.77)}& 62.37\textcolor{gray}{(4.45)}& 74.18\textcolor{gray}{(1.97)}& 83.87\textcolor{gray}{(0.47)}\\ 
&SSVAE-\{z\}& 54.15\textcolor{gray}{(2.46)}& \textbf{56.86}\textcolor{gray}{(1.77)}& 62.15\textcolor{gray}{(2.87)}& 75.42\textcolor{gray}{(1.80)}& 81.90\textcolor{gray}{(5.17)}\\ 
&SSVAE-\{KL, z\}& 53.51\textcolor{gray}{(1.99)}& 56.58\textcolor{gray}{(2.22)}& \textbf{63.24}\textcolor{gray}{(4.15)}& \textbf{75.87}\textcolor{gray}{(1.30)}& \textbf{84.79}\textcolor{gray}{(1.34)}\\ 
\hline
    \hline
    \multirow{5}{*}{AGNEWS}
&Supervised& \textbf{68.60}\textcolor{gray}{(4.88)}& 75.92\textcolor{gray}{(1.74)}& 81.96\textcolor{gray}{(0.83)}& 84.59\textcolor{gray}{(0.67)}& 86.98\textcolor{gray}{(0.74)}\\ 
&SSVAE& 65.79\textcolor{gray}{(5.02)}& 75.95\textcolor{gray}{(1.27)}& 82.47\textcolor{gray}{(0.43)}& 85.50\textcolor{gray}{(0.30)}& 87.89\textcolor{gray}{(0.54)}\\ 
&SSVAE-\{KL\}& 68.56\textcolor{gray}{(1.89)}& 76.25\textcolor{gray}{(2.21)}& 82.76\textcolor{gray}{(0.45)}& 85.73\textcolor{gray}{(0.80)}& \textbf{87.95}\textcolor{gray}{(0.19)}\\ 
&SSVAE-\{z\}& 67.13\textcolor{gray}{(6.55)}& \textbf{77.28}\textcolor{gray}{(1.81)}& \textbf{83.48}$^*$\textcolor{gray}{(0.75)}& \textbf{85.75}\textcolor{gray}{(0.74)}& 87.94\textcolor{gray}{(0.33)}\\ 
&SSVAE-\{KL, z\}& 66.96\textcolor{gray}{(3.42)}& 76.47\textcolor{gray}{(1.24)}& 82.58\textcolor{gray}{(0.97)}& 85.51\textcolor{gray}{(0.57)}& 87.85\textcolor{gray}{(0.29)}\\ 
\hline
    \hline
    \multirow{5}{*}{Yelp}
&Supervised& 70.32\textcolor{gray}{(1.84)}& 76.32\textcolor{gray}{(2.07)}& 83.41\textcolor{gray}{(1.75)}& 87.85\textcolor{gray}{(0.58)}& 92.47\textcolor{gray}{(0.48)}\\ 
&SSVAE& \textbf{71.34}\textcolor{gray}{(1.93)}& 76.96\textcolor{gray}{(1.64)}& 82.96\textcolor{gray}{(0.69)}& 89.35\textcolor{gray}{(0.39)}& 92.85\textcolor{gray}{(0.78)}\\ 
&SSVAE-\{KL\}& 69.85\textcolor{gray}{(2.86)}& 76.82\textcolor{gray}{(1.31)}& 82.90\textcolor{gray}{(2.23)}& 88.33\textcolor{gray}{(0.99)}& 92.90\textcolor{gray}{(0.54)}\\ 
&SSVAE-\{z\}& 68.74\textcolor{gray}{(2.95)}& \textbf{78.26}\textcolor{gray}{(1.70)}& 84.11\textcolor{gray}{(1.25)}& \textbf{90.27}$^*$\textcolor{gray}{(0.28)}& 93.60\textcolor{gray}{(0.74)}\\ 
&SSVAE-\{KL, z\}& 69.21\textcolor{gray}{(1.10)}& 77.30\textcolor{gray}{(2.57)}& \textbf{85.02}$^*$\textcolor{gray}{(1.24)}& 89.74\textcolor{gray}{(1.31)}& \textbf{93.77}\textcolor{gray}{(0.61)}\\ 

\hline
    \hline
    \multirow{5}{*}{DBPedia}
&Supervised& 63.67\textcolor{gray}{(1.74)}& 81.49\textcolor{gray}{(2.25)}& 90.56\textcolor{gray}{(1.21)}& 94.63\textcolor{gray}{(0.32)}& 96.97\textcolor{gray}{(0.28)}\\ 
&SSVAE& 64.42\textcolor{gray}{(1.83)}& 83.16\textcolor{gray}{(1.49)}& 92.95\textcolor{gray}{(0.82)}& 96.26\textcolor{gray}{(0.25)}& \textbf{97.75}\textcolor{gray}{(0.11)}\\ 
&SSVAE-\{KL\}& \textbf{66.09}\textcolor{gray}{(3.05)}& 81.97\textcolor{gray}{(1.54)}& \textbf{93.64}\textcolor{gray}{(0.76)}& 96.32\textcolor{gray}{(0.28)}& 97.58\textcolor{gray}{(0.13)}\\ 
&SSVAE-\{z\}& 62.56\textcolor{gray}{(5.60)}& \textbf{83.40}\textcolor{gray}{(2.42)}& 93.37\textcolor{gray}{(1.00)}& \textbf{96.39}\textcolor{gray}{(0.21)}& 97.40$^\dag$\textcolor{gray}{(0.14)}\\ 
&SSVAE-\{KL, z\}& 62.15\textcolor{gray}{(1.68)}& 82.67\textcolor{gray}{(2.16)}& 93.40\textcolor{gray}{(1.10)}& 96.31\textcolor{gray}{(0.24)}& 97.58\textcolor{gray}{(0.19)}\\ 

\hline
     \end{tabular}}
    \caption{Accuracies on IMDB, AGNEWS, Yelp and DBPedia with varying amount of \textbf{labeled} data. The values are averages over 5 runs with standard deviations between parentheses. The best score for each dataset and each amount of labeled data is given in bold. Each semi-supervised objective that scores above (resp. below) SSVAE with p-value<0.05 is marked with $^*$ (resp. $^\dag$)}
    \label{tab:resultsLabeledSSVAE}
  \end{table*}

\subsection{Results}
\label{RESULTS}
\paragraph{Classification performance}
In Table~\ref{tab:resultsLabeledSSVAE}, we compare the performance of a standard SSVAE, to a SSVAE where we remove the KL-divergence (SSVAE-\{KL\}) another where $z$ is removed (SSVAE-\{z\}) and a third version where both the KL-divergence and $z$ are removed (SSVAE-\{KL, z\}) for amounts of data varying from 1\% to 100\% of the 4K \textit{labeled} samples in each dataset. We measure performance on all datasets using \emph{accuracy}.
As a baseline, we also include the results of an objective that does not use unlabeled data. The architecture we use for this objective is simply the LSTM encoder that we use to obtain $y$ for the SSVAE objectives. This baseline is referred to as \emph{Supervised}. 
 
The aim of our experiment is to see whether we observe that there is a harm to the performance of SSVAEs when applying the proposed simplifications.
In Table~\ref{tab:resultsLabeledSSVAE}, we only find 4 statistically significant differences between SSVAE and its variants: 3 in favor of one of our Simplified SSVAEs, and 1 in favor of the standard SSVAE. 

 \paragraph{Out-of-domain experiments}
 The sentiment analysis tasks we use for these experiments take place in
 different domains (restaurant reviews for Yelp, and movie reviews for IMDB). 
 Using models trained for each domain (with 100\% of the data), we measure performance on the other domain to see whether the changes we study have an effect on out-of-domain generalization. In Table~\ref{tab:SSVAEOOD}, we compare the out-of-domain performances of each of the objectives to that of the baseline that doesn't use unlabeled data (\emph{Supervised}). 
 
\begin{table*}[!ht]
    \centering
    \resizebox{1.0\textwidth}{!}{%
    \begin{tabular}{c||c c c c c}
    \hline
    Dataset&  Supervised& SSVAE& SSVAE-\{KL\} &SSVAE-\{z\} &SSVAE-\{KL, z\}\\
    \hline \hline
    IMDB$\longrightarrow$Yelp
    & 59.07\textcolor{gray}{(1.19)}
    & 61.78\textcolor{gray}{(6.03)}
    & 68.67$^+$\textcolor{gray}{(4.85)}
    & \textbf{71.30}$^+$\textcolor{gray}{(7.67)}
    & 64.69$^+$\textcolor{gray}{(3.84)}
    \\ 
    
    Yelp$\longrightarrow$IMDB 
    & 66.17\textcolor{gray}{(2.62)}
    & \textbf{69.54}\textcolor{gray}{(2.49)}
    & 66.67\textcolor{gray}{(3.26)}
    & 65.15\textcolor{gray}{(2.31)}
    & 66.13\textcolor{gray}{(3.82)}
    \\ 
    \hline
     \end{tabular}}
    \caption{Out-of-domain Accuracies between IMDB and Yelp for the different objectives. The best objective for each out-of-domain inference direction is given in bold. The scores displaying statistically significant improvement compared to the score of the supervised objective are marked with $^+$  }
    \label{tab:SSVAEOOD}
  \end{table*}
 The table shows no statistically significant gains from using unlabeled Yelp training data for inference on IMDB. This is to be expected as reviews from Yelp are drastically shorter than those from IMDB (\textit{cf.} Table~\ref{tab:Data}). However, for out-of-domain inference in the opposite direction, all the semi-supervised objectives except the standard SSVAE show statistically significant gains. Removing the KL-divergence to accumulate more information in $y$, and removing $z$ to have conditional generation exclusively rely on $y$ seem to be effective to help generalization beyond the original domain of the task. 

 \paragraph{Speeding up the learning process} By removing the KL-divergence and the components associated with $z$, an improvement on the speed of the learning process is to be expected. This improvement is highly dependent on the model and on the implementation at hand. As an example, we measure the average speed of an optimization iteration for each dataset, and each version of SSVAE. In Table~\ref{tab:speed}, the speed of each objective is displayed proportionally to the speed of standard SSVAEs.
 The calculations associated with the KL-divergence do not seem to slow down the iterations. However, removing $z$ and its associated components consistently cuts out a considerable proportion of the duration of optimization steps. This proportion ranges from 14\% (DBPedia) to 26\%(AGNEWS).
  \begin{table}[!h]
    \centering
    \resizebox{1.0\textwidth}{!}{%
    \begin{tabular}{c|| c c c }
    \hline
    Dataset& SSVAE-\{KL\} &SSVAE-\{z\} &SSVAE-\{KL, z\}\\
    \hline \hline
    
    AGNEWS& 
    0.911\textcolor{gray}{(0.73)}& \textbf{0.742}\textcolor{gray}{(0.65)}& \textbf{0.742}\textcolor{gray}{(0.81)} \\ 
    DBPedia& 
    1.03\textcolor{gray}{(0.56)}& \textbf{0.861}\textcolor{gray}{(0.61)}& 0.867\textcolor{gray}{(0.56)} \\
    IMDB& 
    1.018\textcolor{gray}{(0.25)}& 0.822\textcolor{gray}{(0.25)}& \textbf{0.816}\textcolor{gray}{(0.25)}\\ 
    
    Yelp& 
    0.986\textcolor{gray}{(1.52)}& \textbf{0.819}\textcolor{gray}{(1.39)} & \textbf{0.819}\textcolor{gray}{(1.52)}\\

    \hline
     \end{tabular}}
    \caption{Training durations for each objective relative to standard SSVAE, averaged over 200 iterations. Standard deviations are given between parentheses. Lowest duration for each dataset is given in bold.}
    \label{tab:speed}
  \end{table}
 
    

\section{Related Works}
After the pioneering work of \citet{Kingma2014a}, SSVAEs
were extended to tasks such as morphological inflections~\cite{Wolf-Sonkin2018AInflection}, controllable speech synthesis~\cite{Habib2019Semi-SupervisedSynthesis}, parsing~\cite{Corro2019DifferentiableAutoencoder}, sequential labeling~\cite{Chen2018VariationalLearning} among many others.
VAE internals have also been \emph{tweaked} in various manners to improve the learning performance.
For instance, \citet{Gururangan2020VariationalClassification} introduce a low resource pretraining scheme to improve transfer with VAEs, while \citet{Zhang2019AdvancesInference} propose to use the deterministic ancestor of a latent
variable to perform classification, and constrain it with an adversarial term to have it abide by the values of the random latent variable.

 While this chapter is dedicated to the theoretical soundness and the practical advantages of two simplifications to the SSVAE framework for text classifications, it could be extendend to other tasks involving text generation as the unsupervised VAE objective. For instance, the work of \citet{Corro2019DifferentiableAutoencoder} shows that semi-supervised dependency parsing scores higher with both the changes we study.

\section{Conclusion}

Starting from the observation that SSVAEs can be viewed as the combination of a supervised learning signal with an unsupervised conditional generation learning signal, we show that this framework needs neither to include a KL-divergence nor an unobserved latent variable ($z$) when dealing with text classification (\S~\ref{ABLATING}). We subsequently perform experimental comparisons between standard SSVAEs and simplified SSVAEs that indicate that our simplifications show no harm to performance both for in-domain classification and out-of-domain classification (\S~\ref{SSVAEEXPSEC}).

Our changes provide a number of practical advantages. First, removing the KL-divergence frees practitioners from choosing priors for the variables they use, and allows information to flow freely into these variables. Second, removing the latent variable $z$ from the computational graph speeds up computation and shrinks the size of the network. Despite their popularity, VAEs are often tedious to train for NLP tasks. In that regard, our simplifications should facilitate their usage in future works.

The inductive bias used to obtain an interpretable latent variable in this chapter is a supervised learning signal that trains $y$ to abide by a certain label. Although semi-supervised learning minimizes the need for labels, it still requires annotated data. As discussed in~\ref{DISENTINDUCSEC}, inductive bias comes in various forms, some of which require no supervised learning signal and only rely on structural patterns shared by the model and the data (\textit{cf.} HoloGAN's example discussed in the same section). The following chapters explore these types of inductive bias where models are built to induce understandable concepts in their latent representations without labeled data. 

\part{Unsupervised Disentanglement of Sentence Representations}}
{\fancyhead[RO]{CHAPTER \theHchapter.  UNSUPERVISED DISENTANGLEMENT VIA SYNTACTIC ROLES}
\chapter{Towards Unsupervised Content Disentanglement in Sentence Representations via Syntactic Roles}\label{chap:SynRoleDisentChap}

  This chapter presents our first contribution on unsupervised disentanglement of sentence representations, \textit{i.e.} the process of separating information in neural representations of sentences along understandable axes without annotated data. As discussed in \S~\ref{DISENTAPPPBGSEC}, disentanglement in NLP has been mostly performed to separate the semantics
(or content) in a sentence from characteristics such as style and structure in order to generate paraphrases~\citep{Chen2019ARepresentations, Bao2020, John2020DisentangledTransfer, huang-chang-2021-generating, huang-etal-2021-disentangling}.
We show in this chapter that the information in the content itself can be separated with a VAE-based model, and that this separation can be accomplished without supervision or input syntactic information. The axes along which we demonstrate content separation are the lexical realizations of core syntactic roles. As explained in \S~\ref{DEPBGSEC}, some syntactic roles, such as subjects and direct objects, are said to be \textit{core} because they strongly relate to the predicate-argument structure of sentences, which makes them compelling axes to separate information in sentences. In order to study this separation, we present a framework including a model designed to disentangle information from the realizations of different core syntactic roles and an evaluation protocol aimed at measuring this disentanglement.

The model we introduce is an Attention-Driven VAE (ADVAE), which we train  on raw text from the Stanford Natural Language Inference (SNLI) dataset 
~\citep{Schmidt2020AutoregressiveLoops}, a dataset where sentences exhibit low syntactic variation. It draws its inspiration from attention-based Machine Translation models~\citep{Bahdanau2015NeuralTranslate,Luong2015EffectiveTranslation}.
 Such models translate sentences between languages with different underlying structures and can be inspected to show a coherent 
 alignment between spans from both languages, as shown in \S~\ref{ATTBGSEC}. Our ADVAE uses Transformers~\citep{Vaswani2017}, an attention-based architecture, to map sentences from
 a language to a fixed number of independent latent variables, then map these variables back to the same sentences. Although ADVAE could be used to study other attributes, we motivate it (\S~\ref{MOTIVATION}) and therefore study it for the alignment of syntactic roles with latent variables.
  
Evaluating disentanglement with regard to spans is challenging. After training the model and only for evaluation, we use linguistic information (from an off-the-shelf dependency parser) to first extract
  syntactic roles from sentences, and then study their relation to latent variables. To study this relation on the
   ADVAE decoder, we repeatedly \textit{i)} generate a sentence from a sampled latent vector
  \textit{ii)} perturb this latent vector at a specific location
  \textit{iii)} generate a sentence from this new vector and observe the difference.
  On the encoder side, we study the attention values to see whether each latent variable is focused on a particular syntactic role 
  in input sentences. 
  The latter procedure is only possible through the way our ADVAE uses attention to produce latent variables. To the best of our knowledge, we are the first to use this transparency mechanism to obtain quantitative results for a latent variable model. 
  
  We first justify our focus on syntactic roles in \S~\ref{SyntacticRoles}, then we go over our contribution, which is threefold: \textit{i)} We introduce the ADVAE, a model that is designed for \emph{unsupervised}
    disentanglement of syntactic roles, and that enables analyzing the interaction
    between latent variables and observations through the values of attention (\S~\ref{PROPOSEDMODEL}), \textit{ii)} We design an experimental protocol for the challenging assessment of disentanglement over realizations of syntactic roles, based on perturbations on the decoder side and attention on the encoder side (\S~\ref{EVALUATIONPROC}), \textit{iii)} Our empirical results show that our architecture disentangles syntactic roles better than standard sequence VAEs and Transformer VAEs and that it is capable of controlling realizations of syntactic roles separately during generation when trained on a dataset with regularly structured sentences (\S~\ref{REGEXPADVAE}). To better characterize the limits of our model, we also provide results for hierarchical version of ADVAE(\S~\ref{HIERARCH}), and for our standard ADVAE when trained on Yelp, a dataset with more challenging syntactic variability (\S~\ref{YELP}).

\section{Syntactic Roles as Targets for Unsupervised Disentanglement}
\label{SyntacticRoles}

As presented in \S~\ref{DEPBGSEC}, dependency parsing yields a tree, where edges are labeled with syntactic roles (or relations or functions) such as \emph{nominal subject} (\emph{nsubj}). The lexical realizations of these syntactic functions are textual spans and correspond to syntactic constituents. For instance, the lexical realization of the \emph{direct object} (\emph{dobj}) of the verb \emph{holds} in our example sentence displayed in Figure~\ref{fig:BGSYNADVAECHAP} is the span \emph{his nice guitar}, with {\em guitar} as head.

\begin{figure*}[!h]
\centering
    \begin{minipage}[b]{0.48\linewidth}
    \begin{adjustbox}{minipage=\linewidth,scale=0.9}
        \begin{dependency}[theme = default]
           \begin{deptext}[column sep=1em, row sep=.1ex]
               DT  \&JJ  \&NN  \&VBZ  \&PRP\$  \&JJ  \&NN \\
              A \& talented \& musician \& holds \& his \& nice \& guitar \\
           \end{deptext}
           \deproot[edge style={orange!70!black},
            label style={orange!70!black, fill=white, text=orange!70!black}]{4}{ROOT}
           \depedge[edge style={blue},
            label style={blue, fill=white, text=blue}]{4}{3}{nsubj}
           \depedge[edge style={green!60!black},
            label style={green!60!black, fill=white, text=green!60!black}]{4}{7}{dobj}
           \depedge{3}{2}{amod}
           \depedge{3}{1}{det}
           \depedge{7}{6}{amod}
           \depedge{7}{5}{poss}
           \wordgroup[style={orange!70!black, fill=white}]{2}{4}{4}{pred}
           \wordgroup[style={blue, fill=white}]{2}{1}{3}{a0}
           \wordgroup[style={green!60!black, fill=white}]{2}{5}{7}{a1}
           \groupedge[edge style={blue},
            label style={blue, fill=white, text=blue}, edge below]{pred}{a0}{ARG0}{2ex} 
           \groupedge[edge style={green!60!black},
            label style={green!60!black, fill=white, text=green!60!black}, edge below]{pred}{a1}{ARG1}{2ex} 
        \end{dependency}
            \end{adjustbox}
    \end{minipage}
    \begin{minipage}[b]{0.48\linewidth}
    \begin{adjustbox}{minipage=\linewidth,scale=0.9}
        \begin{tikzpicture}
            \draw [color=white](0,0) rectangle (3,2);
            \draw[decorate,decoration={brace,raise=0.1cm}]
            (4.5,3.5) -- (4.5,1.6) node[align=center,above=0.2cm,pos=0.8, xshift=1.2cm] {~Dependency \\Parse};
            
            \draw[decorate,decoration={brace,raise=0.1cm}]
            (4.5,1.55) -- (4.5,1.1) node[align=center,above=-0.2cm,pos=0.8, xshift=1.2cm] {PoS Tags};
            \draw[decorate,decoration={brace,raise=0.1cm}]
            (4.5,1.05) -- (4.5,0.0) node[align=center,above=-0.2cm,pos=0.8, xshift=1.2cm] {Predicative\\ Structure};
        \end{tikzpicture}
    \end{adjustbox}
    \end{minipage}

    \caption{A sentence and its syntactic roles. The correspondence between syntactic roles and elements of the predicative structure is highlighted with colors. $ROOT$ is the root node of the dependency tree, $nsubj$ is a nominal subject, $dobj$ is a direct object, $det$ is a determiner, $poss$ is a possessive determiner, and $amod$ is an adjectival modifier. $ARG0$ and $ARG1$ are semantic \textit{proto-roles}~\cite{bonial2012english} designating respectively the entity that performs the action (a.k.a the \textit{agent}) and the entity affected by the action (a.k.a the \textit{patient}).}
    \label{fig:BGSYNADVAECHAP}
\end{figure*}
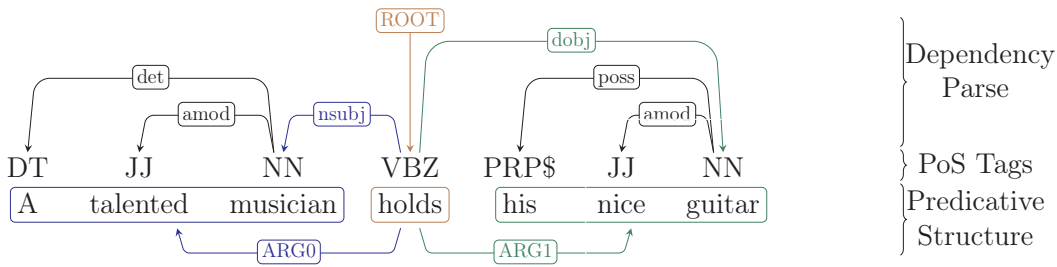

In our work, we focus\footnote{Future research that takes interest in the finer-grained disentanglement of content may simply study a larger array of syntactic roles. Using our current system we display results including all syntactic roles in Appendix~\ref{ENTIREROLERESULTS}.} on verbal roots of sentences, their nominal subjects, and their direct or prepositional objects, \textit{i.e.} \emph{core} (as opposed to \emph{oblique}) syntactic roles. These syntactic roles directly relate to the predicative structure (\textit{cf.} \S~\ref{DEPBGSEC}) which is a decomposition of the information in a sentence into a predicate and its arguments. 
With that in mind, we hypothesize in this work that core syntactic roles could constitute principal axes of variation for the information in sentences. Given the similarity between the way VAEs form latent variables and PCA (\textit{cf.} \S~\ref{ALIGNDISENTBGSEC}), and under the hypothesis that  core syntactic roles are principal axes, we expect our VAE model to separate information pertaining to realizations of these syntactic functions without supervision, which is the goal of our work.

\section{Model Description}
\label{PROPOSEDMODEL}
The usual method to obtain sentence representations from Transformer models uses only a Transformer encoder either by taking an average of the token representations or by using the representation of a special token (\textit{e.g} [CLS] in BERT~\citep{devlin-etal-2019-bert}). Recently, the usage of both Transformer encoders and decoders has also been explored in order to obtain representations whether by designing classical Autoencoders~\citep{Lewis2020BART:Comprehension, siddhant2019evaluating, Raffel2020t5},  or VAEs~\citep{Li2020Optimus:Space}, where training involves Transformer encoders and decoders but representations are obtained with only the encoder. Our model, the ADVAE, differs from these models in that it uses both an encoder and a decoder to produce sentence representations, similar to the way a Machine Translation (MT) Transformers produces translations.

Producing representations with Cross-Attention has been introduced by \citet{Locatello2020Object-centricAttention} as part of the Slot Attention modules in the context of unsupervised object discovery. However, in contrast to \citet{Locatello2020Object-centricAttention}, we simply use Cross-Attention as it is found in \citet{Vaswani2017}, \textit{i.e.} without normalizing attention weights over the query axis, or using GRUs~\citep{cho2014learning} to exchange information between the vectorial representations. As will be shown through our experiments, this is sufficient to disentangle syntactic roles. Concurrently to our work, \citet{jaegle2022perceiver} also develop a model that uses Cross-Attention to encode inputs into latent vectors and to decode them with the intent to produce an architecture that is general enough to apply to all input/output format. Contrary to our model, their model \textit{i)} does not use Self-Attention; \textit{ii)} is not trained as a VAE, meaning it is not generative; \textit{ii)} is not designed or evaluated for disentanglement. 

We explain the observation that motivates our work  
in \S~\ref{MOTIVATION}, we then describe in \S~\ref{ARCHITECTURE} the minimal changes we apply to MT Transformers, and finally,  we present the objective we use in \S~\ref{OBJECTIVE}. The parallel between our model and MT Transformers is illustrated in Figure~\ref{fig:SCHEMES}.

\begin{figure*}[!h]
\centering
    \begin{minipage}[b]{0.33\linewidth}
    \begin{adjustbox}{minipage=\linewidth,scale=0.53}
        \fbox{\begin{tikzpicture}
        \tikzstyle{main}=[rounded corners, minimum size = 10mm, thick, draw =black!80, node distance = 5mm, font=\Large, text centered]
        \tikzstyle{connect}=[-latex, thick]
        
        \node[main,text width=3cm] (source) [] {$s_1, s_2, ..., s_{N_s}$};
        \node[main,text width=2cm] (spe) [above=of source] {Positional Encoding};
        \node[main,text width=2.5cm] (enc) [above=of spe] {Transformer Encoder};
          
        \node[main,text width=4cm] (ptarget) [right=5 mm of source] {$t_1, t_2, ..., t_{N_t-1}$};
        \node[main,text width=2cm] (tpe) [above=of ptarget] {Positional Encoding};
        
        \node[minimum height = 5.8cm, minimum width = 5.5cm, thick, draw =black!80, node distance = 5mm, font=\Large, align=left, dashed, yshift=-0.2cm] (dec) [above=of tpe] {};
        \node[main,text width=5cm, align=left] (att) [above=of tpe] {Masked Self-Attention + Cross-attention to align source information with target};
        \node[main,text width=5cm] (mlp) [above=of att] {MLP};
        \node[minimum width = 2.8cm, thick, draw =black!80, node distance = 5mm, font=\Large, align=left, minimum height = 1cm,  text width=2.8cm, xshift=-12mm, dashed] (declab) [above=of mlp] {Transformer\\ Decoder};
        
        \node[main,text width=5cm] (result) [above=20 mm of mlp] {$t_1, t_2, ..., t_{N_t}$};
        
       \path (source) edge [connect] (spe)
            (spe) edge [connect] (enc)
            (enc) edge [bend left][connect] (att)
            (ptarget) edge [connect] (tpe)
            (tpe) edge [connect] (att)
            (att) edge [connect] (mlp)
            (mlp) edge [bend right][connect] (result);
         \end{tikzpicture}}
    \end{adjustbox}
    \vskip -5px
    \caption*{(a) An MT Transformer}
    \end{minipage}
    \begin{minipage}[b]{0.33\linewidth}
    \begin{adjustbox}{minipage=\linewidth,scale=0.53}
        \fbox{\begin{tikzpicture}
        \tikzstyle{main}=[rounded corners, minimum size = 10mm, thick, draw =black!80, node distance = 5mm, font=\Large, text centered]
        \tikzstyle{connect}=[-latex, thick]
        
        \node[main,text width=3cm] (source) [] {$w_1, w_2, ..., w_{N}$};
        \node[main,text width=2cm] (spe) [above=of source] {Positional Encoding};
        \node[main,text width=2.5cm] (enc) [above=of spe] {Transformer Encoder};
          
        \node[main,text width=4cm, color=blue!100] (ptarget) [right=5 mm of source] {$e_1, e_2, ..., e_L$};
        
        \node[minimum height = 5.7cm, minimum width = 5.5cm, thick, draw =black!80, node distance = 5mm, font=\Large, align=left, dashed, yshift=1.4cm, color=blue!100] (dec) [above=of ptarget] {};
        \node[main,text width=5cm, align=left, yshift=1.6cm] (att) [above=of ptarget] {
        \textcolor{blue}{Cross-attention to align source information with latent variable information}};
        \node[main,text width=5cm] (mlp) [above=of att] {MLP};
        \node[minimum width = 2.8cm, thick, draw =black!80, node distance = 5mm, font=\Large, align=left, minimum height = 1cm,  text width=2.8cm, xshift=-12mm, dashed, color=blue!100] (declab) [above=of mlp] {Transformer\\ Encoder};
        
        \node[main,text width=2.8cm, xshift=-2.0cm, yshift=0.3cm, color=blue!100] (result1) [above=20 mm of mlp] {$\mu_1, \mu_2, ..., \mu_L$};
        \node[main,text width=2.8cm, color=blue!100] (result2) [right=1mm of result1] {$\sigma_1, \sigma_2, ..., \sigma_L$};
        
       \path (source) edge [connect] (spe)
            (spe) edge [connect] (enc)
            (enc) edge [bend left][connect] (att)
            (ptarget) edge [connect, color=blue!100] (att)
            (att) edge [connect] (mlp)
            (mlp) edge [bend right][connect, color=blue!100] (result1)
            (mlp) edge [bend right][connect, color=blue!100] (result2);
        \end{tikzpicture}}
    \end{adjustbox}
    \vskip -5px
    \caption*{(b) Our encoder}
    \end{minipage}\hfill
    \begin{minipage}[b]{0.33\linewidth}
    \begin{adjustbox}{minipage=\linewidth,scale=0.53}
        \fbox{\begin{tikzpicture}
        \tikzstyle{main}=[rounded corners, minimum size = 10mm, thick, draw =black!80, node distance = 5mm, font=\Large, text centered]
        \tikzstyle{connect}=[-latex, thick]
        
        \node[main,text width=3cm, color=blue!100] (source) [] {$z_1, z_2, ..., z_L$};
        \node[main,text width=3cm, color=blue!100] (spe) [above=of source] {z-Identifier};
        \node[main,text width=2.5cm] (enc) [above=of spe] {Transformer Encoder};
          
        \node[main,text width=4cm] (ptarget) [right=5 mm of source] {$w_1, w_2, ..., w_{N-1}$};
        \node[main,text width=2cm] (tpe) [above=of ptarget] {Positional Encoding};
        
        \node[minimum height = 5.7cm, minimum width = 5.5cm, thick, draw =black!80, node distance = 5mm, font=\Large, align=left, dashed, yshift=-0.2cm] (dec) [above=of tpe] {};
        \node[main,text width=5cm, align=left] (att) [above=of tpe] {Masked Self-Attention + Cross-attention to align latent variable information with source};
        \node[main,text width=5cm] (mlp) [above=of att] {MLP};
        \node[minimum width = 2.8cm, thick, draw =black!80, node distance = 5mm, font=\Large, align=left, minimum height = 1cm,  text width=2.8cm, xshift=-12mm, dashed] (declab) [above=of mlp] {Transformer\\ Decoder};
        
        \node[main,text width=5cm, yshift=0.1cm] (result) [above=20 mm of mlp] {$w_1, w_2, ..., w_N$};
        
       \path (source) edge [connect, color=blue!100] (spe)
            (spe) edge [connect] (enc)
            (enc) edge [bend left][connect] (att)
            (ptarget) edge [connect] (tpe)
            (tpe) edge [connect] (att)
            (att) edge [connect] (mlp)
            (mlp) edge [bend right][connect] (result);
        \end{tikzpicture}}
    \end{adjustbox}
    \vskip -5px
    \caption*{(c) Our decoder}
    \end{minipage}
    \caption{
    (a) is a minimalistic representation of the functioning scheme of an MT Transformer (full scheme can be seen in Figure~\ref{fig:TransFigBG}).In blue, we highlight in (b) the difference between our encoder and a source-to-target MT model, and in (c) the difference between our decoder and a target-to-source MT model. The input at the bottom right for the Transformer Decoders in (a) and (c) is the series of previous words for auto-regressive generation. The input to our model is a series of words $w$, at the bottom left of (b), and its output is the reconstruction of these words in the same language, at the top right of (c).
        }
    \label{fig:SCHEMES}
    
\end{figure*}

\subsection{The Intuition Behind our Model}
\label{MOTIVATION}
Consider $s=(s_i)_{1\leq i\leq N_s}$ and $t=(t_j)_{1\leq j\leq N_t}$, two sequences of tokens forming respectively a sentence in 
the source language and a sentence in the target language. Given $s$, attention-based translation models are capable of yielding 
$t$ while also providing information about the alignment between the groups of tokens (of different sizes) in both sentences 
~\citep{Bahdanau2015NeuralTranslate,Luong2015EffectiveTranslation}). 
  This evidence suggests that attention-based architectures are capable of factoring information from groups  of words 
  according to a source structure, and redistributing it according to a target structure.

 The aim of our design is to use, as a target, a set  of $L$ \emph{independent} latent variables that will act as fixed placeholders
  for the information in sentences. We stress that 
   $L$ is fixed and independent of the input sentence size $N$. Combining Transformers, an attention-based MT model, 
 and the VAE framework for disentanglement,
 our ADVAE  is intended to factor information from independent groups of words into separate latent variables. 
 In the following sections, we refer to this set of independent latent variables as the latent \textit{vector} $z=(z_l)_{1\leq l\leq L}$ and to each $z_l$ as a latent \textit{variable}.
 
 \subsection{Model Architecture}
 \label{ARCHITECTURE}
\paragraph{Inference model:}
This is the inference model $q_\phi$ (encoder in Fig.~\ref{fig:SCHEMES}.b) for our latent variables $z=(z_l)_{1\leq l\leq L}$. It differs from an MT Transformer in two ways.
 First it uses as input a sentence $w$, and $L$ learnable vectors $(e_l)_{1\leq l\leq L}$ instead of the source and target tokens $s$ and $t$ used in translation. The learnable vectors $e$ will go through Cross-Attention without Self-Attention. We stress that these learnable vectors are input-independent.
Second its output is not used to select a token from a vocabulary but rather
passed to a linear layer (resp. a linear layer followed by a softplus non-linearity) to yield the mean parameters $(\mu_l)_{1\leq l\leq L}$ (resp.
 the standard deviation parameters  $(\sigma_l)_{1\leq l\leq L}$)  to parameterize the diagonal Gaussian distributions $(q^{(l)}_\phi(z_l|w))_{1\leq l\leq L}$.

Formally, let us define $M^\mu$ and $M^\sigma$ to be linear layers that will respectively be used to obtain the latent variables' means and standard deviations. Given input token sequence $w$, the encoder $q_\phi(z|w)=\prod_{l} q_{\phi}(z_{l}|w)$ yields parameters $\mu_l$ and $\sigma_l$ to be used by the diagonal Gaussian distribution of each of the latent variables $z_l$ as follows\footnote{The $\TransDec$ and $\TransEnc$ modules used here were defined in \S~\ref{TRANSFORMERBGSEC}. To simplify equations, we omit word embedding look-up tables and positional encodings.}:

\begin{flalign}
    \Tilde{z} = \TransDec(e; \TransEnc(&w))&& \nonumber\\
     \forall\hskip 1mm l \hskip 2mm\text{ s.t. }\hskip 1mm 1\leq l\leq& L: &&\nonumber\\
    \mu_{l}=  M^\mu(&\Tilde{z}_l), \hskip 2mm \sigma_{l} =  \SoftPlus(M^\sigma(\Tilde{z}_l)) &&\nonumber\\
     z_l \sim \mathcal{N}(\mu&_l; \sigma_l)&& \label{ADVAEncEq}
\end{flalign}

The distribution of the whole latent vector is simply the product of Gaussians
\\ $q_\phi(z_1,\ldots,z_L |s)=\prod_l^{L}q^{(l)}_\phi(z_l|w)$.

\paragraph{Generation model:}
Our generation model consists of an autoregressive decoder \\(Fig.~\ref{fig:SCHEMES}.c) $p_\theta(w|z_1,\ldots, z_L) = \prod_i^{N}p_\theta(w_i|w_{<i}, z_1,\ldots, z_L)$ where $w_{<i}$ is the series of tokens preceding $w_i$, and
a  prior assuming independent standard Gaussian variables, \emph{i.e.} $p(z_1,\ldots,z_L) =  \prod_l^Lp(z_l)$.
Each latent variable $z_l$ is concatenated with an associated learnable vector $d_l$ (\emph{z-Identifier} in Fig.~\ref{fig:SCHEMES}.c) instead of going through positional encoding. From there on, the latent variables are used like source tokens in an MT Transformer.

For autoregressive decoding, \citet{Vaswani2017} define a version of $\TransDec$ we call $\ARTransDec$.
This version uses Masked Self-Attention (Eq.~\ref{MaskedAttEq} from Section~\ref{TRANSFORMERBGSEC})  so that each word only queries information from the previously generated words.
Even though $\ARTransDec$ yields a sequence of length equal to that of the sentence $w$, in what follows, we consider 
its output to be only the last element of the sequence in order to express auto-regressive generation in a clear manner.

Given the above, Cross-Attention is used by the ADVAE decoder to dispatch information from the \textit{source} latent variable samples to the 
\textit{target} generated sequence. Accordingly, using a beginning-of-sentence token $w_0$,
  $p_\theta(w|z)=\prod_{i}p_\theta(w_{i}|w_{<i},z)$ yields probabilities
for the categorical distribution of the generated tokens $w$ by decoding latent variables $z$ concatenated with their embeddings $d$:
\begin{flalign*}
  y = \Concat&(d; z)\nonumber\\
\forall\hskip 1mm i \hskip 2mm\text{ s.t. }\hskip 0mm &1\leq i\leq |w|: \hskip 2mm\\
    &\Tilde{w}_i =\ARTransDec(w_0, \dots, w_{i-1};
    \TransEnc(y)) \nonumber\\
    &w_i \sim \Categorical(\softmax(M^w(\Tilde{w}_i))) 
\end{flalign*}

\subsection{Optimization Objective}
\label{OBJECTIVE}
We train our ADVAE using the 
$\beta$-VAE~\citep{Higgins2019-VAE:Framework} objective discussed in \S~\ref{INDPLVDISENTBG}:
\begin{equation}
    \log p_\theta(w) \geq \mathbb{E}_{(z) \sim q_\phi(z|w)}\left[ \log p_\theta(w|z) \right] -
    \beta \KL[q_\phi(z|w)||p(z)] \label{BETAVAE}
\end{equation}
In Eq.~\ref{BETAVAE}, $w$ is a sample from our dataset, $z$ is our latent vector and the distributions 
$p_\theta(w) = \int p_\theta(w|z) p(z)dz$ and $q_\phi(z|w)$ are respectively the generation model and the inference model.
We use a standard Gaussian  distribution as prior $p(z)$ and a diagonal Gaussian distribution as the approximate
inference distribution $q_\phi(z|w)$. The weight $\beta$ is
used (as in \citealp{Chen2018c}, \citealp{Xu2020OnSupervision}, \citealp{Li2020ProgressiveRepresentations}) to control
 disentanglement, but also to find a balance between the expressiveness of latent variables 
 and the generation quality.

\section{Evaluation Protocol}
\label{EVALUATIONPROC}
In order to quantify disentanglement, we first measure the interaction between latent variables and syntactic roles. To do so, we extract \emph{core} syntactic roles from sentences according to the procedure we describe in \S~\ref{ROLEEXTRACT}. Subsequently, for the ADVAE decoder, we repeatedly perturb latent variables and measure their influence on the realizations of the syntactic roles in generated sentences (\S~\ref{DECODERMETRICS}). For the ADVAE encoder, we use attention to determine the syntactic role that participates most in  producing the value of each latent variable (\S~\ref{ENCODERMETRICS}). 

Given these metrics, we measure disentanglement taking inspiration from MIG in \S~\ref{DISENTMETRICS}. Recall from \S~\ref{DisentMeasSec} that MIG consists in measuring the difference between the first and second latent variables with the highest mutual information with regard to a target factor. It is intended to quantify the extent to which a target factor is concentrated in a single variable.
This metric assumes knowledge of the underlying distribution of the target information in the dataset.
 However, there is no straightforward or agreed-upon way to set this distribution for text spans, and therefore to calculate MIG in our case. As a workaround, we use the influence metrics defined in \S~\ref{DECODERMETRICS} and \S~\ref{ENCODERMETRICS} as a replacement for mutual information to quantify disentanglement.

\subsection{Syntactic Role Extraction}\label{ROLEEXTRACT}
We use the Spacy\footnote{\url{https://spacy.io/models/en\#en\_core\_web\_sm}} dependency parser~\citep{spacy2}
trained on \\Ontonotes5~\citep{AB2/MKJJ2R_2013}. For each sentence the realization of \emph{verb} is
 the root of the dependency tree if its POS tag is \textit{VERB}. Realizations of \emph{subj}
(subject), \emph{dobj} (direct object), and \emph{pobj} (prepositional object) are \emph{spans}
corresponding to subtrees whose roots are labelled resp. \textit{nsubj}, \textit{dobj}, and \textit{pobj} and which are direct children\footnote{\textit{pobj} syntactic roles are taken to be the direct descendants of a preposition (\textit{prep}) that is directly underneath the root.} of the verbal root.

A realization of a syntactic role in $R=\{verb, subj, dobj, pobj\}$ is empty if no node in the dependency tree satisfies its
extraction condition.\footnote{Examples of syntactic role extractions can be found in Appendix~\ref{EXAMPLESEXTRACTIONS}.}

\subsection{Latent Variable Influence on Decoder}\label{DECODERMETRICS}

Intuitively, we repeatedly compare the text generated from a sampled latent vector to the text generated using the same \emph{vector} where only one latent \emph{variable} is resampled.
Thus we can isolate the effect of each latent $variable$ on output text and gather statistics.

More precisely, we sample $T^{dec}$ latent \emph{vectors} $(z^{(j)})_{1\leq j\leq T^{dec}}=(z_l^{(j)})_{1\leq j\leq T^{dec}, 1\leq l\leq L}$.
Then for each $z^j$, and for each $l$ we create an altered version $\tilde{z}^{(jl)}=(\tilde{z}^{(jl)}_{l'})_{1\leq l'\leq L}$ where we resample only the $l^{\text{th}}$ latent \emph{variable} (\textit{i.e.} $\forall l'\neq l,\  \tilde{z}^{(jl)}_{l'}=z^{(j)}_{l'}$).

Generating the corresponding sentences\footnote{Throughout this chapter, we use greedy sampling, \textit{i.e.} sampling the highest-probability word at each step as explained in \S~\ref{SAMPLMDEFSEC}, for all generated sentences. } with $p_\theta(w|z)$ yields a list of original sentences ${(w^{(j)})}_{1\leq j\leq T^{dec}}$, and a matrix of sentences displaying the effect of modifying each latent variable ${(\tilde{w}^{(jl)})}_{1\leq j\leq T^{dec}, 1\leq l\leq L}$. 
For each syntactic role $r \in R$, we denote the realization extracted from a sentence $w$ with $\rop(w)$.

To measure the influence of a variable $z_l$ on the realization of a syntactic role $r$, denoted $\Gamma^{dec}_{rl}$, we estimate the probability that a change in this latent variable incurs a change in the span corresponding to the syntactic role.
We first discard, for the influence on a role $r$, sentence pairs  $(w^{(j)}, \tilde{w}^{(jl)})$ where one the syntactic role is not realized in one of the sentences\footnote{In the original version of this work published in \citet{Felhi2021TowardsRoles}, we discarded sentence pairs where a syntactic role appears or disappears. This causes sentence pairs that do not contain realizations of the syntactic role to count as sentences where the realization of the syntactic role did not change. This leads to some variations in the decoder-specific measures compared to the original publication. However, these variations do not alter the conclusions drawn from empirical results.} , 
because the presence of a syntactic role is a property of its parent word, 
(\textit{e.g.} the presence or absence  of  a \emph{dobj} is controlled by the \textit{transitivity} of the verb) hence not directly connected to the representation of the role $r$ itself.
As they are out of the scope of our chapter, we report measures of these structural changes (diathesis) in Appendix~\ref{StructHeatMap}, and leave their extensive study to future works. We denote the remaining number of samples $T'^{dec}_{rl}$.

In the following, we use operator $\mathbf{1}{\{.\}}$, which is equal to 1 when the boolean expression it contains is true and to 0 when it is false.
This process yields a matrix $\Gammaopdec$ of shape $(|R|, L)$ which summarizes interactions in the \emph{decoder} between syntactic roles and latent variables:
\begin{align}
    &\Gamma^{dec}_{rl} =
    \sum_{j=1}^{T'^{dec}_{rl}} \frac{ \mathbf{1}{\{\displaystyle\rop(w^{(j)})\neq \rop(\tilde{w}^{(jl)})\}}}{T'^{dec}_{rl}} 
\end{align}
\subsection{Encoder Influence on Latent Variables}\label{ENCODERMETRICS}
We compute this on a held out set of size $T^{enc}$ of sentences ${(w^{(j)}_{i})}_{1\leq j\leq T^{enc}, 1\leq i\leq N_{w^{(j)}}}$.
Each sentence $w^{(j)}$ of size $N_{w^{(j)}}$ generates an attention matrix ${(a^{(j)}_{li})}_{1\leq l\leq L, 1\leq i\leq N_{w^{(j)}}}$. 
Attention values are available in the Transformer encoder with Cross-Attention computing the inference model\footnote{For simplicity, attention values are averaged over attention heads and transformer layers. This also allows drawing conclusions with regard to the tendency of the whole attention network, and not just particular specialized heads as was done in \citet{clark-etal-2019-bert}. For the sake of completeness, we display per-layer results in Appendix~\ref{PerLayerAtt}.}, and quantify the degree to which each latent variable embedding $e_{z_i}^{enc}$ draws information from each token $w_j$ to form the value of $z_i$.

For the encoder, we consider the influence of a syntactic role on a latent variable to be the probability for the attention values
of the latent variable to reach their maximum on the index of a token in that syntactic role's realization. The indices of tokens belonging to a syntactic role $r$ in a sentence $w^{(j)}$ are denoted $\argr(w^{(j)})$. For each syntactic role $r$ and sentence $w^{(j)}$, we discard 
inputs where this syntactic role cannot be found, and denote the remaining number of samples $T'^{enc}_{r}$. The resulting measure of influence of syntactic role $r$ on variable $z_l$ is denoted $\Gamma^{enc}_{rl}$.
The whole process yields matrix $\Gammaopenc$ of shape $(|R|, L)$ which summarizes interactions in the \emph{encoder} between syntactic roles and latent variables:
\begin{align}
    &\Gamma^{enc}_{rl} =
    \sum_{j=1}^{T'^{enc}_{r}} \frac{\mathbf{1}{\{\displaystyle \argmax_{l}(a^{(j)}_{li})\in \argr(w^{(j)})\}}}{T'^{enc}_{ r}}
\end{align}

\subsection{Disentanglement Metrics}\label{DISENTMETRICS}

For $\Gamma^*$ (either $\Gammaopdec$ or $\Gammaopenc$) each line corresponds to a syntactic role $r\in R$.
The disentanglement metric for role $r$ is the following:
\begin{align}
    \Delta\Gamma^*_{r} &= \Gamma^*_{rm_1} -\Gamma^*_{rm_2}\label{DELTA}\\
    s.t. \hspace{0.2cm}
    m_1 &= \argmax_{1\leq l\leq L} \Gamma^*_{rl}, \hspace{0.2cm}
    m_2 =  \argmax_{1\leq l\leq L, l\neq m1} \Gamma^*_{rl}
\end{align}

We calculate total disentanglement scores for syntactic roles using  $\Gammaopdec$, $\Gammaopenc$ as follows:
\begin{align}
    \mathbb{D}_{dec} = \sum_{r \in R} \Delta{\Gamma}^{dec}_{r}\hskip 1mm, \hskip 2mm 
    \mathbb{D}_{enc} = \sum_{r \in R} \Delta{\Gamma}^{enc}_{r} 
\end{align}

In summary, the more each syntactic role's information is concentrated in a single variable, the higher the values of $\mathbb{D}_{dec}$ and $\mathbb{D}_{enc}$.
However, similar to MIG, these metrics do not say whether variables capturing our concepts of interest are \emph{distinct}. Therefore, we also report the number of distinct variables that capture the most
each syntactic role (\textit{i.e} the number of distinct values of $m_1$ in
Eq.~\ref{DELTA} when looping over $r$). This is referred to as $N_{\Gammaopenc}$
for the encoder and $N_{\Gammaopdec}$ for the decoder.

\section{Experiments}
\label{EXPE}

\subsection{Experimenting on Regularly Structured Sentences}
\label{REGEXPADVAE}
\paragraph{Dataset} Previous unsupervised disentanglement works 
~\citep{Higgins2019-VAE:Framework, Kim2018DisentanglingFactorising, Li2020ProgressiveRepresentations} tend to use relatively homogeneous 
and low complexity data.
The data has \textit{low complexity} if it varies along clear factors which correspond to what the model aims 
to disentangle.
Similarly, we use a dataset where samples exhibit low variance in terms of syntactic structure\footnote{Statistics regarding syntactic diversity for SNLI are provided in section~\ref{YELP} with a comparison to a more syntactically diverse dataset.} while providing a high diversity of realizations for the syntactic roles composing the sentences, which is an adequate test-bed for unsupervised disentanglement of syntactic roles' realizations. This dataset is the plain text from the SNLI dataset~\citep{bowman-etal-2015-large}  extracted\footnote{
\url{github.com/schmiflo/crf-generation/blob/master/generated-text/train}}  by \citet{Schmidt2020AutoregressiveLoops}.
The SNLI data is a collection of premises (on average $8.92\pm 2.66$ tokens long) made for Natural Language Inference. We use 90K samples as a training set, 5K for development, and 5K as a test set.

\paragraph{Setup}
Our objective is to check whether the architecture of our ADVAE induces better syntactic role disentanglement. We compare it to standard Sequence VAEs 
~\citep{Bowman2016GeneratingSpace} and to a Transformer-based baseline that doesn't use Cross-Attention. Instead of Cross-Attention, this second baseline uses mean-pooling over the output of a Transformer encoder for encoding. For decoding, it uses the latent variable as a first token in a Transformer decoder, as is done for conditional generation with GPT-2~\cite{santhanam-shaikh-2019-emotional}.
These comparisons are performed using the same $\beta$-VAE objectives, and the decoder disentanglement scores as metrics.  Training specifics and hyper-parameter settings are 
 detailed in Appendix~\ref{TRAINING&HP}.  For each of the two baselines, the latent variables we vary during the decoder's 
 evaluation are the mono-dimensional components of its latent vector. It is easier to pack information about the realizations of multiple syntactic roles into $D_z$ dimensions than into a single dimension. Consequently, the single dimensions we study for the baselines should be at an advantage to separate information into different variables.

Scoring disentanglement on the encoder side will not be possible for the baselines above as it requires attention values. 
To establish that our model effectively tracks syntactic roles, we compare it to a third baseline that locates each syntactic role 
through its median position across the dataset. This baseline is fairly strong on short sentences from a language where word order is rigid (\textit{i.e} a configurational language) such as English. We refer to this Position Baseline as PB.

The scores are given for different values of $\beta$ (Eq.~\ref{BETAVAE}). Raising $\beta$ lowers the expressiveness of
 latent variables, but yields better disentanglement~\citep{Higgins2019-VAE:Framework}. 
Following \citet{Xu2020OnSupervision}, we set $\beta$ to low values to avoid posterior collapse. In our case,
we observed that the models do not collapse for $\beta<0.5$. Therefore, we display results for $\beta \in \{0.3, 0.4\}$. We stop at 0.3 as lower values for $\beta$ result in poorer generation quality.
For our model we report performance for instances with $L=4$ (\emph{ours-4}) and $L=8$ (\emph{ours-8}).

\paragraph{Results}
The global disentanglement metrics are reported in Table~\ref{tab:results}.\footnote{Fine-grained scores are given in Appendix~\ref{FULLSYNRESULTS}.}
  \begin{table}[!htbp]
    \centering
    \caption{Disentanglement quantitative results for the encoder (enc) and the decoder (dec). $N_{\Gamma}$ indicates the number of separated syntactic roles, and $\mathbb{D}$ measures concentration in a single variable. Values are averaged over 5 experiments. The standard deviation is between parentheses. PB refers to the Position Baseline.}
    \resizebox{1.0\textwidth}{!}{%
    \begin{tabular}{|c|c||c|c||c|c|}
    \hline
    Model& $\beta$ & $\mathbb{D}_{enc}\uparrow$ &  $N_{\Gammaopenc}\uparrow$  &  $\mathbb{D}_{dec}\uparrow$ &  $N_{\Gammaopdec}\uparrow$\\
    \hline \hline
    
    \multirow{2}{*}{Sequence VAE}&0.3& -& -& 0.43\textcolor{gray}{(0.18)}& 1.70\textcolor{gray}{(0.48)}\\ 
    &0.4& -& -& 0.91\textcolor{gray}{(0.32)}& 1.40\textcolor{gray}{(0.52)}\\
    \hline
    \multirow{2}{*}{Transformer VAE}&0.3& -& -& 0.08\textcolor{gray}{(0.04)}& 3.00\textcolor{gray}{(0.71)}\\ 
    &0.4& -& -& 0.11\textcolor{gray}{(0.05)}& 3.80\textcolor{gray}{(0.45)}\\
    \hline
    \multirow{1}{*}{PB}& - & 0.98 \textcolor{gray}{(-)}& 3.00\textcolor{gray}{(-)}& - & - \\     \hline
    \hline
    \multirow{2}{*}{ours-4} &0.3& 1.48\textcolor{gray}{(0.15)}& 3.00\textcolor{gray}{(0.00)}& 0.78\textcolor{gray}{(0.10)}& 3.00\textcolor{gray}{(0.00)}\\ 
     &0.4& 1.43\textcolor{gray}{(0.79)}& 3.00\textcolor{gray}{(0.00)}& 0.84\textcolor{gray}{(0.10)}& 3.00\textcolor{gray}{(0.00)}\\
    \hline
    \multirow{2}{*}{ours-8} &0.3& 1.34\textcolor{gray}{(0.18)}& 3.80\textcolor{gray}{(0.45)}& 0.62\textcolor{gray}{(0.17)}& 3.20\textcolor{gray}{(0.45)}\\ 
     &0.4& 1.75\textcolor{gray}{(0.47)}& 2.80\textcolor{gray}{(0.45)}& 0.80\textcolor{gray}{(0.11)}& 3.00\textcolor{gray}{(0.00)}\\

 \hline
     \end{tabular}}
    \label{tab:results}
  \end{table}
  
\begin{figure*}[!h]
\centering
    \begin{minipage}[b]{0.45\textwidth}
            \centering
           \begin{minipage}[b]{\textwidth}
            \begin{adjustbox}{minipage=\textwidth,scale=0.45}
            \hspace{ 1cm} \includegraphics[trim={1.3cm 0.7cm 2.2cm 1.3cm},clip] {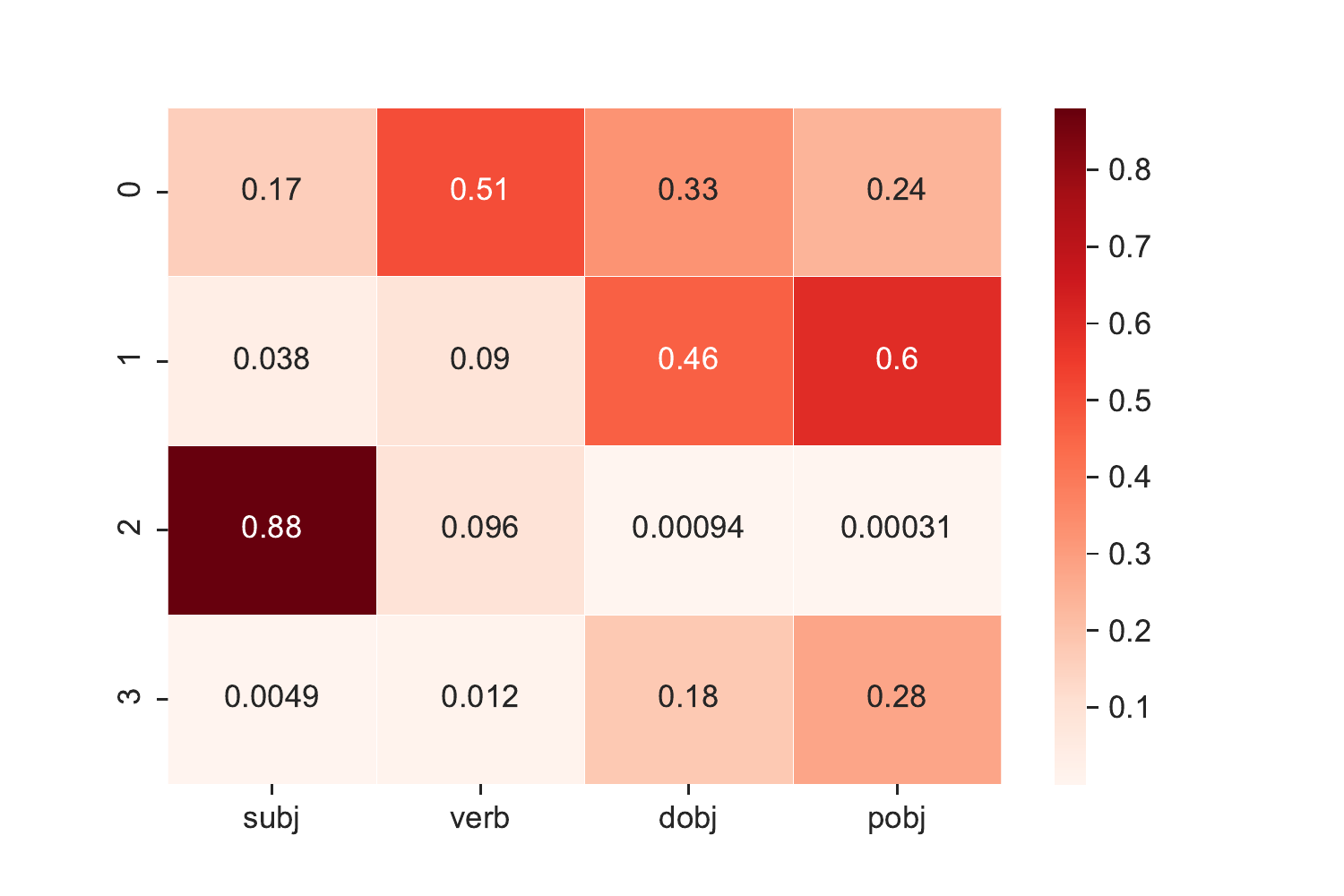}
            \end{adjustbox}
            \end{minipage}
            \caption{\centering Encoder influence heatmap ($\Gammaopenc$).}
            \label{fig:ENCHEAT}
    \end{minipage}
    \begin{minipage}[b]{0.45\textwidth}
            \centering
            \begin{minipage}[b]{\textwidth}
            \begin{adjustbox}{minipage=\textwidth,scale=0.45}
             \hspace{ 1cm} \includegraphics[trim={1.4cm 0.7cm 2.2cm 1.3cm},clip] {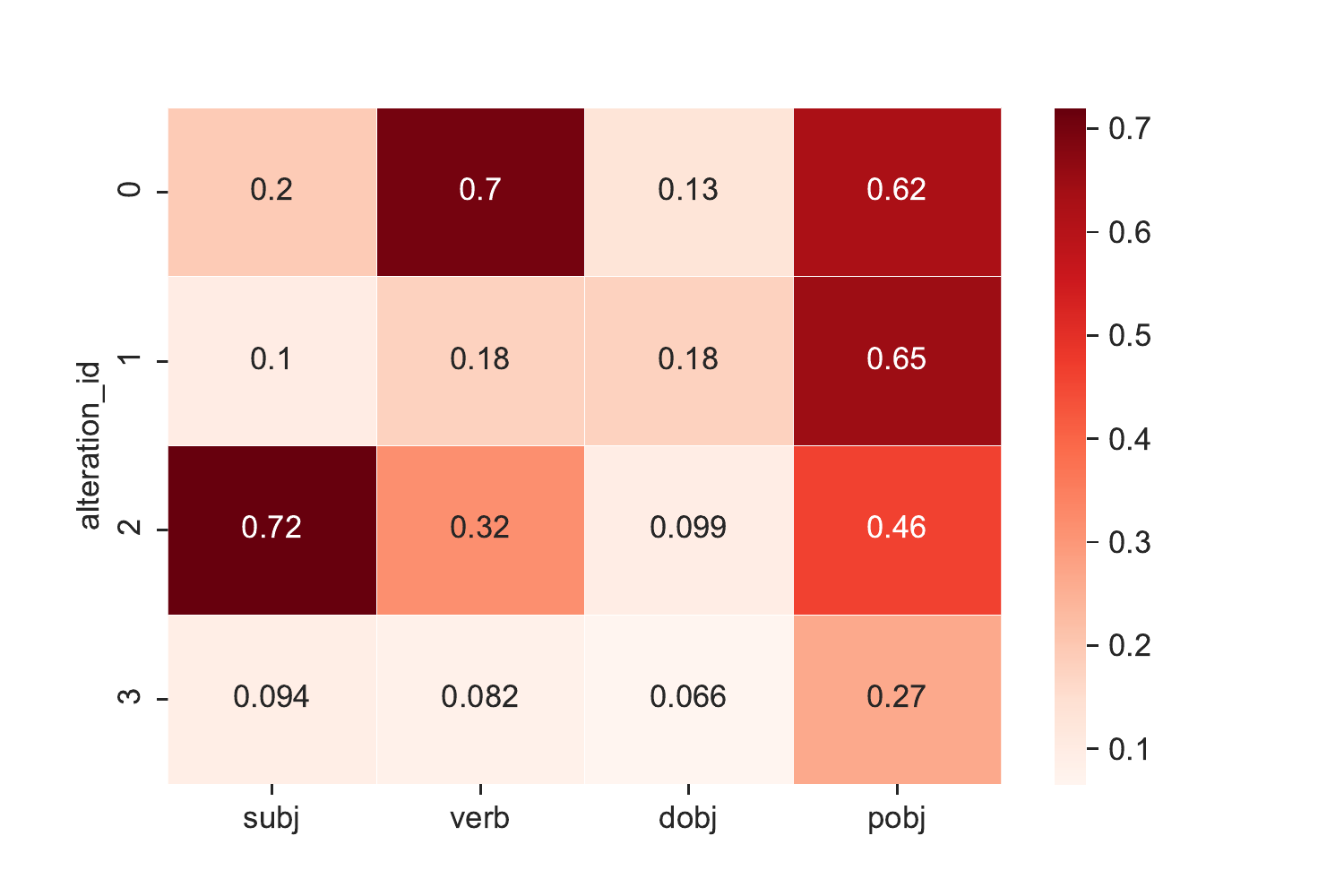}
            \end{adjustbox}
            \end{minipage}
            \caption{\centering Decoder influence heatmap ($\Gammaopdec$).}
            \label{fig:DECHEAT}
    \end{minipage}
\end{figure*} 
On the decoder side, the Sequence VAE exhibits disentanglement scores in the range of those reported for our model for $\beta=0.3$, and higher for $\beta=0.4$.
However, $N_{\Gammaopdec}$ shows that it controls the realizations of the 4 syntactic roles with less than 2 latent variables on average, meaning that it struggles to factor the realizations of different syntactic roles in different latent variables. The higher score shown for $\beta=0.4$ is accompanied by an even lower tendency to separate the information from different syntactic 
 roles (\textit{i.e.} lower $N_{\Gammaopdec}$). The Transformer VAE baseline assigns different latent variables to different syntactic roles (high $N_{\Gammaopdec}$), but suffers from very low specialization for these latent variables (low $\mathbb{D}_{dec}$). 
In contrast, our model is consistently able to separate 3 out of 4 syntactic roles, and a higher $\beta$ raises its $\mathbb{D}_{dec}$.
As \emph{ours-8} has more latent variables, this encourages the model to further split the information in each syntactic role between more latent variables\footnote{Results for a larger grid of $L$ values are reported in Appendix~\ref{NZVARY}.}.
The fact that ADVAEs perform better than both Sequence VAEs and classical Transformer VAEs shows that  its disentanglement capabilities are due to the usage of Cross-Attention to obtain latent variables, and not only to the usage of Transformers. On the encoding side, our models consistently score above the baseline, showing that our latent variables
 actively follow the syntactic roles.

In Figures~\ref{fig:ENCHEAT} and~\ref{fig:DECHEAT}, we display the influence matrices $\Gammaopenc$ and $\Gammaopdec$ for an 
instance of our ADVAE with $L=4$ as heatmaps.
The vertical axes correspond to the latent variables.
As can be seen, our model successfully associates latent variables to verbs and subjects 
but chooses not to separate direct objects and prepositional objects into different latent variables.
Upon further inspection of the same heatmaps for the Sequence VAE baseline, it appears that it most often uses a single latent variable for 
\emph{verb} and \emph{subj}, and another for \emph{dobj} and \emph{pobj}.

One can also notice in Figures~\ref{fig:ENCHEAT} and~\ref{fig:DECHEAT}, that the encoder matrix is sparser than the decoder matrix 
(which is consistent with the higher encoder disentanglement scores in Table~\ref{tab:results}).
This is to be expected since the decoder $p_\theta(w|z)$ adapts the realizations of syntactic roles to each other after they are
 sampled separately from $p(z)$.
The reason for this is that the language modeling objective requires some coherence between syntactic roles
 (conjugating verbs with subjects, changing objects that are semantically inadequate for a verb, etc).
This \emph{co-adaptation}, contradicts the independence of our latent variables.
It is further discussed in the following paragraph.

\begin{table*}[t]
    \small
    \centering
    \caption{Resampling a specific latent variable for a sentence. The ID column is an identifier for the example.}
    \begin{tabularx}{14cm}{|c|X|X|X|X|}
    \hline
     ID&Original sentence& Resampled subject& Resampled verb & Resampled dobj/pobj \\
    \hline \hline
     1& people are sitting on the beach & a young man is sitting on the beach & people are playing in the beach  & people are sitting on a bench
    \\\hline
     2&a man and woman are sitting on a couch & a man is sitting on a park bench &a man and woman are running on a grassy field   & the man and woman are on a beach\\\hline
     3&a man is playing with his dog & a boy is playing in the snow & a man is selling vegetables & a man is playing the game with his goal .\\\hline
     \end{tabularx}
    \label{tab:resultsresample}
\end{table*} 
\begin{table*}[t]
    \small
    \centering
    \caption{Swapping the value of a specific latent variable between two sentences. The SSR (Swapped Syntactic Role) column indicates the syntactic role that has been swapped.}
    \begin{tabularx}{14cm}{|p{0.2cm}|X|X|p{0.8cm}|X|X|}
    \hline
     ID& Sentence 1& Sentence 2& SSR & Swapped Sentence 1 & Swapped Sentence 2 \\
    \hline \hline
      1& a woman is talking on a train & people are sitting on the beach & subj & people are talking on a train & a woman is sitting on the beach \\\hline
      2&people are sitting on the beach  &   a woman is talking on a train& verb& people are talking on a beach&   a woman is standing on a train \\\hline
      3&a woman is talking on a train &   a man is playing with his dog  & \makecell{dobj/\\pobj}  & a man is playing the guitar with a goal & a woman is performing a trick\\\hline
      \end{tabularx}
    \label{tab:resultsswap}
\end{table*} 
\paragraph{Changing the realizations of syntactic roles}
\label{QUALITATIVE}
Here, we display a few qualitative examples of how the realizations of syntactic roles can be separately changed using an instance of our ADVAE.

As a first example, we generate a sentence from a random latent vector, then resample for each syntactic role the corresponding
 disentangled latent variable to observe the change on the subsequently generated altered sentence. The results of this manipulation 
 are in Table~\ref{tab:resultsresample}\footnote{More Examples are available in Appendix~\ref{QUALIAPPEN}}. As can be seen, some examples exhibit changes that only affect the target syntactic role 
 (example 1).
However, the model often produces co-adaptations that go past the target syntactic role either for semantic soundness
 (example 2, resampled verb adapts the object), or simply for lack of generalization 
 from 
 the SNLI data used for training.

A second example we display is a swap of syntactic role  realizations between sentences.
A few examples are given in Table~\ref{tab:resultsswap}. Similar to Table~\ref{tab:resultsresample}, the model often
yields the expected result. Co-adaptation is best seen here, as taking a syntactic role to a sentence with which 
it is incompatible results in unexpected changes (example 3).

The co-adaptation seen here is caused by the independence between our latent variables which leaves it to the the decoder $p_\theta(w|z)$ to correct the incoherence between independently sampled syntactic role realizations\footnote{We stress, here, that the co-adaptation we describe is different from entanglement. In fact, entanglement happens at the representation-level while co-adaptation can re-scramble, with the decoder, information that was separated in the representations.}. Using structured latent variables to learn relations between syntactic roles seems to be the natural solution to this problem. Accordingly, the next section describes an investigation of a hierarchical version of the ADVAE. 

\subsection{A Hierarchical Version of our ADVAE}
\label{HIERARCH}

\begin{figure}[!h]
\centering
    \begin{minipage}[b]{0.33\textwidth}
    \begin{adjustbox}{minipage=\textwidth,scale=0.6}
        \fbox{\begin{tikzpicture}
        \tikzstyle{main}=[rounded corners, minimum size = 10mm, thick, draw =black!80, node distance = 5mm, font=\Large, text centered]
        \tikzstyle{connect}=[-latex, thick]
        
        \node[main,text width=3cm, color=blue!100] (source) [] {$z^{m-1}_{1}, z^{m-1}_{2},$ \\$..., z^{m-1}_{L}$};
        \node[main,text width=2cm, color=blue!100] (spe) [above=of source] {z-identifier};
        \node[main,text width=2.5cm] (enc) [above=of spe] {Transformer Encoder};
          
        \node[main,text width=4cm, color=blue!100] (ptarget) [right=5 mm of source] {$e_{1}^{m}, e_{2}^{m},..., e_{L}^{m}$};
        
        \node[minimum height = 5.7cm, minimum width = 5.5cm, thick, draw =black!80, node distance = 5mm, font=\Large, align=left, dashed, yshift=1.4cm] (dec) [above=of ptarget] {};
        \node[main,text width=5cm, align=left, yshift=1.6cm] (att) [above=of ptarget] {\textcolor{blue}{Self-Attention} + Cross-Attention to align source information with latent variables};
        \node[main,text width=5cm] (mlp) [above=of att] {MLP};
        \node[minimum width = 2.8cm, thick, draw =black!80, node distance = 5mm, font=\Large, align=left, minimum height = 1cm,  text width=2.8cm, xshift=-12mm, dashed] (declab) [above=of mlp] {Transformer\\ Decoder};
        
        \node[main,text width=3.4cm, xshift=-2.3cm, yshift=0.1cm, color=blue!100] (result1) [above=20 mm of mlp] {$\mu^{m}_{1}, \mu^{m}_{2}, ..., \mu^{m}_{L}$};
        \node[main,text width=3.4cm, color=blue!100] (result2) [right=1mm of result1] {$\sigma^{m}_{1}, \sigma^{m}_{2}, ..., \sigma^{m}_{L}$};
        
       \path (source) edge [connect, color=blue!100] (spe)
            (spe) edge [connect] (enc)
            (enc) edge [bend left][connect] (att)
            (ptarget) edge [connect, color=blue!100] (att)
            (att) edge [connect] (mlp)
            (mlp) edge [bend right][connect, color=blue!100] (result1)
            (mlp) edge [bend right][connect, color=blue!100] (result2);
        \end{tikzpicture}}
    \end{adjustbox}
    \vskip -5px
    \caption*{(b)}
    \end{minipage}
    \caption{The conditional inference module linking each of the hierarchy levels in our prior with the next level $p_\theta(z^m|z^{m-1})$. This module treats latent variables from previous layers as they are treated in our original decoder, and generates parameters for latent variables in subsequent hierarchy levels as it is done in our encoder.}
    \label{fig:DEEPSCHEME}
    
\end{figure}
As we stated, our ADVAE aims to factor sentences into independent latent variables.
However, given the dependency structure of sentences,  realizations of syntactic roles are known to be interdependent to some degree in general. 
Therefore one may think that a structured latent variable model would be better suited to model the realizations of syntactic roles. In fact, such a model could absorb the language modeling co-adaptation between syntactic roles. For instance, instead of sampling an object and a verb from $p(z)$ that are inadequate, then co-adapting them through $p_\theta(w|z)$, a structured $p_\theta(z)$ could  produce an \emph{adequate} object for the verb.
For this experiment, rather than using an independent prior $p(z)$, we use a structured prior 
$p_\theta(z) = p(z^0) \prod^M_{m=1} p_\theta(z^m|z^{m-1})$  where $p(z^0)$ is a standard Gaussian, and all subsequent
 $M-1$ hierarchy levels are parameterized by learned conditional diagonal Gaussians. The model used for each $p_\theta(z^m|z^{m-1})$ 
 is shown in Figure~\ref{fig:DEEPSCHEME}.

We display the results for $M=2$ and $M=3$ in Table~\ref{tab:resultsDeep}. For both models, we set $L$ to 4.

\begin{table}[!h]
    \centering
    \caption{Disentanglement results for structured latent variable models on SNLI. }
    \resizebox{1.0\textwidth}{!}{%
    \begin{tabular}{|c|c||c|c||c|c|}
    \hline
    Depth & $\beta$ & $\mathbb{D}_{enc}$ &  $N_{\Gammaopenc}$  &  $\mathbb{D}_{dec}$ &  $N_{\Gammaopdec}$\\
    \hline \hline
    \multirow{2}{*}{$M=2$} &0.3& 0.79\textcolor{gray}{(0.36)}& 3.60\textcolor{gray}{(0.55)}& 0.69\textcolor{gray}{(0.22)}& 2.60\textcolor{gray}{(0.55)}\\
     &0.4& 0.42\textcolor{gray}{(0.23)}& 2.80\textcolor{gray}{(0.45)}& 0.61\textcolor{gray}{(0.08)}& 2.40\textcolor{gray}{(0.55)}\\
    \hline
    \multirow{2}{*}{$M=3$} &0.3& 0.90\textcolor{gray}{(0.25)}& 3.14\textcolor{gray}{(0.69)}& 0.60\textcolor{gray}{(0.18)}& 2.80\textcolor{gray}{(0.45)}\\
     &0.4& 0.32\textcolor{gray}{(0.38)}& 2.75\textcolor{gray}{(0.50)}& 0.59\textcolor{gray}{(0.20)}& 3.20\textcolor{gray}{(0.84)}\\

 \hline
     \end{tabular}}
    \label{tab:resultsDeep}
  \end{table}
The results show lower mean disentanglement scores, and high standard deviations compared to the standard version of our ADVAE.
By inspecting individual training instances of this hierarchical model, we found that some instances achieve disentanglement
with close scores to those of the standard ADVAE, while others completely fail, which results in the high variances observed 
in Table~\ref{tab:resultsDeep}. 
Unfortunately, hierarchical latent variable models are notoriously difficult to train~\citep{Zhao2017LearningModels}. Our independent latent variable model is therefore preferable to the structured one due to these empirical results. More advanced hierarchical latent variable training techniques (such as Progressive Learning and Disentanglement~\citep{Li2020ProgressiveRepresentations}) may, however, provide better results. We plan to investigate this in our future works (\textit{cf.} \S~\ref{CONCPERP}).

\subsection{Experimenting with the Yelp Dataset}
\label{YELP}
 
As this is a first step in this research direction, we conducted this study on a dataset of relatively regular sentences. In this section, we aim to investigate the behavior of our ADVAE on the user-generated reviews from the Yelp dataset used in \citet{li-etal-2018-delete} using the same procedure we used for SNLI.

\begin{figure}[!t]
\centering
    \begin{minipage*}[b]{1.0\textwidth}
            \centering
           \begin{minipage}[b]{\textwidth}
            \begin{adjustbox}{minipage=\textwidth,scale=0.5}
            \hspace{ 1cm} \includegraphics[trim={1.3cm 5.7cm 2.2cm 3cm},clip] {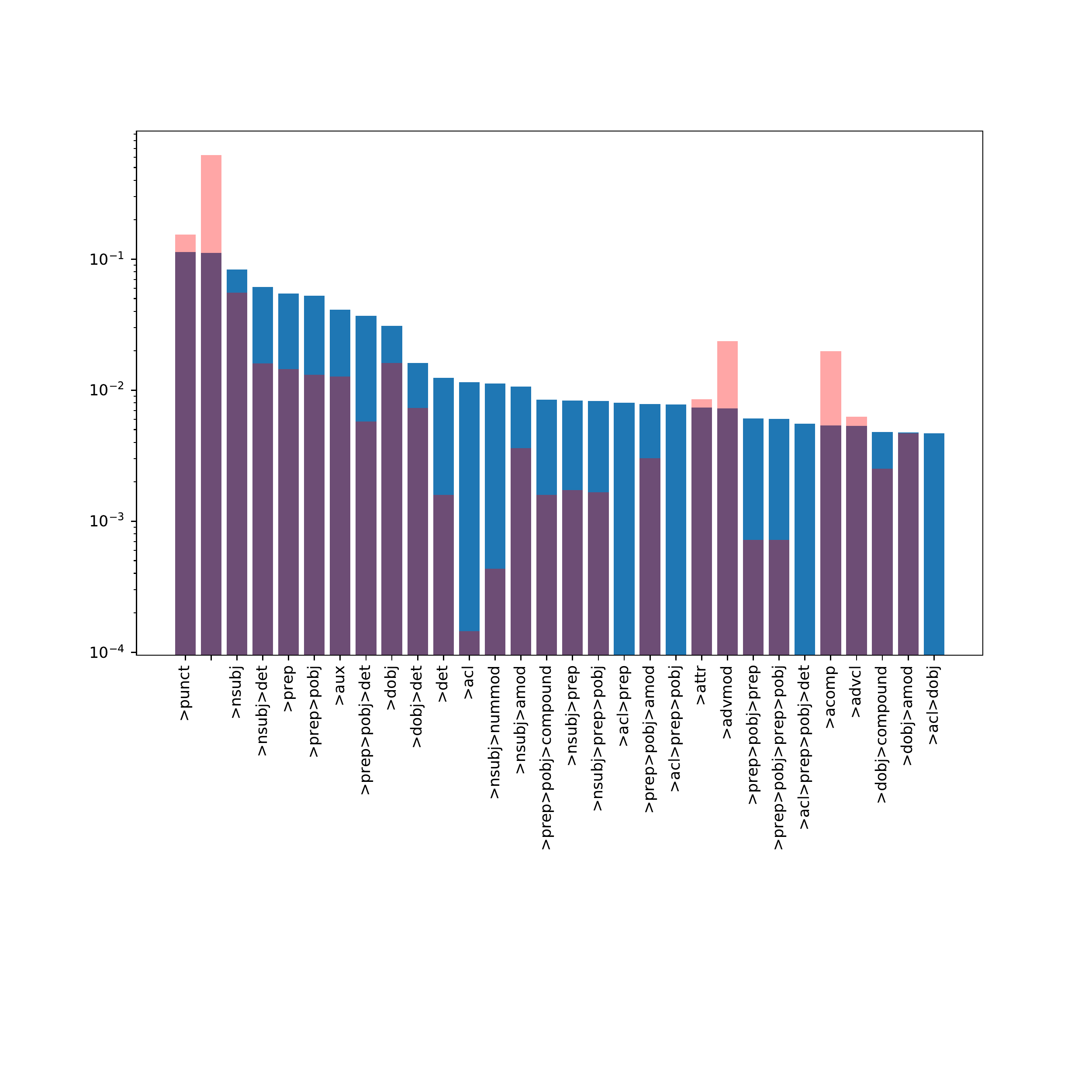}
            \end{adjustbox}
            \end{minipage}
    \end{minipage*}\\
    \begin{minipage*}[b]{1.0\textwidth}
            \centering
            \begin{minipage}[b]{\textwidth}
            \begin{adjustbox}{minipage=\textwidth,scale=0.5}
             \hspace{ 1cm} \includegraphics[trim={1.3cm 5.7cm 2.2cm 2.8cm},clip] {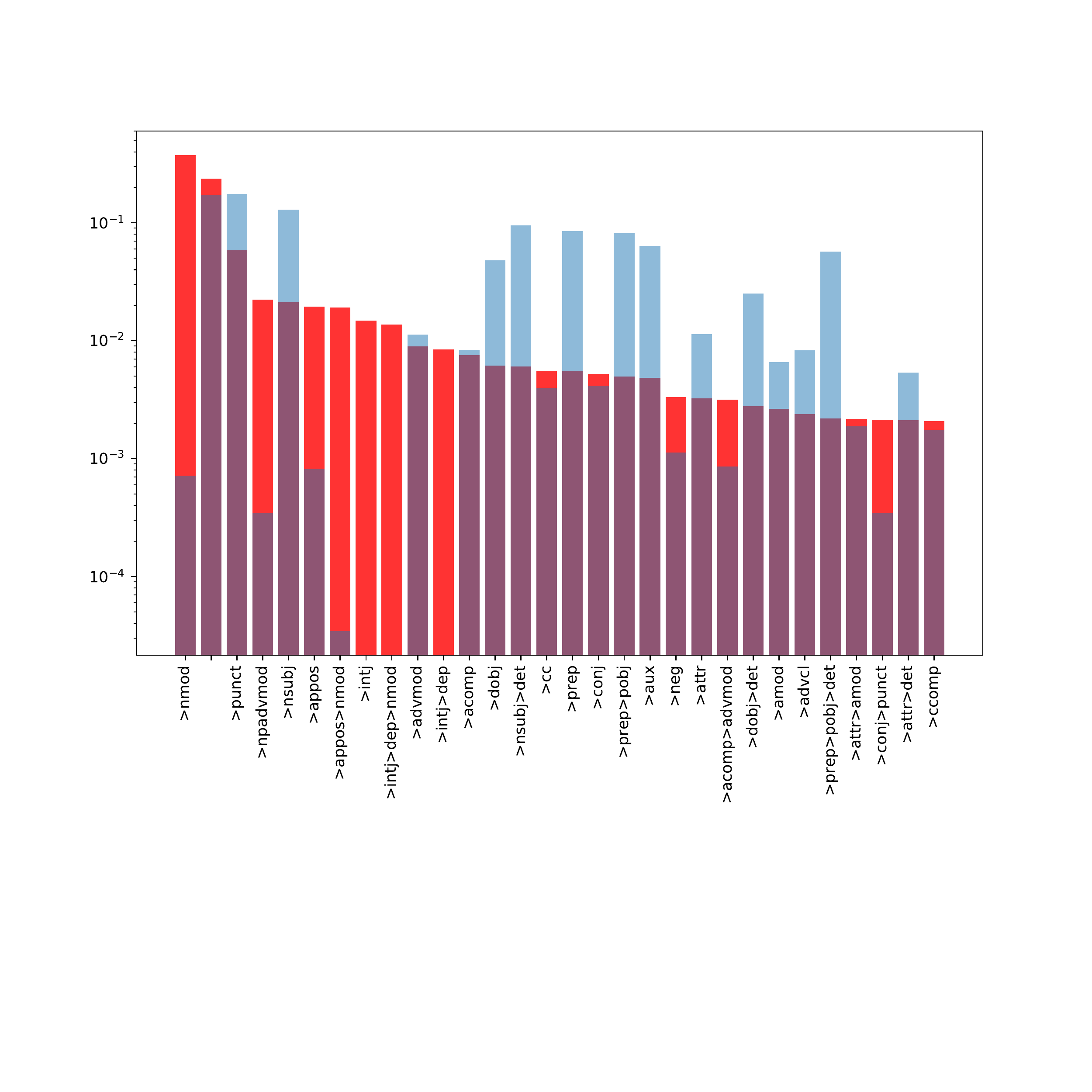}
            \end{adjustbox}
            \end{minipage}
    \end{minipage*}
    \caption{Top 30 dependency paths for each of SNLI (top, blue bars) and Yelp (bottom, red bars) datasets. In the top figure, the frequency of each dependency path among SNLI's top 30 is plotted in red with low opacity for Yelp to ease comparison. Analogously, low opacity blue bars are displayed in the bottom figure for SNLI's frequencies. The columns with an empty label corresponds to the root (empty) dependency path.}
    \label{fig:DEPSTATS}
\end{figure} 

\paragraph{Contrasting SNLI to YELP: } 
The length of sentences from this dataset ($8.88\pm3.64$) is similar to the length of sentences from the SNLI dataset. However, contrary to SNLI, sentences presenting the regular Subject-Verb-Object (SVO) structure are much less common. To emphasize this difference, we label tokens in each dataset by the path that leads to them in the dependency tree and calculate the frequency (\textit{i.e.} chances it appears in a sentence) of each one of such labels in SNLI and Yelp datasets. For example, in the sentence \textit{"The man plays Football"}, \textit{The} is labeled with \textit{>nsubj>det}, meaning it is a determinant to a nominal subject beneath the root\footnote{The root is ommitted in these labels for brevity.} of the sentence. In Figure~\ref{fig:DEPSTATS}, we display the top 30 labels in each dataset ranked by their frequency.

As can be seen in the figure, the frequency of core syntactic roles in SNLI is markedly higher than that of other remaining syntactic roles. Yelp, however, displays very different statistics where the core syntactic roles are much more rare, and where distributions of syntactic roles is much flatter, making it more difficult to learn syntactic regularities.

\paragraph{Results}
Similar to the experiments in the main body of the paper, we display the disentanglement scores in Table~\ref{tab:resultsYelp}, 
and the influence metrics of one of the instances of our model as heatmaps in Figures~\ref{fig:ENCHEATYELP} and~\ref{fig:DECHEATYELP}.

\begin{table}[!h]
    \centering
    \caption{Disentanglement results for the Yelp dataset}
    \resizebox{1.0\textwidth}{!}{%
    \begin{tabular}{|c|c||c|c||c|c|}
    \hline
    Model& $\beta$ & $\mathbb{D}_{enc}$ &  $N_{\Gammaopenc}$  &  $\mathbb{D}_{dec}$ &  $N_{\Gammaopdec}$\\
    \hline \hline
    \multirow{2}{*}{Sequence VAE}& 0.3 & - & - & 0.49\textcolor{gray}{(0.04)}& 2.00\textcolor{gray}{(0.00)}\\
    & 0.4&- & - & 0.49\textcolor{gray}{(0.06)}& 1.8\textcolor{gray}{(0.84)}\\
    \hline
    \multirow{2}{*}{Transformer VAE}&0.3& -& -& 0.11\textcolor{gray}{(0.02)}& 2.80\textcolor{gray}{(0.45)}\\ 
    &0.4& -& -& 0.11\textcolor{gray}{(0.05)}& 3.00\textcolor{gray}{(0.71)}\\
    \hline
    \multirow{1}{*}{PB}& - & 0.33\textcolor{gray}{(-)} & 2.00\textcolor{gray}{(-)} & - & - \\
    \hline \hline
    \multirow{2}{*}{ours-4} &0.3& 0.48\textcolor{gray}{(0.07)}& 2.00\textcolor{gray}{(0.00)}& 0.23\textcolor{gray}{(0.09)}& 2.20\textcolor{gray}{(0.45)}\\
     &0.4& 0.54\textcolor{gray}{(0.04)}& 3.00\textcolor{gray}{(0.00)}& 0.22\textcolor{gray}{(0.08)}& 2.20\textcolor{gray}{(0.45)}\\
    \hline
    \multirow{2}{*}{ours-8} &0.3& 0.44\textcolor{gray}{(0.04)}& 3.80\textcolor{gray}{(0.45)}& 0.17\textcolor{gray}{(0.09)}& 3.00\textcolor{gray}{(0.00)}\\
     &0.4& 0.57\textcolor{gray}{(0.26)}& 3.40\textcolor{gray}{(0.55)}& 0.25\textcolor{gray}{(0.10)}& 2.80\textcolor{gray}{(0.84)}\\

 \hline
     \end{tabular}}
    \label{tab:resultsYelp}
  \end{table}

\begin{figure*}[!h]
\centering
    \begin{minipage}[b]{0.48\textwidth}
            \centering
           \begin{minipage}[b]{\textwidth}
            \begin{adjustbox}{minipage=\textwidth,scale=0.5}
            \hspace{ 1cm} \includegraphics[trim={1.3cm 0.7cm 2.2cm 1.3cm},clip] {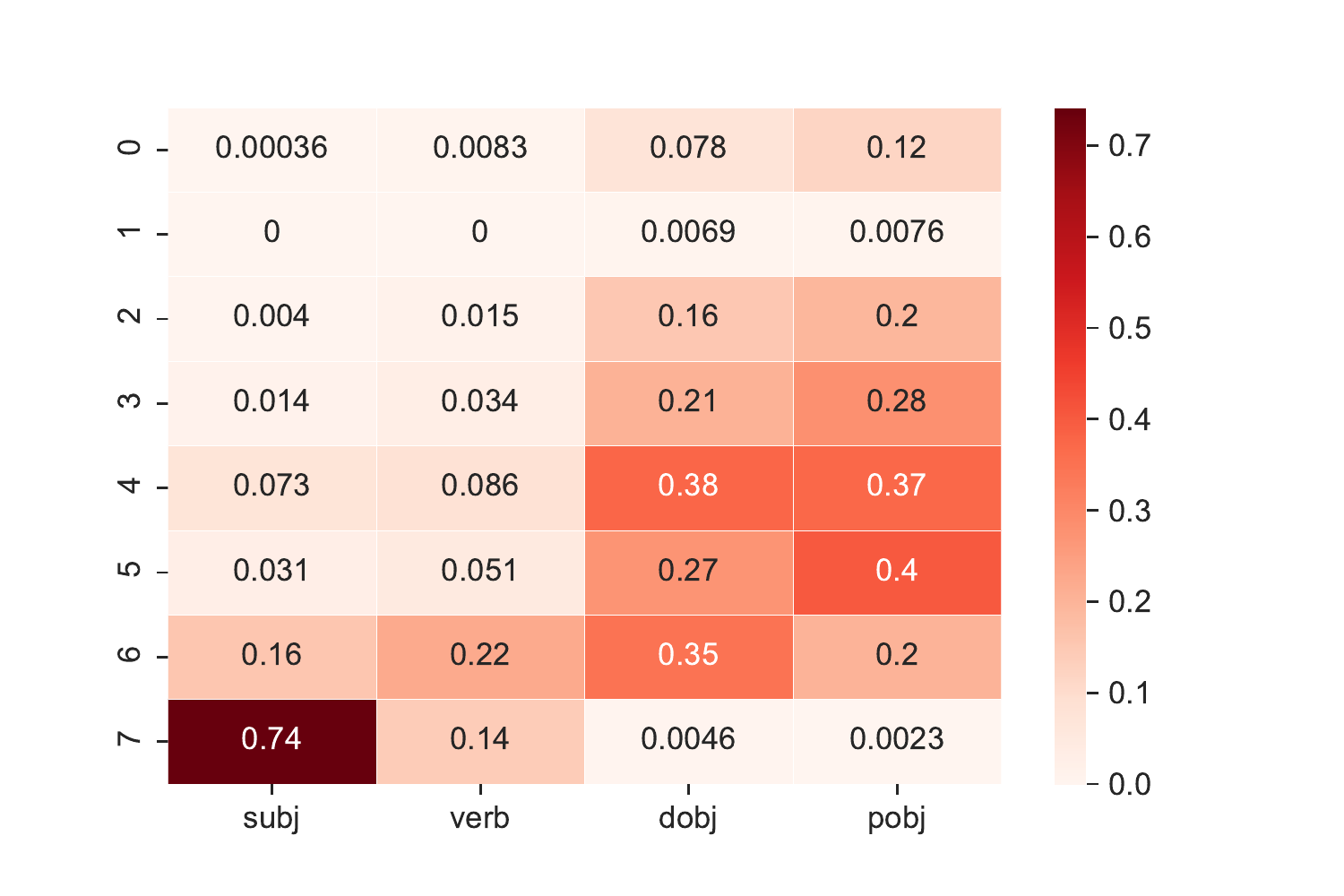}
            \end{adjustbox}
            \end{minipage}
            \caption{\centering Encoder influence heatmap for Yelp($\Gammaopenc$).}
            \label{fig:ENCHEATYELP}
    \end{minipage}
    \begin{minipage}[b]{0.48\textwidth}
            \centering
            \begin{minipage}[b]{\textwidth}
            \begin{adjustbox}{minipage=\textwidth,scale=0.5}
             \hspace{ 1cm} \includegraphics[trim={1.3cm 0.7cm 2.2cm 1.3cm},clip] {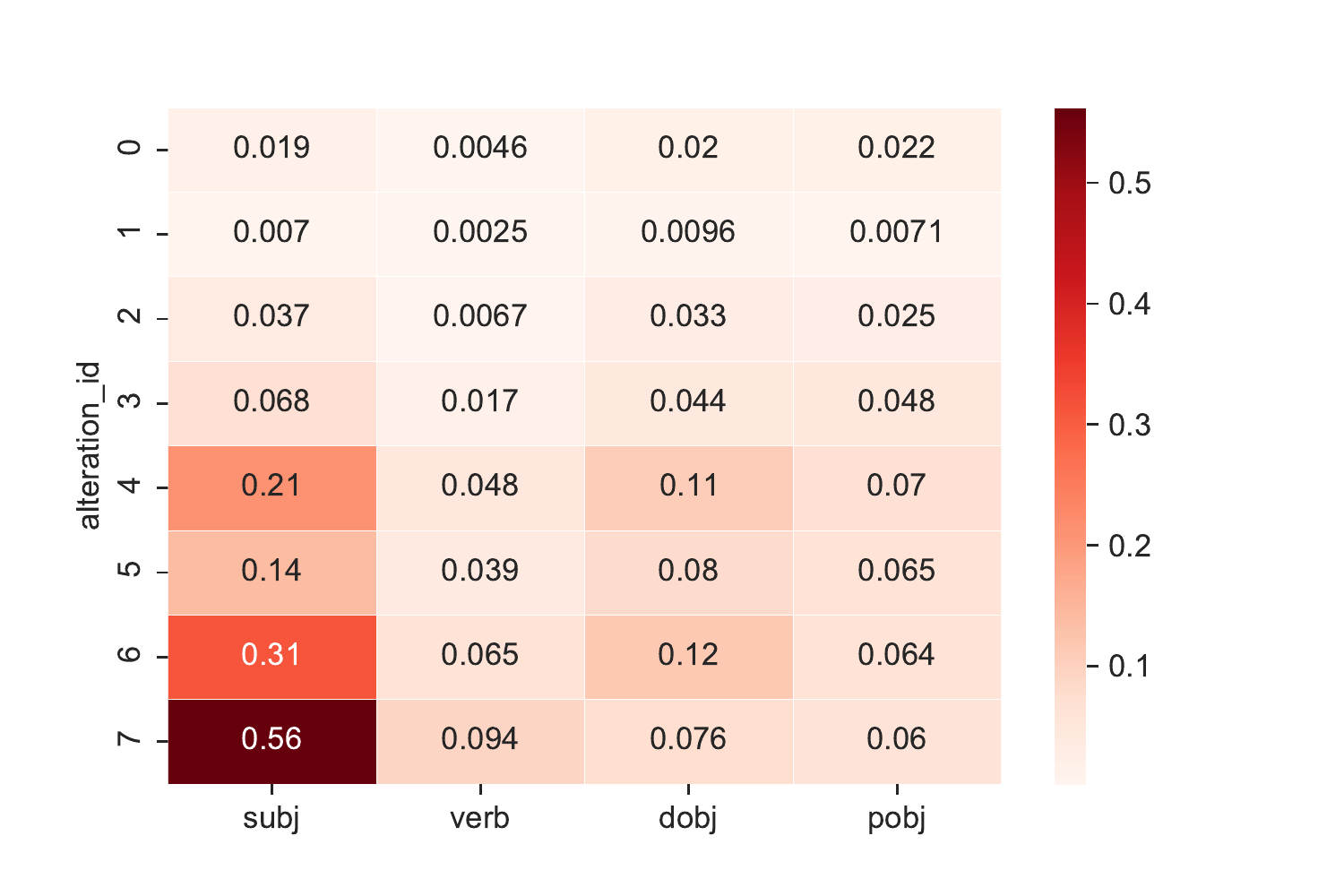}
            \end{adjustbox}
            \end{minipage}
            \caption{\centering Decoder influence heatmap for Yelp($\Gammaopdec$).}
            \label{fig:DECHEATYELP}
    \end{minipage}
\end{figure*}

Although the results show similar trends, they are weaker than what we obtained for SNLI. Given the difference between SNLI and Yelp (also discussed in
Appendix~\ref{EXAMPLESEXTRACTIONS}) there are two clear reasons for this decrease. The first is that Yelp is a dataset where it is
 harder to locate the 
syntactic roles. This is illustrated by the fact that the PB baseline obtains a much lower score. The second is that 
our syntactic role extraction heuristics are tailored for regular sentences with verbal roots, which subjects the evaluation metrics on Yelp 
to a considerable amount of noise. Nevertheless, the comparisons between a Sequence VAE, a Transformer VAE, an ADVAE, and PB retain the same conclusions, but with lower margins and some 
overlapping standard deviations.


Through manual inspection of examples, we observed that the occurrence of various syntactic phenomena (enumerations, sentences with nominal roots, presence of coordinating conjunctions, etc) was controlled by different latent variables. This calls for a model that provide ways to separate structural information, \textit{i.e.} information pertaining to variations in syntax, from content-related information, \textit{i.e.} information pertaining to variations in the lexical realization of the observed syntactic functions.

\section{Related Works}

\label{RELATED}
\paragraph{Linguistic information in neural models} 
Accounting for linguistic information so as to design neural networks with  beneficial inductive bias has been a successful trend in NLP system design during recent years. For instance, successful attempts at capturing linguistic information with neural models helped improve grammar induction (RNNG;  \citealp{Dyer2016RecurrentGrammars}),
 constituency parsing and language modeling (ON-LSTM; \citealp{Shen2019OrderedNetworks}, ONLSTM-SYD; \citealp{Du2020ExploitingApproach}), as well as controllable generation (SIVAE; \citealp{Zhang2020Syntax-infusedGeneration}). 
 The evaluation protocol we present also relates our work to research that
 dives into the linguistic capabilities of neural NLP models (\textit{cf.}~\ref{TRANSFORMERLMBGANAL}). However, 
   such studies most often rely on structural probes~\citep{jawahar-etal-2019-bert, liu-etal-2019-linguistic, hewitt-manning-2019-structural} to explain representations, probes which are not without issues, as shown by  \citet{pimentel-etal-2020-information}.
In that regard, the generative capabilities and the attention mechanism of our model offer an alternative to probing: analysis is performed directly on sentences generated by the model and on internal attention values.

\paragraph{Disentanglement in NLP}
As discussed in \S~\ref{DISENTAPPPBGSEC}, the main line of work in this area revolves around using multitask learning to 
separate concepts in neural representations (\textit{e.g.} style vs content \citealp{John2020DisentangledTransfer}; syntax vs 
semantics \citealp{Chen2019ARepresentations, Bao2020}) and literature on \emph{unsupervised} disentanglement in NLP remains scarce~\cite{Xu2020OnSupervision, Behjati2021InducingAttention}. The work of \citet{Behjati2021InducingAttention} is closest to ours as it uses Slot Attention~\citep{Locatello2020Object-centricAttention}, a Cross-Attention-based representation technique,  to induce meaningful units from character sequences. Although developed with a different goal in mind, Perceiver IO~\citep{jaegle2022perceiver}, a work concurrent to ours, also shares similarities with our work given that it uses Cross-Attention to encode and decode latent vectors. 

Our contribution differs from previous work in that \emph{i)} syntactic parses are not used as learning signals but as a way to interpret our model, and \emph{ii)} Cross-Attention enables our model to link a fixed number of latent variables to text spans.
   
\section{Conclusion}

The work presented in this chapter is the first part of our contribution on unsupervised disentanglement of sentence representations. Specifically, we study the hypothesis that lexical realizations of \textit{core} syntactic roles can be disentangled \textit{without supervision} (\S~\ref{SyntacticRoles}). Our framework includes: \textit{i)} our model, the ADVAE (\S~\ref{PROPOSEDMODEL}), which maps sentences to vectorial latent variables and allows for the use of attention to study the interaction between latent variables and spans; \textit{ii)} an evaluation protocol to quantify disentanglement between latent variables and spans both in the encoder and in the decoder (\S~\ref{EVALUATIONPROC}).

Using our evaluation protocol we show in Section~\ref{REGEXPADVAE} that, when trained on a dataset of regularly structured sentences, ADVAE learns  representations of sentences which exhibit a significant separation in the realizations of core syntactic functions without supervision. We also show that it separates syntactic roles to more latent variables than standard Sequence VAEs and with better concentration than standard Transformer VAEs.

This study constitutes a first step in a promising process towards \emph{unsupervised} explainable modeling and fine-grained control over the lexical realizations of core syntactic roles in sentences. Although we focused on syntactic roles realizations, this architecture as well as the evaluation method are generic and could be applied to other tasks. For example, the architecture could be used at the document level (\emph{e.g.} disentangling discourse relations), while the evaluation protocol could be applied to other spans such as constituents.

The limitations of ADVAE revealed in the last sections are twofold: \textit{i)} a phenomenon we called \textit{co-adaptation} where the independently sampled syntactic roles are subject to corrections by the decoder, which hurts our disentanglement metrics; \textit{ii)} a degradation of disentanglement performance when dealing with a dataset that does not exhibit regular syntactic structures such as Yelp (\S~\ref{YELP}). We have shown in section~\ref{HIERARCH} that simply adding structure to our probabilistic graphical model so as to learn relations between our latent variables is not sufficient to deal with limitation \textit{i)}. As for limitation \textit{ii)}, it is likely to originate from the fact that our model does not model syntactic variation. In the next chapter, we introduce a new latent variable that models syntactic information (again without supervision) and addresses this limitation.
}
\chapter{Exploiting Inductive Bias in Transformers for Unsupervised Disentanglement of Syntax and Semantics with VAEs}\label{chap:QKVCHAP}
{\fancyhead[RO]{CHAPTER \theHchapter.  UNSUPERVISED DISENTANGLEMENT OF SYNTAX AND SEMANTICS}

This chapter is the continuation of our work on unsupervised disentanglement of sentence representations. It focuses on a form of disentanglement that we discussed in Section~\ref{DISENTAPPPBGSEC} and that received a lot of interest from the
  NLP community: the separation between syntax and semantics in neural representations~\cite{Chen2019ARepresentations, chen-etal-2019-controllable, Bao2020, 
  Zhang2020Syntax-infusedGeneration, Huang2021GeneratingPairs, Huang2021DisentanglingModelsb}.
  The interest in this separation is mainly justified by the fact that purely semantic representations enable better paraphrase detection, and separation between syntax and semantics in an Seq2seq models enables paraphrase generation by changing syntax while keeping the same semantics.  
Previous works perform disentanglement using paraphrase pairs as information for semantics, and/or constituency parses as information
 for syntax.
In general, the dependence of models on labeled data is known to often entail high cost \citep{bohmova2003prague, martinez-alonso-etal-2016-noisy,choudhary2018cost,Seddah2020BuildingHell}, and to often require new labels to handle problems such as concept drift (\textit{i.e.} changes in the relation between observation and label; \citealp{Lu2019LearningReview}) and domain adaptation (changes in the distribution of the observations; \citealp{Farahani2021AAdaptation}).
 
In light of the the difficulties incurred by the use of annotated data, we propose in this chapter an unsupervised model which directs syntax and semantics into different 
neural representations without semantic or syntactic information. In the Transformer architecture~\cite{Vaswani2017} presented in chapter~\ref{TRANSCHAP}, the attention 
mechanism is built upon a \emph{query} from a set $Q$, which pools \emph{values} $V$ through \emph{keys} $K$. For each query, 
values are selected according to their matching score computed by the similarity between their corresponding keys and 
the query. Building on an analogy between the $(K, V)$ couple and syntactic roles with their lexical realizations (explicited in 
\S~\ref{Behavior}) we present QKVAE\footnote{A contraction of the $(Q, K, V)$ triplet with the VAE acronym.}, a Transformer-based 
VAE.

To build our model, we modify ADVAE, the model presented in the previous chapter.
Using Cross-Attention, QKVAE encodes sentences into two latent variables: $z^{sem}$ to infer values for $V$, and $z^{syn}$ to
 assign keys in $K$ for values in $V$.
These keys and values are then used in the attention mechanism of a Transformer Decoder to generate sentences.
We show that $z^{syn}$ tends to contain syntactic information, while $z^{sem}$ tends to represent semantic information.
Additionally, comparisons with a supervised model show that it needs a considerable amount of data to outperform our model
 on syntactic and semantic transfer metrics. Finally, we confirm the hypothesis formulated about syntactic information hindering syntactic role disentanglement with ADVAE's latent variables in the previous chapter, and show that QKVAE displays better syntactic role disentanglement metrics on the decoder side.

The contributions presented in this chapter can be summarized as follows:
\begin{itemize}
    \item We describe QKVAE (\S~\ref{QKVDESIGNSEC}), a model designed to disentangle syntactic information from semantic information by using separate 
    latent variables for keys and values in Transformers attention.
    \item We run experiments on a dataset for English which empirically show that the two types of latent variables have strong preferences respectively for syntax and semantic (\S~\ref{QKVEXPDISENTSEC}).
    \item We also show that our model is capable of transferring syntactic and semantic information between sentences by using their 
    respective latent variables (\S~\ref{QKVEXPTRANSSEC}).
    Moreover, we show that our model's ability to transfer syntax is competitive with supervised models
    when they use their full training set (more than 400k sentences), and that a supervised model needs a fairly large amount 
    of labeled data (more than 50k samples) to outperform it on both semantic and syntactic transfer (\S~\ref{QKVEXPMINTRANSSEC}).
    \item Looping back to syntactic role disentanglement (\S~\ref{QKVEXPSYNRSEC}), we show that semantic latent variables in QKVAE, when freed of the syntactic information modeled by the syntactic variable, display better syntactic role disentanglement numbers on the decoder side  than ADVAE's latent variables across 3 datasets which include datasets with non-regular syntactic structures\footnote{This contribution came after the publication corresponding to this chapter~\citep{felhi-etal-2022-exploiting}, and was therefore not included in the publication.}. 
\end{itemize}

\section{QKVAE: Using Separate Latent Variables for Keys and Values}
\label{QKVDESIGNSEC}
In this section, we describe the architecture of our model, the behavior it entails, and  how we deal with the optimization 
challenges it poses.

\subsection{QKVAE Architecture}

The modification we bring to ADVAE is aimed at controlling how information is selected from the latent space with the value of a
 newly introduced latent variable.
We call this latent variable $z^{syn}$, and refer to the latent
 variables already formulated in ADVAE as  $z^{sem}=\{z^{sem}_1, \dots, z^{sem}_{L}\}$.
$z^{syn}$ is obtained with the same process as each $z^{sem}_l$ (Eq.
~\ref{ADVAEncEq}), \textit{i.e.} by adding an additional
 identifier embedding $e_{s}$, and matrices $M^{\mu s}$ and $M^{\sigma s}$ to obtain its mean and standard-deviation parameters.
 
For the QKVAE Decoder, we modify the Transformer Decoder $\TransDec$ introduced in \S~\ref{TRANSFORMERBGSEC} into $\QKVDec$ so as to use Multi-Head Attention with separate
inputs for keys and values instead of Cross-Attention :

 \[
  \hskip -10mm \QKVDec(T; S_K; S_V) =\tilde{T}_{D^{QKV}}, \ \text{s.t. :}
\] 
\[\tilde{T}_{d} =
  \left \{
    \begin{array}{l}
      T \text{ if } d=0,\text{ else:}\\
      \FF(\MHA(\SA(\tilde{T}_{d-1}), S_K, S_V)
    \end{array}
    \right.
\]
\noindent where $D^{QKV}$ is the number of layers. Recall, here, that theres a fixed number $L$ of keys $k$ and values $v$ for all sentences regardless of their length.
Similar to $\ARTransDec$, we define $\ARQKVDec$ to be the auto-regressive version of $\QKVDec$.
The QKVAE decoder yields probabilities for the generated tokens by using this operator on values given by $z^{sem}$ concatenated with
embeddings $d$, and keys given by a linear transformation on $z^{syn}$: 
\begin{flalign}
  v = \Concat(d;& z^{sem})\nonumber, \hskip 2 mm k = M^s(z^{syn})\nonumber\\
\forall\hskip 1mm i\hskip 2mm\text{ s.t. }\hskip 2mm 1&\leq i\leq |w|: \hskip 2mm\nonumber\\
    \Tilde{w}_i =& \ARQKVDec(w_0, \dots, w_{i-1};k ; v) \nonumber\\
w_i \sim& \Categorical(\softmax(M^w(\Tilde{w}_i))) 
\end{flalign}

\begin{figure*}[!h]
\centering
\hspace{1cm}
    \begin{minipage}[b]{0.4\textwidth}
            \centering
            \hspace{-5cm}
           \begin{minipage}[b]{\textwidth}
            \begin{adjustbox}{minipage=\textwidth,scale=0.45}
            \hspace{ 1cm} \includegraphics
            {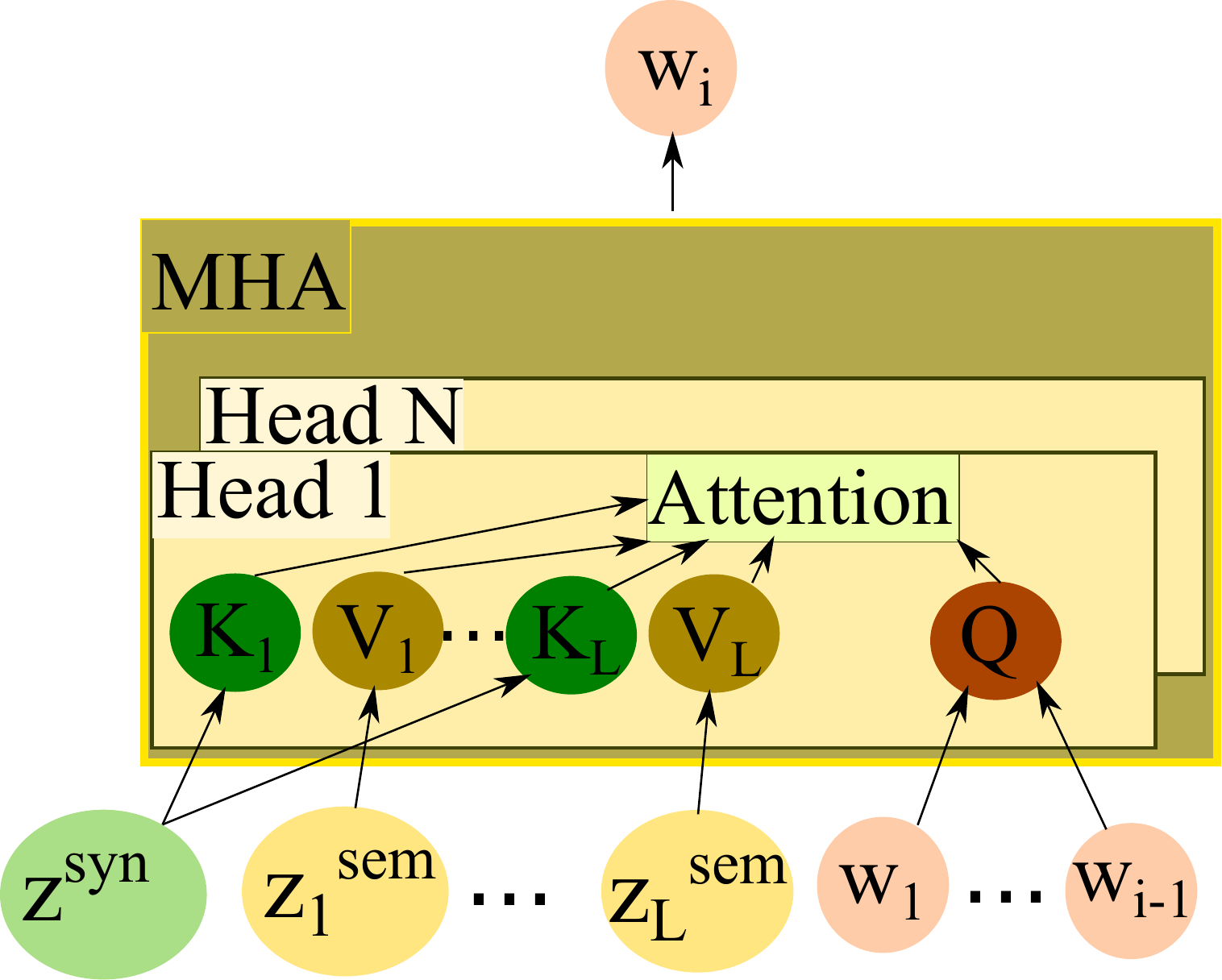}
            \end{adjustbox}
            \end{minipage}
    \end{minipage}
    \begin{minipage}[b]{0.4\textwidth}
            \centering
            \begin{minipage}[b]{\textwidth}
            \begin{adjustbox}{minipage=\textwidth,scale=0.45}
             \hspace{ 1cm} \includegraphics
             {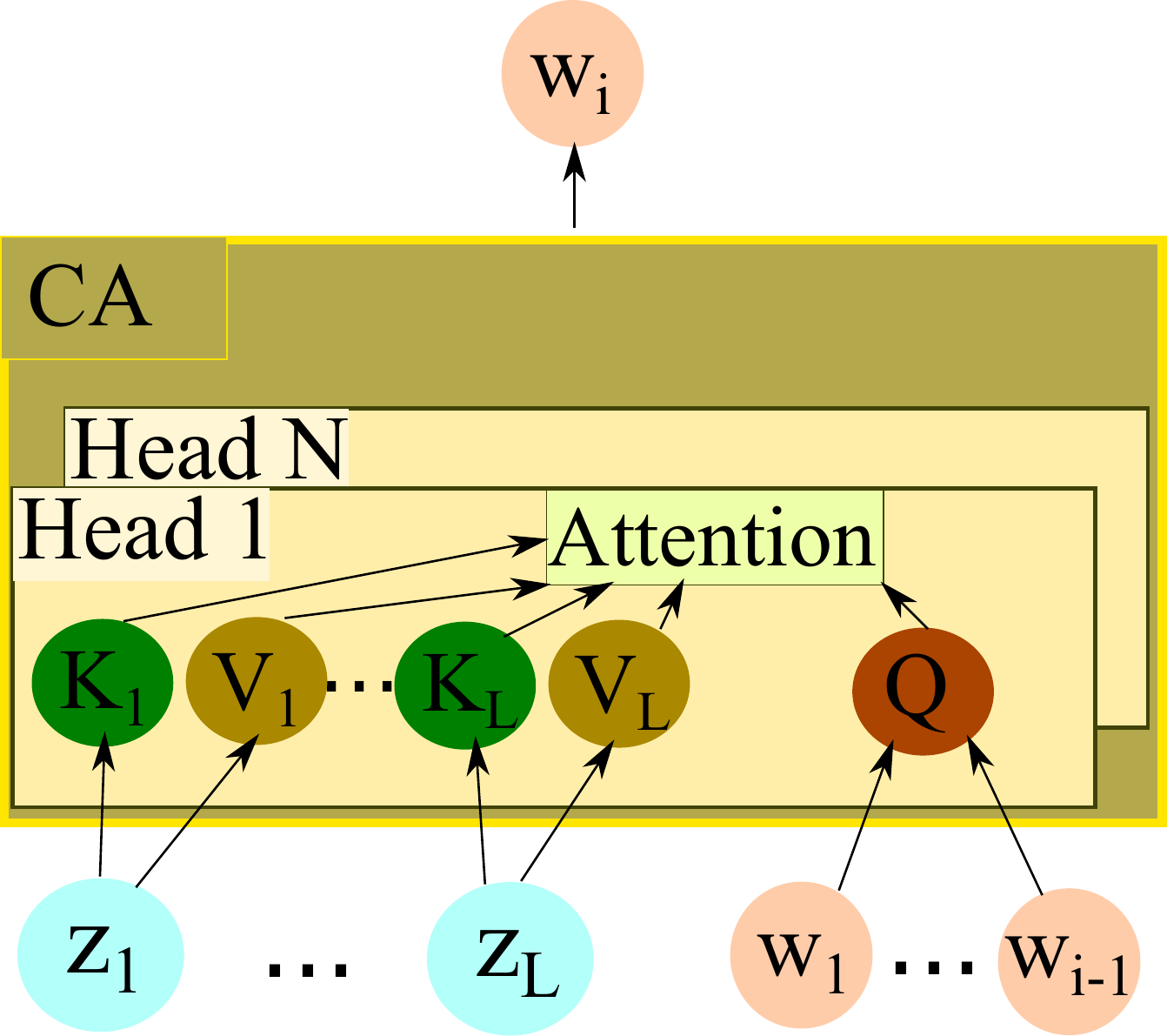}
            \end{adjustbox}
            \end{minipage}
    \end{minipage}
    \caption{\centering The usage of latent variables within ADVAE's decoder (right) and QKVAE's decoder (left). In contrast to ADVAE, all keys used during decoding in QKVAE come from a single latent variable that is separate from the latent variables used to obtain values.}
    \label{fig:ARCHIQK}
\end{figure*} 
\noindent where $M^s$ is a linear layer.\footnote{The output of $M^s$ is reshaped to obtain a matrix of keys.}
While ADVAE already uses Cross-Attention to encode and decode latent variables,
our model uses separate variables to obtain keys and values for Multi-Head Attention in its decoder. To better envision the difference between QKVAE and ADVAE, Figure~\ref{fig:ARCHIQK} depicts the usage of latent variables in QKVAE and ADVAE decoders.


\subsection{QKVAE Behavior}
\label{Behavior}

\begin{table*}[h]
    \small
    \centering
    \hspace{-2 cm}
    \begin{tabularx}{12.5cm}{X X p{1cm} p{1cm} X p{1cm} p{2.4cm} p{3cm}}
    \cline{1-5} 
    \cline{1-5} 
      $v$ & child & to wear & cloak & winter & & & \\
      
      $k1$ & nsubj & root & dobj & $\emptyset$ &$\longrightarrow$ & decoded ($v$, $k1$): & A child wears a cloak.\\
    
      $k2$ & agent & root &  nsubjpass & pobj  &$\longrightarrow$ & decoded ($v$, $k2$):& A cloak is worn, in winter, by a child\\
    \cline{1-5}     \cline{1-5} 
      \end{tabularx}
      \caption{Example of interpretable values for the $v$ and $k$ in our model with $L=4$.
We display a sentence transiting from the active form to the passive form, to illustrate how different \textit{keys} arranging the 
same \textit{values} can lead to the same minimal semantic units being rearranged according to a different syntactic structure.
We also stress that a different set of \textit{keys} may omit or bring forth an element from the \textit{values} vector 
(\textit{e.g.} "winter" here above).}
    \label{tab:QKexpect}
\end{table*} 
In the Multi-Head Attention of our decoder, $z^{syn}$ controls keys, and $z^{sem}$ controls values.
In other words, the value of each 
$z^{sem}_l$ is called to be passed to the target sequence according to its key $k_l$ which is given by the variable $z^{syn}$.
Therefore, for each query, $z^{syn}$ decides which content vector $z^{sem}_l$ participates most to the value of the generated token 
at each generation step.
To better get a gist of the kind of behavior \emph{intended} by this construction, we assume in Table~\ref{tab:QKexpect} for 
explanatory purposes, that our decoder has one layer and one attention head, that the value of each $k^l$  in key matrices $k_1$
and $k_2$ corresponds to syntactic roles, and that each $v^l$ informs on the realization of the corresponding syntactic role.
 Table~\ref{tab:QKexpect} displays the resulting sentence when each of $k1$ and $k2$ are coupled with $v$.

In the examples in Table~\ref{tab:QKexpect}, the generator uses a query at each generation step to pick a word in a manner 
that would comply with English syntax. Therefore, the key of each value should inform on its role in the target structure, which 
justifies syntactic roles as an adequate meaning for keys.
 
Although our model may stray from this possibility and formulate non-interpretable values and keys, keys will still inform on the \emph{roles} of values in the target structure, 
and therefore influence the way values are injected  
into the target sequence.
And given the fact that our model uses multiple layers and attention
heads, and the continuous nature of keys in attention (as opposed to discrete syntactic role labels), our model performs a multi-step and
continuous version of the behavior described in Table~\ref{tab:QKexpect}.

Injecting values into the structure of a sentence requires the decoder to model this structure.
Previous works have shown that
this is well within the capabilities of Transformers.
Specifically, \citet{Hewitt2019ARepresentations} showed that 
Transformers embed syntactic trees in their inner representations, \citet{clark-etal-2019-bert}
showed that numerous attention heads attend to specific syntactic roles, and we
showed in the previous chapter that Transformer-based VAEs can capture the realizations of syntactic roles in latent variables obtained with Cross-Attention.

\subsection{Balancing the Learning of \texorpdfstring{$z^{sem}$}{z\^sem} and \texorpdfstring{$z^{syn}$}{z\^syn}}
Similar to ADVAE, we use a standard Normal distribution as a prior \\$p(z)=p(z^{sem})p(z^{syn})$ and
 train QKVAE with the $\beta$-VAE objective.
To avoid posterior collapse, we follow the strategy from \citet{Li2020AText} presented in \S~\ref{PColSec}: \textit{i)} We pretrain our model as an autoencoder
by setting $\beta$ to 0; \textit{ii)} We linearly increase $\beta$ to its final value ($\KL$ annealing; 
\citealp{Bowman2016GeneratingSpace}) and we threshold each dimension of the $\KL$ term with a factor $\lambda$ (Free-Bits
strategy; \citealp{Kingma2016ImprovedFlow}).

In preliminary experiments with our model, we observed that it tends to encode sentences using only $z^{sem}$.
As  we use conditionally independent posteriors\footnote{These posteriors are ADVAE encoders (Eq.~\ref{ADVAEncEq}).} 
$q(z^{syn}|w)$ and $q(z^{sem}|w)$ for our latent variables, their $\KL$ terms can be written separately, and they can therefore be weighted separately with different values of $\beta$.
Using a lower $\beta$ for $z^{syn}$ as was done by \citet{chen-etal-2019-controllable}
\footnote{Although not explicitly mentioned in the paper, this is performed in their companion source code.}
  did not prove effective in making it informative for the model.
Alternatively, linearly annealing $\beta$ for $z^{sem}$ before $z^{syn}$ did solve the issue. This intervention on the learning process
was inspired by the work of \citet{Li2020ProgressiveRepresentations} which shows that latent variables used at different parts
 of a generative model should be learned at different paces.

\section{Experiments}
\subsection{Setup}
\paragraph{Data}
To compare our model to its supervised counterparts, we train it with data from the English machine-generated paraphrase pairs dataset 
ParaNMT~\cite{Wieting2018ParanMT-50M:Translations}.
More specifically, we use the 493K samples used by 
\citet{chen-etal-2019-controllable}\footnote{\href{https://drive.google.com/open?id=1HHDlUT\_-WpedL6zNYpcN94cLwed\_yyrP}
  {https://drive.google.com/open?id=1HHDlUT\_-WpedL6zNYpcN94cLwed\_yyrP}} to train their model VGVAE.
Since our model is unsupervised, we only use the reference sentences (half the training set) to train our model. Using the development and test sets of
 ParaNMT, \citet{chen-etal-2019-controllable} also provide a curated set of triplets formed by a target sentence (\emph{target}),
   a semantic source (\emph{sem\_src}),and a syntactic source (\emph{syn\_src}).
The semantic source is a paraphrase of the target sentence, while the syntactic source is selected by finding a sentence that is syntactically close to the target (\textit{i.e.} edit distance between the sequence of PoS Tags of both sentences is low\footnote{We follow \citet{chen-etal-2019-controllable} by using this evaluation data, although edit distance between PoS tags might not be a good proxy for syntactic similarity.}) and semantically different from the paraphrase (has low BLEU score with it).
Contrary to paraphrases in the training set of ParaNMT, paraphrases from this set were manually curated.
These triplets are divided into a development set of 500 samples and a test set of 800 samples.
We display results on the test set in the main body of the paper. The results on the development set, which lead to the same conclusions, are reported in Appendix~\ref{DEVRES}.

\paragraph{Training details \& hyper-parameters}
Encoders and Decoders in QKVAE are initialized with parameters from BART~\cite{Lewis2020BART:Comprehension}. After
manual trial and error on the development set, we set 
the sizes of $z^{syn}$ and $z^{sem}$ to 768, and $L$ to 4. Further Hyper-parameters are in Appendix~\ref{HPAPPEN}.
We train 5 instances of our model and report the average scores throughout all experiments.
\paragraph{Baselines}
We compare our system to 4 previously published models, where 2 are supervised and 2 are unsupervised: 
\textit{i) VGVAE~\cite{chen-etal-2019-controllable}: } a VAE-based paraphrase 
generation model with an LSTM 
architecture. This model is trained using paraphrase pairs and PoS Tags to separate syntax and semantics into two latent variables.
This separation is used to separately specify semantics and syntax to the decoder in order to produce paraphrases; \textit{ii) SynPG~\cite{Huang2021GeneratingPairs}:} A paraphrase generation Seq2Seq model based on a Transformer architecture
 which also separately encodes syntax and semantics for the same purpose as VGVAE.
This model is, however, trained using only source sentences with their syntactic parses, without paraphrases; \textit{iii) Optimus~\cite{Li2020Optimus:Space}:} A large-scale VAE based on a fusion between BERT~\cite{devlin-etal-2019-bert} and GPT-2~\cite{Radford2018LanguageLearners} with competitive performance on various NLP benchmarks; \textit{iv) ADVAE: } This model is QKVAE
 without its syntactic variable. The size of its latent variable is set to 1536 to equal the total size of latent variables in QKVAE.

Official open-source instances\footnote{
VGVAE: \href{https://github.com/mingdachen/syntactic-template-generation/}{github.com/mingdachen/syntactic-template-generation/};
SynPG: \href{https://github.com/uclanlp/synpg}{github.com/uclanlp/synpg};
Optimus: \href{https://github.com/ChunyuanLI/Optimus}{github.com/ChunyuanLI/Optimus};
ADVAE: \href{https://github.com/ghazi-f/ADVAE}{github.com/ghazi-f/ADVAE}
}
 of the 4 models above are available,
 which ensures accurate comparisons. The off-the-shelf instances of 
VGVAE and SynPG are trained on ParaNMT with GloVe\footnote{Gains could be observed with better embeddings for supervised models, 
but we stick to the original implementations.}~\cite{Pennington2014} embeddings.
We fine-tune a pre-trained Optimus on our training set following instructions from the authors. Similar to QKVAE,
 we initialize ADVAE with 
parameters from BART~\cite{Lewis2020BART:Comprehension} and train  5 instances of it on ParaNMT with $L=4$.
\subsection{Syntax and Semantics Separation in the Embedding Space}
\label{QKVEXPDISENTSEC}
We first test whether $z^{syn}$ and $z^{sem}$ respectively specialize in syntax and semantics.
A syntactic (resp. semantic) embedding should place syntactically (resp. semantically) similar sentences close to each other in the embedding space.

Using the (\textit{target, sem\_src, syn\_src}) triplets, we calculate for each embedding the proportion of \textit{target} sentences closer
 to \textit{sem\_src} than they are to \textit{syn\_src} in the embedding space.
For simplicity, we refer to the syntactic and semantic embeddings of all models as $z^{syn}$ and $z^{sem}$.
For Gaussian latent variables, we use the mean parameter as a representation (respectively the mean direction parameter from 
the von Mises-Fisher distribution of the semantic variable of VGVAE).
We use an Euclidean distance for Gaussian variables and a cosine distance for the others. 

Since Optimus and ADVAE do not have separate 
embeddings for syntax and semantics \textit{i)} We take the whole embedding for Optimus; \textit{ii)}For ADVAE, we 
 measure the above proportion on the development set for each latent variable $z_l$ (Eq.~\ref{ADVAEncEq}).
 Then, we choose the latent variable that places \emph{target} sentences closest 
 to their \emph{sem\_src} (resp. \emph{syn\_src}) as a semantic (resp. syntactic) variable.
The results are presented in Table~\ref{enc_res}.

\begin{table}\normalsize
\centering
\begin{tabular}{l c c}
\hline
  &$z^{sem} \uparrow$ & $z^{syn} \downarrow$\\
\hline\multicolumn{3}{c}{\textit{Supervised Models}}\\\hline
VGVAE & 99.9& 14.8\\
SynPG & 93.4& 26.5\\
\hline\multicolumn{3}{c}{\textit{Unsupervised Models}}\\\hline
Optimus & 91.8 & -\\
ADVAE & 39.5 & 40.0\\
QKVAE & 89.2& 26.4\\

\hline
\end{tabular}
\caption{\label{enc_res} 
The proportion*100 of embeddings that place a target sentence closer to its semantic source than it is to its syntactic 
source in the embedding space. Arrows ($\uparrow$/$\downarrow$) indicate whether higher or lower scores are better.}
\end{table}

Table~\ref{enc_res} clearly shows for QKVAE, SynPG, and VGVAE that the syntactic (resp. semantic) variables lean towards positioning 
sentences in the embedding space according to their syntax (resp. semantics).
Surprisingly, the syntactic variable of our model specializes in syntax (\textit{i.e.} has low score) as much as that of SynPG.
The generalist latent variable of Optimus seems to position sentences in the
latent space according to their semantics. Accordingly, we place its score in the $z^{sem}$ column. Interestingly, the variables in
 ADVAE have very close scores and score well below 50, which shows that the entire ADVAE embedding leans more towards syntax. This means
that, without the key/value distinction in the attention-based decoder, the variables specialize more in
 structure than in content.

\subsection {Syntactic and Semantic Transfer}
\label{QKVEXPTRANSSEC}
Similar to \citet{chen-etal-2019-controllable}, we aim to produce sentences that take semantic content from \emph{sem\_src} 
sentences and syntax from \emph{syn\_src} sentences.
For each of SynPG, VGVAE, and QKVAE we simply use the syntactic embedding of \emph{syn\_src}, and the semantic embedding of 
\emph{sem\_src} as inputs to the decoder to produce new sentences. Using the results of the specialization test in the previous experiment,
we do the same for ADVAE by taking the 2 latent variables that lean most to semantics (resp. syntax) as semantic (resp. syntactic) variables.
The output sentences are then scored in terms of syntactic and semantic similarity with \emph{sem\_src}, \emph{syn\_src} and \emph{target}.

\paragraph{Control and reference baselines} Beside model outputs, 
  we also use our syntactic and semantic comparison metrics, explicited below, to compare \emph{syn\_src} and \emph{sem\_src} sentences to one
   another and to  \emph{target} sentences. 
Additionally, using Optimus, we embed \emph{sem\_src} and \emph{syn\_src}, take the dimension-wise average of both embeddings, 
and decode it.
As VAEs are known to produce quality sentence interpolations~\cite{Bowman2016GeneratingSpace, Li2020Optimus:Space}, 
the scores for sentences from Optimus help contrast a naïve fusion of features in the embedding space with a composition of well 
identified disentangled features.

\paragraph{Transfer metrics} We measure the syntactic and semantic transfer from source sentences to output sentences.
\textit{i) Semantics:} For semantics, previous works~\cite{chen-etal-2019-controllable, Huang2021GeneratingPairs} rely on 
lexical overlap measures such as BLEU~\cite{Papineni2001BLEU:Translation}, ROUGE~\cite{lin2004rouge}, and Meteor
~\cite{denkowski-lavie-2014-meteor}. As will be shown in our results, the lexical overlap signal does not capture semantic 
transfer between sentences when this transfer is too weak to produce paraphrases. Therefore, we use Meteor (\emph{M}) in conjunction with ParaBART~\cite{Huang2021DisentanglingModelsb} a model where BART~\cite{Lewis2020BART:Comprehension} is fine-tuned using syntactic information to produce neural representations that represent semantics maximally and syntax minimally.
We measure the cosine similarity between sentences according to ParaBART embeddings (\emph{PB}). \textit{ii) Syntax:} We use the script of \citet{chen-etal-2019-controllable} to produce a syntactic tree edit distance (STED) between the constituency trees of sentences, as was done to assess VGVAE.
Additionally, following the evaluation procedure designed by \citet{Huang2021GeneratingPairs} for SynPG, we measure the Template
 Matching Accuracy 
between sentences, where the template is the constituency tree truncated at the second level (TMA2).
TMA2 is the percentage of sentence pairs where such templates match exactly.
We extend this measure by also providing it at the third level (TMA3)\footnote{For example cuts of constituency trees at the second or third level, refer to Figure~\ref{fig:CONSTITCUT} in Chapter~\ref{SYNBGCHAP}.}. Transfer results are presented in Tables~\ref{syn_res} and~\ref{sem_res}. 

\begin{table*}[t]
  \normalsize
\centering
\resizebox{14cm}{!} 
{ 
\begin{tabular}{l c c c|| c c c|| c c c }
\hline
&\multicolumn{3}{c}{\textit{sem\_src}} & \multicolumn{3}{c}{\textit{syn\_src}}& \multicolumn{3}{c}{\textit{target}}\\
&\emph{STED}$\uparrow$ & \emph{TMA2}$\downarrow$ & \emph{TMA3}$\downarrow$ & 
\emph{STED}$\downarrow$ & \emph{TMA2}$\uparrow$ & \emph{TMA3}$\uparrow$ &
\emph{STED}$\downarrow$ & \emph{TMA2}$\uparrow$ & \emph{TMA3}$\uparrow$ \\
\hline\multicolumn{10}{c}{\textit{Control and Reference baselines}}\\ 
\hline
\textit{sem\_src} &0.0 & 100 & 100 & 13.0& 40.3& 4.8 & 12.0& 39.6& 7.0 \\
\textit{syn\_src} & 13.0& 40.3& 4.8& 0.0& 100& 100& 5.9& 84.3& 45.8\\
Optimus & 11.6& 50.0& 15.9& 9.2& 61.6& 23.6& 10.2& 58.9& 21.8\\
\hline\multicolumn{10}{c}{\textit{Supervised Models}}\\
\hline
VGVAE & 13.1& 39.9& 5.4& 3.3& 86.4& 64.1& 6.7& 80.4& 44.6\\
SynPG & 11.7& 41.9& 18.0& 13.5& 74.1& 10.5& 13.1& 69.1& 13.3\\
\hline\multicolumn{10}{c}{\textit{Unsupervised Models}}\\\hline
ADVAE    & 11.9& 47.3& 14.0& 10.3& 54.3$^\dag$& 19.2$^\dag$& 11.1& 52.3& 17.0\\
QKVAE    & 12.7& 40.2& 7.8& 7.2& 68.2& 39.5& 8.9& 63.9& 28.1\\
\hline
\end{tabular}}
\caption{Syntactic transfer results. \emph{STED} is the Syntactic Tree Edit Distance, and \emph{TMA2/3} is the exact matching
 between constituency trees truncated at the $2^{nd}$/$3^{rd}$ level. The comparison 
scores between sentences and \emph{syn\_src} that are not significantly different from the same scores produced with regard to 
\emph{sem\_src} are marked with $^\dag$. We consider differences to be significant 
if their associated $t$-test yields a $p$-value<0.01.\label{syn_res}}
\end{table*}

\begin{table}[t]
  \small
\centering
\begin{tabular}{p{0.045\textwidth} c| c|| c| c|| c| c}
\hline
&\multicolumn{2}{c}{\textit{sem\_src}} & \multicolumn{2}{c}{\textit{syn\_src}}& \multicolumn{2}{c}{\textit{target}}\\
& \emph{M}$\uparrow$&\emph{PB}$\uparrow$
& \emph{M}$\downarrow$&\emph{PB}$\downarrow$
& \emph{M}$\uparrow$& \emph{PB}$\uparrow$ \\
\hline
\multicolumn{7}{c}{\textit{Control and Reference baselines}}
\\
\hline
\textit{sem\_src}    & 100 & 1.0 &  6.9& 0.14& 28.8&  0.84\\
\textit{syn\_src}    & 6.9& 0.14&  100 & 1.0 &  12.1& 0.16\\
Optimus & 12.4& 0.34& 15.9& 0.39&  10.8 & 0.32\\
\hline
\multicolumn{7}{c}{\textit{Supervised Models}}\\\hline
VGVAE    &17.6 & 0.58&  15.3& 0.18& 24.9 & 0.58\\
SynPG  & 45.9 & 0.87& 8.0& 0.13& 25.2& 0.75\\
\hline
\multicolumn{7}{c}{\textit{Unsupervised Models}}\\
\hline
ADVAE & 8.0& 0.19& 8.3$^\dag$& 0.17& 7.4& 0.19\\
QKVAE & 12.8& 0.35&  11.0& 0.19 & 12.6 &0.34\\
\hline
\end{tabular}
\caption{Semantic transfer results. \emph{M} is the Meteor score, and \emph{PB} is the ParaBart cosine similarity. The comparison 
scores between sentences and \emph{syn\_src} that are not significantly different from the same scores produced with regard to 
\emph{sem\_src} are marked with $^\dag$.\label{sem_res}}
\end{table}
 
\paragraph{Sanity checks with metrics and baselines} We notice in Table~\ref{sem_res} that using Meteor as a semantic similarity measure results in various inconsistencies.
For instance, paraphrases \textit{target} have a higher Meteor score with the syntactic
sources than with interpolations from \textit{Optimus}.
It can also be seen that the Meteor score between outputs from VGVAE and both syntactic and semantic sources are rather close
\footnote{This was not observed by \citet{chen-etal-2019-controllable}, as they only compared outputs 
from VGVAE to the target paraphrases.}.
In contrast, ParaBART score behaves as expected across comparisons in Table~\ref{sem_res}.
Consequently, we retain ParaBART score as a semantic similarity measure. 
In the following, we use the scores between \textit{sem\_src}, \textit{syn\_src}, and
\textit{target} (first two rows in Tables~\ref{syn_res} and~\ref{sem_res}) as reference scores for unrelated sentences, paraphrase pairs, and syntactically similar sentences.  
\paragraph{Comparing the supervised baselines} 
VGVAE and SynPG greatly differ in scores.
It can be seen that SynPG copies a lot of lexical items from its semantic input (high Meteor score) which allows for higher semantic similarity scores.
However, Table~\ref{syn_res} shows that SynPG transfers syntax from \textit{syn\_src} at a high level (high TMA2, but low TMA3). 
In contrast, VGVAE transfers syntax and semantics in a balanced way and achieves the best syntax transfer scores overall 
(lowest STED with \emph{syn\_src} and \emph{target}).

\paragraph{Analysing the scores of QKVAE} The semantic similarity scores \textit{PB} of QKVAE outputs with  \textit{target} and \textit{sem\_src} are close to those of Optimus outputs.
Although these scores are low compared to supervised models, they are notably higher
 than semantic similarity scores between unrelated sentences (\textit{e.g.} \textit{syn\_src} and \textit{sem\_src}). 
However, in contrast to Optimus, QKVAE outputs display low PB scores 
with \textit{syn\_src}, which show that they draw very little semantic information from the syntactic sources. 
Concerning syntactic transfer in Table~\ref{syn_res}, QKVAE outputs
share syntactic information with \textit{syn\_src} on all levels (low STED, and high TMA2 and TMA3).
Our model is even competitive with SynPG on TMA2, and better on TMA3 and STED.
As expected, the scores comparing QKVAE outputs to \textit{sem\_src} show that they share very little syntactic information.
On the other hand, ADVAE shows poor transfer performance on syntax and semantics, with only slight differences
between scores w.r.t \textit{syn\_src} and scores w.r.t \textit{sem\_src}.

\subsection{Comparison with a Supervised Model with Less Data}
\label{QKVEXPMINTRANSSEC}
Since VGVAE displays balanced syntactic and semantic transfer capabilities,
 we use it for this experiment where we train it on subsets of sizes in $\{10K, 25K, 50K, 100K\}$ 
from its original training data.
Our goal is to find out how much labeled data is needed for VGVAE to outperform our
unsupervised model on both transfer metrics.
 
\begin{figure}[!h]
  \centering
  \hskip -0.6cm
  \begin{minipage}[b]{\textwidth}
  \begin{adjustbox}{minipage=\textwidth,scale=0.8}
  \hspace{ 1cm} \includegraphics[trim={0.7cm 0.2cm 1.5cm 1.2cm},clip] {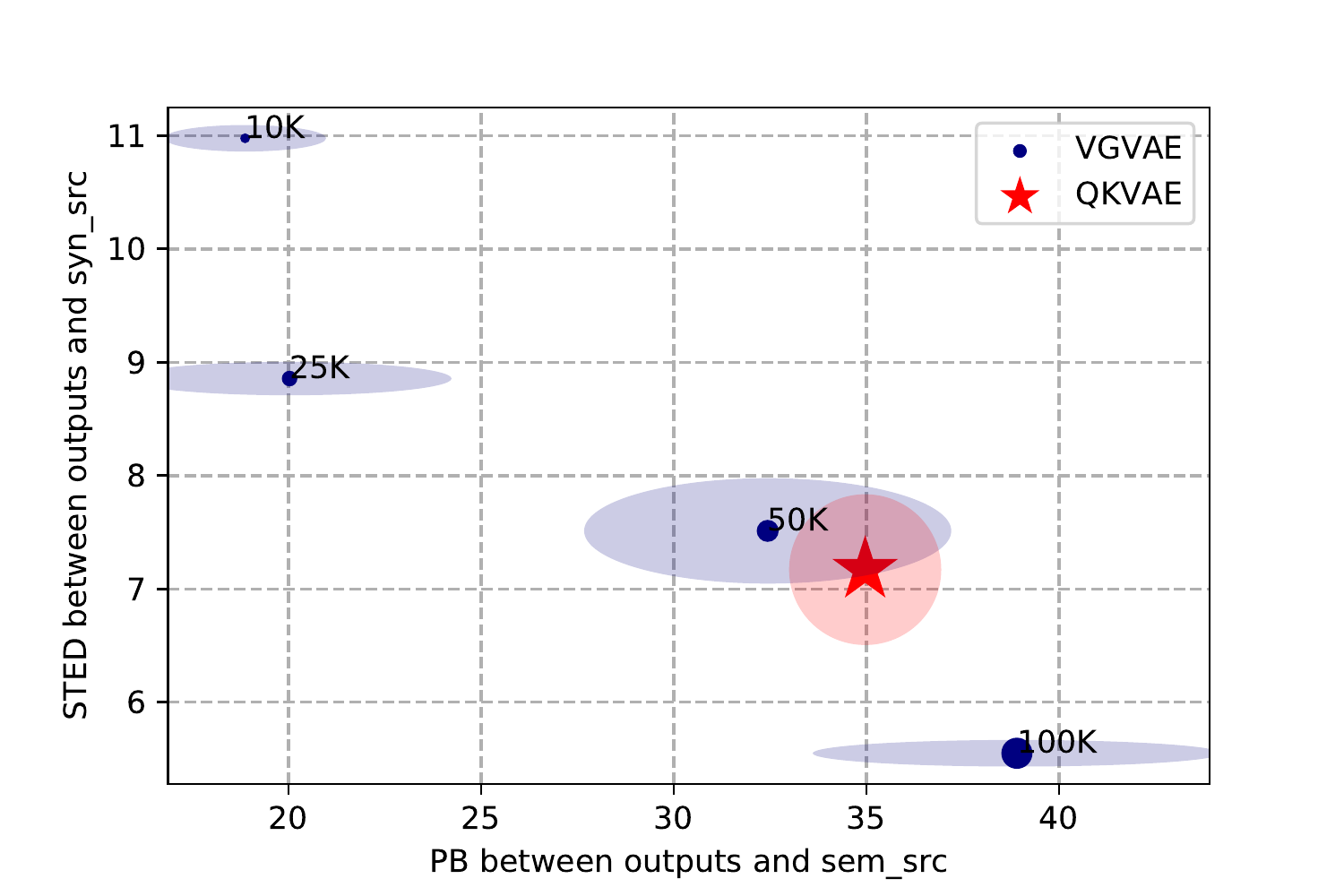}
  \end{adjustbox}
  \end{minipage}
  \caption{Plotting STED w.r.t \emph{syn\_ref} and the PB cosine similarity w.r.t \emph{sem\_ref} for VGVAE with different amounts of labeled data and for QKVAE.
   Points are scaled proportionally to the amount of training data. The vertical and horizontal diameters of each ellipse are equal
   to the standard deviation of the associated data points and axes.}
  \label{fig:STEDvsPB}
\end{figure} 
In Figure~\ref{fig:STEDvsPB}, we plot for QKVAE and instances of VGVAE the \emph{STED} of their outputs w.r.t
 \emph{syn\_src} and the \emph{PB} of these outputs w.r.t \emph{sem\_src}. All values are averages over 5 runs, with standard
 deviations plotted as ellipses.
 Figure~\ref{fig:STEDvsPB} shows that to outperform QKVAE on  syntactic and semantic transfer,  VGVAE needs more than 50K labeled samples. 

\section{A Look-Back to Syntactic Role Disentanglement}
\label{QKVEXPSYNRSEC}
In the previous chapter, we saw that ADVAE performed poorly when it came to syntactic role disentanglement outside of the case where the dataset it was trained on was regularly structured, \textit{e.g.} on the Yelp dataset instead of the SNLI dataset. We speculated that latent variables in ADVAE also had to model syntax, and that accordingly, syntactic variation coming from re-sampling the latent variables made it very difficult to observe change focused on fixed syntactic roles. 

In this chapter, we partly confirmed that speculation through the previous investigations. In fact, Table~\ref{enc_res} showed that the vectorial latent variables in ADVAE leaned slightly more to syntax than they did to semantics. This shows that syntactic information does indeed hinder encoding semantic content in ADVAE. In the same table, we can see that those latent variables, when accompanied with a new  variable which deals with syntactic information as is done in QKVAE, display a much higher specialization in semantics. Given that the information encoded by QKVAE's $z^c$ is much more semantic than the information encoded by ADVAE's $z$, chances are that QKVAE can better disentangle information about realizations of syntactic roles than ADVAE.

\paragraph{Setup: }
We run the same quantitative syntactic role disentanglement experiment as the one displayed in the previous chapter, while setting QKVAE to use the same hyper-parameters and architecture as ADVAE from the previous chapter. We use $L=4$ since results from the last chapter have shown that splitting information into more latent variables than the target syntactic roles generally leads to worse disentanglement.  We train and measure performance on SNLI, Yelp and and a 1.5 Million sentences extract from Wikipedia. 

\paragraph{Results: }
The results are displayed in Table~\ref{tab:resultsSynRoleQKV}.

\begin{table}[!h]
    \centering
    \resizebox{1.0\textwidth}{!}{%
    \begin{tabular}{|c|c|c||c|c||c|c|}
    \hline
    data & model      & beta      & $\mathbb{D}_{enc}$ &  $N_{\Gammaopenc}$  &  $\mathbb{D}_{dec}$ &  $N_{\Gammaopdec}$\\
    \hline
\multirow{4}{*}{SNLI}&\multirow{2}{*}{ADVAE}& 0.3& 1.48\textcolor{gray}{(0.15)}& 3.00\textcolor{gray}{(0.00)}& 0.78\textcolor{gray}{(0.10)}& 3.00\textcolor{gray}{(0.00)}\\ 
     &&0.4& 1.43\textcolor{gray}{(0.79)}& 3.00\textcolor{gray}{(0.00)}& 0.84\textcolor{gray}{(0.10)}& 3.00\textcolor{gray}{(0.00)}\\
&\multirow{2}{*}{QKVAE}& 0.3& 1.06\textcolor{gray}{(0.09)}& 3.80\textcolor{gray}{(0.45)}& 1.09\textcolor{gray}{(0.17)}& 3.00\textcolor{gray}{(0.00)}\\
&& 0.4& 1.06\textcolor{gray}{(0.18)}& 3.00\textcolor{gray}{(0.00)}& 1.31\textcolor{gray}{(0.14)}& 3.00\textcolor{gray}{(0.00)}\\
    \hline
\multirow{4}{*}{Wiki}&\multirow{2}{*}{ADVAE}& 0.3& 0.41\textcolor{gray}{(0.30)}& 2.00\textcolor{gray}{(0.00)}& 1.72\textcolor{gray}{(0.61)}& 1.00\textcolor{gray}{(0.00)}\\
&& 0.4& 0.25\textcolor{gray}{(0.29)}& 2.67\textcolor{gray}{(0.58)}& 0.44\textcolor{gray}{(0.50)}& 2.00\textcolor{gray}{(1.00)}\\
&\multirow{2}{*}{QKVAE}& 0.3& 0.32\textcolor{gray}{(0.06)}& 2.80\textcolor{gray}{(0.45)}& 0.69\textcolor{gray}{(0.18)}& 2.20\textcolor{gray}{(0.45)}\\
&& 0.4& 0.41\textcolor{gray}{(0.03)}& 3.00\textcolor{gray}{(0.00)}& 0.62\textcolor{gray}{(0.07)}& 2.33\textcolor{gray}{(0.58)}\\
    \hline
\multirow{4}{*}{Yelp}&\multirow{2}{*}{ADVAE}& 0.3& 0.48\textcolor{gray}{(0.07)}& 2.00\textcolor{gray}{(0.00)}& 0.23\textcolor{gray}{(0.09)}& 2.20\textcolor{gray}{(0.45)}\\
 &&0.4& 0.54\textcolor{gray}{(0.04)}& 3.00\textcolor{gray}{(0.00)}& 0.22\textcolor{gray}{(0.08)}& 2.20\textcolor{gray}{(0.45)}\\
&\multirow{2}{*}{QKVAE}& 0.3& 0.45\textcolor{gray}{(0.07)}& 2.80\textcolor{gray}{(0.45)}& 1.05\textcolor{gray}{(0.11)}& 2.40\textcolor{gray}{(0.55)}\\
&& 0.4& 0.46\textcolor{gray}{(0.03)}& 2.60\textcolor{gray}{(0.55)}& 0.87\textcolor{gray}{(0.15)}& 2.40\textcolor{gray}{(0.55)}\\
 \hline
     \end{tabular}}
    \caption{Syntactic role disentanglement results for QKVAE vs ADVAE. For QKVAE, the indicated $\beta$ value is used for both latent variables $z^s$ and $z^c$.}
    \label{tab:resultsSynRoleQKV}
  \end{table}

On the encoder side, the table displays comparable or better performance for ADVAE compared to QKVAE across the 3 datasets. This is to be expected as the semantic latent variables of QKVAE will look at different syntactic roles for sentences with different syntax. 

On the decoder side, the disentanglement measure for QKVAE involves only resampling one $z^c_i$ while keeping $z^s$ and any other $z^c_j$ s.t. $j\neq i$ fixed, which should enable it to vary separately syntactic role realizations while better maintaining a fixed syntax for sentences. The numbers in Table~\ref{tab:resultsSynRoleQKV} confirm this hypothesis by displaying consistently higher\footnote{The only exception is ADVAE with $\beta=0.3$ for Wiki, but it concentrates all syntactic roles in a single latent variable; \textit{i.e.} $N_{\Gammaopdec}=1$.} disentanglement concentration scores $\mathbb{D}_{dec}$ for QKVAE than it does for ADVAE across the 3 datasets with a considerable margin, especially for Yelp. The number of disentangled latent variables $N_{\Gammaopdec}$ is also either equal or slightly higher for QKVAE than it is for ADVAE.

\section{Related Work}

We broadly divide recent works on explainability in NLP into two research directions, where
the first seeks \emph{post hoc} explanations for black-box models (\textit{cf.}  \S~\ref{TRANSFORMERLMBGANAL})
and the second seeks to build models that are explainable by design.
This led to models with explicit linguistically informed mechanisms such as the induction of grammars (RNNG;
 \citealp{Dyer2016RecurrentGrammars}, URNNG; \citealp{Kim2019UnsupervisedGrammars}) or constituency trees (ON-LSTM; 
 \citealp{Shen2019OrderedNetworks}, ONLSTM-SYD; \citealp{Du2020ExploitingApproach}).

As a work on disentangled representation learning, this work belonds to this second research direction. As explained in \S~\ref{DISENTAPPPBGSEC},
disentanglement in NLP was performed on various characteristics in text such as style~\cite{John2020DisentangledTransfer,
 Cheng2020ImprovingGuidance}, sentiment and topic~\cite{Xu2020OnSupervision}, or word morphology~\cite{Behjati2021InducingAttention}, with a particular focus on  the separation between syntax and semantics, whether merely to obtain 
an interpretable specialization in the embedding space~\cite{Chen2019ARepresentations, Bao2020, ravfogel-etal-2020-unsupervised, Huang2021DisentanglingModelsb}, or for
 controllable generation~\cite{chen-etal-2019-controllable, Zhang2020Syntax-infusedGeneration, Huang2021GeneratingPairs, Hosking2021FactorisingParaphrasing, Li2021ARepresentation, Hosking2022HierarchicalGeneration}.
However, all these works rely on syntactic information (constituency parses and PoS tags) or semantic information (paraphrase pairs).
To the best of our knowledge, our work is the first to present a method that directs syntactic and semantic information into assigned
embeddings in the challenging unsupervised setup.

From a broader machine learning perspective, using knowledge of the underlying phenomena in our data, we design our model QKVAE 
with an inductive bias that induces understandable behavior in an unsupervised fashion.
Among the existing line of applications of this principle~\cite{Rezende2016UnsupervisedImages, Hudson2018, 
Locatello2020Object-centricAttention, Tjandra2021UnsupervisedRepresentation}, ADVAE~\cite{Felhi2021TowardsRoles}, the model presented in Chapter~\ref{chap:SynRoleDisentChap} which constitutes the basis for QKVAE, is designed to separate information from the realizations of different syntactic roles without 
 supervision on a dataset of regularly structured sentences.
 
\section{Qualitative Results and Discussion}

\begin{table*}[h]
  \small
  \centering
  \begin{tabularx}{14cm}{|X|X|X|X|X|X|}
  \hline
   \emph{sem\_src}& \emph{syn\_src}& SynPG & VGVAE& QKVAE& \emph{target}\\
  \hline \hline
we have destroyed the 49th armored division.&
	concomitant usage is not recommended.&
  	we have destroyed the 49th armored division.&
    	armored division hasn't destroyed.&
      	this military force will be destroyed. &
        	49th armored division has been destroyed .\\\hline																
let the fire burn and put a piece of hot iron in it.&
	sing a song. sing a song for boys.&
  	don't put the fire in it burn a hot piece of iron and fire.&
    	burn the fire. put the iron on burns.&
      	come on fire. get a fire on it.&
        	keep this fire going. keep a piece of hot iron on it.\\\hline																				
they took the lunch boxes ?&
	have you given me your hands ?&
  	do they boxes took the lunch ?&
    	have they taken them your snacks ?&
      	have you heard of some lunch ?&
        	have they taken the lunch boxes ?	\\\hline																		
does it have a coach ?&
	that's a phone switcher, right ?&
  	how does it have a coach ?&
    	that's a coach coach, right ?& 
      	that's a warden, huh?&
        	it has a coach, no ?\\\hline																			
an old lady in a cemetery.&
	that is a bad time for a war.&
  	there's a lady in an old cemetery.&
    	that's an old lady in the cemetery.&
      	this is a strange place for a woman.&
        	there is an old lady in the cemetery.\\\hline
        	
don't be afraid.&
    there are still many places to go.&
        you don't be afraid.&
            there aren't be afraid to be.&
                there will be no need to worry.&
                    there is no need to be afraid .\\\hline
                    
isn't there a door open ?&
	the machines are still good, right ?&
  	a isn't open door there ?&
    	the doors aren't open, right ?&
      	the door will be open, okay?&
        	there is a door open, right ?\\\hline
   \end{tabularx}
  \caption{ Syntactic sources (\emph{syn\_src}), semantic sources (\emph{sem\_src}), the sentences produced when using them with different models, and the corresponding correct paraphrases (\emph{target}).}
  \label{tab:qualiRes}
\end{table*} 
In Table~\ref{tab:qualiRes}, we display example outputs of SynPG, VGVAE, and QKVAE along with their syntactic sources, 
semantic sources, and targets. We generally observed that the outputs of QKVAE range from paraphrases (line 6) to broadly 
related sentences (line 3). 

As was shown by our quantitative results, outputs from VAE-based models (VGVAE and QKVAE) share relatively few lexical items
   with the semantic input. This can be seen in the qualitative examples where they often swap words in the semantic
   source with closely related words (\textit{e.g.} "armored division" to "military force" in line 1, or "lunch boxes" to 
   "snacks" in line 2). We attribute this quality to the smoothness of the latent space of VAEs which places 
   coherent alternative lexical choices in the same vicinity. The examples above also show that
 our model is capable of capturing and transferring various syntactic characteristics such
  as the passive form (line 1), the presence of subject-verb inversion (lines 3, 4, and 7),
   or interjections (lines 4 and 6).

\section{Conclusion}
As a continuation to the work on unsupervised disentanglement of sentence representations presented in the previous chapter, we presented in this chapter QKVAE, an unsupervised model which is designed to disentangle syntax from semantics without syntactic or semantic information (\S~\ref{QKVDESIGNSEC}).
Our experiments show that its latent variables effectively position sentences in the latent space according to these attributes (\S~\ref{QKVEXPDISENTSEC}).
Additionally, we show that QKVAE displays clear signs of disentanglement in transfer experiments (\S~\ref{QKVEXPTRANSSEC}).
Although the semantic transfer is moderate, syntactic transfer with QKVAE is competitive with SynPG, one of its supervised counterparts.
We also show that VGVAE, a supervised model, needs more than 50K samples to outperform QKVAE on both syntactic and semantic transfer (\S~\ref{QKVEXPMINTRANSSEC}). Finally, we show (\S~\ref{QKVEXPSYNRSEC}) that QKVAE moderates the shortcomings of ADVAE when it comes to syntactic role disentanglement outside of the regularly structured dataset SNLI as speculated in the previous chapter .

   We plan to extend this work in three directions: 
\text{i)} Finding ways to bias representations of each $z^{sem}_l$ towards semantic proto-roles (\textit{cf.} \S~\ref{DEPBGSEC}) instead of syntactic roles; \textit{ii)} Applying QKVAE 
to non-text data since it is data agnostic (\text{e.g.} to rearrange elements of a visual landscape.); \textit{iii)} Investigating the behavior of QKVAE on other languages. These extensions are elaborated upon in the next chapter.

\part{Conclusion and Perspectives}}
\chapter{Conclusion and Perspectives}
\label{CONCCHAPT}
The objective of the work presented in this thesis is to help remedy the dire need in NLP for explainable Deep Learning techniques. It was conducted while keeping in mind the rarity of annotated data that plagues explainability. In light of these elements, we present methods that ease obtaining explainable representations with recent Deep Learning components while requiring little-to-no annotated text data. As a conclusion to this thesis, we summarize our contributions in Section~\ref{CONCSUMM}, and outline a few perspective research directions which could extend our work in Section~\ref{CONCPERP}.   

\section{Summary of Contributions}
\label{CONCSUMM}

In the last 3 chapters, we detailed contributions made to data-efficient explainable Deep Learning-based NLP. In Chapter~\ref{chap:SSVAEchap}, we have shown that the Semi-Supervised VAE framework, in the case of sentence classification, was impeded by unnecessary components, namely the Kullback-Leibler divergence in its loss, and the unobserved latent variable in its architecture. Removing these components displayed no degradation to the performance of SSVAEs, improved their speed, and made them easier to design (\textit{i.e.} no prior to specify) and to train (\textit{i.e.} no posterior collapse to counteract). 

The remainder of our contributions pertained to inducing understandable representations without annotations through unsupervised disentanglement with VAEs and Transformers attention. First, in Chapter~\ref{chap:SynRoleDisentChap}, we presented our Attention-Driven VAE (ADVAE), which is the first VAE to use Cross-Attention to encode and decode vectorial latent variables. We have shown that these vectorial latent variables are able to spontaneously align with the realizations of core syntactic roles when trained on a dataset of regularly structured sentences. The latent variables in ADVAE actively follow separate syntactic roles using Cross-Attention, and are able to separately change the realizations of syntactic roles. Subsequently in Chapter~\ref{chap:QKVCHAP}, we have shown that, when different latent variables are asssigned to produce keys and values in the Cross-Attention of a VAE-based Transformer decoder, the keys naturally lean towards encoding syntactic information while the values lean towards semantic information. QKVAE, the neural network we designed to verify this hypothesis, has shown clear signs of successful syntactic and semantic transfer between sentences. Moreover, we showed that a previous supervised disentanglement model needs more than 50K samples to perform better than QKVAE, which shows that QKVAE allows for a considerable gain in annotation effort for disentanglement. Looking back to syntactic role disentanglement with this last model, we finally show that its separation between syntax and semantics also allows for better scores on different datasets when it comes to separately varying the realizations of core syntactic roles in generated sentences.

\section{Perspectives}
\label{CONCPERP}

\subsection{Extending our Investigations to Other Languages}
As is the case for the large majority of works on disentanglement in NLP, such as the ones discussed in \S~\ref{DISENTAPPPBGSEC}, the works presented in this thesis in Chapters~\ref{chap:SynRoleDisentChap} and~\ref{chap:QKVCHAP} only exhibits results on English corpora. We expect extensions to other languages to be interesting, especially for free word order language such as Arabic~\citep{bassam2014formal}. The richer morphology of free word order language allows swapping the positions of syntactic roles in sentences, \textit{e.g.} from Subject-Verb-Object (SVO) to Verb-Subject-Object (VSO), while keeping the sentences grammatical. However, although word order is more flexible for these languages, a preferred word order exists and is usually largely dominant over other word orders \cite[Chapter~4]{song2014linguistic}.

Since our work pertains to disentanglement of syntactic roles and to disentanglement of syntax from semantics, results on free word order languages should be informative about the extent to which ADVAE and QKVAE use word order information to achieve disentanglement.

\subsection{A Structured Latent Variable Version of QKVAE with an Account for Compositional Semantics }

The last model presented in this work, QKVAE, relies on some simplifying assumptions which hinder its ability to accurately model languages. Here are the main simplifications that we think future works could address to extend our work:
\begin{enumerate}
    \item \textit{Syntax and Semantics are modeled as independent generative factors}: It is clearly inaccurate to model language with independent syntax and semantics. For instance, if the semantics of a sentence are expressed by a verbal predicate that subcategorizes multiple arguments, the sentence cannot be syntactically realized through only a subject and a verb. In other words, syntax and semantics in a sentence should be bound by a common \textit{frame}\footnote{\textit{Frame} here, refers to frames in Frame Semantics theory~\cite{fillmore1976frame}. This theory is a theory of meaning which defines for each word a semantic frame, \textit{i.e.} a set of typed semantic arguments that may accompany the word. From a compter science perspective, semantic frames can be understood as function signatures, where the functions are linguistic predicates. Example semantic frames can be seen on the lexical database FrameNet~\cite{ruppenhofer2016framenet} accessible through this URL: \href{https://framenet.icsi.berkeley.edu/fndrupal/frameIndex}{https://framenet.icsi.berkeley.edu/fndrupal/frameIndex}.}, which is not modeled in QKVAE.
    \item  \textit{Semantics is modeled as a set of independent latent variables}: Decomposing the semantics of sentences is largely understood as identifying predicates and their arguments, which are not independent. As a matter of fact, predicates are known to perform argument \textit{selection} by applying paradigmatic restrictions. Furthermore, some arguments to the predicates may narrow down the selection process:
    \begin{enumerate}
        \item $[ARG_0]$ is using a \textit{stick}.
        \item $[ARG_0]$ is using a \textit{computer}.
    \end{enumerate}
    For instance, while sentences (1) and (2) display the same verbal predicate, agent selection in sentence (1) is restricted to the set of animate agents while the agent of sentence (2) should normally be in \textit{human}, a subset of \textit{animate}. Since semantic roles are tightly linked to core syntactic roles, this flaw is also behind syntactic role co-adaptation observed in Chapter~\ref{chap:SynRoleDisentChap}. 
    \item \textit{Sentence-level semantics are only trained using the sentences themselves}: when rearranging the content of a sentence in a new syntactic structure, QKVAE often changes its semantics. This is not surprising since QKVAE does not feature design choices that are targeted at well estimating, an thus, preserving sentence-level semantics (or \textit{compositional} semantics in general). More specifically, sentence representations are not trained using the context in which sentences appear.

\end{enumerate}

Future works could improve upon the above simplifications by investigating the solutions below:

\paragraph{Binding all latent variables to a common underlying frame}
In QKVAE, $p(z)= p(z^{syn})\prod_i p(z^{sem}_i)$. To deal with the first two simplifications listed above, one could introduce a \textit{frame} variable $z^f$ to the probabilistic model as follows  $p(z)= p(z^{syn}|z^f) \prod_i p(z^{sem}_i|z^f)$. This variable would be able to conditionally bind syntax $z^{syn}$ to semantics $z^{sem}$, but also the different component of $z^{sem}$ so as to absorb syntactic role co-adaptation into this minimalistic structured latent variable model. As shown in~\ref{HIERARCH}, hierarchical versions of our Cross-Attention-based models are not trivial to train. In that regard, the recently introduced DELLA~\cite{hu-etal-2022-fuse}, a successfully trained structured latent variable Transformer for language modeling, should help guide the design we describe here.   

\paragraph{Training for effective unsupervised estimation of sentence-level semantics}
The work of \citet{kiros2015skip} on Skip-Thought vectors has shown that sentence level semantic vectors can be effectively estimated, without supervision, simply through next sentence prediction. As discussed in Chapter~\ref{TRANSCHAP}, this method continued to thrive for sentence-level representation learning over the following years and propagated to BERT~\cite{devlin-etal-2019-bert} and other Bert-like models. 
Using our semantic variable $z^{sem}$ and our newly defined frame variable $z^f$, one could design a sentence representation process which takes inspiration in how syntagms compose the meaning of a sentence according to the compositional semantics governing the sentence at hand\footnote{According to our preliminary investigations, Generative Lexicon theory~\cite{pustejovsky1998generative} constitutes a reasonable candidate to inspire such work.}. This linguistically inspired sentence-level representation, together with the well established next sentence prediction method could make for a method to obtain semantic representations which \textit{i)} better model sentence-level semantics since they make use of next sentence prediction; \textit{ii)} are more understandable since they feature a variable $z^f$ modeling the semantic frame, other variables $z^{sem}_i$ modeling the arguments to this frame, and linguistically interpretable interactions between these variables.

\subsection{Applying this Work to Multi-Modal Data }
\label{OtherMedia}

As highlighted in the conclusion of Chapter~\ref{chap:QKVCHAP}, QKVAE is input agnostic, and could therefore as well be used for non-linguistic data. We think an interesting research direction would be to investigate the behavior of QKVAE on other modalities so as to observe the information that will be disentangled analogously to the linguistic concepts studied in our work. The idea of applying linguistically-inspired analysis to all types of data is the basis for Semiotics\footnote{Readers may refer to \citet{Chandler1994} for a beginner friendly introduction to Semiotics.}, a discipline which studies signs and symbols of all natures. In a similar fashion, the research presented in this thesis could lead, for instance, to models that could learn to extract the core syntagms\footnote{The term Syntagm, here, refers to an element in the structure of an observation, where the observation may or may not be textual.} in pictures, regenerate the picture while changing only one of these syntagms, or produce a version of this picture where these syntagms are rearranged. As a matter of fact, the work of \citet{jaegle2022perceiver} shows that Cross-Attention can be used to build an encoder-decoder model that can generalize to any input format with good scaling and generalization performance, and the work of \citet{Locatello2020Object-centricAttention} shows that Cross-Attention based architectures are capable of unsupervised object segmentation on images, which encourages pursuing this research direction.

Further down the road, one could investigate our models in the multi-modal context, where multiple information channels are aggregated in order to model an event syntactically and semantically as one would model the sentence describing the event.


\clearpage
\bibliographystyle{plainnat}  
\bibliography{references, bib2}

\appendix
\chapter{Background}
\section{Derivations for the TC-VAE decomposition}
\label{TCDERIVAPP}
Here, we lay out the derivations necessary to obtain the decomposition displayed in Eq.~\ref{TCVAEKLEQ} in Section~\ref{INDPLVDISENTBG}. We start from the 3 terms on the left-hand-side:

\begin{align}
     &KL[q_\phi(z, n)||q_\phi(z)p(n)] + KL[q_\phi(z)||\prod_j q_\phi(z_j)] + \sum_j KL[q_\phi(z_j)||p(z_j)]\\
  = &\mathbb{E}_{(z; n)\sim q_\phi(z, n)}\left[\log\frac{q_\phi(z, n)}{q_\phi(z)p(n)}\right] + 
     \mathbb{E}_{z\sim q_\phi(z)}  \left[\log \frac{q_\phi(z)}{\prod_j q_\phi(z_j)}\right] \nonumber\\
     &+ \sum_j \mathbb{E}_{z_j\sim q_\phi(z_j)}\left[\log \frac{q_\phi(z_j)}{p(z_j)}\right]\\
  = &\sum_n\int_z q_\phi(z, n)\log\frac{q_\phi(z, n)}{q_\phi(z)p(n)}dz + 
     \int_z q_\phi(z)  \log \frac{q_\phi(z)}{\prod_j q_\phi(z_j)}dz \nonumber\\
     &+ \sum_j \int_{z_j} q_\phi(z_j)\log \frac{q_\phi(z_j)}{p(z_j)}dz_j
\end{align}
We first unify all three terms under the expectation over $q_\phi(z,n)$. For the second term we can use $q_\phi(z)=\sum_n q_\phi(z,n)$:
\begin{align}
     \int_z q_\phi(z)  \log \frac{q_\phi(z)}{\prod_j q_\phi(z_j)}dz=
     \sum_n \int_z q_\phi(z, n)  \log \frac{q_\phi(z)}{\prod_j q_\phi(z_j)}dz 
\end{align}

For the third term, we apply an expectation over $q_\phi(z_1, \dots, z_{j-1}, z_{j+1}, \dots, z_{|z|}|z_j)$ on each element of the sum over j \footnote{The expectation of an expression over the distribution of variables which are not involved in the expression is the expression itself, \textit{i.e.}  $\mathbb{E}_{x\sim p(x)}[f(y)]=f(y)$.}, then also use  $q_\phi(z)=\sum_n q_\phi(z,n)$:

\begin{align}
&\sum_j \int_{z_j} q_\phi(z_j)\log \frac{q_\phi(z_j)}{p(z_j)}dz_j\\
=&\sum_j \int_{z_1, \dots, z_{j-1}, z_{j+1}, \dots, z_{|z|}}\int_{z_j}\nonumber\\ &q_\phi(z_1, \dots, z_{j-1}, z_{j+1}, \dots, z_{|z|}|z_j)q_\phi(z_j)\log \frac{q_\phi(z_j)}{p(z_j)}dz_jdz_1, \dots, dz_{j-1}, dz_{j+1}, \dots, dz_{|z|}\\
=& \sum_j\int_z q_\phi(z)\log \frac{q_\phi(z_j)}{p(z_j)}dz\\
=& \int_z q_\phi(z)\log \frac{\prod_j q_\phi(z_j)}{\prod_j p(z_j)}\\
=& \sum_n\int_z q_\phi(z, n)\log \frac{\prod_j q_\phi(z_j)}{\prod_j p(z_j)}
\end{align}
Finally, fusing the three terms under the same expectation yields:
\begin{align}
&\mathbb{E}_{(z; n)\sim q_\phi(z, n)}\left[\log\frac{q_\phi(z, n)}{q_\phi(z)p(n)}+\log \frac{q_\phi(z)}{\prod_j q_\phi(z_j)}+\log \frac{\prod_j q_\phi(z_j)}{\prod_j p(z_j)}\right]\\
=& \mathbb{E}_{(z; n)\sim q_\phi(z, n)}\left[\log\frac{q_\phi(z, n)}{\cancel{q_\phi(z)}p(n)}\frac{\cancel{q_\phi(z)}}{\cancel{\prod_j q_\phi(z_j)}}\frac{\cancel{\prod_j q_\phi(z_j)}}{\prod_j p(z_j)}\right]
\end{align}

Since $\frac{q_\phi(z, n)}{p(n)}=q_\phi(z|n)$ and $p(z)$ has independent components (\textit{i.e.} $p(z)=\prod_j p(z_j)$), the derivation continues as follows:
\begin{align}
\mathbb{E}_{(z; n)\sim q_\phi(z, n)}\left[\log\frac{q_\phi(z| n)}{ p(z)}\right]=\KL\left[ q_\phi(z| n)||p(z)\right]
\end{align}
which proves the equality in Equation~\ref{TCVAEKLEQ}.


 \chapter{Unsupervised Disentanglement of Syntactic Roles}
This appendix contains the supplementary materials for our work on unsupervised disentanglement of syntactic roles. In section~\ref{StructHeatMap}, we measure the effect of varying latent variables on the appearance/disappearance of syntactic roles to analyze the influence of these latent variables on the \textit{structure} of sentences rather than their content. Appendix~\ref{EXAMPLESEXTRACTIONS} displays a few sentences from both SNLI and Yelp datasets together with the syntactic role extractions obtained with our extraction heuristic to get an idea of the successes/failures of this heuristic. In Appendix~\ref{TRAINING&HP}, we provide the training details and hyper-parameters  for our experiments. In Appendix~\ref{FULLSYNRESULTS} we display the fine-grained syntactic role-wise disentanglement scores corresponding to the global scores we display in the core text. Appendix~\ref{ENTIREROLERESULTS} shows influence heatmaps produced by our model when measured for a wide range of syntactic roles over Stanford Dependency-type annotations and Universal Dependency-type annotations, and also for PoS tags. In Appendix~\ref{QUALIAPPEN} we give a larger array of random examples demonstrating the controlled generation on syntactic roles realizations enabled by ADVAE. Appendix~\ref{RECKLADVAEAPPEN} provides standard VAE language modeling metrics reported for the runs we conducted with the different architectures we compare. Since for our encoder-related metrics we average attention values over all the layers, we display Appendix~\ref{PerLayerAtt} layer-wise encoder influence  heatmaps to show that that trends we described can also be observed on individual layers. Finally, Appendix~\ref{NZVARY} display disentanglement results for ADVAE over a larger grid of values for $L$, the number of vectorial latent variables used.

\section{Measuring the effect of latent variables on the structure of sentences}
\label{StructHeatMap}
\begin{figure*}[!h]
\centering
    \begin{minipage}[b]{0.48\textwidth}
            \centering
            \begin{minipage}[b]{\textwidth}
            \begin{adjustbox}{minipage=\textwidth,scale=0.5}
             \hspace{ 1cm} \includegraphics[trim={1.3cm 0.7cm 2.2cm 1.3cm},clip] {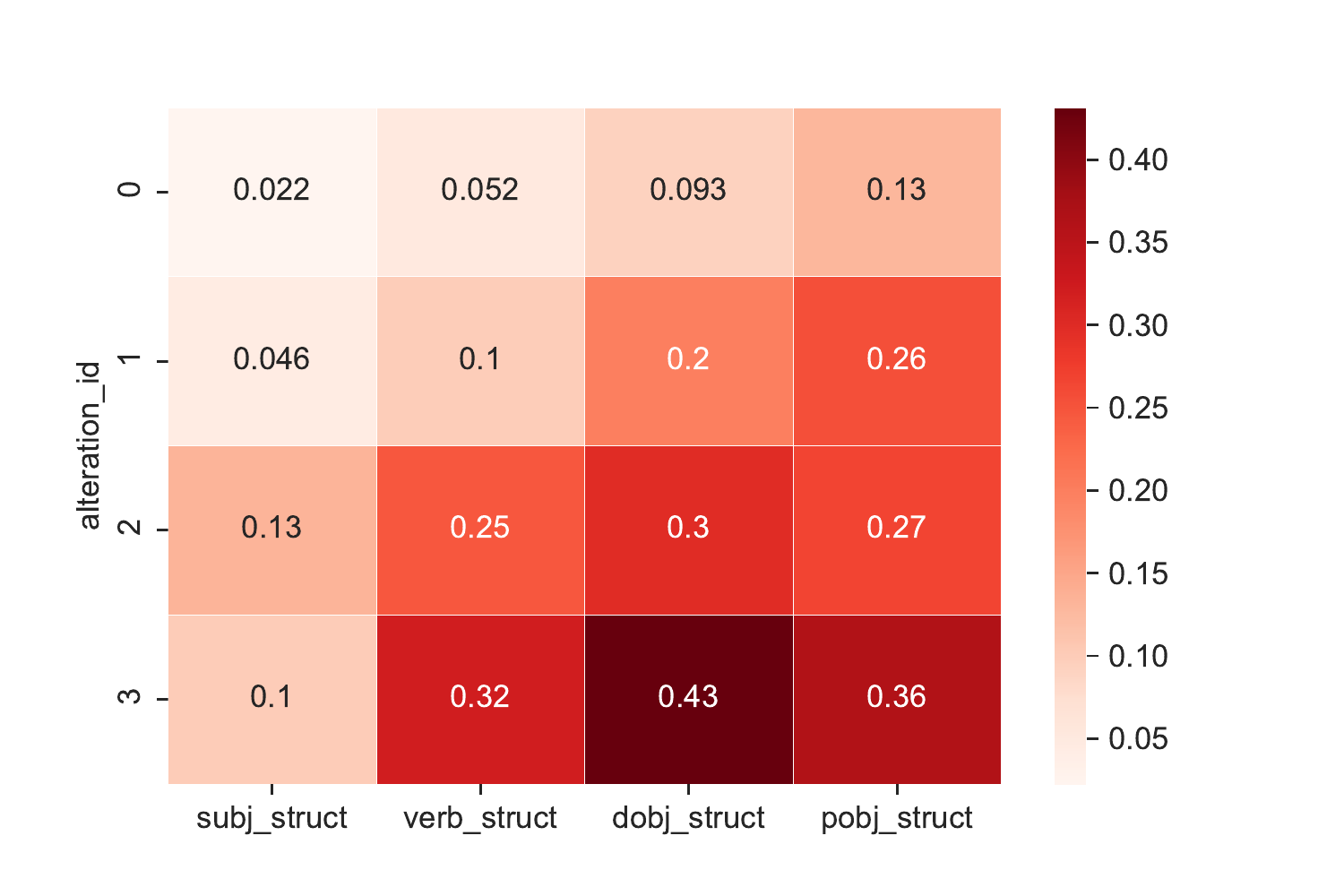}
            \end{adjustbox}
            \end{minipage}
            \caption{\centering The influence of latent variables on the appearance or disappearance of syntactic roles.}
            \label{fig:DECHEATStruct}
    \end{minipage}
\end{figure*}
In Figure~\ref{fig:DECHEATStruct}, for each latent variable and each syntactic role, we report the probability that resampling the latent variable causes the appearance/disappearance of the syntactic role. The instance we use here is the same as the one we use for the heatmaps in the main body of the paper. According to the heatmaps in Figures~\ref{fig:ENCHEAT} and~\ref{fig:DECHEAT}, latent variable 3 is the one associated with the verb. As can be seen in the present heatmap in Figure~\ref{fig:DECHEATStruct}, this same variable is the one that has the most influence on the appearance/disappearance of direct and prepositional objects, and this is a pattern that proved to be consistent across our different runs. This constitutes empirical justification  for our choice of discarding these cases from our decoder influence metrics. 

\section{Example Sentences from Yelp and SNLI and their Corresponding Syntactic Extractions}
\label{EXAMPLESEXTRACTIONS}
Table~\ref{tab:EXAMPLESEXTRACTIONS} shows some samples from SNLI and Yelp reviews. Samples from 
Yelp Reviews exhibit a clearly higher structural diversity. On the other hand, most SNLI samples are highly similar in structure.\\ 
Our syntactic role extraction heuristics were tailored for sentences with verbal roots. As a result,
it can be seen that they struggle with sentences with nominal roots as well as other forms of irregular utterances present in Yelp. For SNLI, 
our extractions mostly yield the expected results, allowing for a reliable global assessment of our models.\\

\begin{table*}[!h]
    \small
    \centering
    \caption{Example syntactic role extractions from both SNLI and Yelp}
    \begin{tabularx}{14cm}{|c|p{4cm}|X|X|X|X|}
    \hline
     Source & Sentence & subj & verb & dobj & pobj \\
    \hline \hline
    Yelp &  i was originally told it would take \_num\_ mins . &  it& told& \_ num \_ mins&  \\\hline
    Yelp &  slow , over priced , i 'll go elsewhere next time . &  i& go& &  \\\hline
    Yelp &  we will not be back &  we& & &  \\\hline
    Yelp &  terrible . &  & & &  \\\hline
    Yelp &  at this point they were open and would be for another hour . &  they& & & this point \\\hline
    SNLI &  people are outside playing baseball . &  people& & baseball& \\\hline
    SNLI &  two dogs pull on opposite ends of a rope . &  two dogs& pull& opposite ends of a rope& a rope \\\hline
    SNLI &  a lady lays at a beach . &  a lady& lays& & a beach\\\hline
    SNLI &  people are running through the streets while people watch . &  people& running& & the streets   \\\hline
    SNLI &  someone prepares food into bowls  &  someone& prepares& food& bowls \\\hline
    \end{tabularx}
    \label{tab:EXAMPLESEXTRACTIONS}
\end{table*} 

\section{Training Details and Hyper-Parameter Settings}
\label{TRAINING&HP}
\paragraph{Our ADVAE's hyper-parameters}
Our model has been set to be large enough to reach a low reconstruction error during the initial reconstruction 
phase of the training. We use 2-layer Transformers with 4 attention heads and a hidden size of 192. Contrary to Vanilla VAEs, our model
seems to perform better with high values of $L$. Therefore, we set our latent vector to a size of 768, and divide it into 96-dimensional
 variables for our $L=8$ model and to 192-dimensional latent variables for our $L=4$ model.
 No automated hyper-parameter selection has been done afterward. 

\paragraph{Sequence VAE hyper-parameters}
As is usually done for this baseline~\citep{Xu2020OnSupervision}, we set both the encoder and the decoder to be 2-layer LSTMs.\\
We run this model for hidden LSTM sizes in [256, 512], and latent vector sizes in [16, 32]. The results for the model scoring the
 highest $\mathbb{D}_{dec}$ are then reported. Even though selection has been done according to $\mathbb{D}_{dec}$,
 we checked the remaining instances of our baselines and they also yielded low $N_{\Gammaopdec}$ values.\\
  
\paragraph{Transformer VAE hyper-parameters}
We set the hidden sizes and number of layers for this baseline similarly to ADVAE, since it is also a Transformer. We run this model for latent vector sizes in [16, 32] and display the highest scoring model, as is done for the Sequence VAE. 
\paragraph{Training phases}
All our models are trained using ADAM~\citep{Kingma2015} with a batch size of 128 and a learning rate of 2e-4 for 20 epochs. The
dropout is set to 0.3.
To avoid posterior collapse, we train all our models for 3000 steps with $\beta=0$ (reconstruction phase), then we linearly increase 
$\beta$ to its final value for the subsequent 3000 steps. Following \citet{Bowman2016GeneratingSpace}, we also use
 word-dropout. We set its probability to 0.1.
\paragraph{Evaluation}
For the evaluation, $T^{dec}$ is set to 2000, and $T^{enc}$ is equal to the size of the test set.

\section{Disentanglement Scores for each Syntactic Role}
\label{FULLSYNRESULTS}
The full disentanglement scores are reported in Table~\ref{tab:results1} for the decoder, and in Table~\ref{tab:results2} for the encoder.
\begin{table*}[h]
    \centering
    \caption{Complete decoder disentanglement scores for SNLI}
    \resizebox{\textwidth}{!}{%
    \begin{tabular}{|c|c||c|c||c|c|c|c|}
    \hline
    Model& $\beta$ & $\mathbb{D}_{dec}$ &  $N_{\Gammaopdec}$  &  $\Delta\Gamma_{dec,verb}$ &  $\Delta\Gamma_{dec,subj}$ & $\Delta\Gamma_{dec,dobj}$ & $\Delta\Gamma_{dec,pobj}$ \\
    \hline \hline
    \multirow{3}{*}{ours-4} &0.3& 0.78\textcolor{gray}{(0.10)}& 3.00\textcolor{gray}{(0.00)}& 0.41\textcolor{gray}{(0.17)}& 0.33\textcolor{gray}{(0.09)}& 0.03\textcolor{gray}{(0.01)}& 0.02\textcolor{gray}{(0.02)}\\ 
     &0.4& 0.84\textcolor{gray}{(0.10)}& 3.00\textcolor{gray}{(0.00)}& 0.47\textcolor{gray}{(0.04)}& 0.31\textcolor{gray}{(0.07)}& 0.04\textcolor{gray}{(0.01)}& 0.01\textcolor{gray}{(0.01)}\\
    \hline
    \multirow{3}{*}{ours-8} &0.3& 0.62\textcolor{gray}{(0.17)}& 3.20\textcolor{gray}{(0.45)}& 0.32\textcolor{gray}{(0.08)}& 0.23\textcolor{gray}{(0.14)}& 0.04\textcolor{gray}{(0.01)}& 0.03\textcolor{gray}{(0.03)}\\
     &0.4& 0.80\textcolor{gray}{(0.11)}& 3.00\textcolor{gray}{(0.00)}& 0.45\textcolor{gray}{(0.06)}& 0.27\textcolor{gray}{(0.05)}& 0.04\textcolor{gray}{(0.03)}& 0.04\textcolor{gray}{(0.03)}\\

\hline
    \hline
    \multirow{3}{*}{Sequence VAE}&0.3& 0.43\textcolor{gray}{(0.18)}& 1.70\textcolor{gray}{(0.48)}& 0.07\textcolor{gray}{(0.05)}& 0.26\textcolor{gray}{(0.10)}& 0.04\textcolor{gray}{(0.05)}& 0.06\textcolor{gray}{(0.05)}\\
    &0.4& 0.91\textcolor{gray}{(0.32)}& 1.40\textcolor{gray}{(0.52)}& 0.24\textcolor{gray}{(0.13)}& 0.45\textcolor{gray}{(0.13)}& 0.05\textcolor{gray}{(0.05)}& 0.16\textcolor{gray}{(0.13)}\\
    \hline
    \multirow{3}{*}{Transformer VAE}&0.3& 0.08\textcolor{gray}{(0.04)}& 3.00\textcolor{gray}{(0.71)}& 0.05\textcolor{gray}{(0.06)}& 0.01\textcolor{gray}{(0.01)}& 0.01\textcolor{gray}{(0.01)}& 0.01\textcolor{gray}{(0.01)}\\ 
    &0.4& 0.11\textcolor{gray}{(0.05)}& 3.80\textcolor{gray}{(0.45)}& 0.04\textcolor{gray}{(0.03)}& 0.04\textcolor{gray}{(0.03)}& 0.02\textcolor{gray}{(0.01)}& 0.01\textcolor{gray}{(0.02)}\\
   
 \hline
     \end{tabular}}
    \label{tab:results1}
  \end{table*}
  \begin{table*}[h]
    \centering
    \caption{Complete encoder disentanglement scores for SNLI}
    \resizebox{\textwidth}{!}{%
    \begin{tabular}{|c|c||c|c||c|c|c|c|}
    \hline
    Model& $\beta$ & $\mathbb{D}_{enc}$ &  $N_{\Gammaopenc}$  &  $\Delta\Gamma_{enc,verb}$ &  $\Delta\Gamma_{enc,subj}$ & $\Delta\Gamma_{enc,dobj}$ & $\Delta\Gamma_{enc,pobj}$ \\
    \hline \hline
    \multirow{3}{*}{ours-4} &0.3& 1.30\textcolor{gray}{(0.09)}& 3.00\textcolor{gray}{(0.00)}& 0.28\textcolor{gray}{(0.05)}& 0.65\textcolor{gray}{(0.02)}& 0.08\textcolor{gray}{(0.03)}& 0.29\textcolor{gray}{(0.03)}\\ 
     &0.4& 1.46\textcolor{gray}{(0.33)}& 3.00\textcolor{gray}{(0.00)}& 0.38\textcolor{gray}{(0.12)}& 0.64\textcolor{gray}{(0.10)}& 0.14\textcolor{gray}{(0.04)}& 0.30\textcolor{gray}{(0.10)}\\
     \hline
    \multirow{3}{*}{ours-8} &0.3& 1.36\textcolor{gray}{(0.13)}& 3.40\textcolor{gray}{(0.89)}& 0.44\textcolor{gray}{(0.12)}& 0.60\textcolor{gray}{(0.18)}& 0.21\textcolor{gray}{(0.08)}& 0.11\textcolor{gray}{(0.06)}\\ 
     &0.4& 1.44\textcolor{gray}{(0.79)}& 3.40\textcolor{gray}{(0.55)}& 0.42\textcolor{gray}{(0.23)}& 0.61\textcolor{gray}{(0.34)}& 0.17\textcolor{gray}{(0.10)}& 0.23\textcolor{gray}{(0.16)}\\
     
\hline
    \hline
    \multirow{1}{*}{Average Position}& - & 0.98 \textcolor{gray}{(-)}& 3.00\textcolor{gray}{(-)}& 0.12\textcolor{gray}{(-)} & 0.70\textcolor{gray}{(-)} & 0.12\textcolor{gray}{(-)} & 0.04\textcolor{gray}{(-)}

\\ 
   
 \hline
     \end{tabular}}
    \label{tab:results2}
  \end{table*}

\section{Disentanglement Heatmaps Over the Entire Range of Syntactic Roles and PoS Tags}
\label{ENTIREROLERESULTS}

\begin{figure*}[!h]
    \centering
    \begin{minipage}[b]{\textwidth}
    \begin{adjustbox}{minipage=\textwidth,scale=0.3}
    \includegraphics[trim={8.9cm 0.7cm 19cm 1.3cm},clip] {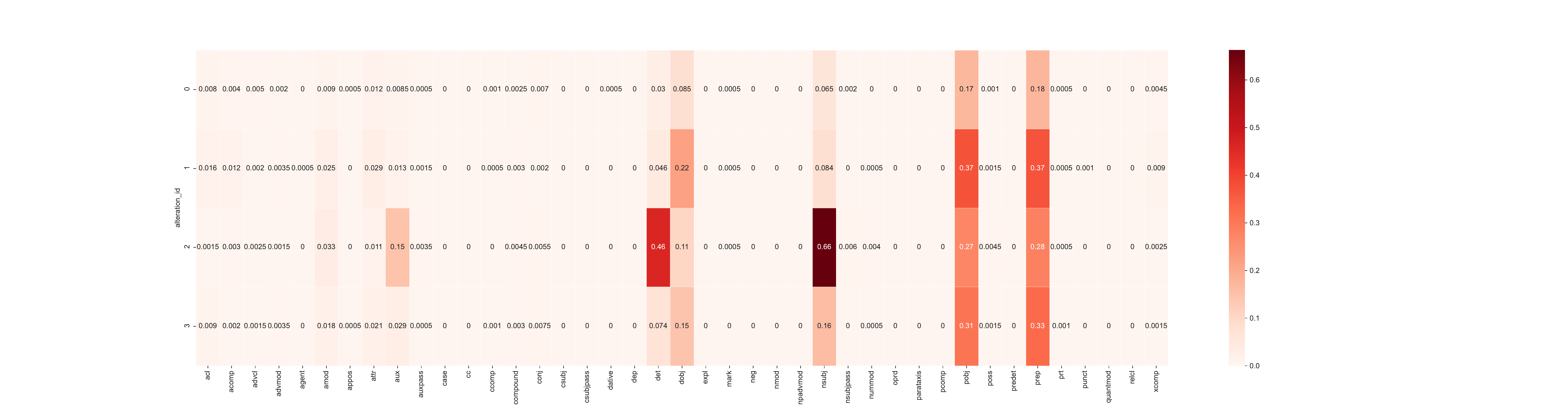}
    \end{adjustbox}
    \end{minipage}
    \caption{Decoder influence heatmap for all SD syntactic roles.}
    \label{fig:ENCHEATLDC}
\end{figure*}

\begin{figure*}[!h]
    \centering
    \begin{minipage}[b]{\textwidth}
    \begin{adjustbox}{minipage=\textwidth,scale=0.3}
    \includegraphics[trim={8.9cm 0.7cm 19cm 1.3cm},clip] {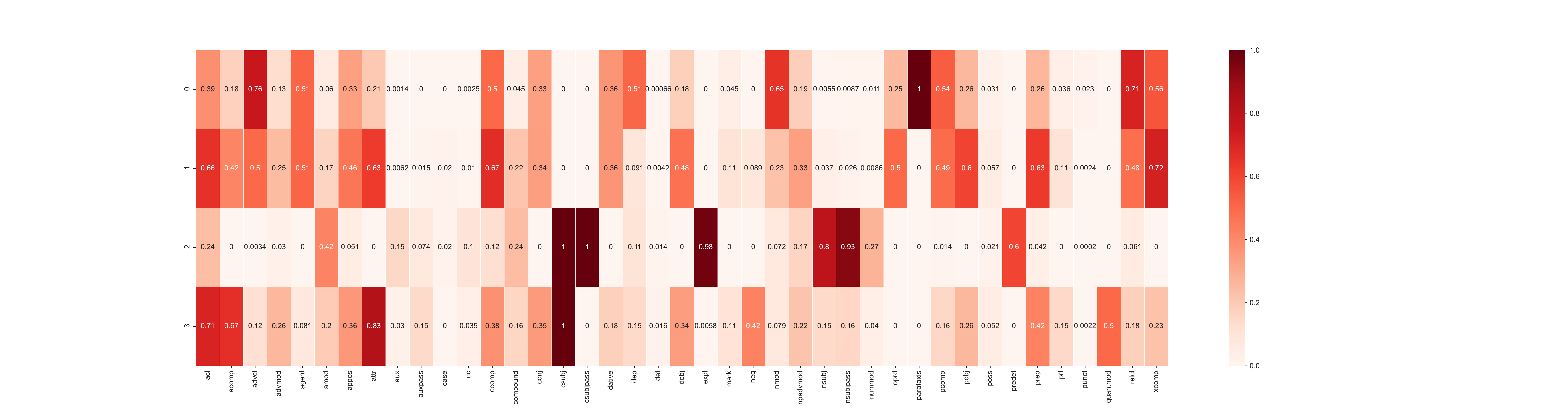}
    \end{adjustbox}
    \end{minipage}
    \caption{Encoder influence heatmap for all SD syntactic roles.}
    \label{fig:DECHEATLDC}
\end{figure*}

\begin{figure*}[!h]
    \centering
    \begin{minipage}[b]{\textwidth}
    \begin{adjustbox}{minipage=\textwidth,scale=0.33}
    \includegraphics[trim={8.9cm 0.7cm 25cm 1.3cm},clip] {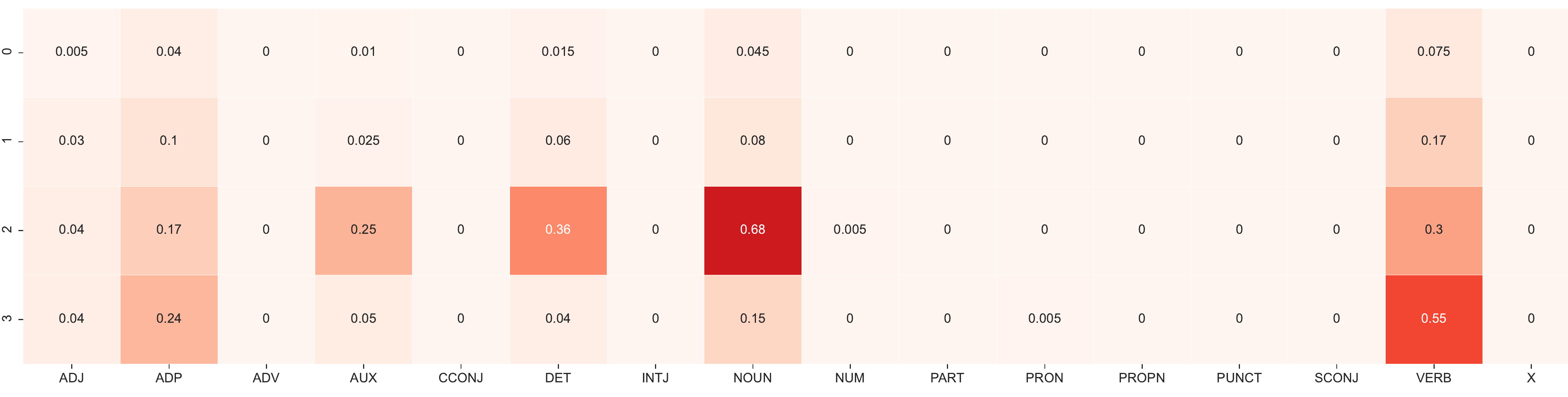}
    \end{adjustbox}
    \end{minipage}
    \caption{Decoder influence heatmap for all PoS Tags.}
    \label{fig:ENCHEATPOS}
\end{figure*}

\begin{figure*}[!h]
    \centering
    \begin{minipage}[b]{\textwidth}
    \begin{adjustbox}{minipage=\textwidth,scale=0.3}
    \includegraphics[trim={8.9cm 0.7cm 19cm 1.3cm},clip] {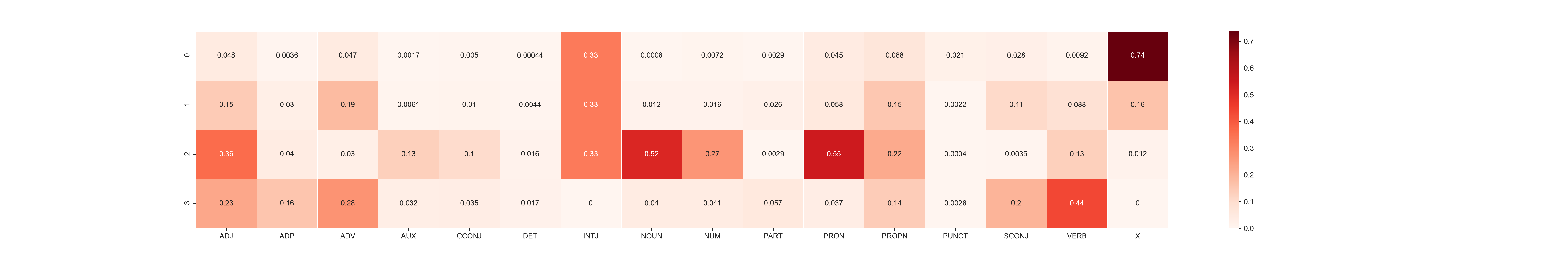}
    \end{adjustbox}
    \end{minipage}
    \caption{Encoder influence heatmap for all PoS Tags.}
    \label{fig:DECHEATPOS}
\end{figure*}

\begin{figure*}[!h]
    \centering
    \begin{minipage}[b]{\textwidth}
    \begin{adjustbox}{minipage=\textwidth,scale=0.3}
    \includegraphics[trim={8.9cm 0.7cm 19cm 1.3cm},clip] {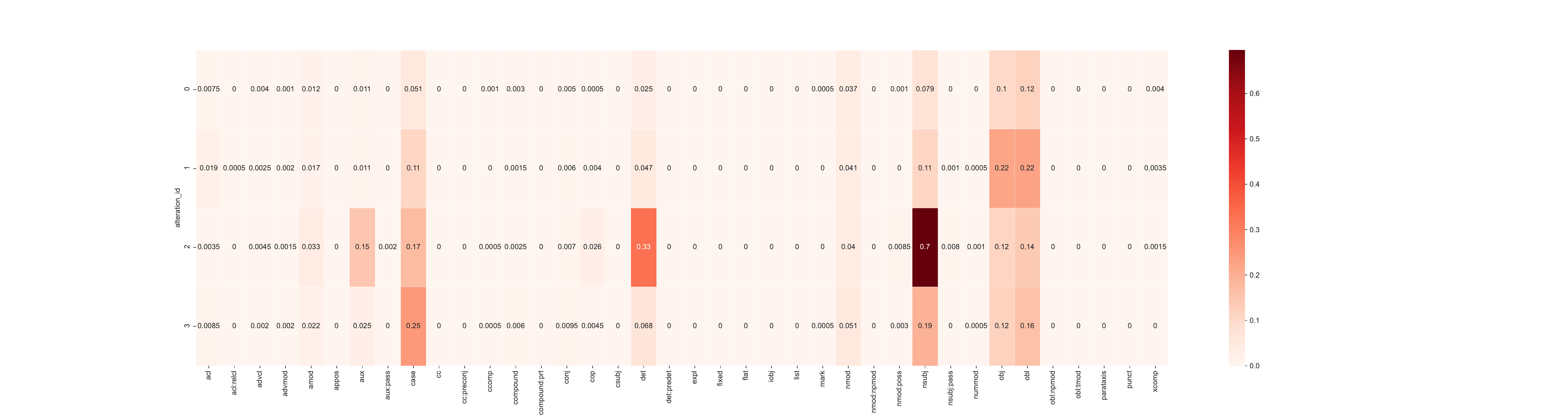}
    \end{adjustbox}
    \end{minipage}
    \caption{Decoder influence heatmap for all UD syntactic Roles.}
    \label{fig:ENCHEATUD}
\end{figure*}

\begin{figure*}[!h]
    \centering
    \begin{minipage}[b]{\textwidth}
    \begin{adjustbox}{minipage=\textwidth,scale=0.3}
    \includegraphics[trim={8.9cm 0.7cm 19cm 1.3cm},clip] {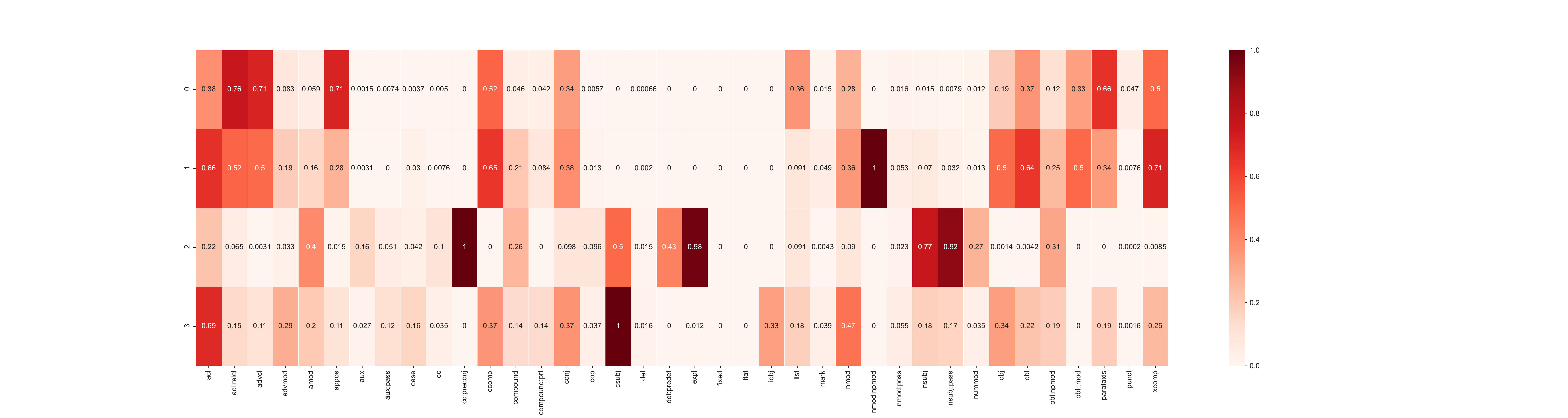}
    \end{adjustbox}
    \end{minipage}
    \caption{Encoder influence heatmap for all UD syntactic Roles.}
    \label{fig:DECHEATUD}
\end{figure*}

We report decoder and encoder heatmaps for all the syntactic roles following the Stanford Dependencies (SD; \citealp{de2008stanford}) annotation scheme of Ontonotes, which was used to train our Spacy2 parser, in Figures~\ref{fig:ENCHEATLDC} and~\ref{fig:DECHEATLDC}. For the sake of extensiveness and to make sure we did not draw results from some parser biases, we also report the same heatmaps but using UDPipe 2.0~\citep{straka-2018-udpipe}, which uses UD type annotations\footnote{A widely adopted annotation scheme derived from Stanford Dependencies.}, in Figures~\ref{fig:ENCHEATUD} and~\ref{fig:DECHEATUD}. Finally, we also report heatmaps for interaction with PoS Tags extracted with Spacy2 in Figures~\ref{fig:ENCHEATPOS} and~\ref{fig:DECHEATPOS}. As was done in the main body of the paper, the span corresponding to each syntactic role (in both annotation schemes) was taken to be the series of words included in its corresponding subtree. In contrast, the span corresponding to each PoS tag was just taken to be the tagged word. Results from UD parsing extraction lead to the same conclusions as from our initial SD results.

The instance of our ADVAE for which we display the above heatmaps is the same one for which we display the heatmaps in Figures~\ref{fig:ENCHEAT} and~\ref{fig:DECHEAT} in the main body of the paper. As shown in those Figures, it mostly uses variable 3 for verbs, variables 2 for subjects, and variable 1 for objects. The remaining variable (0) also seems to capture some interaction with objects.  The heatmaps show that our ADVAE tends to group syntactic roles into latent variables in a way that aligns with the predicative structure of sentences. In fact, variable 2 displays the highest influence on the PoS tag VERB as well as its surroundings as a predicate argument such as adverbs and adverbial phrases. Similarly, latent variable 2 displays a high influence on subjects (nominal or clausal), numeral modifiers, adjectival modifiers, and auxiliaries (for conjugation). Moreover, Variable 1 highly influences the direct and prepositional objects, which we study in the main body of the paper, but also diverse clausal modifiers and obliques which often play similar roles to direct and prepositional objects in a predicate structure.

\section{Additional Examples of Resampled realizations for each syntactic role}
\label{QUALIAPPEN}
Table~\ref{tab:resultsresamplebig} contains a wide array of examples where the latent variable corresponding to each syntactic role is resampled.
\begin{table}[h!]
    \small
    \centering
    \caption{More examples where we resample a specific latent variable for a sentence.}
    \begin{tabularx}{14cm}{|X|X|X|X|}
    \hline
     Original sentence& Resampled subject& Resampled verb & Resampled dobj/pobj \\
    \hline \hline
     the woman is riding a large brown dog  & two men are riding in a large city  & the woman is wet  & the woman is riding on the bus \\ \hline
 the police are running in a strategy  & a man is looking at a date  & the police are at an arid  & the police are running in a wooded area \\ \hline
 a man is holding a ball  & a man is holding a ball  & a man is , and a woman are talking on a road  & a man is sitting on a cellphone outside \\ \hline
 everyone is watching the game  & some individuals are watching tv  & everyone is a man  & everyone is watching the game in the air \\ \hline
 there is a man in the air  & a man is sitting in the air  & there is no women wearing swim trunks  & there is a man in a red shirt \\ \hline
 a group of friends are standing on a beach  & an elderly father and child are standing on the beach  & a group of people are standing on a beach  & a group of friends are looking at the beach \\ \hline
 the women are in a store  & a man is playing a game  & two women are on a break  & two women are sitting on a bench \\ \hline
 a man is playing a game  & a little girl is playing with a ball  & a man is clean  & a man is sitting on a lake to an old country \\ \hline
 a man is playing a game  & some dogs are playing in the pool  & a man is preparing to chase himself  & a man is playing a game \\ \hline
 the memorial woman is happy  & a dog is happy  & the memorial workers are in a room  & the memorial is happy \\ \hline
 a man is wearing a green jacket and a ship  & a boy sitting in a green device  & a man is dancing for the camera  & the man is wearing a hat \\ \hline
 a man is playing a game  & a man is playing a game  & two men are tripod  & a man is playing with a guitar \\ \hline
 a man is wearing a brown sweater and green shirt  & a karate dog is swimming in a chair  & a man is bought a brown cat in an airplane  & a man is wearing a dress and talks to the woman \\ \hline
 the woman is about to visitors  & three people are working at a babies  & the woman is wearing a sewer  & the woman is about to sell a tree \\ \hline
 a man is sitting in the snowy field  & a man is sitting in the snowy field  & a man is wearing electronics  & a man is sitting on a park bench \\ \hline
 two people are playing in the snow  & the motorcycle is a woman on the floor  & two people play soccer in the snow  & two people are playing in a concert \\ \hline
 a man is standing next to another man  & a boy is standing next to another man  & a man is standing  & a man is standing next to a man \\ \hline
 a man is on his bike  & a man is on his bike  & a dog is showing water  & a man is on his bike \\ \hline
 a man is sitting in front of a tree , taking a picture  & a man is sitting in front of a tree  & a man is holding a red shirt and climbing a tree  & a man is sitting on a suburban own \\ \hline
 a man is sitting with a dog  & the children are sitting with the dog  & a man is playing with a dog  & a man is sitting with an umbrella \\ \hline
     \end{tabularx}
    \label{tab:resultsresamplebig}
\end{table} 

\clearpage
\section{Reconstruction and Kullback-Leibler Values Across Experiments}
\label{RECKLADVAEAPPEN}
\begin{table}[!h]
    \caption{Reconstruction loss and Kullback-Leibler values on SNLI.}
    \centering
    \resizebox{1.0\textwidth}{!}{%
    \begin{tabular}{|c|c||c|c|c|}
    \hline
    Model& $\beta$ & $-\mathbb{E}_{(z) \sim q_\phi(z|x)}\left[ \log p_\theta(x|z) \right] $& $\KL[q_\phi(z|x)||p(z)]$& Perplexity Upper Bound\\
    \hline \hline
    
    \multirow{2}{*}{Sequence VAE}&0.3&  31.38\textcolor{gray}{(0.12)}& 2.80\textcolor{gray}{(0.25)}&22.02\textcolor{gray}{(0.30)}\\ 
    &0.4& 32.19\textcolor{gray}{(0.13)}& 1.22\textcolor{gray}{(0.04)}&21.08\textcolor{gray}{(0.22)}\\
    \hline
    
    \multirow{2}{*}{Transformer VAE}&0.3&  24.35\textcolor{gray}{(0.14)}& 13.38\textcolor{gray}{(0.19)}&25.07\textcolor{gray}{(0.27)}\\ 
    &0.4& 26.57\textcolor{gray}{(0.27)}& 8.36\textcolor{gray}{(0.32)}&20.68\textcolor{gray}{(0.16)}\\
    \hline
    \hline
    \multirow{2}{*}{ours-4} &0.3&  10.75\textcolor{gray}{(0.94)}& 42.63\textcolor{gray}{(1.16)}&68.49\textcolor{gray}{(5.96)}\\ 
    &0.4& 16.01\textcolor{gray}{(0.64)}& 27.93\textcolor{gray}{(1.52)}&36.16\textcolor{gray}{(2.20)}\\
    \hline
    \multirow{2}{*}{ours-8} &0.3& 8.83\textcolor{gray}{(1.66)}& 46.99\textcolor{gray}{(2.99)}&77.26\textcolor{gray}{(9.02)}\\ 
     &0.4& 16.84\textcolor{gray}{(8.50)}& 27.34\textcolor{gray}{(14.99)}&39.23\textcolor{gray}{(11.27)}\\

 \hline
     \end{tabular}}
    \label{tab:resultsRecKL}
  \end{table}
\label{RECKL}

\begin{table}[!h]
    \caption{Reconstruction loss and Kullback-Leibler values on Yelp.}
    \centering
    \resizebox{1.0\textwidth}{!}{%
    \begin{tabular}{|c|c||c|c|c|}
    \hline
    Model& $\beta$ & $-\mathbb{E}_{(z) \sim q_\phi(z|x)}\left[ \log p_\theta(x|z) \right] $& $\KL[q_\phi(z|x)||p(z)]$& Perplexity Upper Bound\\
    \hline \hline
    \multirow{2}{*}{Sequence VAE}&0.3&  32.55\textcolor{gray}{(0.27)}& 4.26\textcolor{gray}{(0.57)}&36.97\textcolor{gray}{(0.82)}\\ 
    
    &0.4& 33.35\textcolor{gray}{(0.11)}& 1.42\textcolor{gray}{(0.15)}&32.42\textcolor{gray}{(0.13)}\\
    \hline
    \multirow{2}{*}{Transformer VAE}&0.3&  23.64\textcolor{gray}{(0.11)}&
    19.24\textcolor{gray}{(0.32)}&53.94\textcolor{gray}{(1.14)}\\ 
    &0.4& 26.41\textcolor{gray}{(0.11)}& 12.79\textcolor{gray}{(0.20)}&41.25\textcolor{gray}{(0.55)}\\
    \hline
    \hline
    \multirow{2}{*}{ours-4} &0.3&  7.30\textcolor{gray}{(0.27)}& 55.19\textcolor{gray}{(0.30)}&121.44\textcolor{gray}{(5.16)}\\ 
    &0.4& 18.19\textcolor{gray}{(4.38)}& 29.85\textcolor{gray}{(6.52)}&58.11\textcolor{gray}{(8.40)}\\
    \hline
    \multirow{2}{*}{ours-8} &0.3& 5.36\textcolor{gray}{(0.48)}& 59.24\textcolor{gray}{(0.61)}&129.54\textcolor{gray}{(7.63)}\\ 
     &0.4& 15.36\textcolor{gray}{(5.75)}& 34.16\textcolor{gray}{(12.40)}&63.62\textcolor{gray}{(16.56)}\\

 \hline
     \end{tabular}}
    \label{tab:resultsRecKLYelp}
  \end{table}
The values for the reconstruction loss, the $\KL$ divergence, and the upper bound on perplexity concerning the experiments in the main body of the paper are reported in Table~\ref{tab:resultsRecKL}. The same value for the Yelp experiments are in Table~\ref{tab:resultsRecKLYelp}. Since our models are VAE-based, one can only obtain the upper bound on the perplexity and not its exact value. These upper bound values are obtained using an importance sampling-based estimate of the negative log-likelihood, as was done in \citet{Wu2020OnBeyond}. We set the number of importance samples to 10.

It can be seen that the behavior of ADVAEs is very different from classical Sequence VAEs and Transformer VAEs. On the plus side, they are capable of sustaining much more information in their latent variables as shown by their higher $\KL$, and they do better at reconstruction. The upper bound estimate of their perplexity is however higher. A high $\KL$ makes it more difficult for the importance sampling-based perplexity estimate to reach the true value of the model's perplexity. This may be the reason behind the higher values observed for ADVAEs. 

\section{Layer-wise Encoder Attention}
\label{PerLayerAtt}
\begin{figure*}[!h]
\centering
    \begin{minipage}[b]{0.30\textwidth}
            \centering
           \begin{minipage}[b]{\textwidth}
            \begin{adjustbox}{minipage=\textwidth,scale=0.35}
            \hspace{ 1cm} \includegraphics[trim={1.3cm 0.7cm 2.2cm 1.3cm},clip] {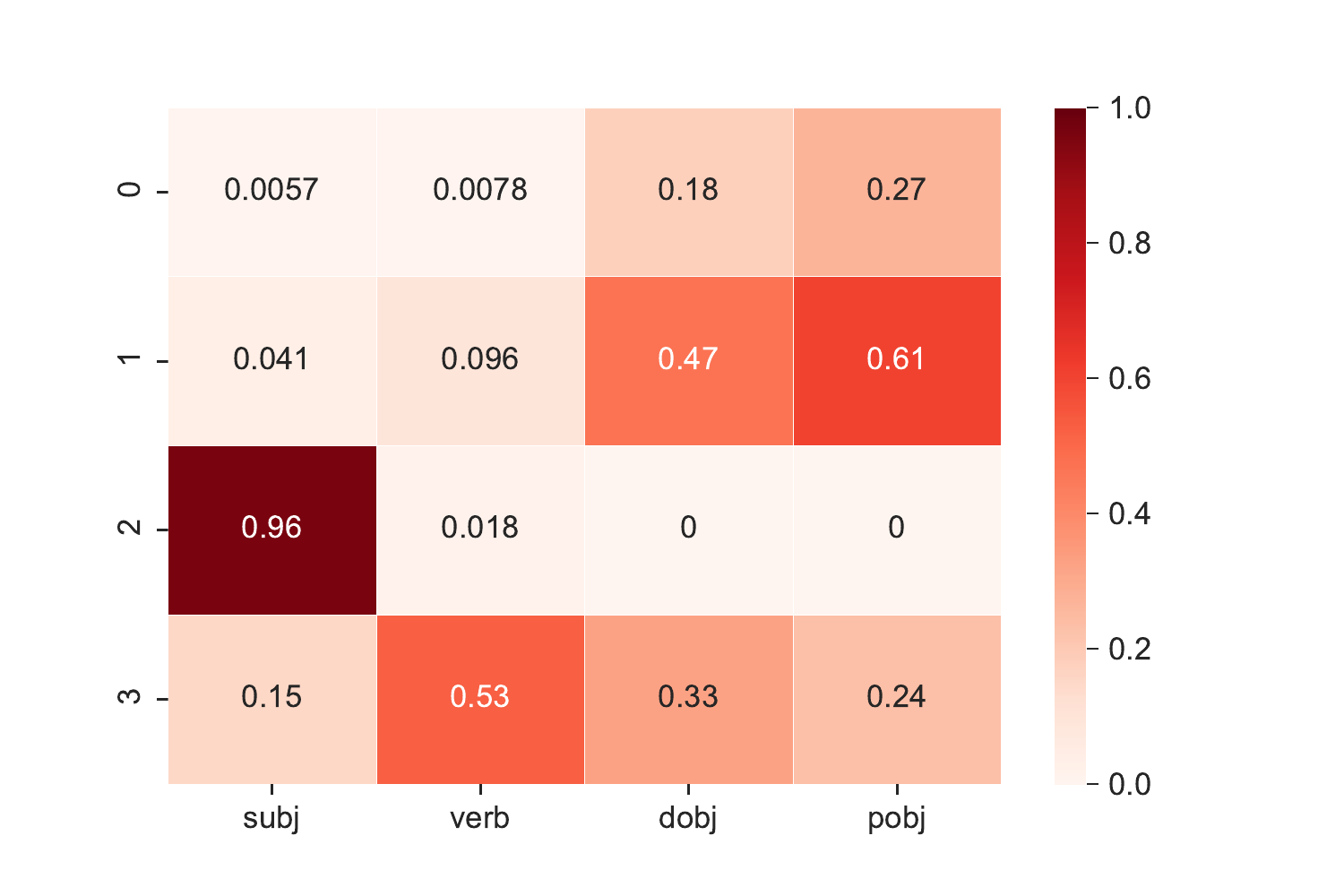}
            \end{adjustbox}
            \end{minipage}
            \caption{\centering Encoder influence heatmap ($\Gammaopenc$) when only using the \emph{first} layer.}
            \label{fig:ENCHEATLAYER0}
    \end{minipage}
    \begin{minipage}[b]{0.30\textwidth}
            \centering
            \begin{minipage}[b]{\textwidth}
            \begin{adjustbox}{minipage=\textwidth,scale=0.35}
             \hspace{ 1cm} \includegraphics[trim={1.3cm 0.7cm 2.2cm 1.3cm},clip] {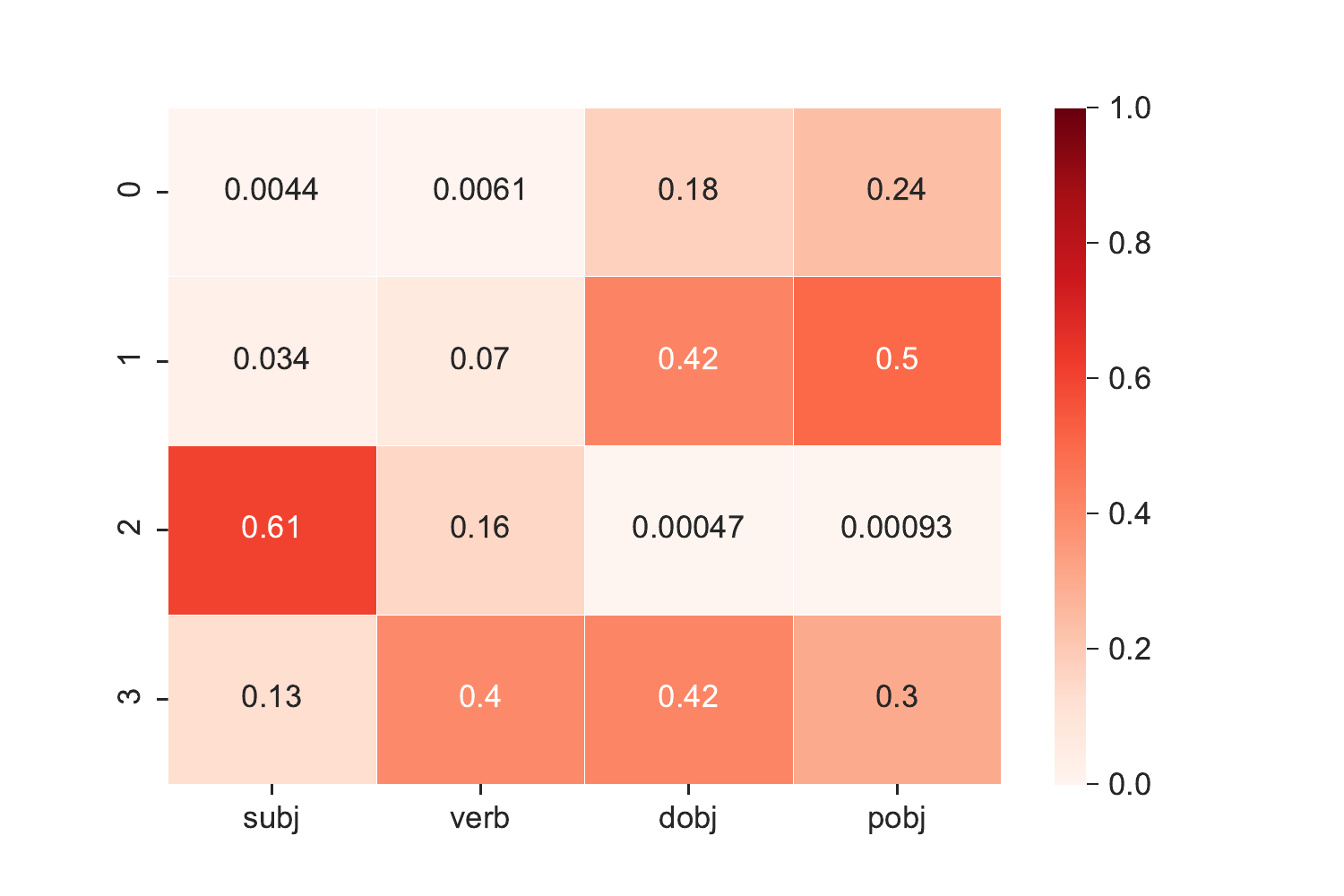}
            \end{adjustbox}
            \end{minipage}
            \caption{\centering Encoder influence heatmap ($\Gammaopenc$) when only using the \emph{second} layer.}
            \label{fig:ENCHEATLAYER1}
    \end{minipage}
    \begin{minipage}[b]{0.30\textwidth}
            \centering
           \begin{minipage}[b]{\textwidth}
            \begin{adjustbox}{minipage=\textwidth,scale=0.35}
            \hspace{ 1cm} \includegraphics[trim={1.3cm 0.7cm 2.2cm 1.3cm},clip] {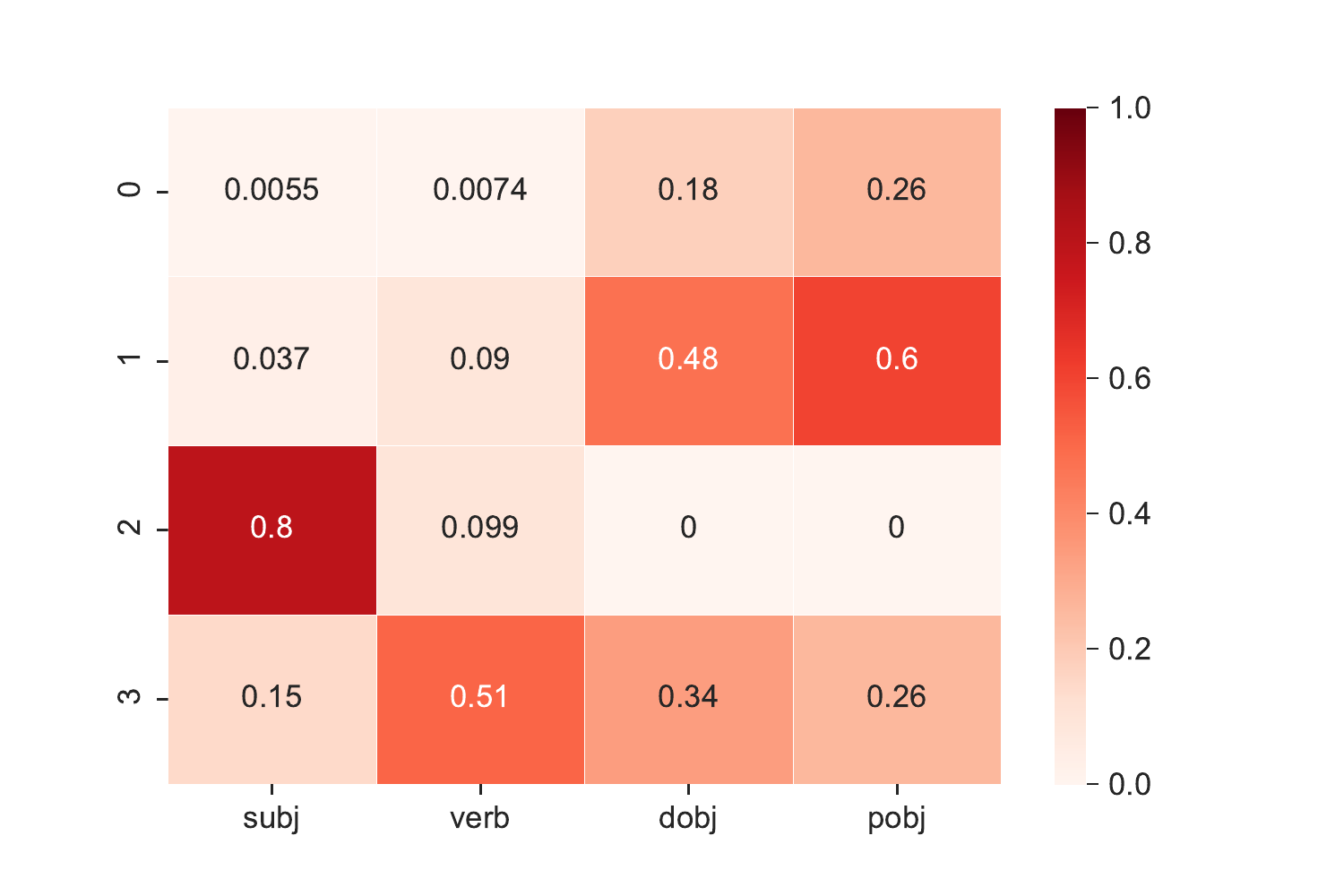}
            \end{adjustbox}
            \end{minipage}
            \caption{\centering Encoder influence heatmap ($\Gammaopenc$) when \emph{averaging} over both layers.}
            \label{fig:ENCHEATLAYERMEAN}
    \end{minipage}
\end{figure*}

In the main body of the paper, we use attention values that are averaged throughout the network. We display the encoder heatmaps obtained by using attention values from the first layer (Fig.~\ref{fig:ENCHEATLAYER0}), the second layer (Fig.~\ref{fig:ENCHEATLAYER1}), or an average on both layers (Fig.~\ref{fig:ENCHEATLAYERMEAN}) for comparison.

As can be seen, the first layer alone provides the most sparse heatmap, and thus, the clearest correspondence between syntactic roles and latent variables. This corroborates the claims of \citet{Tenney2020BERTPipeline} about syntax being most prominently processed in the early layers of Transformers.

\section{ADVAE Results for a larger grid of \texorpdfstring{$L$}{L} values}
\label{NZVARY}

We display in Table~\ref{tab:NzVAry} the quantitative results of ADVAE on SNLI for $L$ in $\{2, 4, 6, 8\}$.  For ours-2, it is normal that it only separates syntactic role realizations into a maximum of 2 latent variables, as seen from the values of $N_{\Gammaopenc}$ and $N_{\Gammaopenc}$ , since 2 is its total number of latent variables.

\begin{figure*}[!h]
\hspace{-1cm}
    \begin{minipage}[b]{0.48\textwidth}
            \centering
           \begin{minipage}[b]{\textwidth}
            \begin{adjustbox}{minipage=\textwidth,scale=0.6}
            \hspace{ 1cm} \includegraphics[trim={1.3cm 0.7cm 2.2cm 1.3cm},clip] {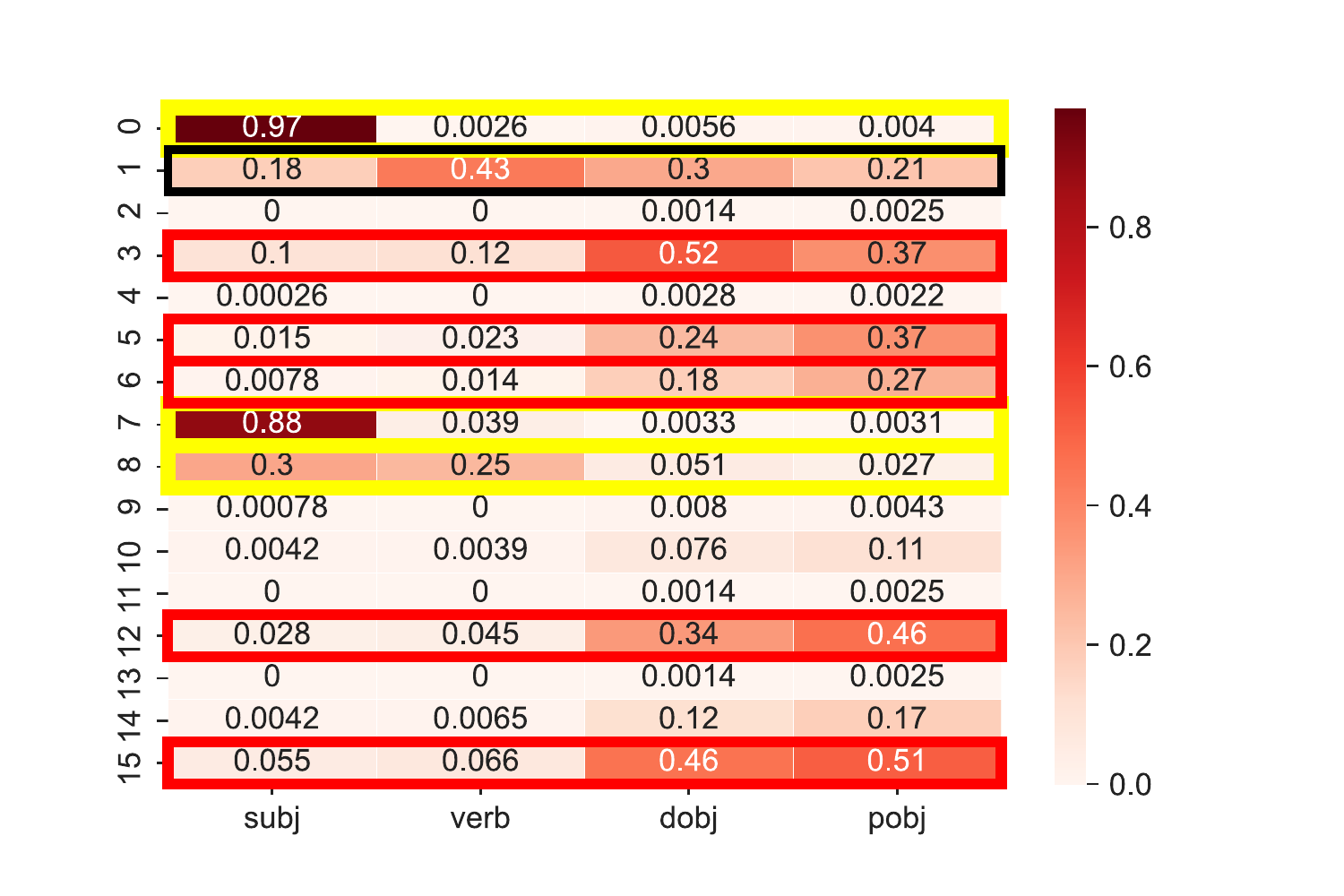}
            \end{adjustbox}
            \end{minipage}
            \caption{\centering Encoder influence heatmap for ADVAE with 16 latent variables on SNLI ($\Gammaopenc$). Squares with similar colors highlight groups of latent variables that relate to the same syntactic role.}
            \label{fig:ENCHEAT16}
    \end{minipage}
\hspace{0.7cm}
    \begin{minipage}[b]{0.48\textwidth}
            \centering
            \begin{minipage}[b]{\textwidth}
            \begin{adjustbox}{minipage=\textwidth,scale=0.6}
             \hspace{ 1cm} \includegraphics[trim={1.3cm 0.7cm 2.2cm 1.3cm},clip] {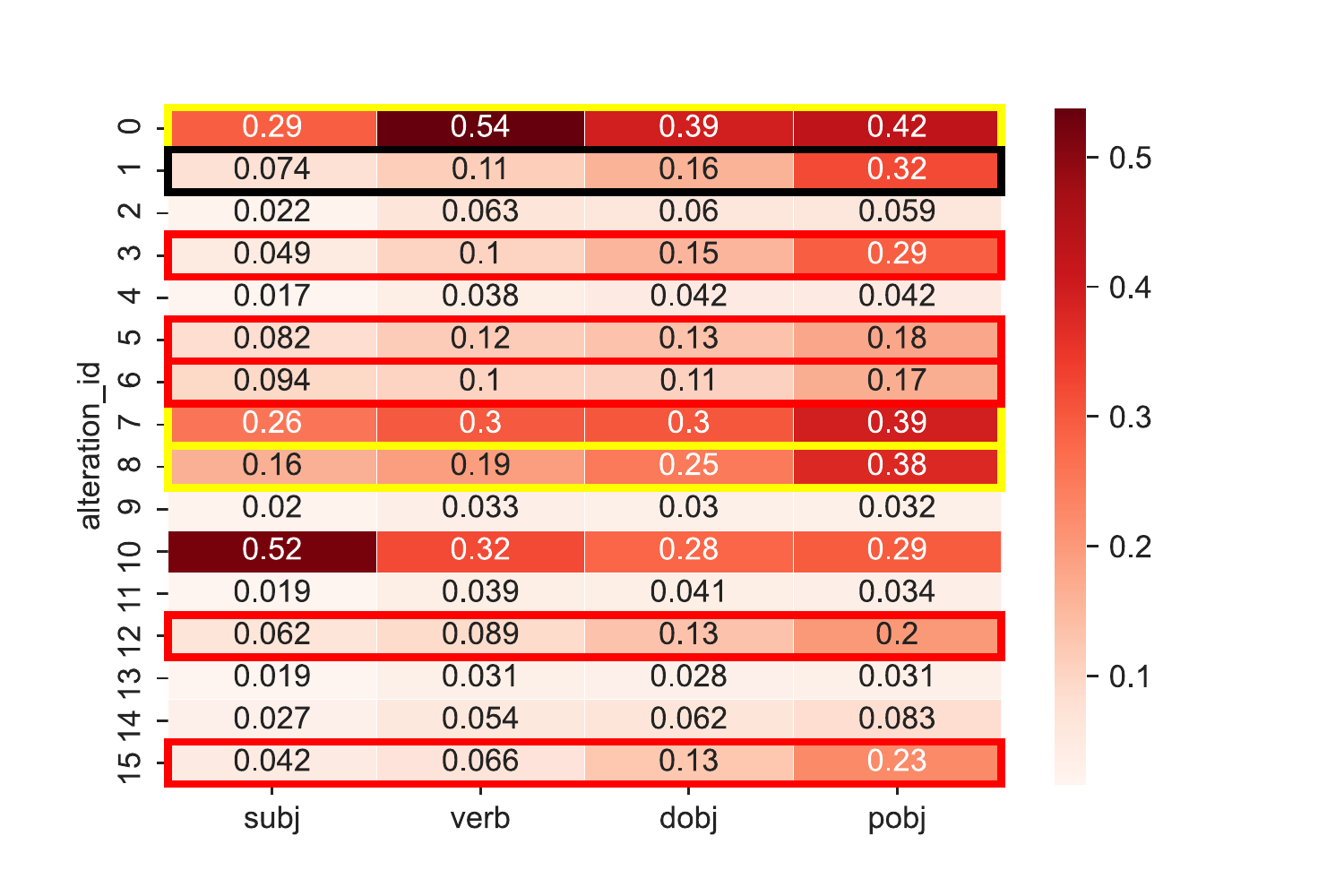}
            \end{adjustbox}
            \end{minipage}
            \caption{\centering Decoder influence heatmap for ADVAE with 16 latent variables on SNLI ($\Gammaopdec$). Squares with similar colors highlight groups of latent variables that relate to the same syntactic role. }
            \label{fig:DECHEAT16}
    \end{minipage}
\end{figure*}

As observed in the main body of the thesis, the increase of the number of latent variables used in ADVAE leads to dispatching the influence on the realization of a single syntactic role to multiple latent variables. This in turn,leads to the decrease observed for $\mathbb{D}_{enc}$ and $\mathbb{D}_{dec}$. In Figures~\ref{fig:ENCHEAT16} and~\ref{fig:DECHEAT16}, we respectively display the encoder and decoder heatmaps of ADVAE with 16 latent variables. As can be seen in these figures, latent variables still specializes in specific syntactic roles. This is seen more clearly on the encoder heatmap due to co-adaptation harming the clarity of the decoder heatmap. This specialization seems to be shared among groups (\textit{e.g.} variables 0, 7 and 8 specialize in the subject, as indicated by the green squares on the figure). This causes the difference of influence between the most influential  variable and the second most influential one to be low, and thus decreases the values of $\mathbb{D}_{enc}$ and $\mathbb{D}_{dec}$.
\begin{table}[!h]
    \centering
    \caption{Disentanglement quantitative results on SNLI for a larger grid of $L$ values.}
    \resizebox{0.7\textwidth}{!}{%
    \begin{tabular}{|c|c||c|c||c|c|}
    \hline
    Model& $\beta$ & $\mathbb{D}_{enc}$ &  $N_{\Gammaopenc}$  &  $\mathbb{D}_{dec}$ &  $N_{\Gammaopdec}$\\
    \hline \hline
    
    \multirow{2}{*}{ours-2} &0.3& 2.01\textcolor{gray}{(0.07)}& 2.00\textcolor{gray}{(0.00)}& 0.89\textcolor{gray}{(0.10)}& 2.00\textcolor{gray}{(0.00)}\\ 
     &0.4& 0.33\textcolor{gray}{(0.15)}& 1.60\textcolor{gray}{(0.55)}& 0.94\textcolor{gray}{(0.03)}& 2.00\textcolor{gray}{(0.00)}\\
    \hline
    \multirow{2}{*}{ours-4} &0.3& 1.30\textcolor{gray}{(0.09)}& 3.00\textcolor{gray}{(0.00)}& 0.78\textcolor{gray}{(0.10)}& 3.00\textcolor{gray}{(0.00)}\\ 
     &0.4& 1.46\textcolor{gray}{(0.33)}& 3.00\textcolor{gray}{(0.00)}& 0.84\textcolor{gray}{(0.10)}& 3.00\textcolor{gray}{(0.00)}\\
    \hline
    \multirow{2}{*}{ours-8} &0.3& 1.36\textcolor{gray}{(0.13)}& 3.40\textcolor{gray}{(0.89)}& 0.62\textcolor{gray}{(0.17)}& 3.20\textcolor{gray}{(0.45)}\\ 
     &0.4& 1.44\textcolor{gray}{(0.79)}& 3.40\textcolor{gray}{(0.55)}& 0.80\textcolor{gray}{(0.11)}& 3.00\textcolor{gray}{(0.00)}\\
    \hline
    \multirow{2}{*}{ours-16} &0.3& 0.60\textcolor{gray}{(0.31)}& 3.60\textcolor{gray}{(0.55)}& 0.38\textcolor{gray}{(0.20)}& 2.80\textcolor{gray}{(0.45)}\\ 
     &0.4& 0.65\textcolor{gray}{(0.16)}& 3.40\textcolor{gray}{(0.55)}& 0.50\textcolor{gray}{(0.28)}& 3.00\textcolor{gray}{(0.71)}\\
    \hline

 \hline
     \end{tabular}}
    \label{tab:NzVAry}
  \end{table}

\chapter{Unsupervised Disentanglement of Syntax and Semantics}
In this Appendix, we provide supplemental materials related to our work on unsupervised disentanglement of syntax and semantics. Appendix~\ref{DEVRES} displays results on the development set of ParaNMT to confirm our findings, and to enable future works to compare to our work without experimenting on the test set. The hyper-parameters used for our model as well as the strategy to find them are explicited in Appendix~\ref{HPAPPEN}.
\section{Results on the development set}
\label{DEVRES}

\begin{table}[b!]\small
  \centering
  \begin{tabular}{l c c}
  \hline
  &$z^{sem}\uparrow$ & $z^{syn}\downarrow$\\
  \hline\multicolumn{3}{c}{\textit{Supervised Models}}\\\hline
  VGVAE & 99.0& 16.4\\
  SynPG & 91.6& 31.2\\
  \hline\multicolumn{3}{c}{\textit{Unsupervised Models}}\\\hline
  Optimus & 89.4 & -\\
  ADVAE & 41.0& 40.3\\
  QKVAE & 86.7& 27.0\\
  \hline
  \end{tabular}
  \caption{\label{enc_res_dev} 
  The probability*100 that an embedding places a target sentence closer to its semantic source than it is to its syntactic source 
  in the embedding space. (development set results)}
  \end{table}
\begin{table}[b!]\small
\centering
\begin{tabular}{p{0.045\textwidth} c| c|| c| c|| c| c}
\hline
&\multicolumn{2}{c}{\textit{sem\_src}} & \multicolumn{2}{c}{\textit{syn\_src}}& \multicolumn{2}{c}{\textit{target}}\\
& \emph{M}$\uparrow$&\emph{PB}$\uparrow$
& \emph{M}$\downarrow$&\emph{PB}$\downarrow$
& \emph{M}$\uparrow$& \emph{PB}$\uparrow$ \\
\hline
\multicolumn{7}{c}{\textit{Control and Reference baselines}}
\\
\hline
\textit{sem\_src}    & 100 & 1.0 &  7.4& 0.13& 27.4&  0.82\\
\textit{syn\_src}    & 7.4& 0.13&  100 & 1.0 &  12.0& 0.16\\
Optimus & 13.00& 0.35& 13.4& 0.34$^\dag$&  10.5 & 0.32\\
\hline
\multicolumn{7}{c}{\textit{Supervised Models}}
\\
\hline
VGVAE    & 18.3& 0.58&  15.2& 0.17& 23.0 & 0.57\\
SynPG  & 47.6 & 0.86& 7.8& 0.11& 24.4& 0.73\\
\hline
\multicolumn{7}{c}{\textit{Unsupervised Models}}
\\
\hline
ADVAE    & 9.0& 0.20& 8.1& 0.17& 7.7& 0.19\\
QKVAE    & 13.4& 0.36&  11.3& 0.19 & 12.9 &0.35\\
\hline

\end{tabular}
\caption{\label{sem_res_dev} 
Semantic transfer results (development set results). The comparison 
scores between sentences and \emph{syn\_src} that are not significantly different from the same scores produced with regard to 
\emph{sem\_src} are marked with $^\dag$.
}
\end{table}

\begin{figure}[!h]
  \begin{minipage}[b]{\textwidth}
    \begin{adjustbox}{minipage=\textwidth,scale=0.59}
    \hspace{ 5cm} \includegraphics[trim={0.7cm 0.2cm 1.5cm 1.2cm},clip] {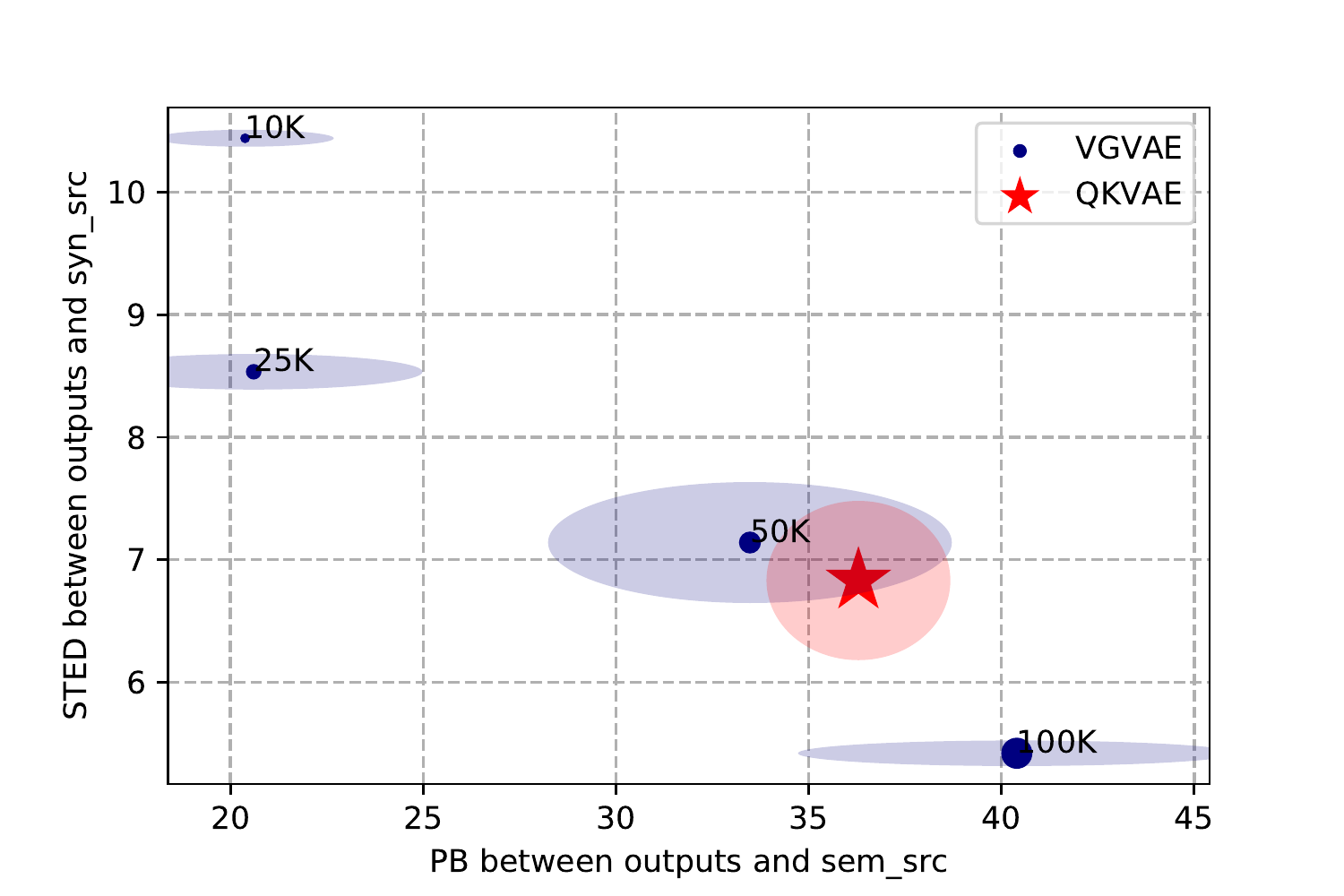}
  \end{adjustbox}
  \end{minipage}
  \caption{Plotting STED w.r.t \emph{syn\_ref} and the PB cosine similarity w.r.t \emph{sem\_ref} for VGVAE with different amounts of labeled data and for QKVAE.
   Points are scaled proportionally to the amount of training data. The vertical and horizontal diameters of each ellipse are equal
   to the standard deviation of the associated data points and axes.
   }
  \label{fig:STEDvsPBdev}
\end{figure}

\begin{table*}[b!]
  \normalsize
\centering
\resizebox{14cm}{!} 
{ 
\centering
\begin{tabular}{l c c c|| c c c|| c c c }
\hline
&\multicolumn{3}{c}{\textit{sem\_src}} & \multicolumn{3}{c}{\textit{syn\_src}}& \multicolumn{3}{c}{\textit{target}}\\
&\emph{STED}$\uparrow$ & \emph{TMA2}$\downarrow$ & \emph{TMA3}$\downarrow$ & 
\emph{STED}$\downarrow$ & \emph{TMA2}$\uparrow$ & \emph{TMA3}$\uparrow$ &
\emph{STED}$\downarrow$ & \emph{TMA2}$\uparrow$ & \emph{TMA3}$\uparrow$ \\
\hline
\multicolumn{10}{c}{\textit{Control/Ceiling baselines}}
\\ 
\hline
\textit{sem\_src} &0.0 & 100 & 100 & 11.9& 46.4& 6.8 & 10.9& 47.0&7.3 \\
\textit{syn\_src} & 11.9& 46.4& 6.8& 0.0& 100& 100& 6.0& 81.6& 45.0\\
Optimus & 9.7& 58.2& 20.6& 9.2$^\dag$& 61.6$^\dag$& 22.6$^\dag$& 9.9& 59.6& 18.4\\
\hline
\multicolumn{10}{c}{\textit{Supervised Models}}
\\
\hline
VGVAE    & 11.9& 45.4& 6.8& 3.2& 84.2& 58.2& 6.7& 77.6& 39.0\\
SynPG & 9.3& 49.4& 21.4& 12.2& 73.0& 12.2& 12.2& 68.6&13.0\\
\hline
\multicolumn{10}{c}{\textit{Unsupervised Models}}
\\
\hline
ADVAE    & 10.1& 53.4& 18.6& 9.8$^\dag$& 55.0$^\dag$& 17.4$^\dag$& 10.5& 52.8& 15.4\\
QKVAE    & 11.4& 45.0& 9.1& 6.8& 66.4& 37.4& 8.6& 63.0&26.9\\
\hline

\end{tabular}}
\caption{\label{syn_res_dev}
Syntactic transfer results (development set results). The comparison 
scores between sentences and \emph{syn\_src} that are not significantly different from the same scores produced with regard to 
\emph{sem\_src} are marked with $^\dag$.
}
\end{table*}
We display, here, the scores on the development set.
The encoder scores concerning the specialization of latent variables are in Table~\ref{enc_res_dev}, while the transfer scores
 are in Table~\ref{sem_res_dev} for semantics, and Table~\ref{syn_res_dev} for syntax. The values on the development set concerning
 the comparison of QKVAE with VGVAE trained on various amounts of data is in Figure~\ref{fig:STEDvsPBdev}.

\section{Hyper-parameters}
\label{HPAPPEN}
\paragraph{Hyper-parameter values}
The $\beta$ weight on the $\KL$ divergence is set to 0.6 for $z^c$ and to 0.3 for $z^s$, and 
the $\lambda$ threshold for the Free-Bits strategy is set to 0.05.
$\KL$ annealing is performed between steps 3K and 6K for $z^{sem}$, and between steps 7K and 20K for $z^{syn}$.
The model is trained using Adafactor~\cite{Shazeer2018Adafactor:Cost}, a memory-efficient version of Adam~\cite{Kingma2015}.
Using a batch size of 64, we train 
for 40 epochs, which takes about 30 hours on a single Nvidia GEForce RTX 2080 GPU. We use 4 layers for both Transformer encoders and decoders.
The encoders (resp. decoders) are initialized with parameters from the 4 first layers (resp. 4 last layers) of BART encoders (resp. decoders).
 In total, our model uses 236M parameters. 

 \paragraph{Manual hyper-parameter search} Given that the architecture for Transformer layers  is fixed by BART,
  we mainly explored 3 parameters: number of latent variables $L$, number of Transformer layers, and values for $\beta$. Our first 
  experiments have shown that setting $L$ to 8 or 16 does not yield good results, which is probably due to the fact that a high 
  $L$ raises the search space for possible arrangements of values with keys, and consequently makes convergence harder. Concerning
  the number of layers, we observed that results with the full BART model (6 layers) have high variance over different runs. Reducing
  the number of layers to 4 solved this issue. In regards to $\beta$, we observed that it must be $0.6$ or less for the model to 
  produce adequate reconstructions and that it is beneficial to set it slightly lower for $z^{syn}$ than for $z^{sem}$ so
   as to absorb more syntactic information with $z^{syn}$.

\end{document}